\documentclass{article}

\usepackage[final, nonatbib]{neurips_2022}

\usepackage[utf8]{inputenc} 
\usepackage[T1]{fontenc}    
\usepackage{hyperref}       
\usepackage{url}            
\usepackage{booktabs}       
\usepackage{amsfonts}       
\usepackage{nicefrac}       
\usepackage{microtype}      
\usepackage{xcolor}         
\usepackage{graphicx}
\usepackage{subcaption}
\usepackage{algorithm}
\usepackage{algorithmic}
\usepackage[numbers, compress, sort]{natbib}
\usepackage{graphicx}      
\usepackage{blkarray}      
\usepackage{multirow}
\usepackage[figuresright]{rotating}

\usepackage{amsmath}
\usepackage{amssymb}
\usepackage{mathtools}
\usepackage{amsthm}
\usepackage{thmtools, thm-restate}

\DeclareMathOperator*\argmin{arg\,min}
\DeclareMathOperator*\argmax{arg\,max}

\theoremstyle{plain}
\newtheorem{theorem}{Theorem}[section]

\theoremstyle{definition}
\newtheorem{definition}[theorem]{Definition}

\newtheorem{remark}[theorem]{Remark}

\title{SnAKe: Bayesian Optimization via Pathwise Exploration}

\author{%
  Jose Pablo Folch \thanks{Corresponding author: \texttt{jose.folch16@imperial.ac.uk}} \\
  Imperial College London \\
  London, UK \\
  \And
  Shiqiang Zhang \\
  Imperial College London \\
  London, UK \\
  \And
  Robert M Lee \\
  BASF SE \\
  Ludwigshen, Germany \\
  \And
  Behrang Shafei \\
  BASF SE \\
  Ludwigshafen, Germany \\
  \And
  David Walz \\
  BASF SE \\
  Ludwigshafen, Germany \\
  \And
  Calvin Tsay \\
  Imperial College London \\
  London, UK \\
  \And
  Mark van der Wilk \\
  Imperial College London \\
  London, UK \\
  \And
  Ruth Misener \\
  Imperial College London \\
  London, UK \\
}

\begin{document}

\maketitle

\begin{abstract}
  Bayesian Optimization is a very effective tool for optimizing expensive black-box functions. Inspired by applications developing and characterizing reaction chemistry using droplet microfluidic reactors, we consider a novel setting where the expense of evaluating the function can increase significantly when making large input changes between iterations. We further assume we are working asynchronously, meaning we have to select new queries before evaluating previous experiments. This paper investigates the problem and introduces `Sequential Bayesian Optimization via Adaptive Connecting Samples' (SnAKe), which provides a solution by considering large batches of queries and preemptively building optimization paths that minimize input costs. We investigate some convergence properties and empirically show that the algorithm is able to achieve regret similar to classical Bayesian Optimization algorithms in both synchronous and asynchronous settings, while reducing input costs significantly. We show the method is robust to the choice of its single hyper-parameter and provide a parameter-free alternative.
\end{abstract}


\section{Introduction} \label{sec: intro}

We introduce a method for black-box optimization that keeps step-wise input variations as small as possible. A black-box function is expensive to evaluate (with respect to time or resources), and we do not have access to gradients. Classically, black-box optimization finds an optimum by sequentially querying the function. We study a variation of this problem, with two important differences. First, we introduce the idea that large changes in inputs, between iterations, cause the function to become more expensive to evaluate. Second, we do not assume observations are available immediately: delay between querying the function and getting a result leads to asynchronous decision making. 

As a motivating example, consider a droplet microfluidic reactor \citep{teh2008droplet} (see Figure \ref{fig: reactor_fig} in Appendix). In such a reactor, we can quickly pump in chemicals, expose them to certain conditions, and collect the results of our experiments as they exit. However, large changes in temperature mean that the reactor is no longer in steady-state, and this makes the evaluations unreliable until the system stabilizes. Smaller changes can maintain a ``quasi-steady state'', as the system is easier to stabilize. Further, we have to wait for droplets to exit the reactor before obtaining observations.

More broadly, consider optimization requiring spatially continuous exploration. Black-box optimization has been investigated for optimal placement of pollution sensors \cite{hellan2021bayesian}, and machine learning methods for detecting the sources of pollution have been developed \cite{leu2019air}. Mobile pollution readers, such as those used by AirQo [\url{https://www.airqo.net/about}], would face similar challenges if black-box optimizers were used to track the pollution sources. \citet{samaniego2021bayesian} show a more concrete example where autonomous vehicles use black-box optimizers to monitor and track water contamination sources in the Ypacarai lake in Paraguay.

Classical Bayesian Optimization (BO) \citep{jones1998efficient, shahriari2016bo} provides effective solutions to black-box optimization; however, physics-based limitations are usually not taken into account \cite{thebelt2022maximizing}. Typically BO follows a myopic approach, in that BO chooses the next query based only on the current state of the surrogate model. Classical BO reduces uncertainty in unexplored areas and then returns to promising areas with no regard for the distance between consecutive query points. This means that BO will incur very high \textit{input costs}.

However, having zero changes in input space is obviously not a good solution. After all we want to explore the search space to find the optimum point. We seek an algorithm that preserves the essence of Bayesian Optimization. Consider a scenario where we know a large number of inputs we want to query. In this case, we could simply order the queries to attain the smallest input cost. A good solution would simply require selecting a large number of queries, creating an ordering, and then following the \textit{path} defined by the ordering. Once we start obtaining new information, we could update our beliefs and update our optimization path.

This paper proposes \textit{\textbf{S}eque\textbf{n}tial Bayesian Optimization via \textbf{A}daptive (\textbf{K})Connecting Sampl\textbf{e}s} (SnAKe). Just as the snake grows by carefully eating items in the classic arcade game, a SnAKe optimization path grows from carefully adding queries to the evaluation path.

\section{Related Work}

`Process-constrained Batch Bayesian Optimization' \citep{vellanki2017proccess-constrained}, is the closest analog to our setting of BO with input costs. \citet{vellanki2017proccess-constrained} navigate the physical limitations in changing the input space by \textit{fixing} the complicated inputs for every batch. Rather than fixing the difficult-to-change inputs, we \textit{penalize} large input variations in line with the costs (time and resources) of changing conditions. SnAKe decides when to make expensive input changes. Recently, \citet{ramesh2022movement} consider a similar cost setting to ours, inspired by applications to wind energy systems. \citet{waldron2019autonomous} compare the use of transient variable ramps (small input changes in a pre-determined manner) against a full steady-state design of experiments approach for learning parameters of chemical kinetic models. They show that the transient approaches give less precise estimates, but much faster. SnAKe combines the best of both worlds, automatically designing experiments while keeping input changes small.

`Cost-aware Bayesian Optimization' \citep{snoek2012practical, lee2020cost, luong2021adaptive} optimizes regret with respect to cost per iteration. These costs, which are fixed throughout the BO process, arise when some regions of the input domain are more expensive to evaluate than others. Similar ideas appear in Multi-fidelity BO \citep{kandasamy2019multi, poloczek2017multi} where cheap approximations for the objective function are used. \citet{bogunovic2016truncated} combines cost-aware and level-set estimation approaches to solve environmental monitoring problems. Diverging from prior works, our setting does not assume that different regions incur different costs. Instead, SnAKe addresses the setting where the \textit{difference} between adjacent inputs defines costs. The query order changes the cost, therefore we focus on optimization paths instead of optimization sets.

The baseline of Cost-aware BO, Expected Improvement per unit Cost (EIpu), defined as $\mathrm{EI}(x)/\mathrm{Cost}(x)$, is not directly applicable to our setting because the cost of not changing inputs, i.e.,\ choosing $x_t = x_{t-1}$, is zero.  We can instead use:
\begin{equation}
\label{eq: EIpu}
    \gamma \mathrm{EIpu}(x) = \mathrm{EI}(x) / (\gamma + \mathrm{Cost}(x, x_{t-1}))
\end{equation}
Three main issues: (a) Eq.\ (\ref{eq: EIpu}) introduces new hyper-parameter $\gamma$ with no obvious way to choose it, (b) Eq.\ (\ref{eq: EIpu}) effectively penalizes the acquisition function far away from the data, so near local optima we only penalize exploration and may over-exploit, (c) asynchronous extensions have a tendency to encourage batch diversity, so we introduce a new trade-off between penalizing locally and far away. As we shall see, SnAKe shows robust results with just one hyper-parameter.

\textit{Look ahead} Bayesian Optimization \citep{ginsbourger2010towards, lam2016bayesian, gonzalez16glasses} considers possible future queries to make the current choice less myopic. We could use it to select the inputs to query, as we could simply look-ahead at what any classical BO would choose and then order the queries accordingly. However, one is only able to look ahead for a few iterations, due to the computational complexity of looking far ahead into the future \citep{bertsekas2012dynamic, yue2020why-non-myopic-bo}. For example, \citet{lee2021nonmyopic} combines Markov Decision Processes \cite{puterman2014markov}, look-ahead and cost-aware ideas, but is limited by the short time horizons rollout can handle. Reinforcement Learning \cite{sutton2018reinforcement} can be used, but it requires access to a dedicated training environment. \citet{mutny2022active} show how similar ideas can be used for trajectory optimization in environmental monitoring.

A computationally cheaper alternative that allows us to select many inputs to query is Batch Bayesian Optimization (BBO) \citep{gonzalez2016batch, azimi2010batch}. BBO is the setting where we are able to parallelize function evaluations, and as such we want to select multiple queries simultaneously. \citet{gonzalez16glasses} and \citet{jiang20binoculars} link BBO with look-ahead BO, using a Local Penalization method \citep{gonzalez2016batch} and Expected Improvement (q-EI) \citep{ginsbourger2010qei}, respectively. Unfortunately, they restrict themselves to smaller batch sizes ($q \leq 15$) due to computational expense. \textit{Asynchronous} Bayesian Optimization \citep{pmlr-v84-kandasamy18a, alvi2019localpen} addresses the problem of choosing queries while waiting for delayed observations.


\section{Methods}

\subsection{Problem Set-up}

We consider finding the maximum, $x^* = \argmax_{x \in \mathcal{X}}f(x)$, of a black-box function, $f$, where $\mathcal{X}$ is a compact subset of $\mathbb{R}^d$. We assume $f$ is continuously differentiable and expensive to evaluate. We seek the optimum point while keeping the number of evaluations small. We evaluate the function sequentially, over a discrete and finite number of samples, $t = 1, ..., T$. For every query, $x_t$, we obtain a noisy observation of the objective, $y_t = f(x_t) + \eta_t$, where $\eta_t \sim \mathcal{N}(0, s^2)$ is Gaussian noise. 

We assume there is delay, $t_{delay}$, between choosing a query and getting an observation. So our data-set at iteration $t$ is given by $D_t = \{(x_i, y_i) : i = 1, ..., t - t_{delay} - 1\}$. If we set $t_{delay} = 0$, we revert to classical sequential Bayesian Optimization, otherwise we are in an \textit{asynchronous} setting.

Finally, we assume there is a \textit{known} cost to changing the inputs to our evaluation, $\mathcal{C}(x_t, x_{t+1})$. We use simple regret, $SR_t = f(x^*) - \max_{i = 1, ..., t}f(x_i)$ as the performance metric. We want to minimize regret, or equivalently, maximize $f$, for the smallest possible cumulative cost, $\sum_{t = 1}^{T-1}\mathcal{C}(x_{t}, x_{t+1})$.

We note that the problem definition is invariant to cost scaling relative to the objective, because an optimal trajectory will be defined solely by the shape of the function. Current cost-aware methods are not scale invariant, and as we shall see, SnAKe first chooses which points to query and then finds the most cost-efficient way of selecting them, making the method invariant to scaling of the cost.

As is common in Bayesian Optimization, we will model the black-box function by putting a Gaussian Process (GP) prior on $f \sim \mathcal{GP}(\mu_0, \sigma^2_0)$. Since we have Gaussian noise, the posterior, $f | D_t$ is also a GP, whose mean function, $\mu_t(\cdot)$, and covariance function, $\kappa_t(\cdot, \cdot)$, can be calculated analytically \citep{rasmussen2005gps}.

\subsection{General Approach}

For our general approach, we will seek to create a large \textit{batch} of queries that we want to evaluate, and then \textit{plan-ahead} a whole optimization path. This is useful for three reasons: (1) it allows us to order the queries in a way that reduces input cost. (2) It allows us to deal with any delay in getting observations, because we can pre-select future queries. (3) It remains computationally feasible to plan-ahead even for very large time horizons. We can follow this ordering or path until new information is available, after which we will update our path. 

While we will provide a specific way of creating, planning and updating the paths, we note that the general ideas can be extended to suit different settings. For example, Appendix \ref{sec: appendix_ypacarai} shows how to alter SnAKe to simultaneously optimize multiple, independent, black-box functions.

\subsection{Creating a Batch Through Thompson Sampling}

We need to produce batches that are representative of the current state of the surrogate model. In addition, the method should allow for big batch sizes, since we want to produce batches as big as our budget (which is usually much larger than the batch size most methods consider). For example, in a micro-reactor, we might be interested in batches that contain hundreds of points \citep{teh2008droplet}.

\citet{pmlr-v84-kandasamy18a} offers a promising solution where every point in the batch is independent. The method is based on Thompson Sampling, which uses the GP's inherent randomness to create a batch. Each batch point is chosen by drawing a realization of the GP, and optimizing it.
The queries will fill out the space, and they are more likely to be on promising, and unexplored areas. It should work very well in our context given we expect our initial batch sizes to be very large, so the sample should be representative of the current state of our surrogate model.

\subsection{Creating a Path via the Travelling Salesman Problem}

After selecting a batch of queries, $\mathcal{P}_t = \{x_t^{(i)}\}_{i = 1}^{\tilde{T}}$, we order them. We do this by embedding a graph into the batch, where the edge weights are the cost for changing one input to another. We then find the shortest path that visits every point, i.e., we solve the Travelling Salesman Problem (TSP) \citep{bellman1962dynamic, dorigo1997ant}. 
Section \ref{sec:computational_considerations} discusses the computational cost.

Mathematically, we define the graph $G = (V,\ E,\ W)$, with $V = \{i \in 1, ..., \tilde{T} : x_t^{(i)} \in \mathcal{P}_t \}$, $E = \{(i, j) : i,j \in 1, ..., \tilde{T}\}$, and $W = \{w_{ij} = \mathcal{C}(x_t^{(i)}, x_t^{(j)}) :  (i, j) \in E\}$, where $\tilde{T}$ is the number of batch samples. We solve the TSP in $G$ to obtain our latest optimization path. A simple example would be to try to minimize the total distance travelled in input space, by selecting the Euclidean norm as cost $\mathcal{C}(x_{t}, x_{t+1}) = || x_t - x_{t+1} ||$.

\subsection{Naively Updating the Optimization Path} \label{sec: naive_sampling}

After updating the GP with new observations, we want to use this information to update our path. We propose updating our strategy by creating a new batch of points.

\begin{figure}[ht]
	\centering
	\begin{subfigure}[t]{0.32\textwidth}
	\includegraphics[width = \textwidth]{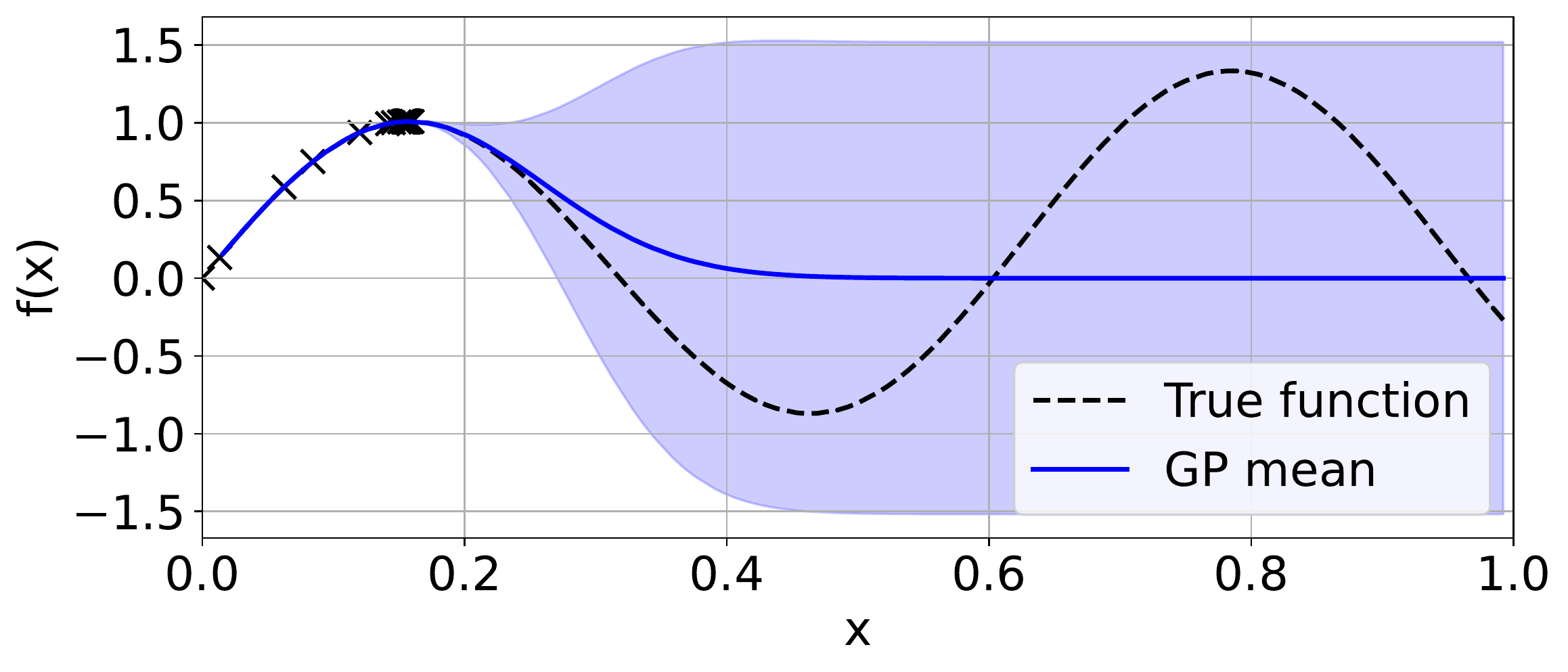}	
	\caption{Naively Resampling}
	\label{fig: no_escape_example}
	\end{subfigure}
	\hfill
	\begin{subfigure}[t]{0.32\textwidth}
	\includegraphics[width = \textwidth]{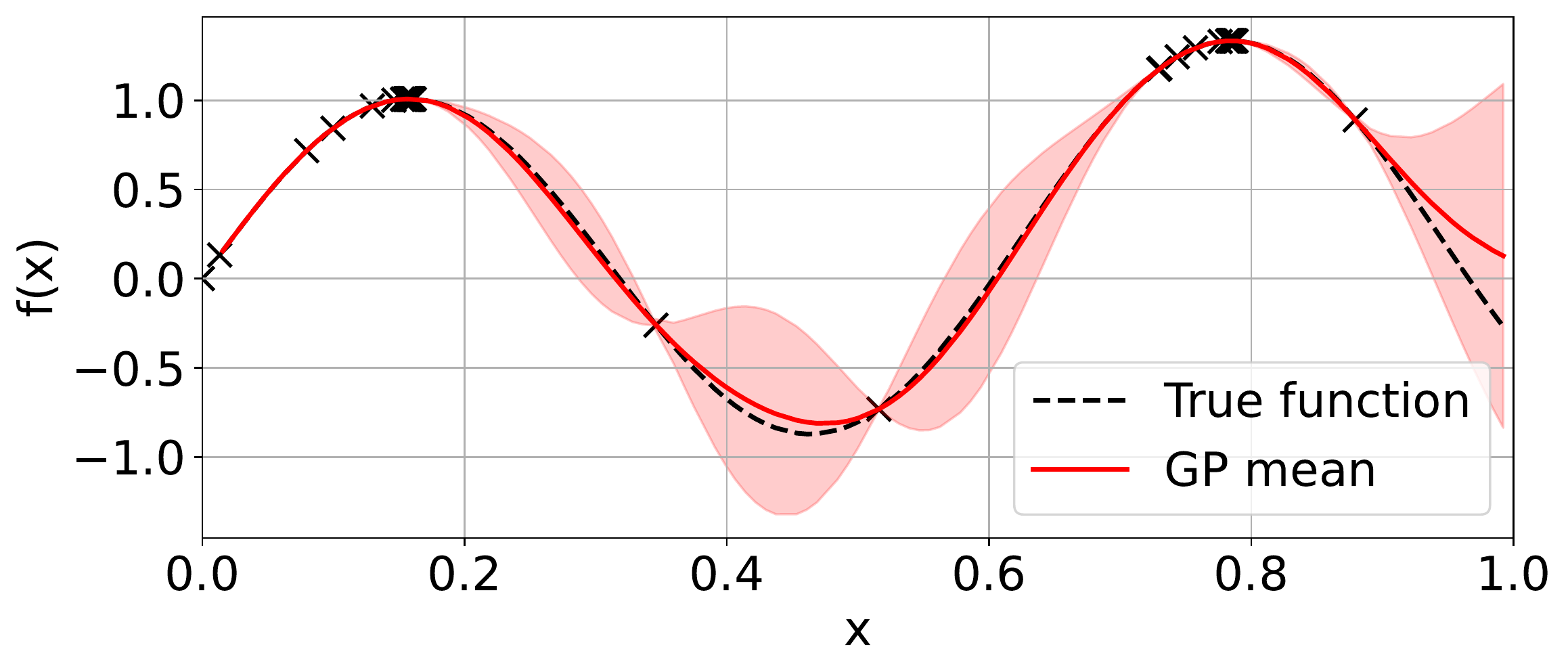}	
	\caption{With Point Deletion}
	\end{subfigure}
	\hfill
	\begin{subfigure}[t]{0.32\textwidth}
	\includegraphics[width = \textwidth]{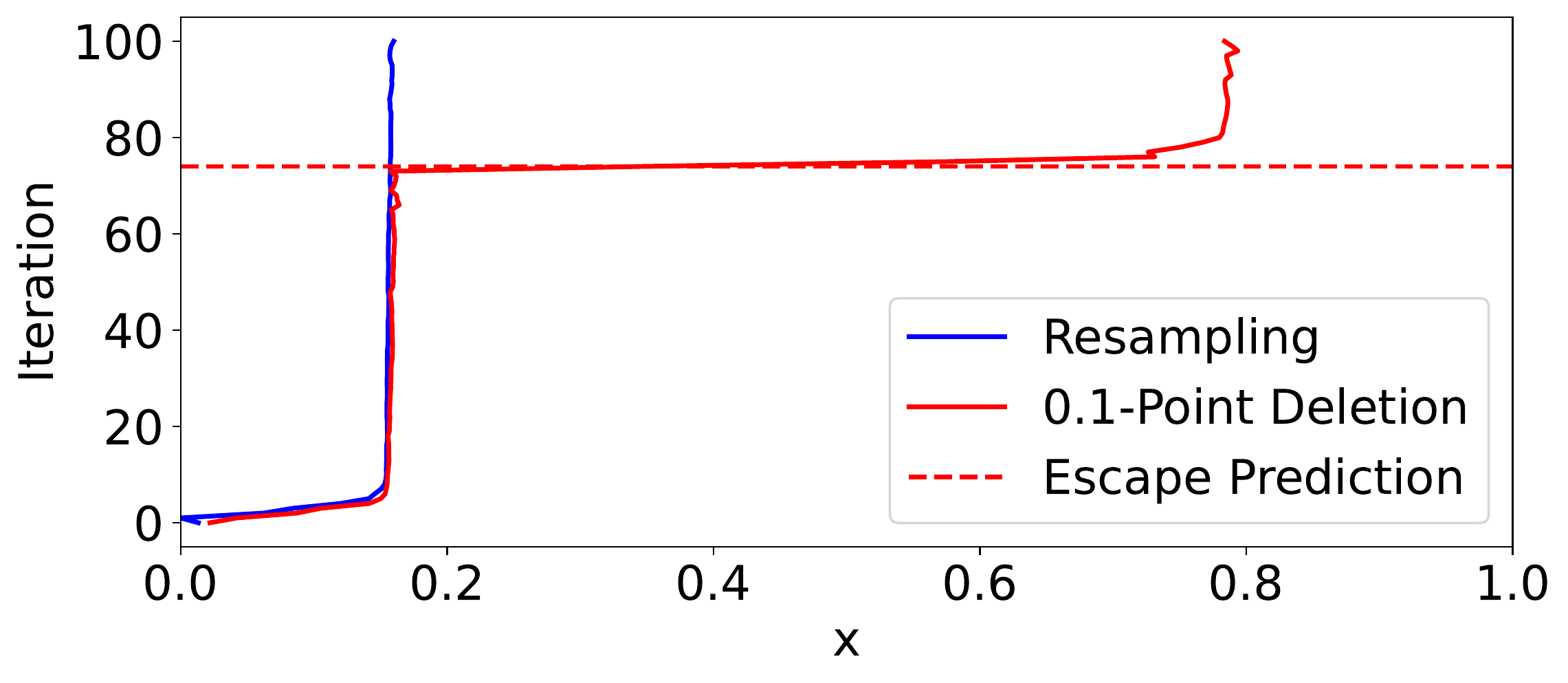}	
	\caption{Optimization paths.}
	\label{fig: point_del_escape_example}
	\end{subfigure}
	\caption{Effect of Point Deletion for ordering-based BO. (a) and (b) show the maximization objective and the underlying surrogate model, with black crosses representing queries and observations. (c) shows the optimization paths. Naive resampling gets stuck in the local optimum $(x = 0.16)$ whereas Point Deletion escapes the local solution. We also show the predicted escape iteration, $T p$ (after estimating $p \approx 0.74$ in Appendix \ref{sec: appendix_empirical_stationary}), which accurately predicts the algorithm behavior in this simple example.}
	\label{fig: escape_dynamics}
\end{figure}

At iteration $t$, the remaining budget has size $T - t$. We first propose sampling $T - t$ queries through Thompson Sampling, and then solving the Travelling Salesman Problem. However, we show this leads to the algorithm getting `stuck' in local optima. 
This is because every time we re-sample, we naturally include \textit{some exploitation} in the batch, and this exploitation will always be the next point chosen by the TSP--we will never reach the \textit{exploration} algorithm steps. 
Figure \ref{fig: no_escape_example} shows an example.

\subsection{Escape Analysis} \label{sec: importance_point_deletion}

To try and solve the convergence problem introduced by naive resampling, we briefly analyze it. This analysis assumes we receive noise-less observations. However, all sampling can be done in the presence of noise by calculating the corresponding posterior. We note that this section will provide intuition for the next step of the algorithm, however, we are not providing rigorous theoretical justification for the step or proving any type of regret bounds. We focus on Thompson Sampling.
\begin{definition}{(Thompson Sample)}
We say $x_t^{(i)}$ is Thompson Sampled, and we write $x^{(i)}_t \sim \tau_t$, if:
$$
    x_t^{(i)} = \argmax_{x \in \mathcal{X}}f^{(i)}_t(x), \ \ \text{where} \ \ f^{(i)}_t(\cdot) \sim \mathcal{GP}(\mu_t,\ \kappa_t | D_t)
$$
Note that the sampling distribution $\mathcal{GP}(\mu_t,\ \kappa_t | D_t)$ changes at each iteration.
\end{definition}

One particular concern, diagrammed in Figure \ref{fig: no_escape_example}, is the possibility of the method being stuck in a certain area. Let $B_{\delta}(a)$ be a Euclidean $\delta$-ball centered at $a$, and assume that $x_{t-1} \in B_{\delta}(a)$. We further assume two more things: (a) we resample the batch at every iteration, and (b) the TSP solution, or any approximation, will always choose a point in $B_{\delta}(a)$ to be the next step, if there are any. These assumptions represent the worst-case scenario for trying to escape a local optima.

\begin{definition}{(Non-escape Probability)} \label{def: non-escape probability}
We define the non-escape probability, $p_t$, at iteration $t$, as the probability of a Thompson Sample falling into $B_{\delta}(a)$. That is, for $x_t^{(i)} \sim \tau_t$:
\begin{equation}
    p_t = \mathbb{P}(x_t^{(i)} \in B_{\delta}(a)) = \mathbb{P}(||x_t^{(i)} - a|| \leq \delta)
\end{equation}
\end{definition}

Let $\mathcal{P}_t = (x_t^{(1)}, ..., x_t^{(T-t)}) \sim \tau_t \ i.i.d.$. Of particular interest to us, is the number of `non-escapes' in the sample, $N_t = | \mathcal{P}_t \cap B_{\delta}(a) |$. We are guaranteed to escape if no sample falls in $B_{\delta}(a)$, so we say we have \textit{fully} escaped if $x^{(i)}_t \notin B_{\delta}(a) \ \ \forall i$.

\begin{remark}
Note that $N_t \sim \text{ Binomial}(T-t,\ p_t)$ as all the samples are mutually independent, therefore the probability of fully escaping is $\mathbb{P}(N_t = 0) = (1 - p_t)^{T-t}$. This means we can only expect to fully escape if $p_t$ is very small.
\end{remark} 

Therefore we are interested in the behavior of $p_t$ as we gain more information about $f$ in $B_{\delta}(a)$. We now consider the circumstances under which $p_t$ becomes very small:

\textbf{Areas without stationary points:} Consider the case when our objective $f$ does not contain a stationary point in $B_{\delta}(a)$. Then the maximum of $f$ on the closure of the ball, $\bar{B}_{\delta}(a)$, must lie on the boundary of the ball. In particular, if we assume that our GP model has no error in $B_{\delta}(a)$, and assuming continuity of sample paths, then the non-escape probability must be zero, $p_t = 0$.

Intuitively, the area itself contains enough information to ensure, with complete certainty, that the global optimum does not lie in the area. We hope that, as we collect information inside areas without stationary points, $p_t \rightarrow 0$, and we will eventually leave them with small probability of returning. Appendix \ref{sec: appendix_empirical_no_stationary} shows an example of this happening very fast.

\textbf{Areas with stationary points:}
Areas with stationary points pose a much bigger problem. We will restrict our arguments to local maxima, since this is where we have observed the problem. Assume that $f$ has a local optimum in $B_{\delta}(a)$, higher than any other we have observed before. In this case, any sample taken from a Gaussian Process with no error in $B_{\delta}(a)$ will have a local maximum inside $B_{\delta}(a)$, and therefore it is possible that this local maximum is the global solution.

As we increase the information inside the area, $p_t$ is not guaranteed go down to zero. This makes intuitive sense; the only way of knowing if a local optimum is not a global optimum is by sampling away from it--therefore with limited information we will allocate a certain probability to the global optimum being inside $B_{\delta}(a)$. We include a clear example where $p_t \rightarrow p > 0$ in Appendix \ref{sec: appendix_empirical_stationary}.

Sampling consistently in a promising area is not necessarily a bad thing, indeed we want to exploit near possible global optimum candidates. However, the question then becomes, how long will it take us to leave a local optimum? Recall the probability of fully escaping is $(1 - p_t)^{T-t}$, and therefore it will be increasing as $t$ increases, even if $p_t$ is (almost) constant.

\begin{remark} \label{remark: escape_prob}
Assume that $p_t \rightarrow p > 0$, and that we have a high escape probability after $t_{e}$ iterations, i.e., $(1 - p)^{T - t_{e}}$ is large, leaving us $T - t_{e}$ iterations to explore the rest of the space. If we increase our budget from $T$ to $T'$, we will not have a high probability of escape until $(1 - p)^{T' - t} = (1 - p)^{T - t_{e}}$, i.e., $t = T' - T + t_{e}$. This leaves us with $T' - (T' - T + t_{e}) = T - t_{e}$ iterations to explore the rest of the space. 
Note that this is independent of $T'$, meaning that increasing our budget \textit{does not} increase our budget after leaving $B_{\delta}(a)$! Rather, it only means we will be stuck in $B_{\delta}(a)$ for a longer time. This is very concerning as the method will be very myopic; if it finds a local optimum, it is likely that it will spend a very large amount of the budget exploiting it.
\end{remark}

\subsection{Escaping with $\epsilon$-Point Deletion} \label{sec: theory_epsilon_point_deletion}

Intuitively, the convergence problem stems from the constant resampling, because in every iteration resampling is reintroducing exploitative points. We seek a way of making sure that the samples used by the algorithm to create paths take into account the previously evaluated designs. Penalizing the samples via distances to previous queries seems like a simple solution, e.g., similar to local penalization \cite{gonzalez2016batch}.
However, penalization introduces the need to carefully tune the strength of the penalization so that we can still exploit optima and remain cost-effective -- the parameterization is nontrivial.

Around a local maximum, if we assume that $p_t \rightarrow p > 0$, then we can interpret $p$ as the probability that the global maximum lies on the ball $B_\delta(a)$, as such, we focus on a method that has the property of exploiting the promising area for $T p$ iterations (that is, the total budget weighted by the probability). To achieve this, we propose creating a total of $T$ samples, and then deleting batch points that are similar to previously-explored points (i.e., we remove excess exploitation introduced by resampling). Algorithm \ref{algo: point_deletion}, which we term $\epsilon$-Point Deletion, still allows us to exploit local optima if we sample many points near them, however, it should eventually move on.

Two points to note: (a) Point Deletion uses the Euclidean norm, and it is independent of the cost function. This is because we are trying to escape local minima of simple regret. (b) After the deterministic removals, it is likely the batch size is still larger than our remaining budget, so we balance it by randomly removing points from the batch. 

\begin{algorithm}
\caption{$\epsilon$-Point Deletion} \label{algo: point_deletion}
\begin{algorithmic}
 \STATE \textbf{input:} New proposed batch $\mathcal{P}_t$ (size $T$), set of already queried points $Q_t$ (size $t$), and deletion distance $\epsilon$
 \FOR{$x \in Q_t$}
 \STATE $\Tilde{d} \leftarrow \min_{x' \in \mathcal{P}_t} || x - x' ||$
 \IF{$\Tilde{d} < \epsilon$}
 \STATE \texttt{\# find the closest point to the query $x$ in the new batch}
 \STATE $ \Tilde{x} \leftarrow \argmin_{x' \in \mathcal{P}_t} || x - x' || $ 
 \ELSE
 \STATE \texttt{\# else pick a random sample \tiny}
 \STATE $\Tilde{x} \leftarrow \text{Random}(\mathcal{P}_t)$
 \ENDIF
 \STATE \texttt{\# remove said point from the batch}
 \STATE $\mathcal{P}_t \leftarrow \mathcal{P}_t$ \textbackslash $\{ \Tilde{x} \}$
\ENDFOR
\STATE \textbf{output:} A batch $\mathcal{P}_t$ (size $T - t$)
\end{algorithmic}
\end{algorithm}

$\epsilon$-Point Deletion allows us to escape local optima by directly increasing the probability of fully escaping without changing $p_t$. Let $q_t$ be the number of previously queried points inside the ball, and set $\epsilon \geq 2 \delta$. Indeed, it follows that we will escape if $N_t \leq q_t$, (as $q_t$ samples are guaranteed to be deleted due to previously queried points) which is much better than requiring $N_t = 0$.

Due to over-sampling, the expected number of `non-escapes' is given by $\mathbb{E}[N_t] = p_t T$. Around a local maximum, if we assume that $p_t \rightarrow p > 0$, and hence, $p_t T \approx pT$, then we can reasonably expect an escape when $q_t = pT$, that is, we will exploit the local maximum for approximately $pT$ iterations. This time we leave $T - q_t = T(1-p)$ extra iterations to explore the remaining space! Increasing the budget will benefit \textit{both} the exploitation and the exploration instead of only the former (in contrast with Remark \ref{remark: escape_prob}).

Figure \ref{fig: escape_dynamics} gives an empirical example where we use Point Deletion, with $\epsilon = 0.1$, to escape a local optimum. We observe the expected behavior from our brief analysis. For Point Deletion, we calculate the escape prediction as $p T \approx 74$, using $\hat{p} \approx 0.74$, which we estimated in Appendix \ref{sec: appendix_empirical_stationary}. We can see that without Point Deletion, we remain stuck in the first local optimum.

\subsection{SnAKe}

\begin{figure}[ht]
	\centering
	\begin{subfigure}[t]{0.3\textwidth}
	\includegraphics[width = \textwidth]{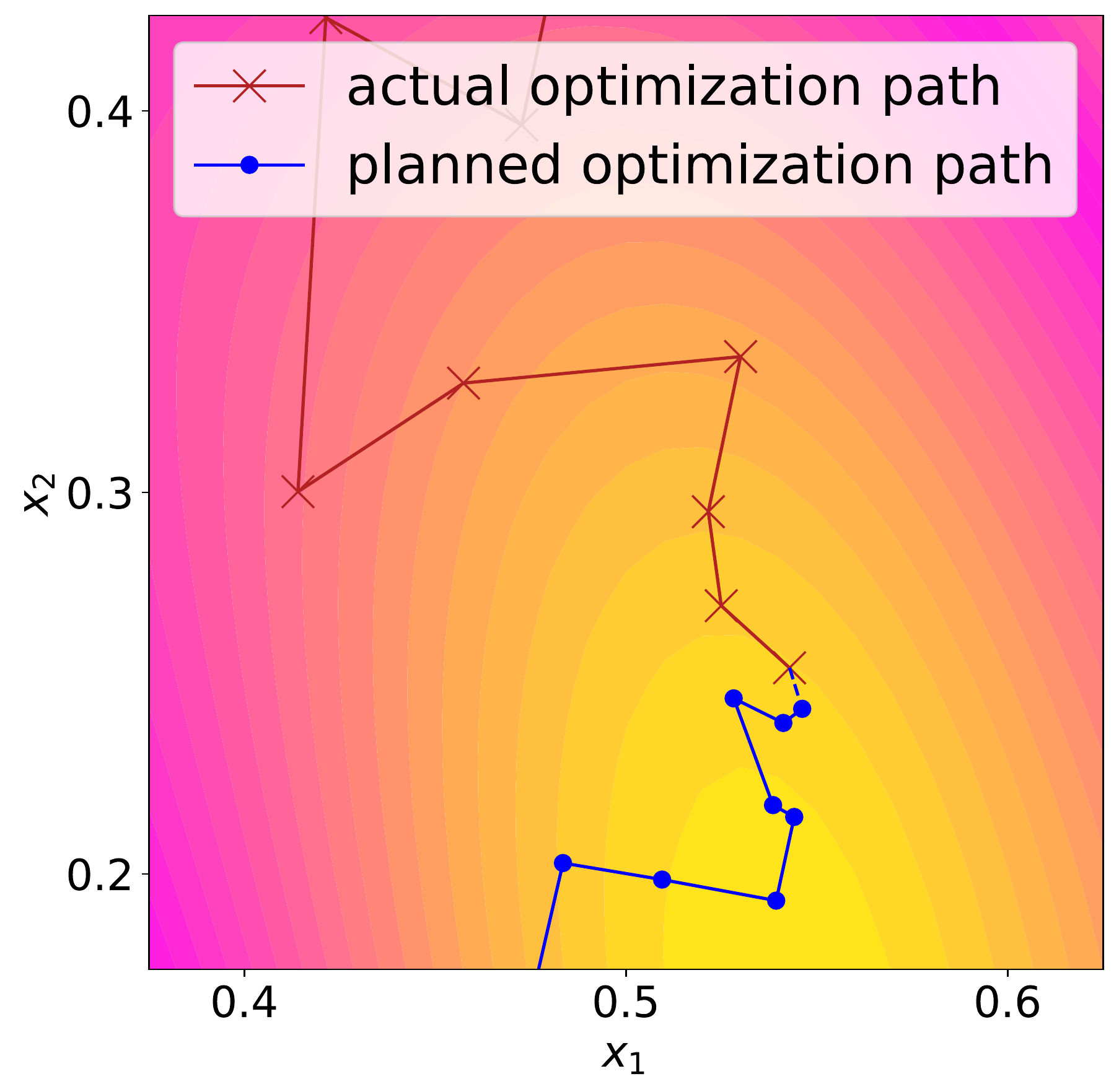}	
	\caption{Optimization path at $t = 10$}
	\end{subfigure}
	\hfill
	\begin{subfigure}[t]{0.3\textwidth}
	\includegraphics[width = \textwidth]{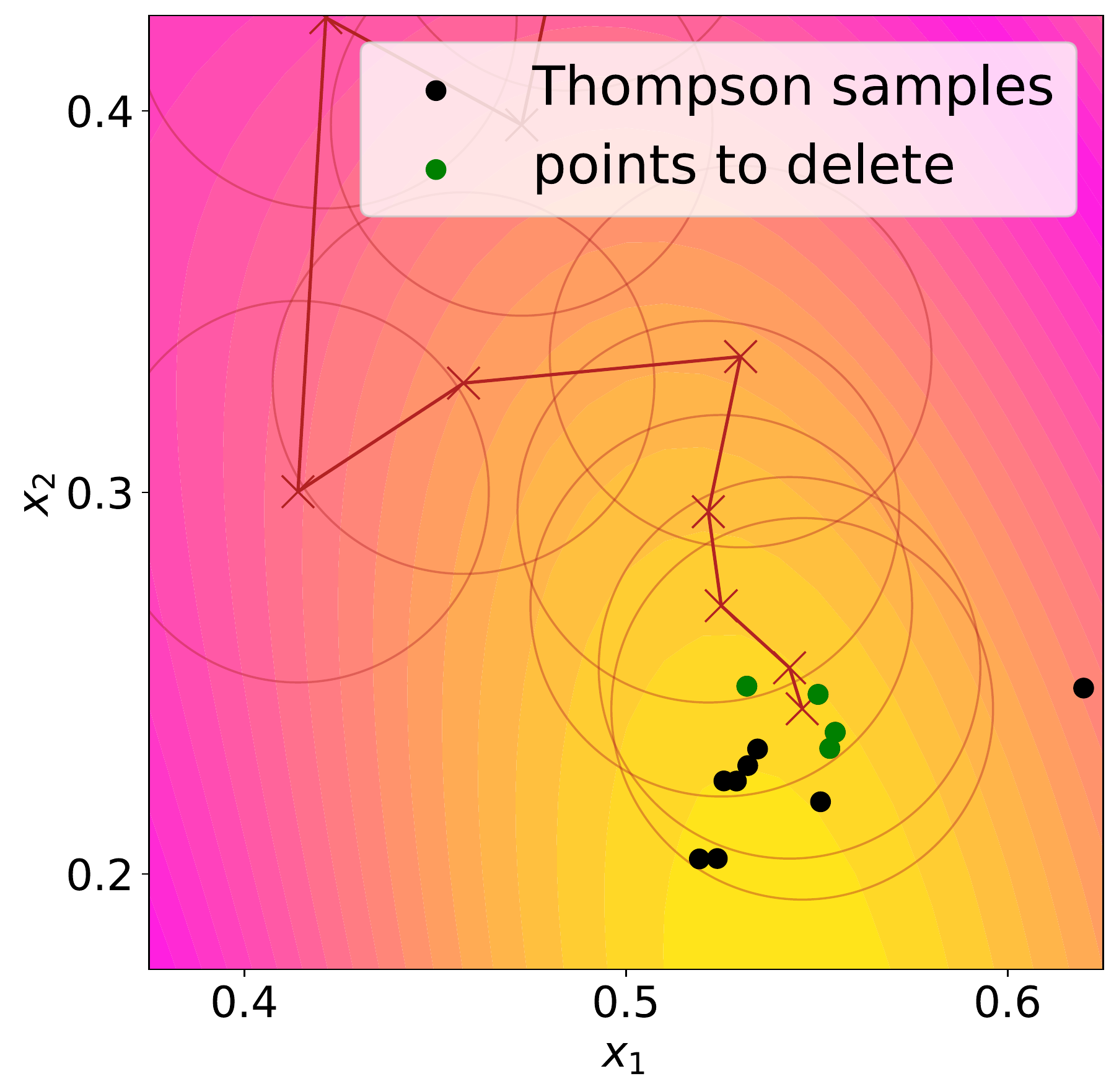}	
	\caption{Sampling \& Deletion Step}
	\end{subfigure}
	\hfill
	\begin{subfigure}[t]{0.3\textwidth}
	\includegraphics[width = \textwidth]{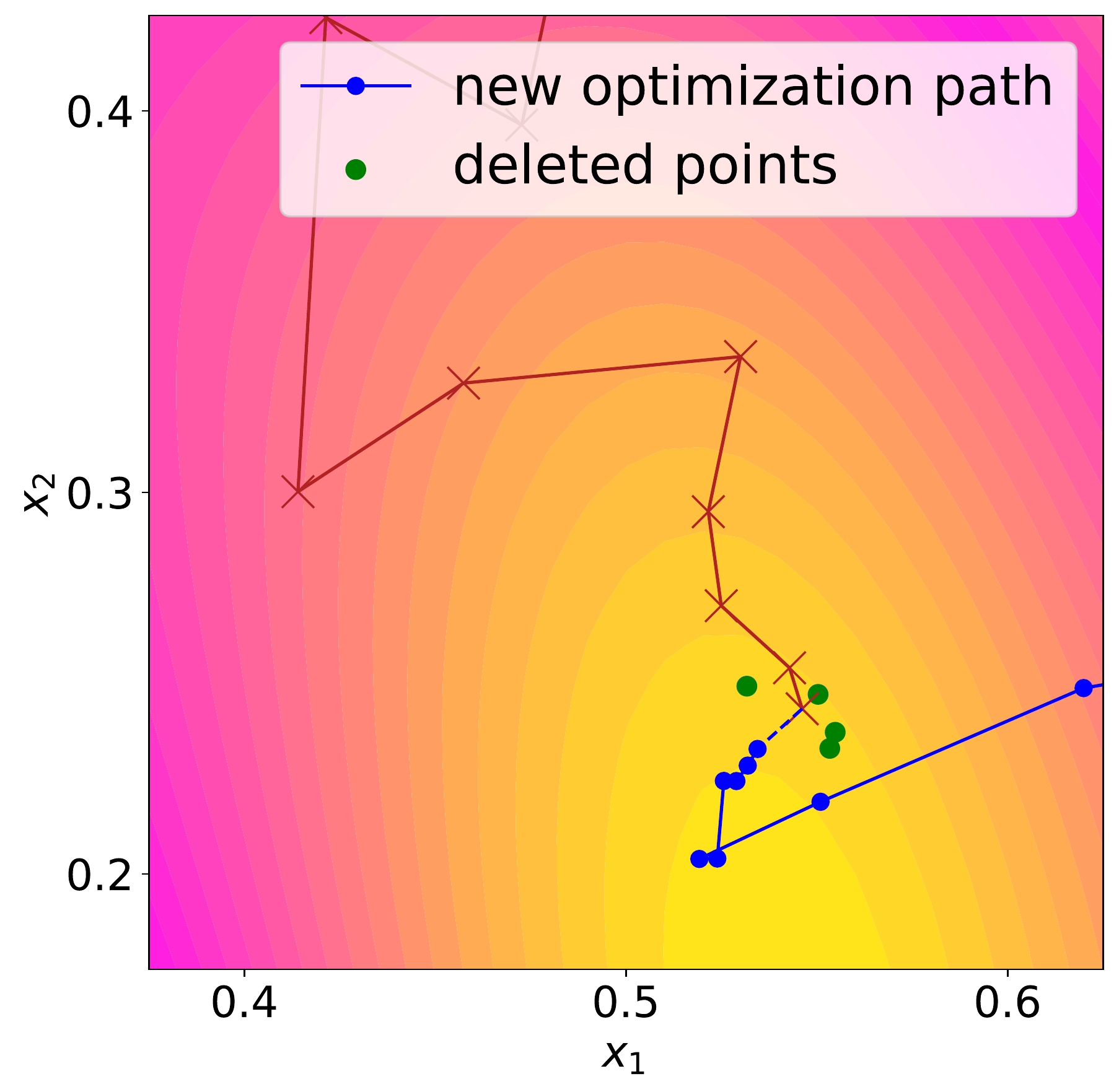}	
	\caption{Optimization path at $t = 11$}
	\end{subfigure}
	\caption{Graphical example of SnAKe behavior in a full iteration if new information is available. The underlying function is Branin 2D. The feasible set is $[0, 1]^2$, so we do not see all the samples or the complete path with this zoomed-in view. (a) The red line shows the already-queried path. The blue path shows our future plans. (b) For each query, we can see the $\epsilon$-ball under which the deletion step is deterministic. We plot the Thompson Samples as dots, the accepted ones in black and the deleted ones in green. (c) The new path is in blue and the points ignored (due to Point Deletion) are in green. Note the higher concentration of samples in what the model considers a promising area.}
	\label{fig: EaS-graphical-demo}
\end{figure}

Algorithm \ref{algo: EaS} which we dub `Sequential Bayesian Optimization via Adaptive Connecting Samples' (SnAKe), combines the ideas of previous sections. Figure \ref{fig: EaS-graphical-demo} diagrams the most important steps of SnAKe. Section \ref{sec: experiments} develops an effective, parameter-free alternative to the choice of $\epsilon$.

Note there is no requirement for data to be available immediately following querying. If $t_{delay} > 0$, we can simply stick to the latest path. It works without modification on the asynchronous setting. This is vital since we were inspired by chemical experiment design which can exhibit asynchronicity. We also note that the problem is invariant to cost scaling, as the batch creation and point deletion steps are independent of the cost, and the solution to the Travelling Salesman Problem is also scale-invariant.

\begin{algorithm}
\caption{SnAKe} \label{algo: EaS}
\begin{algorithmic}
 \STATE \textbf{input}: Optimization Budget, $T$. Deletion constant, $\epsilon$.
 \STATE \textbf{begin}: Create initial batch, $\mathcal{P}_0$, uniformly. Choose starting point $x_0$. $Q_0 \leftarrow \{x_0\}$. Create initial path, $S_0$, by solving TSP on $\mathcal{P}_0$.
 \FOR{$t = 1, 2, 3, ..., T$}
 \STATE Check if any running evaluations are finished
 \IF{there are new observations}
 \STATE Update surrogate model
 \STATE Create batch of size $T$ using Thompson Sampling, $\mathcal{P}_{t-1}$
 \STATE $\mathcal{P}_{t-1} \leftarrow $ $\epsilon$-Point Deletion($\mathcal{P}_{t-1}$, $Q_{t-1}$)
 \STATE $\Tilde{S} \leftarrow $ TSP($\mathcal{P}_t$, source = $x_{t-1}$) \textbackslash $\{ x_{t-1} \}$
 \STATE $S \leftarrow Q_{t-1} \cup \Tilde{S}$
 \ENDIF
 \STATE Choose next query point from schedule: $x_{t} \leftarrow S_{t}$
 \STATE $Q_{t} \leftarrow Q_{t-1} \cup \{ x_{t} \}$
 \STATE Evaluate $f(x_t)$
 \ENDFOR
\end{algorithmic}
\end{algorithm}

\subsection{Computational Considerations}\label{sec:computational_considerations}

We use \citet{wilson2020efficiently} to efficiently sample the GP and optimize the samples using Adam \cite{kingma2014adam}. The optimization step may be expensive, but it is highly parallelizable if required. We heuristically solve the Travelling Salesman Problem with Simulated Annealing \cite{kirkpatrick1983simmulated} and use an adaptive grid to reduce the number of samples away from our current input. The grid consists of the closest $N_l$ samples to our current input; the remaining samples are assigned to one of $N_g$ points in a global grid. Coarsifying the placement of the far-away points effectively ignores detailed paths far away from our current input. These far-away points are not important to the problem, as long as we resample frequently enough. We introduce two hyper-parameters ($N_l, N_g$), but they should not impact the solutions assuming $N_l$ is reasonably larger than any possible delay. With these modifications, we were able to comfortably run all experiments in a CPU (2.5 GHz Quad-Core Intel Core i7), where SnAKe shared a wall-time similar to Local Penalization methods. Appendix \ref{sec: appendix_computational} gives more details.

\section{Experimental Results} \label{sec: experiments}

For all experimental results we report the mean and the standard deviation over 25 experimental runs. We give the full implementation details and results in Appendix \ref{sec: appendix_implement_details} and \ref{sec: appedix_full_results} respectively. Classical BO methods are implemented using BoTorch \citep{balandat2020botorch} and GPyTorch \citep{gardner2018gpytorch}. The code to replicate all results is available online at \url{https://github.com/cog-imperial/SnAKe}.

In all experiments, we examine SnAKe for $\epsilon = 0, 0.1,$ and $1$. We further introduce a parameter-free alternative by adaptively selecting $\epsilon$ to be the smallest length scale from the GP's kernel, and denote it $\ell$-SnAKe. SnAKe proves to be robust to non-zero choices of $\epsilon$. We conjecture this happens because of Thompson Sampling's exploitative nature \citep{nava2022diversified}, so all excess exploitation is concentrated on a very small area. For $\epsilon = 0$ we observe very low cost, at the expense of some regret.

Across all experiments we can see how SnAKe can traverse the search space effectively, constantly achieving good regret at low cost. While other cost-aware methods may also achieve good performance, they are very inconsistent, suggesting they require careful tuning. SnAKe achieves good, robust results with a single hyper-parameter. We do highlight that SnAKe performs best when there are enough iterations for the samples to adequately fill the search space: this is highlighted by poor performance in Perm10D and by ordinary performance in low-budget experiments (see Appendix \ref{sec: appedix_full_results}, where we include performance comparisons for different budget sizes, $T$).

\begin{figure*}[ht]
	\centering
	\begin{subfigure}[t]{0.32\textwidth}
	\includegraphics[width = \textwidth]{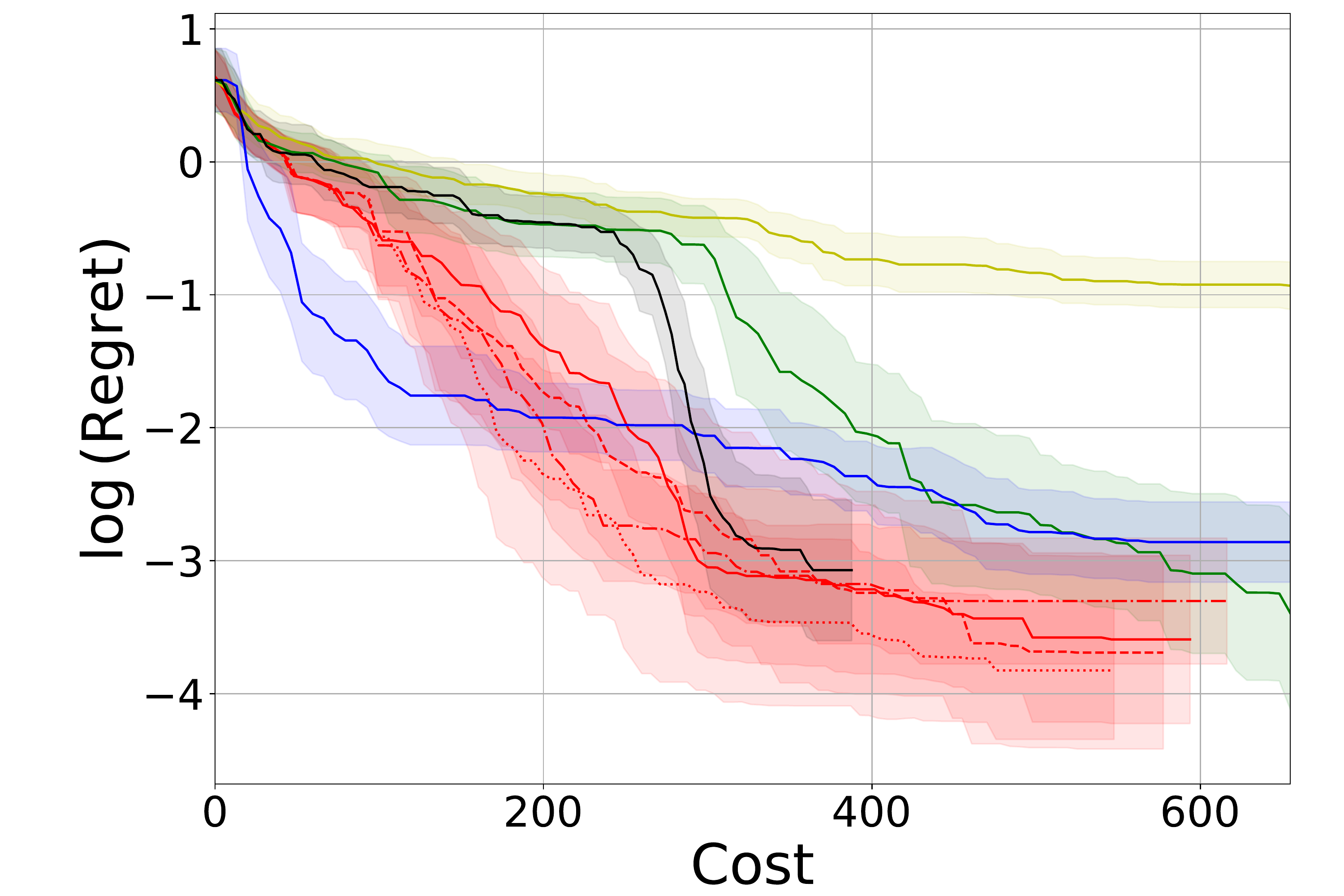}
	\caption{Evolution of regret against input cost. We can see that SnAKe is able to achieve the best regret for low cost.}
	\end{subfigure}
	\hfill
	\begin{subfigure}[t]{0.32\textwidth}
	\includegraphics[width = \textwidth]{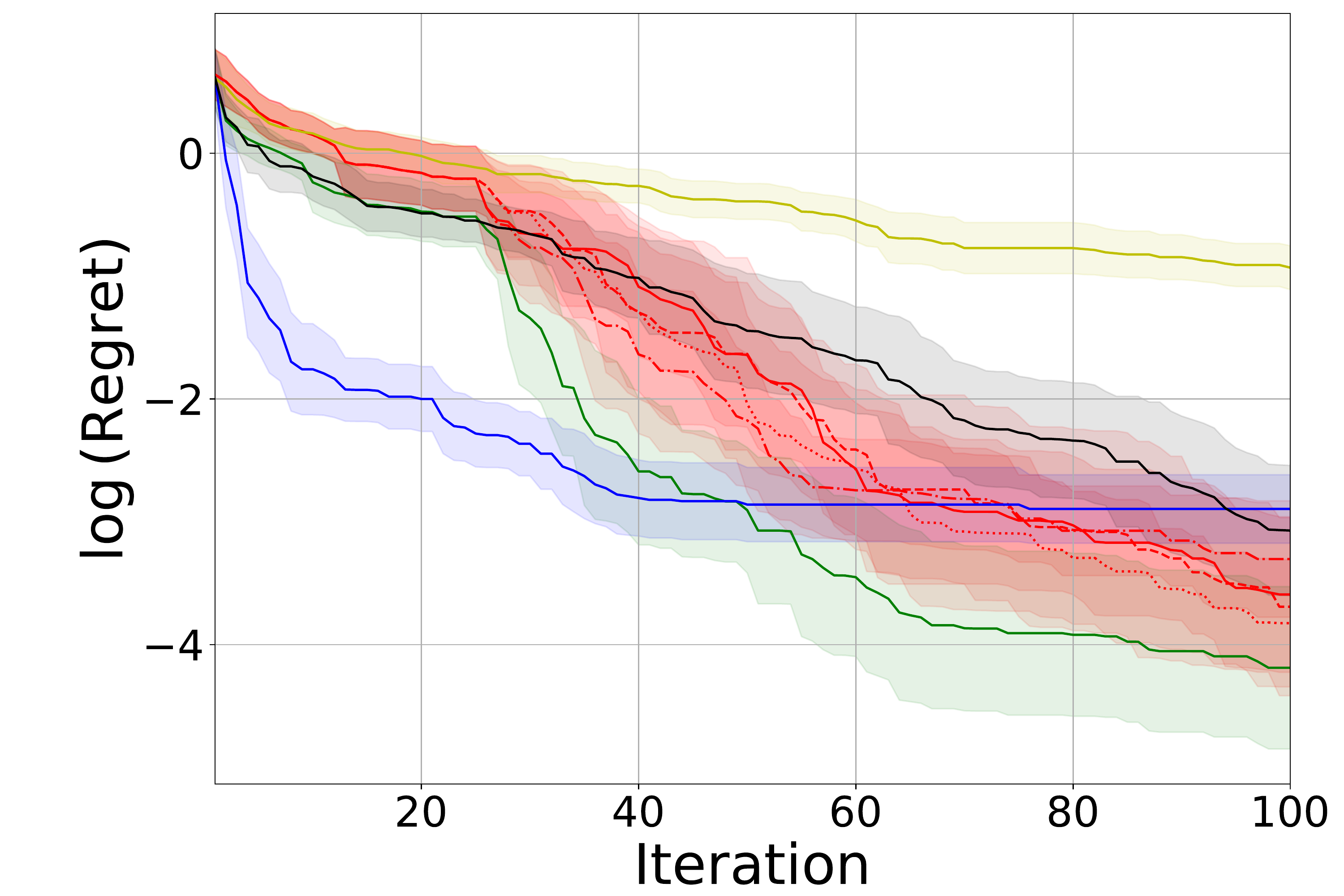}
	\caption{Evolution of regret with iteration number. SnAKe's final regret is comparable with classic BO methods, and better than EIpuLP.}
	\end{subfigure}
	\hfill
	\begin{subfigure}[t]{0.32\textwidth}
	\includegraphics[width = \textwidth]{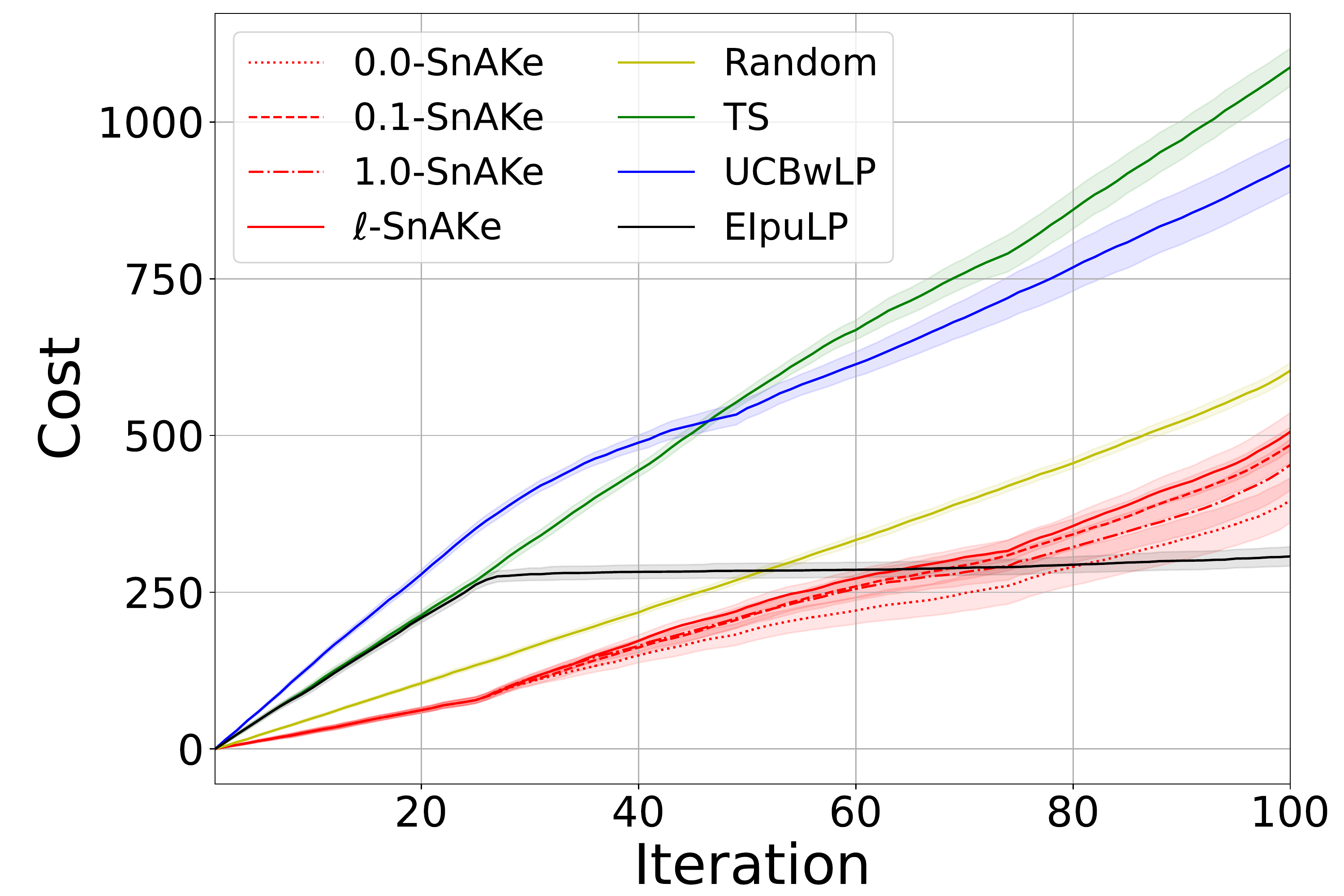}
	\caption{Evolution of input cost with iteration number. Low cost is achieved by EIpuLP, SnAKe and the TSP-ordered Random optimization.}
	\end{subfigure}
	\caption{Results of experiments on the SnAr chemistry Benchmark. SnAKe achieves the best regret out of all low-cost methods. The bounds are created from $\pm$ half the standard deviation of all runs. The best performer, $\ell$-SnAKe, is parameter-free.}
	\label{fig: snar_experiment}
\end{figure*}

\subsection{Synthetic Functions} \label{subsec: synthetic_experiments}

\begin{figure*}[ht]
    \begin{subfigure}{0.32\textwidth}
	\includegraphics[width = \textwidth]{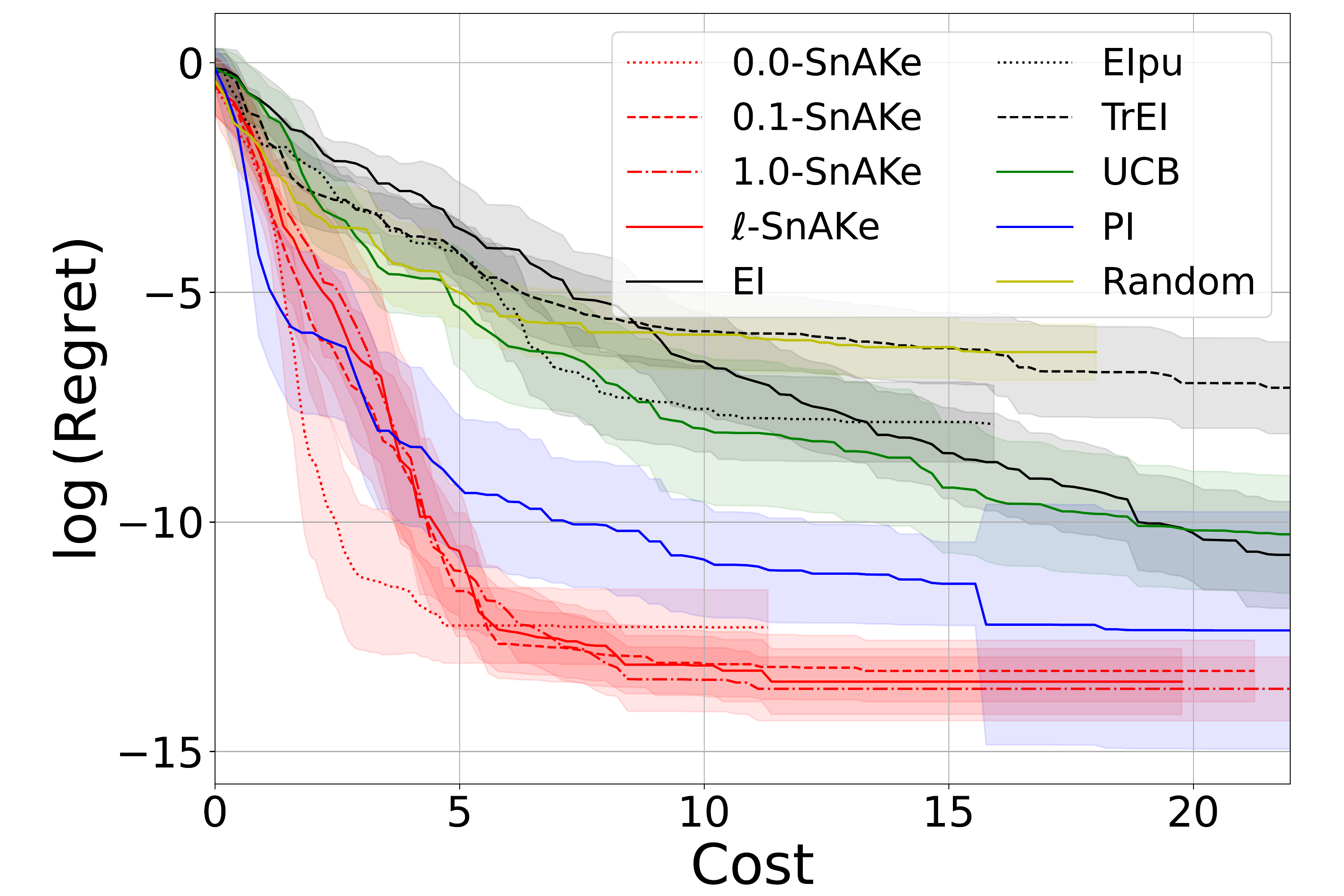}
	\caption{Branin2D}
	\end{subfigure}
	\hfill
	\begin{subfigure}{0.32\textwidth}
	\includegraphics[width = \textwidth]{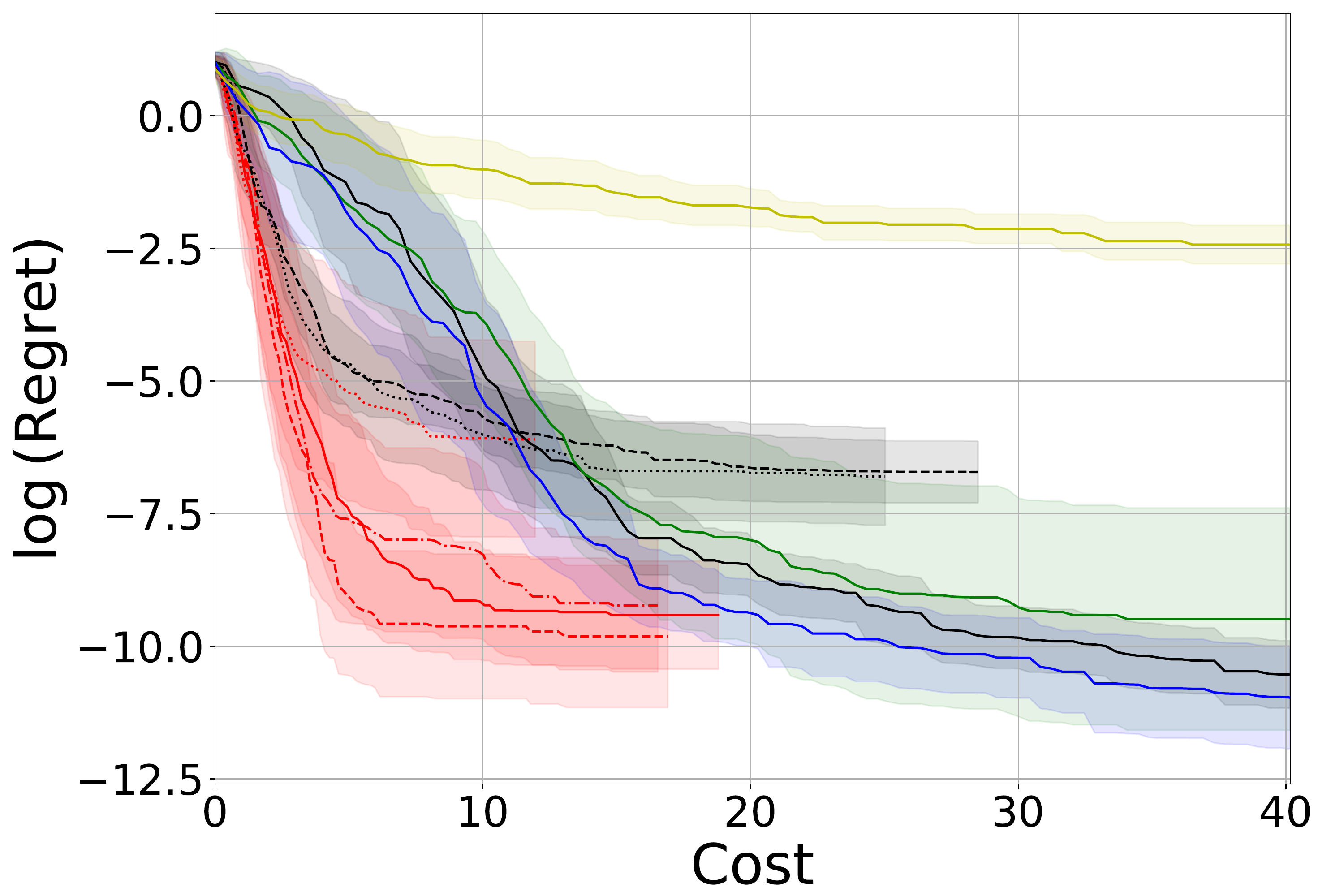}	
	\caption{Hartmann3D}
	\end{subfigure}
	\hfil
	\begin{subfigure}{0.32\textwidth}
	\includegraphics[width = \textwidth]{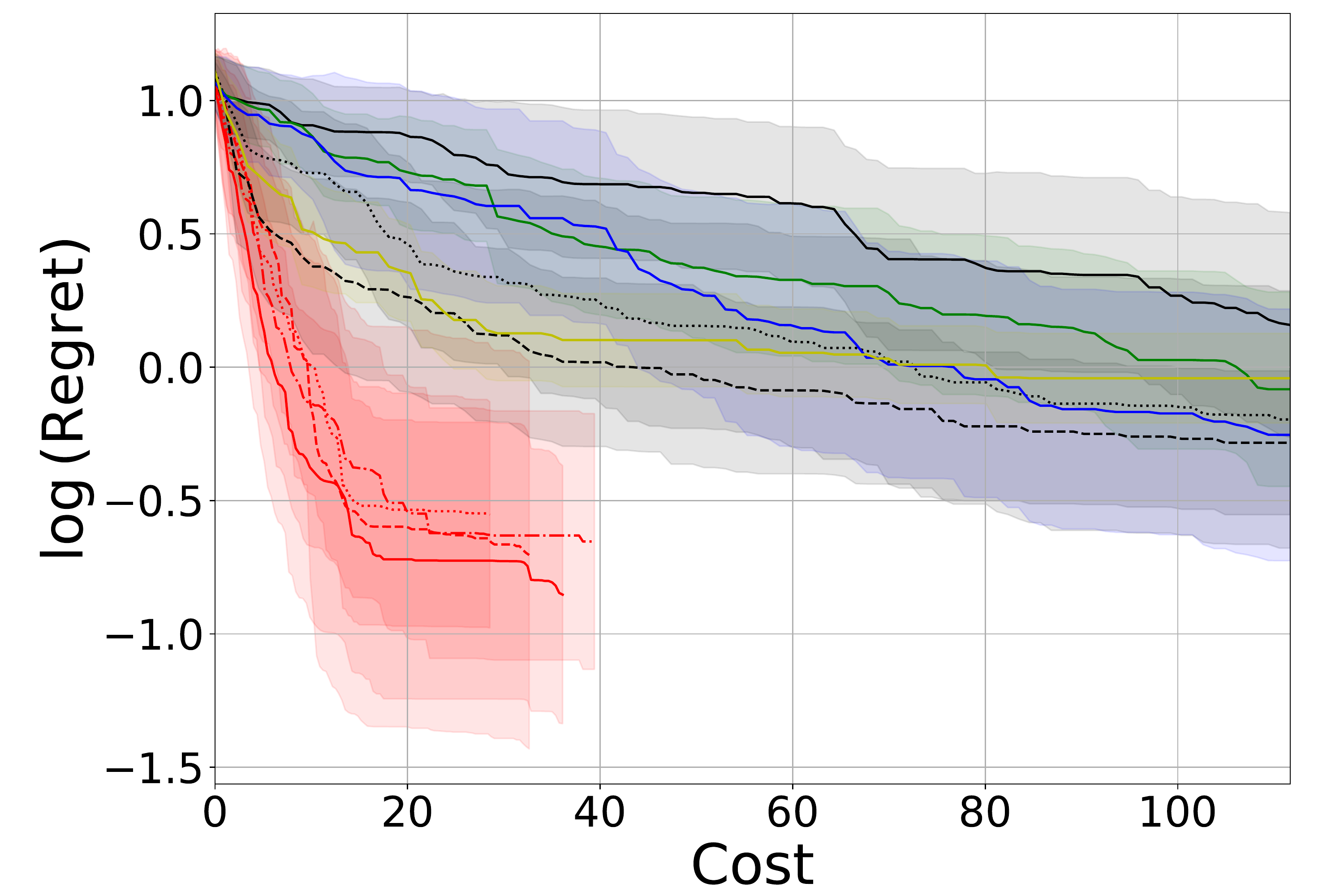}	
	\caption{Hartmann6D}
	\end{subfigure}
	\hfill
    \begin{subfigure}{0.32\textwidth}
	\includegraphics[width = \textwidth]{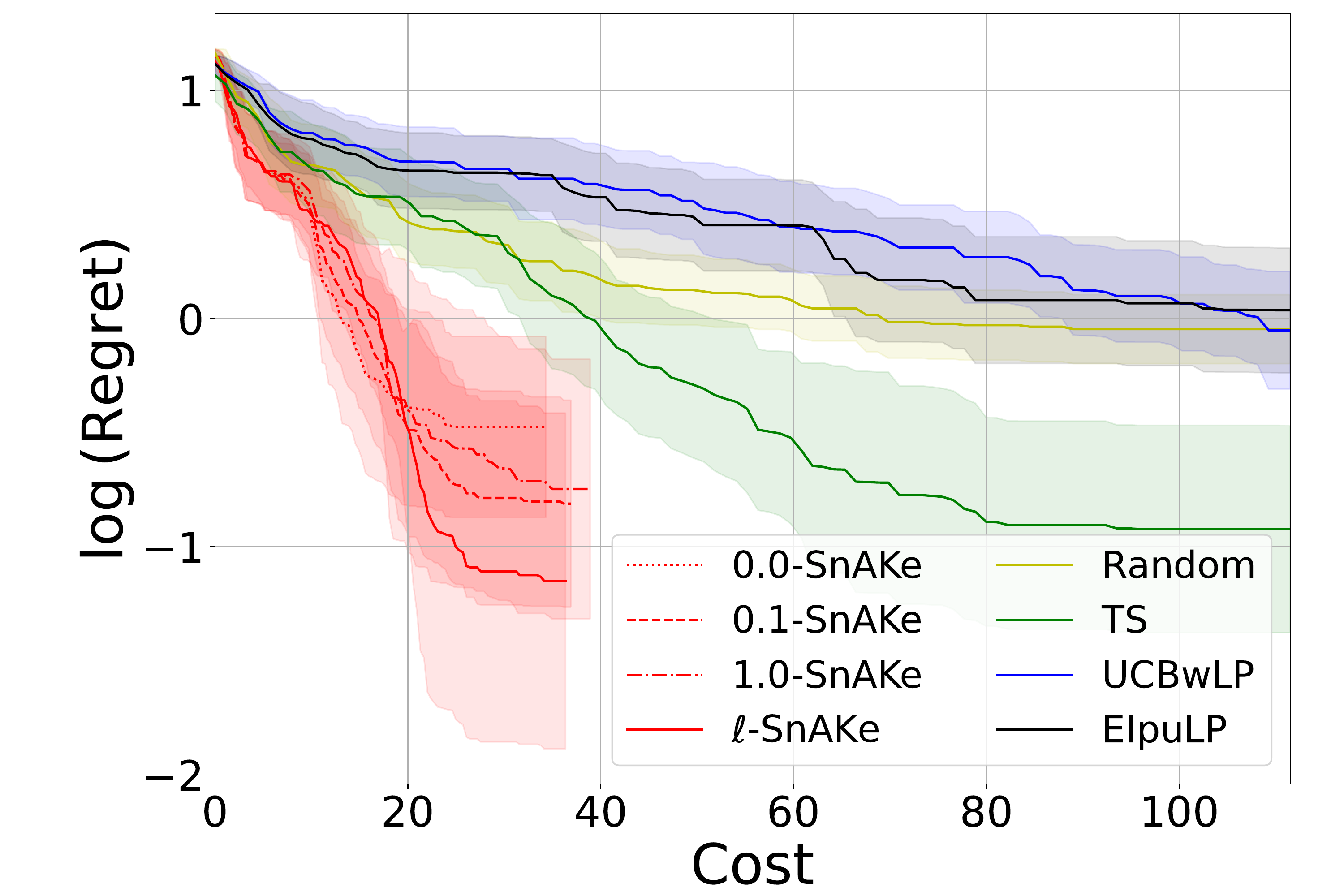}	
	\caption{(async)Hartmann6D}
	\end{subfigure}
	\hfill
	\begin{subfigure}{0.32\textwidth}
	\includegraphics[width = \textwidth]{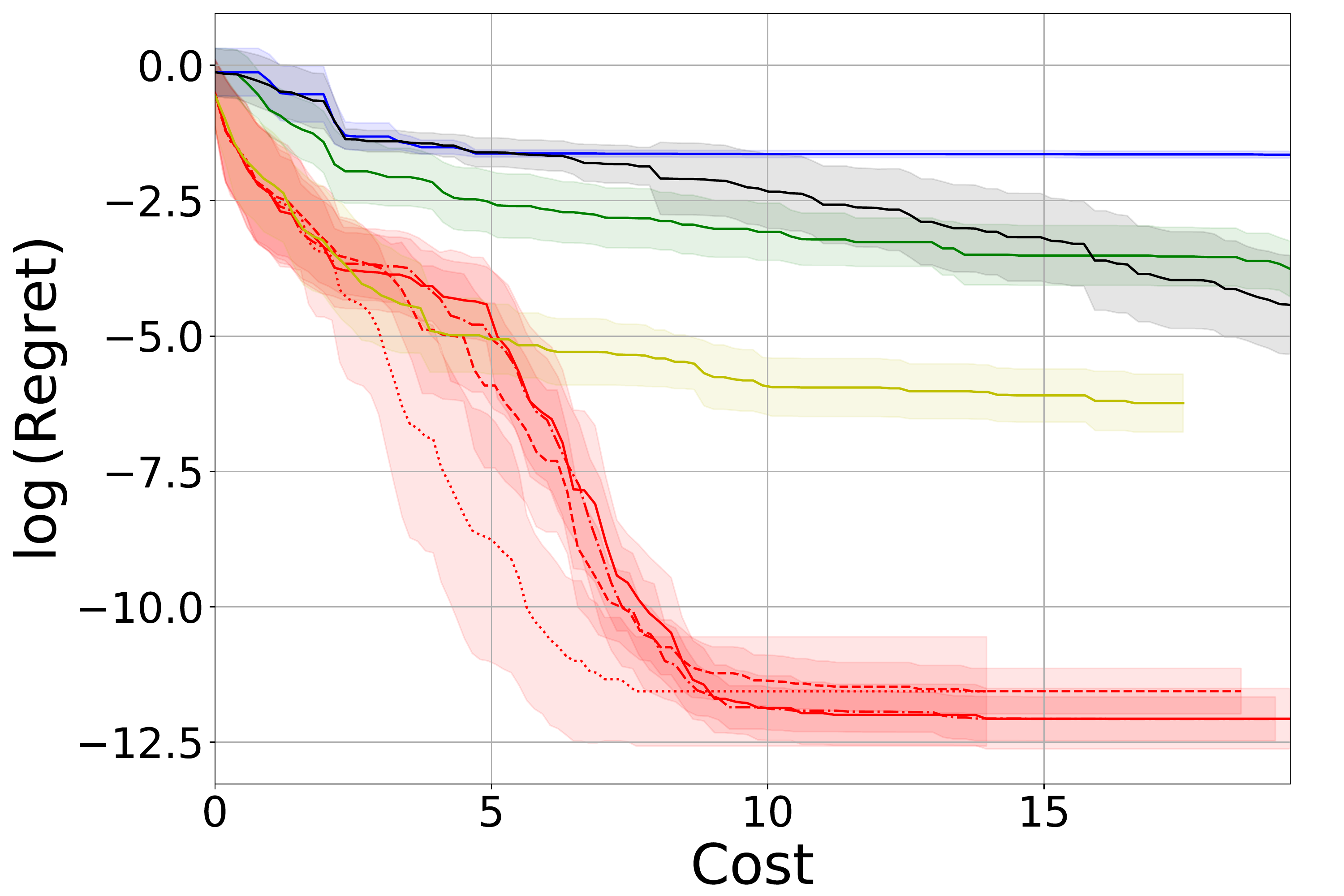}
	\caption{(async)Branin2D}
	\end{subfigure}
	\hfill
	\begin{subfigure}{0.32\textwidth}
	\includegraphics[width = \textwidth]{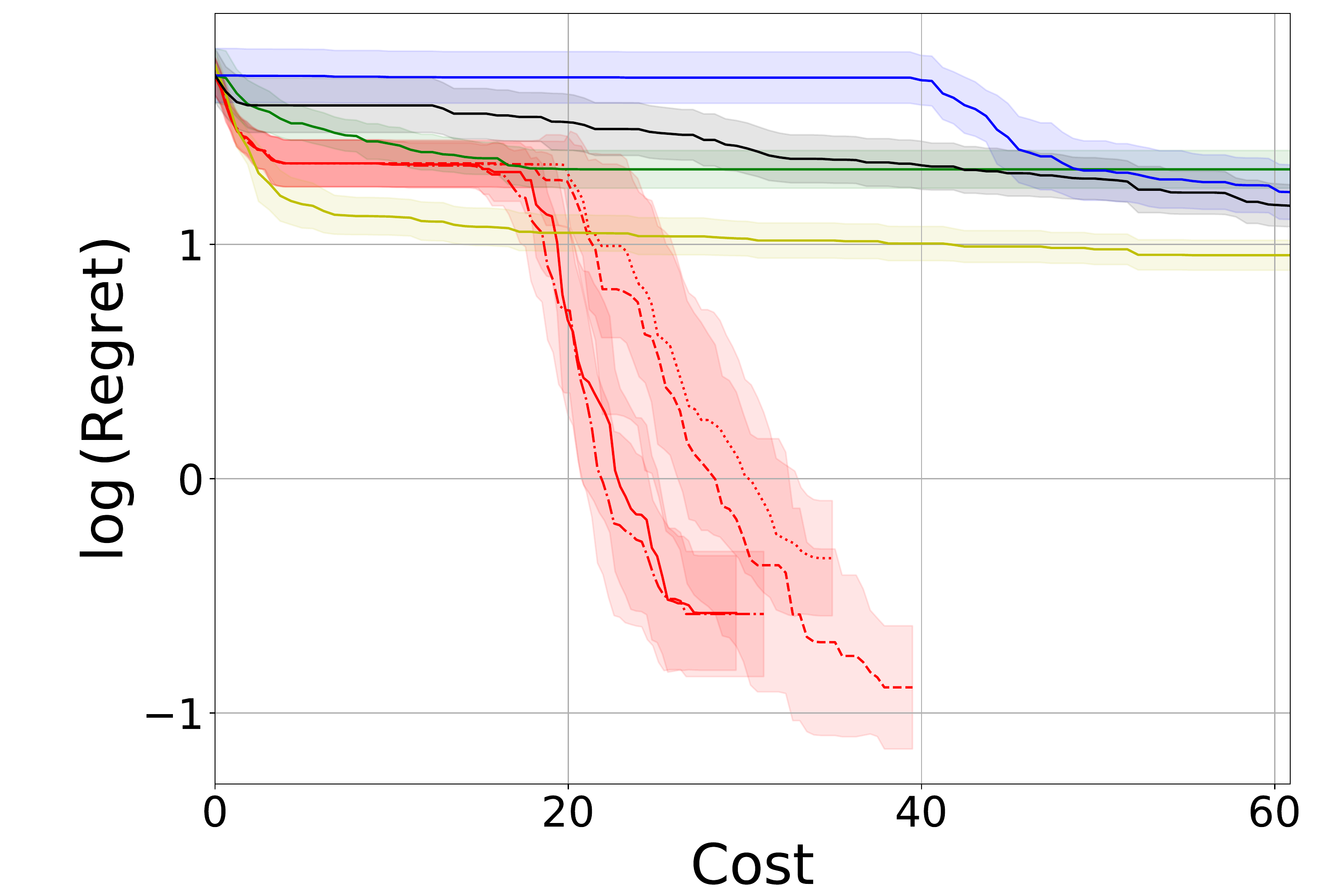}
	\caption{(async)Ackley4D}
	\end{subfigure}
	\hfill
	\begin{subfigure}{0.32\textwidth}
	\includegraphics[width = \textwidth]{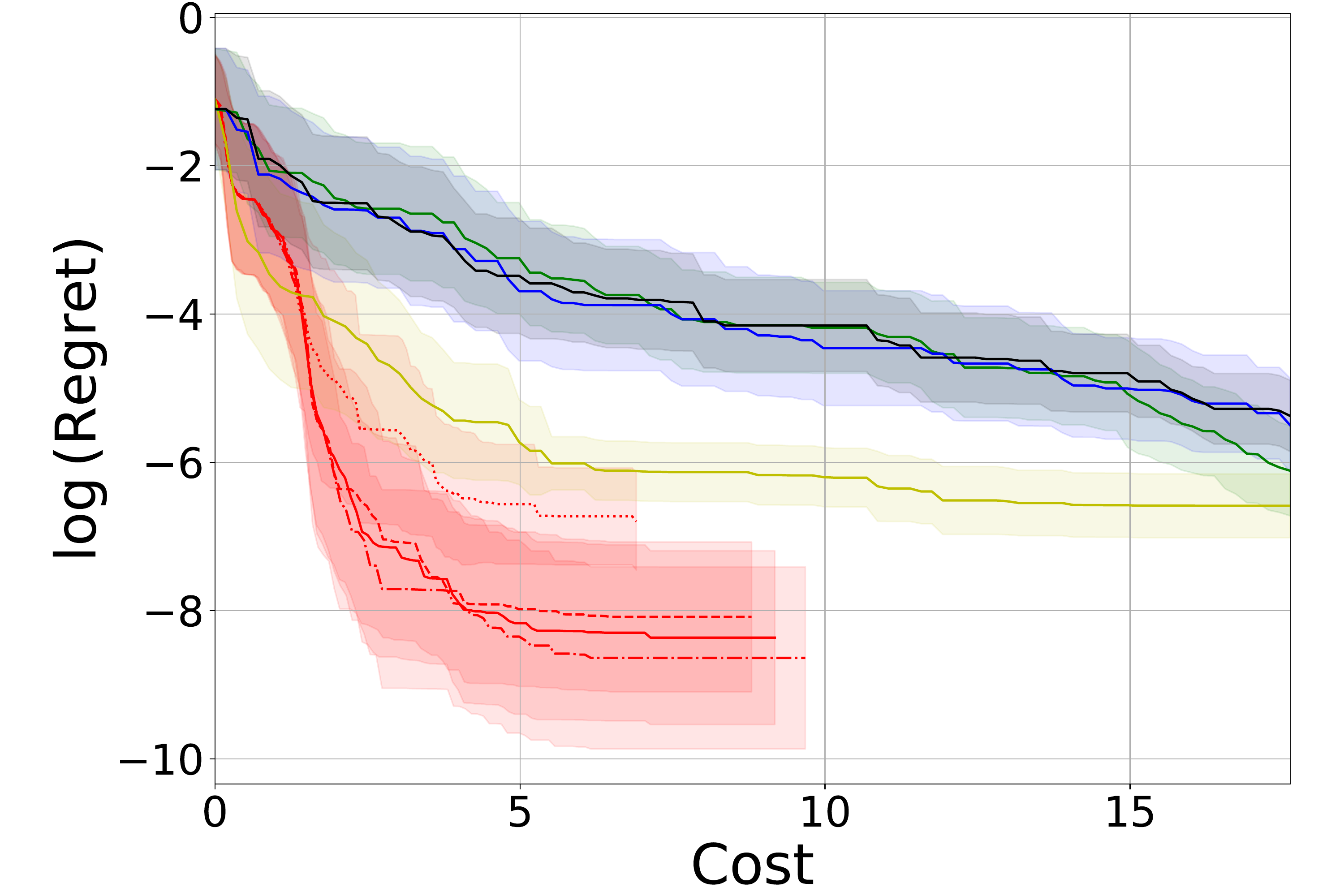}	
	\caption{(async)Michaelwicz2D}
	\end{subfigure}
	\hfill
	\begin{subfigure}{0.32\textwidth}
	\includegraphics[width = \textwidth]{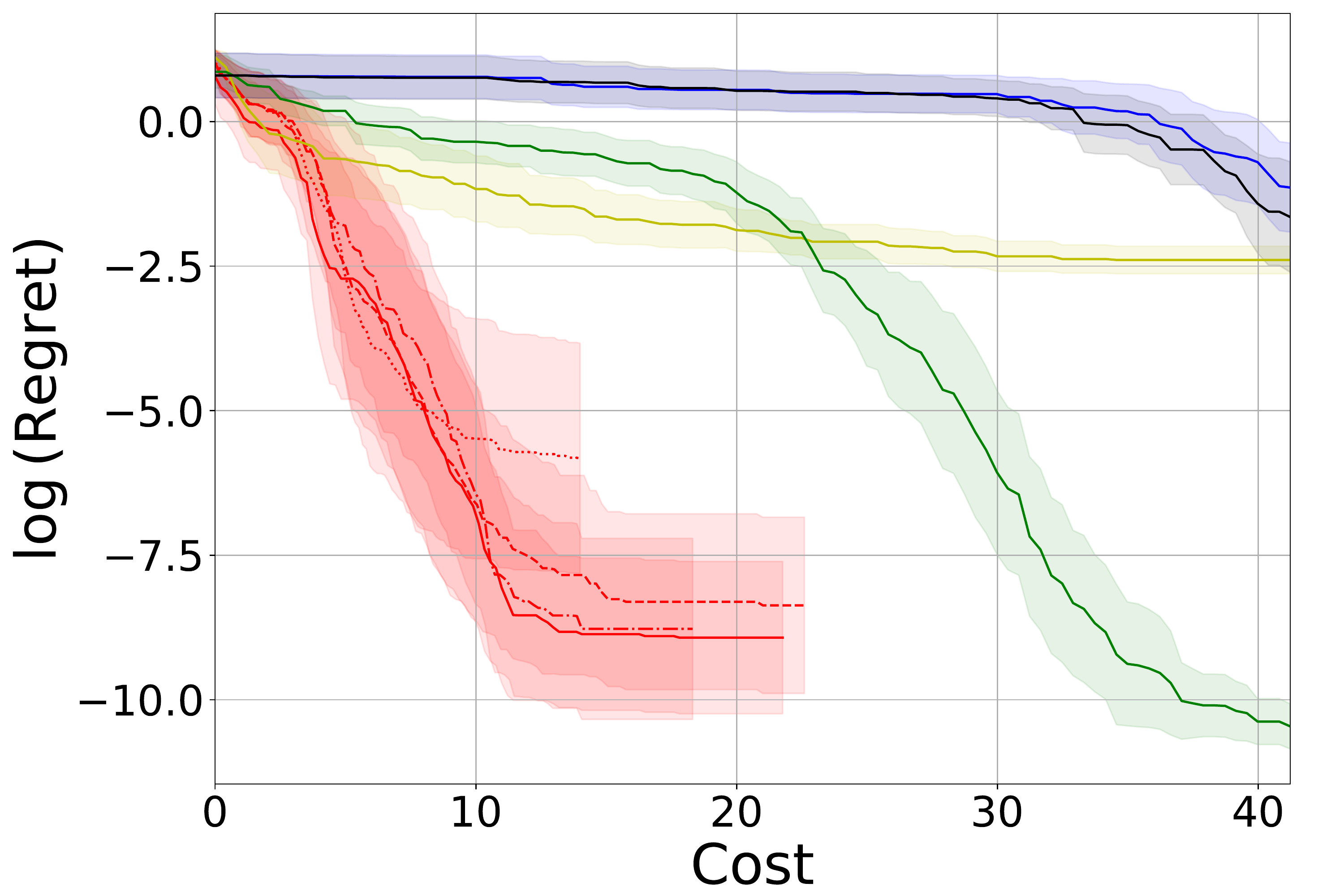}	
	\caption{(async)Hartmann3D}
	\end{subfigure}
	\hfil
	\begin{subfigure}{0.32\textwidth}
	\includegraphics[width = \textwidth]{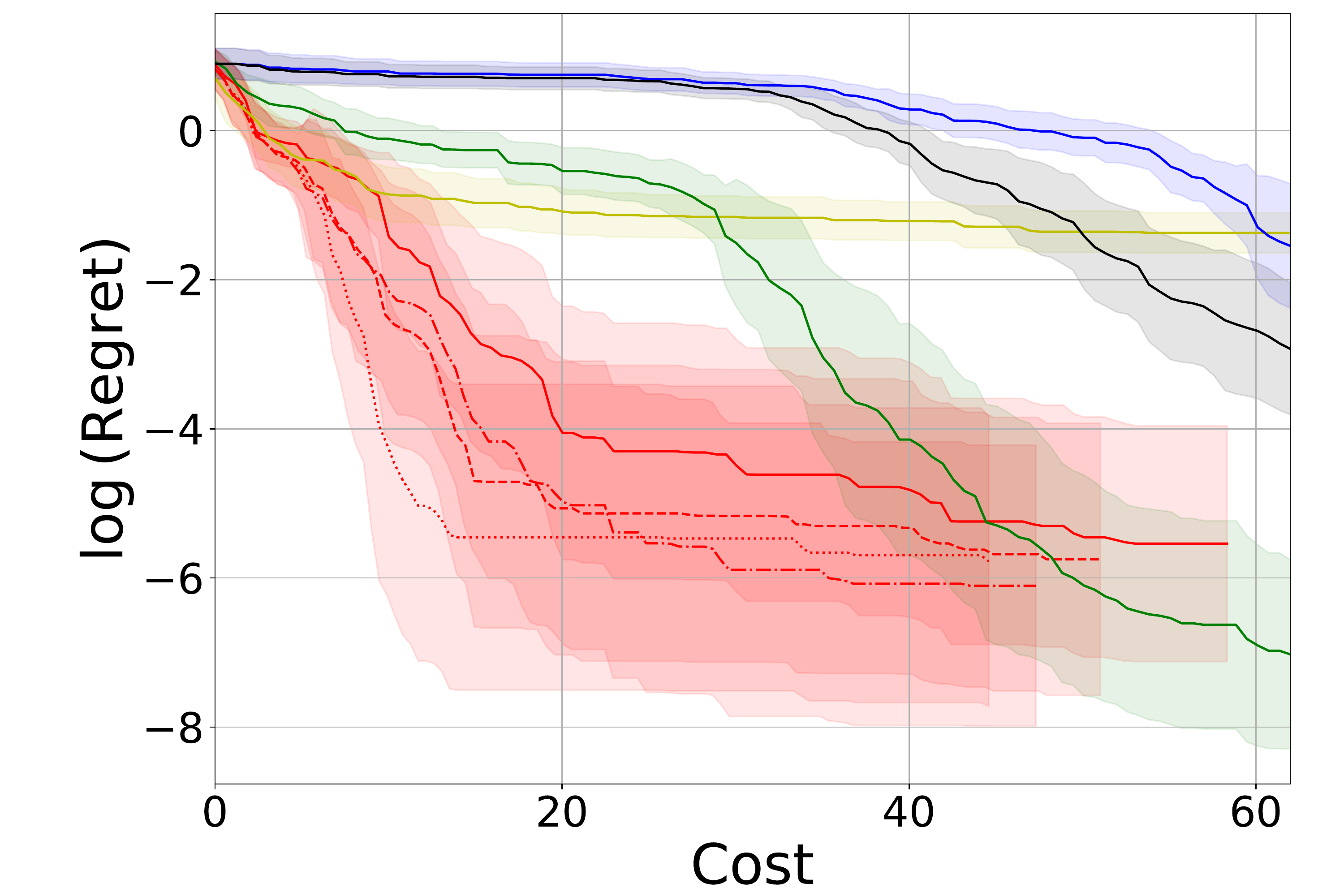}	
	\caption{(async)Hartmann4D}
	\end{subfigure}
	\caption{Synthetic experimental results. The first row represents three of the synchronous benchmarks. The last two rows correspond to asynchronous experiments with $t_{delay} = 25$. We plot the average $\log(\text{regret})$ achieved against the cost spent. For every experiment, $T = 250$, and we limit the $x$-axis to the maximum cost achieved by SnAKe or Random. $\ell$-SnAKe is the parameter-free version. SnAKe consistently achieves good regret against at cost. The bounds are $\pm$ half the standard deviation of 25 different runs.}
	\label{fig: synthetic_results}
\end{figure*}

\textbf{Sequential BO}
This section examines the performance of SnAKe against Classical BO algorithms. We compare Expected Improvement (EI) \citep{mockus2014EI}, Upper Confidence Bound (UCB) \citep{srinivas2009gaussian}, and Probability of Improvement (PI) \citep{kushner1964PI}. We compare against Expected Improvement per Unit Cost \cite{snoek2012practical} (EIpu) as introduced in Eq. (\ref{eq: EIpu}), with $\gamma = 1$, and with Truncated Expected Improvement \cite{samaniego2021bayesian} (TrEI). We also introduce a simple baseline, where we create a random Sobol sample, and then arrange an ordering by solving the TSP, and never update the path again. We do this for three classical benchmark functions, and six total in the Appendix \ref{sec: appendix_synch_experiments}. We set the cost function to be the 2-norm distance between the inputs. Figure \ref{fig: synthetic_results} shows the results in the top row.

\textbf{Asynchronous BO}
We explore the asynchronous setting, comparing Local Penalisation with UCB (UCBwLP) \citep{gonzalez2016batch, alvi2019localpen}, Thompson Sampling (TS) \citep{pmlr-v84-kandasamy18a}, and the same Random baseline from the sequential setting. We also test on EIpu with Local Penalisation (EIpuLP), with $\gamma = 1$. The results are shown in last two rows of Figure \ref{fig: synthetic_results}.

\subsection{Reaction Control on SnAr Benchmark} \label{subsec: snar}

We test our method on a real-world, SnAr chemistry benchmark \citep{hone2017kinetic}. We control four variables; equivalents of pyrrolidine, concentration of 2,4 dinitrofluorobenenze at reactor inlet, reactor temperature, and residence time. We assume changing temperature, concentration and residence time incur an input cost, owing to the response time required for the reactor to reach a new steady state. We assume the reactor as a first-order dynamic system, where the response to changes in input is given by:
\begin{equation} \label{eq: step_control_eq}
    (x_s)_i = (x_t)_i + (1 - e^{-s/\alpha_i})(\Delta x_t)_i
\end{equation}
where $s$ denotes the time after experiment $x_t$ is finished, $(x_t)_i$ denotes the $i$th variable of the $t$th experiment, $(\Delta x_t)_i = (x_{t+1})_i - (x_t)_i $, and $\alpha_i$ is the system time constant. We define quasi-steady state using an absolute residual $\beta_i$, i.e., $|(x_{s})_i - (x_{t+1})_i| \leq \beta_i$. For input changes smaller than $\beta_i$, we assume a linear cost, defined by a parameter $\gamma_i$. Combining with the response time from (\ref{eq: step_control_eq}):
\begin{equation}
\mathcal{C}_i(x_t, x_{t+1}) = \gamma_i \min\{\beta_i, |\Delta^{(i)}x_t | \} + \max\left\{0, \alpha_i \log\left(|\Delta^{(i)} x_t| / \beta_i \right)\right\}
\end{equation}
We assume that we can change variables simultaneously, so the total input cost is simply the longest response time within a given set of input changes. That is, $C(x_t, x_{t+1}) = \max_{i \in I_c}C_i(x_t, x_{t+1})$, where $I_c$ is the index set of the control variables. We implement the simulation using the Summit package \citep{felton2021summit}. More details can be found in Appendix \ref{sec: appendix_implement_details_snar}. We set a delay of $t_{delay} = 25$, and optimize for $T = 100$ iterations. The results of the experiment can be seen in Figure \ref{fig: snar_experiment}.

\subsection{Finding Contamination Sources in Ypacarai Lake} \label{sec: exp_ypacarai}

\begin{figure}[ht]
    \centering
	\includegraphics[width = 0.85 \columnwidth]{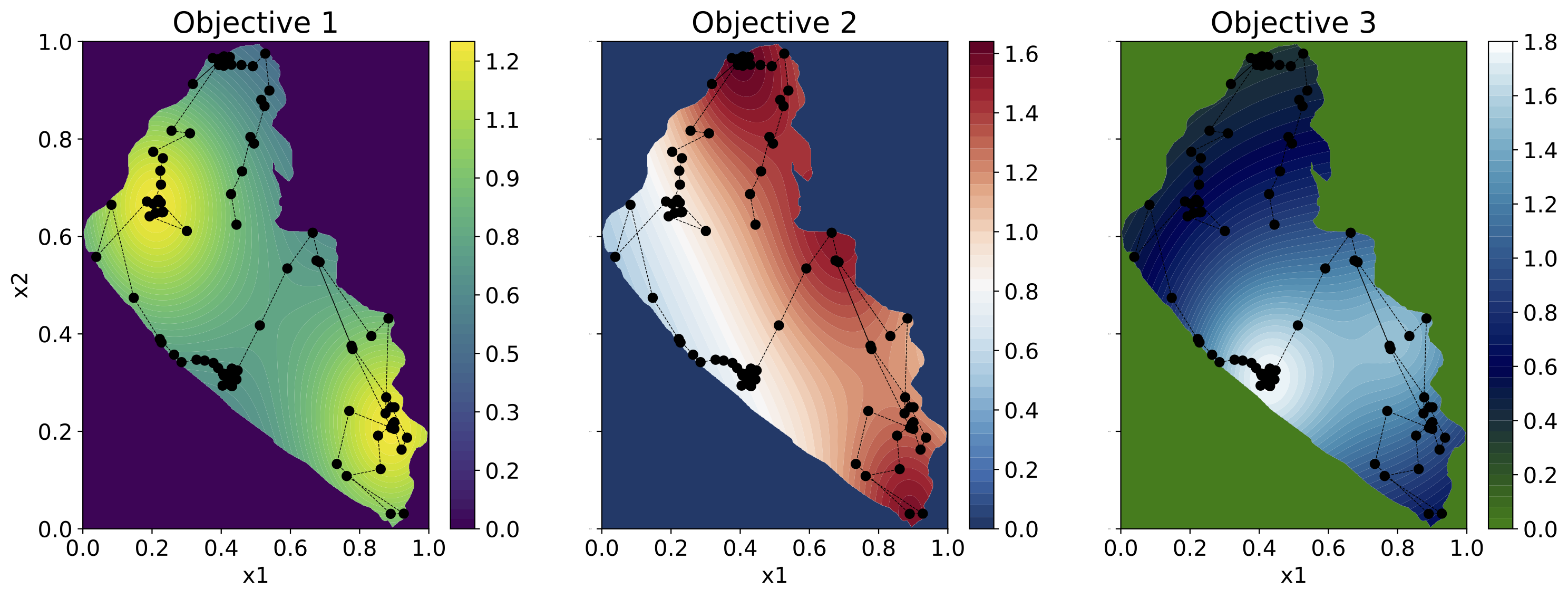}
	\caption{Visualization of Ypacarai Lake, and the objectives we optimize over. We show an example of a SnAKe optimization path on the simultaneous optimization problem (see Appendix \ref{sec: appendix_ypacarai}).}
	\label{fig: ypacarai_example_run}
\end{figure}

As a second real-world example that exhibits costly input changes, we investigate an alternative to the case study introduced by \citet{samaniego2021bayesian}. Autonomous boats are used to monitor water quality in Ypacarai Lake in Paraguay. We consider three different objectives, illustrated in Figure \ref{fig: ypacarai_example_run}, created using the Schekel benchmark function (as in \cite{samaniego2021bayesian}). Each objective could correspond to a measure of water quality, such as pH, turbidity, CO2 levels, or many more. We consider the problem of finding the \textit{largest} source of contamination in the lake by maximizing these objectives. The cost between queries relates to the boat physically moving from one place to another. We assume that the boats can travel 10 units, which corresponds to approximately 100km  (fuel limitations), and no observation time delay. We run optimization runs on each objective and shows the results in Table \ref{tab: ypacarai_results}.

\begin{table}
    \caption{Ypacarai results. We present the average regret from 10 runs $\pm$ the standard deviation, multiplied by $10^3$. O1, O2, O3 correspond to the different objectives. All runs are terminated after the cost exceeds 10 units (approximately 100 km). SnAKe is the best performer in 2 out of 3 objectives.}
	\begin{center}
	\resizebox{0.59 \columnwidth}{!}{
    \begin{tabular}[width = \columnsize]{| c | c | c | c |} \hline
    	& Regret O1 & Regret O2 & Regret O3 \\ \hline
    	TrEI & $0.475 \pm 0.596$ & $0.118 \pm 0.104$ & $0.076 \pm 0.050$\\ \hline
    	EIpu & $0.249 \pm 0.225$ & $\mathbf{0.068 \pm 0.073}$ & $0.071 \pm 0.049$ \\ \hline
    	SnAKe & $\mathbf{0.074 \pm 0.063}$ & $0.109 \pm 0.186$ & $\mathbf{0.036 \pm 0.048}$ \\ \hline
    \end{tabular}}
    \end{center}
    \label{tab: ypacarai_results}
\end{table}

\section{Conclusion and Discussion}

This paper introduces and proposes a solution to the problem of optimizing black-box functions under costs to changes in inputs. We empirically show that the regret achieved by our method is comparable to those of classical Bayesian Optimization methods and we succeed at achieving considerably lower input costs, while being particularly well suited to asynchronous problems. Examples of further work include extending the algorithm so multiple `SnAKes' can run in parallel, or extending it to the classical multi-objective setting, e.g., using variations of Thompson Sampling \cite{bradford2018efficient}.

This setting, with input costs penalizing experimental changes, makes a major step towards automating new reaction chemistry discovery, e.g., in line with the vision of \citet{lazzari2020toward}. We substantially decrease experimental cost with respect to classical black-box optimization, e.g., as used by \citet{fath2021flowoptimization} and \citet{mcmullen2011microfluidic}. In the real-life SnAR benchmark, SnAKe spends 45-55\% of the cost while making similarly strong predictions to classical BO. The synthetic benchmarks and Ypacarai Lake example offer similar advantages.

\section*{Acknowledgments \& Disclosure of Funding}
JPF is funded by EPSRC through the Modern Statistics and Statistical Machine Learning (StatML) CDT (grant no. EP/S023151/1) and by BASF SE, Ludwigshafen am Rhein. SZ was supported by an Imperial College Hans Rausing PhD Scholarship. The research was funded by Engineering \& Physical Sciences Research Council (EPSRC) Fellowships to RM and CT (grant no. EP/P016871/1 and EP/T001577/1). CT also acknowledges support from an Imperial College Research Fellowship.

We would like to thank the reviewers and meta-reviewers throughout the whole peer review process for their time, attention, and ideas, which led to many improvements in the paper. Discussions with colleagues in the Imperial Department of Computing led to the ideas of $\epsilon$-Point Deletion and the nonparametric $\ell$-SnAKe. Linden Schrecker improved our understanding of the chemistry application.


\clearpage

\bibliography{main.bib}

\newpage

\section*{Checklist}

\begin{enumerate}

\item For all authors...
\begin{enumerate}
  \item Do the main claims made in the abstract and introduction accurately reflect the paper's contributions and scope?
    \answerYes{We believe the claims in the abstract are backed by the paper's content}
  \item Did you describe the limitations of your work?
    \answerYes{See the last paragraph of Section \ref{sec: experiments}'s introduction.}
  \item Did you discuss any potential negative societal impacts of your work?
    \answerNo{We do not believe there are any obvious negative societal impacts.}
  \item Have you read the ethics review guidelines and ensured that your paper conforms to them?
    \answerYes{}
\end{enumerate}

\item If you are including theoretical results...
\begin{enumerate}
  \item Did you state the full set of assumptions of all theoretical results?
    \answerNA{See below.}
        \item Did you include complete proofs of all theoretical results?
    \answerNo{While we include some mathematical analysis in Sections \ref{sec: importance_point_deletion} and \ref{sec: theory_epsilon_point_deletion}, it is not fully rigorous and it is only meant to illustrate the importance of the Point Deletion heuristic.}
\end{enumerate}

\item If you ran experiments...
\begin{enumerate}
  \item Did you include the code, data, and instructions needed to reproduce the main experimental results (either in the supplemental material or as a URL)?
    \answerYes{The code is included in the supplementary material, and will be made public after peer-review. We also give details about how the random seeds for each run were chosen, in the appendix.}
  \item Did you specify all the training details (e.g., data splits, hyperparameters, how they were chosen)?
    \answerYes{See Appendix \ref{sec: appendix_implement_details}.}
        \item Did you report error bars (e.g., with respect to the random seed after running experiments multiple times)?
    \answerYes{See Section \ref{sec: experiments}.}
        \item Did you include the total amount of compute and the type of resources used (e.g., type of GPUs, internal cluster, or cloud provider)?
    \answerYes{See Section \ref{sec:computational_considerations}}
\end{enumerate}

\item If you are using existing assets (e.g., code, data, models) or curating/releasing new assets...
\begin{enumerate}
  \item If your work uses existing assets, did you cite the creators?
    \answerYes{See Appendix \ref{sec: appendix_implement_details}}.
  \item Did you mention the license of the assets?
    \answerNo{The licenses can be found in the documentation of each asset.}
  \item Did you include any new assets either in the supplemental material or as a URL?
    \answerNo{But the GitHub repository to generate the paper's results can be found in the following link: \url{https://github.com/cog-imperial/SnAKe}}
  \item Did you discuss whether and how consent was obtained from people whose data you're using/curating?
    \answerNA{}
  \item Did you discuss whether the data you are using/curating contains personally identifiable information or offensive content?
    \answerNA{}
\end{enumerate}

\item If you used crowdsourcing or conducted research with human subjects...
\begin{enumerate}
  \item Did you include the full text of instructions given to participants and screenshots, if applicable?
    \answerNA{}
  \item Did you describe any potential participant risks, with links to Institutional Review Board (IRB) approvals, if applicable?
    \answerNA{}
  \item Did you include the estimated hourly wage paid to participants and the total amount spent on participant compensation?
    \answerNA{}
\end{enumerate}

\end{enumerate}



\newpage
\appendix
\onecolumn

\setcounter{page}{1}

\section{Motivating Example}

\begin{figure}[ht]
	\centering
	\includegraphics[width = 0.9\textwidth]{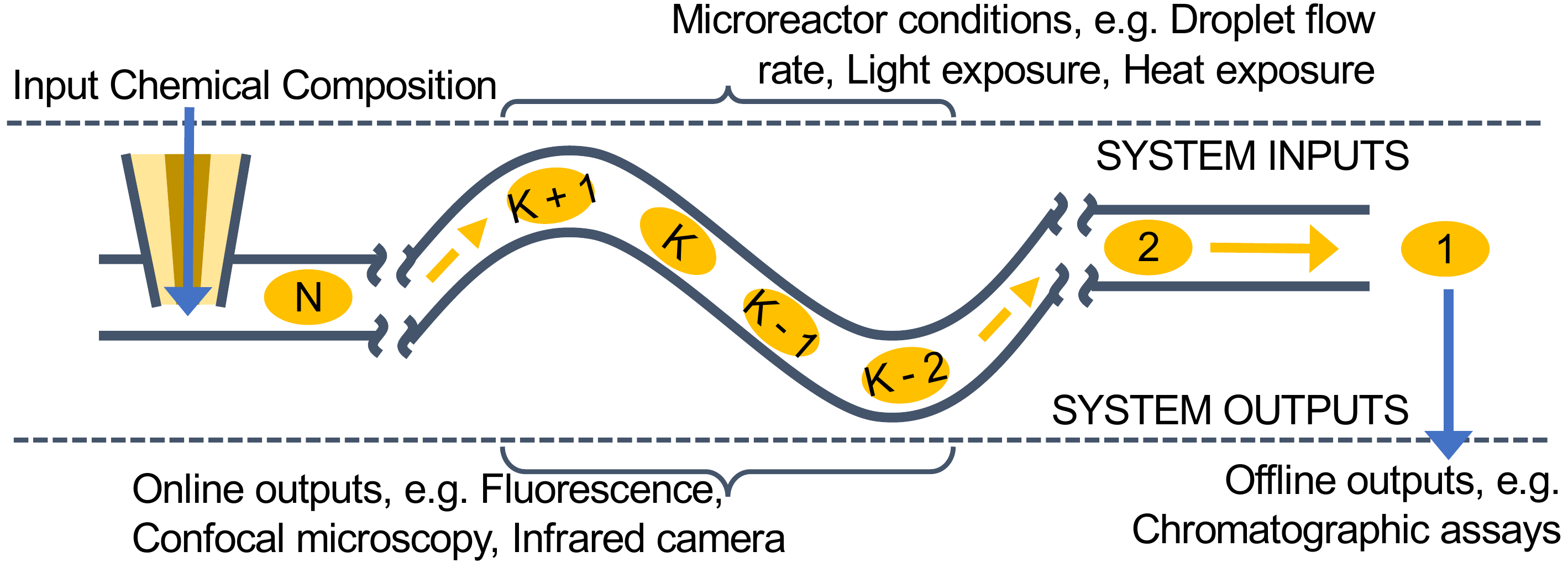}
	\caption{Motivating example. Droplets flow into the micro-reactor where we control conditions such as temperature and flow rate. The cost to change function inputs arises from how adjacent droplets are coupled, e.g. rapidly changing the temperature after droplet $K$ means waiting for system equilibration before taking new measurements. Asynchronicity arises from choosing drops $2, 3, ..., N$ before getting the results of droplet 1.}
	\label{fig: reactor_fig}
\end{figure}

Figure \ref{fig: reactor_fig} shows a graphic of a droplet microfluidic reactor \citep{teh2008droplet} which is our main motivating example.

\section{General Approach}

While we present SnAKe as a detailed algorithm, the ideas behind it are far more general. For example, in Section \ref{sec: appendix_ypacarai} we create our batch by taking samples from three different GPs. Algorithm \ref{algo: general_algo} presents a more general algorithm that could be used for any similar task.

\begin{algorithm}
\caption{General Ordering-Based Optimization} \label{algo: general_algo}
\begin{algorithmic}
 \STATE \textbf{input}: Optimization budget of $T$ samples. Method for creating batch of queries. Method for creating an ordering from a batch of queries. Method for updating paths.
 \STATE \textbf{begin}: Create initial batch of size $T$ and ordering, $S$ 
 \FOR{$t = 1, 2, 3, ..., T$}
 \IF{there is new information}
 	\STATE Update surrogate model
 	\STATE	Choose a batch of new points to query
 	\STATE	Create a new path, $\Tilde{S}$
 	\STATE $S \leftarrow \ \Tilde{S}$
 \ENDIF
 \STATE Choose next query point from ordering: $x_t \leftarrow S_t$
 \STATE Evaluate $f(x_t)$
\ENDFOR
\end{algorithmic}
\end{algorithm}

\section{Ypacarai Lake Experiment} \label{sec: appendix_ypacarai}

A second motivating example for our research comes from optimization requiring spatially continuous exploration. In this section, we examine the problem more closely, and show that SnAKe can easily be extended to optimize multiple black-box functions \textit{simultaneously} in the same search space.

\begin{figure}[ht]
    \centering
	\includegraphics[width = \columnwidth]{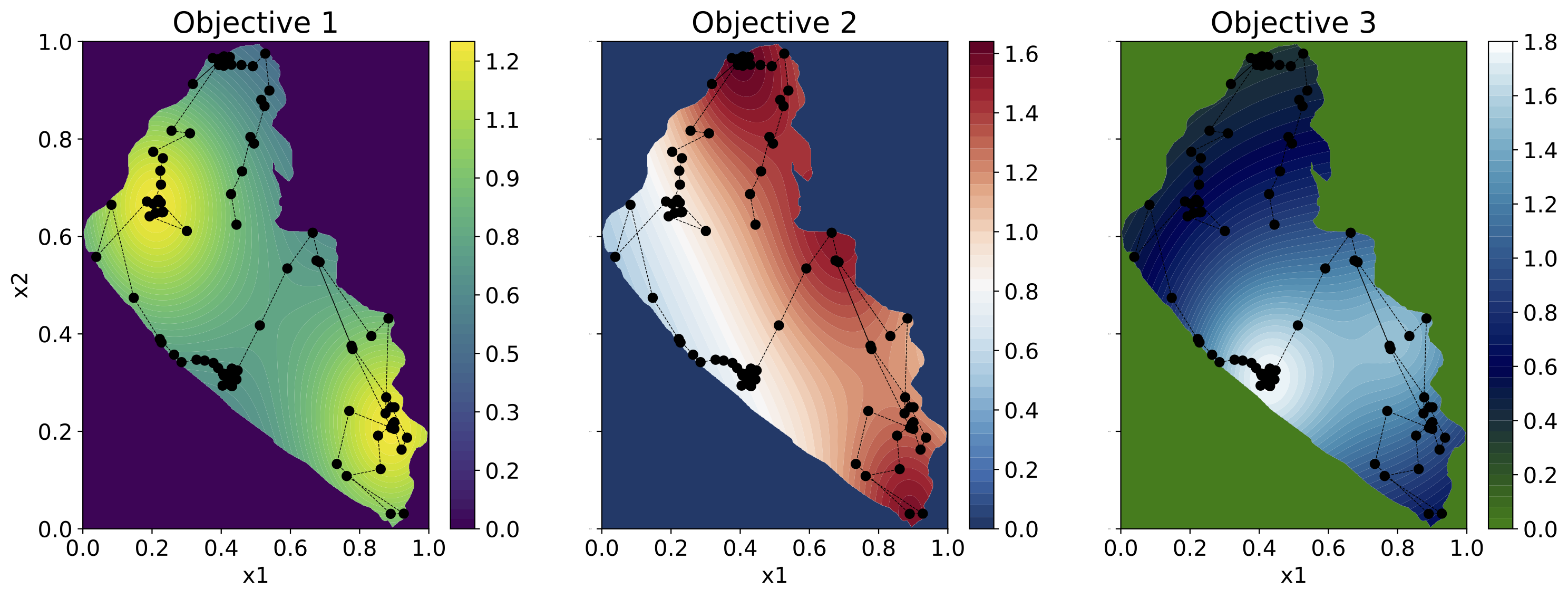}
	\caption{Visualization of Ypacarai Lake, and the corresponding objectives we optimize over. We show an example of an optimization path.}
	\label{fig: ypacarai_example_run_appendix}
\end{figure}

The problem in question is an adaptation of the case study introduced by \citet{samaniego2021bayesian}. The application is to use autonomous boats to find the \textit{largest} source of contamination in Lake Ypacarai in Paraguay. Figure \ref{fig: ypacarai_example_run_appendix} shows a visualization of the Lake, and three objectives we want to optimize over. Each objective corresponds to a different measure of water contamination, examples might include pH, turbidity, CO2 levels, and more. In Section \ref{sec: exp_ypacarai} we explored optimizing over these examples individually. 

In this section, we extend the problem to optimize over all of these \textit{simultaneously}. This is important because it is inefficient to run multiple objective runs when we are exploring the same search space. At every iteration we choose one query $x_t$, and obtain three outputs $f_1(x_t)$, $f_2(x_t)$, and $f_3(x_t)$ one for each objective. 

We model each objective using an independent GP, meaning we can \textit{create Thompson Samples from any objective}. Therefore we extend SnAKe by creating the batch using a mixture of samples from each objective. For the experiment we use the ratio 1:1:1. For TrEI and EIpu, we adapt to the multi-objective case by optimizing the first objective until we have travelled 3.3 units (approximately 33km), after which we change to the subsequent objective. Table \ref{tab: ypacarai_results_appendix} shows the results.

\begin{table}
    \caption{Ypacarai experimental results. We present the average regret from 10 runs $\pm$ the standard deviation multiplied by $10^3$. MO1, MO2, MO3 represent the different maximums (one for each function). All runs are terminated after the cost exceeds 10 units (approximately 100 km). SnAKe is the only method with good regret in all metrics, as TrEI performs poorly in the first objective, and EIpu performs poorly in the second.}
	\begin{center}
	\resizebox{0.6 \columnwidth}{!}{
    \begin{tabular}[width = \columnsize]{| c | c | c | c |} \hline
    	 & Regret MO1 & Regret MO2 & Regret MO3 \\ \hline
    	TrEI & $13.636 \pm 35.842$ & $0.802 \pm 0.828 $ & $0.162 \pm 0.108 $\\ \hline
    	EIpu & $0.995 \pm 0.661$ & $19.308 \pm 55.901$ & $0.212 \pm 0.137$ \\ \hline
    	\textbf{SnAKe} & $0.996 \pm 2.743$ & $1.740 \pm 4.813$ & $0.064 \pm 0.053$ \\ \hline
    \end{tabular}}
    \end{center}
    \label{tab: ypacarai_results_appendix}
\end{table}

\section{Empirical Analysis of Escape Probability}

\subsection{Areas with stationary points} \label{sec: appendix_empirical_stationary}

\begin{figure}[ht]
	\centering
    \includegraphics[width = \textwidth]{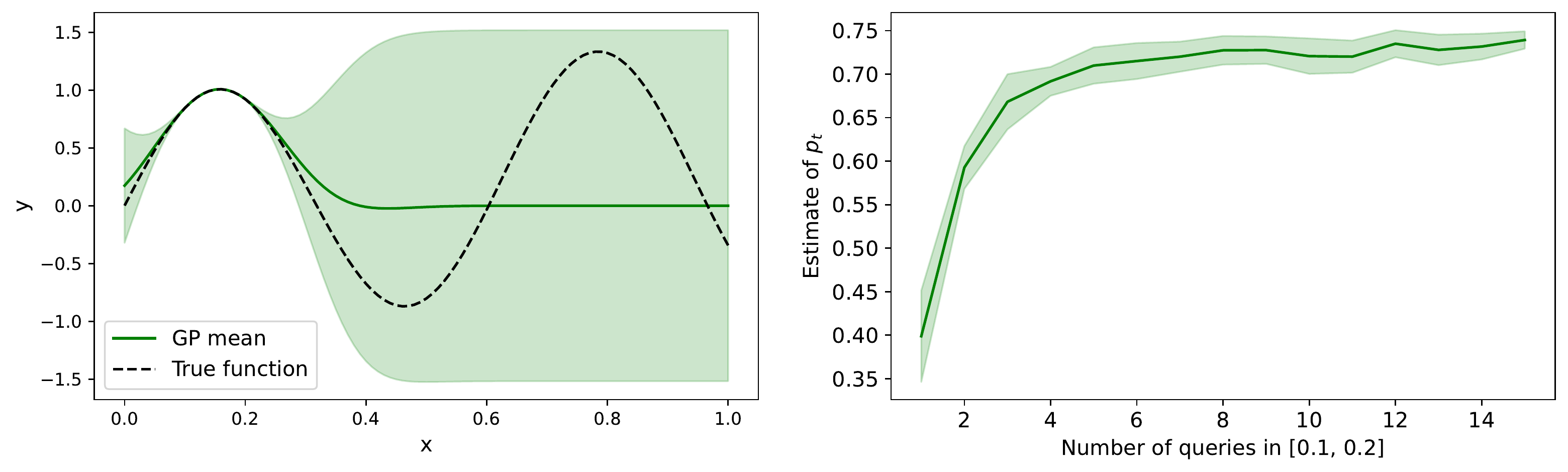}
    \caption{We estimate the probability of non-escape by taking 5000 independent Thompson Samples and counting the number of samples inside $A$ (i.e. the MLE estimator of the Bernoulli distribution). We do this for increasing number of training points in $A$ (which are chosen randomly with a uniform distribution in $A$). We repeat the experiment 10 times. The left plot shows the underlying function and the Gaussian Process for 15 training points. The right plot shows the evolution of our estimate as we increase training points inside $A = [0.1, 0.2]$ (we plot the mean of each run $\pm$ the standard deviation). This example makes it clear that $p_t$ does not converge to zero. Furthermore, it seems to converge to just over 0.7 which a very large probability. This will make fully escaping the local minimum very difficult without Point Deletion.}
    \label{fig: prob_of_escape_with_optimum}
\end{figure}

Figure \ref{fig: prob_of_escape_with_optimum} estimates the non-escape probability (see Definition \ref{def: non-escape probability}) from the interval $A = [0.1, 0.2]$. The optimization objective is a bi-modal function. Once we have 15 samples in the interval $[0.1, 0.2]$, we estimate the escape probability to converge to $p \approx 0.74$.

\subsection{Areas without stationary points} \label{sec: appendix_empirical_no_stationary}

We now repeat the same experiment as in section \ref{sec: appendix_empirical_stationary}, this time we change the interval to $A = [0.0, 0.1]$ which does not contain any stationary points. One can observe a clear difference in the behavior of $p_t$ as we include more information. This time, $p_t \rightarrow 0$ very fast. 

\begin{figure}[ht]
	\centering
    \includegraphics[width = \textwidth]{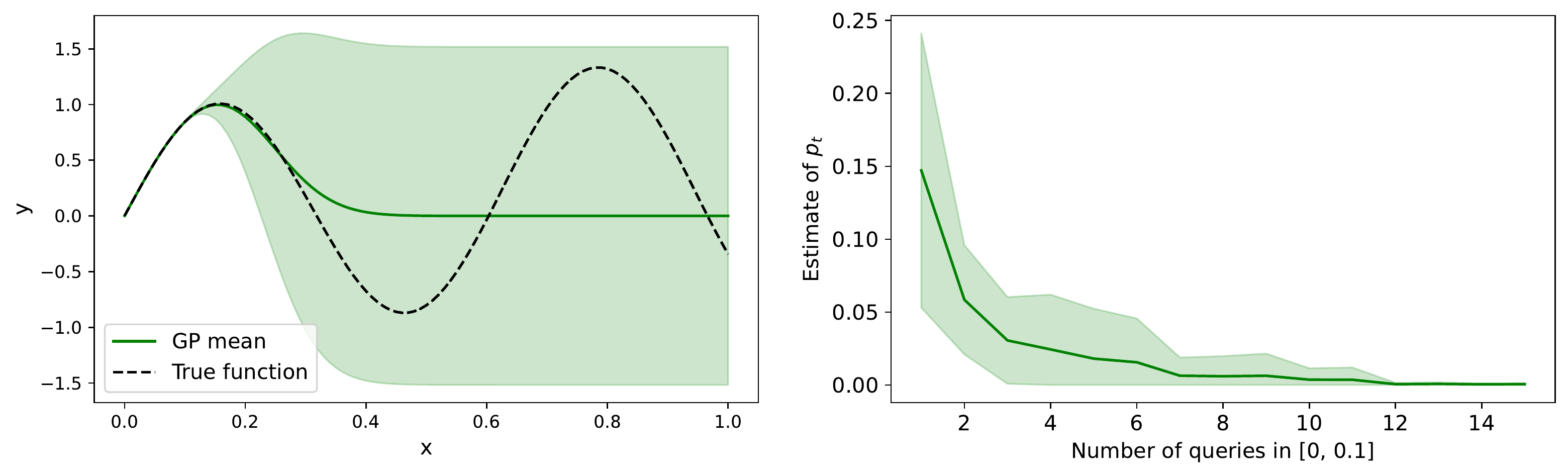}
    \caption{We estimate the probability of non-escape by taking 5000 independent Thompson Samples and counting the number of samples inside $A$ (i.e. the MLE estimator of the Bernoulli distribution). We do this for increasing number of training points in $A$ (which are chosen randomly with a uniform distribution in $A$). We repeat the experiment 10 times. The left plot shows the underlying function and the Gaussian Process for 15 training points. The right plot shows the evolution of our estimate as we increase training points inside $A = [0, 0.1]$ (we plot the mean of each run $\pm$ the standard deviation).  We can see that $p_t$ quickly converges to (almost) zero. We almost guaranteed to fully escape the area after 15 time-steps, even for very large budgets.}
    \label{fig: prob_of_escape_without_optimum}
\end{figure}

\subsection{Resampling vs Point Deletion} \label{sec: appendix_resampling_vs_deletion}

\begin{figure}[ht]
	\centering
	\begin{subfigure}[t]{0.475\textwidth}
	\includegraphics[width = \textwidth]{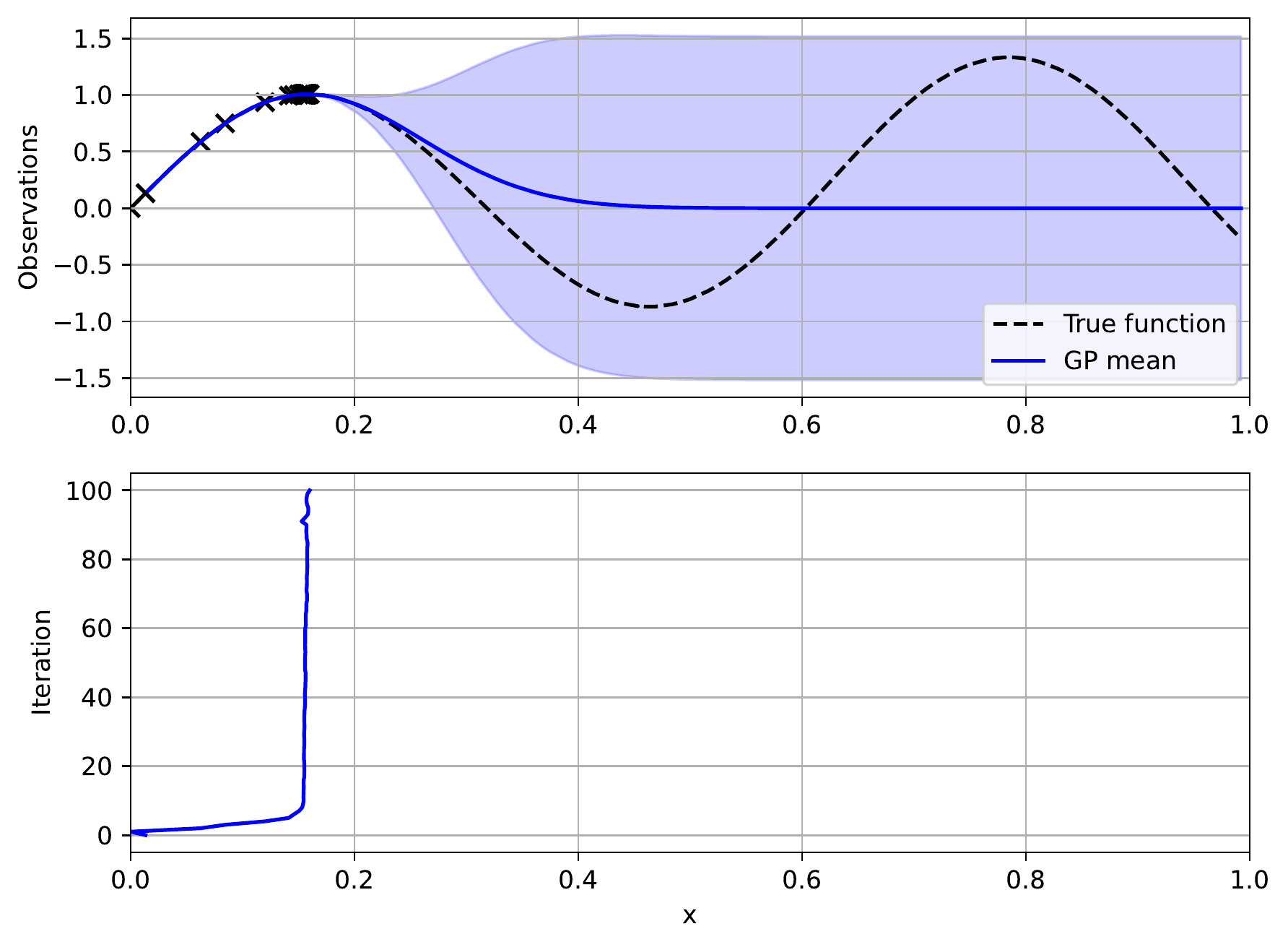}	
	\caption{Optimization with Resampling. $T = 100$.}
	\end{subfigure}
	\hfill
	\begin{subfigure}[t]{0.475\textwidth}
	\includegraphics[width = \textwidth]{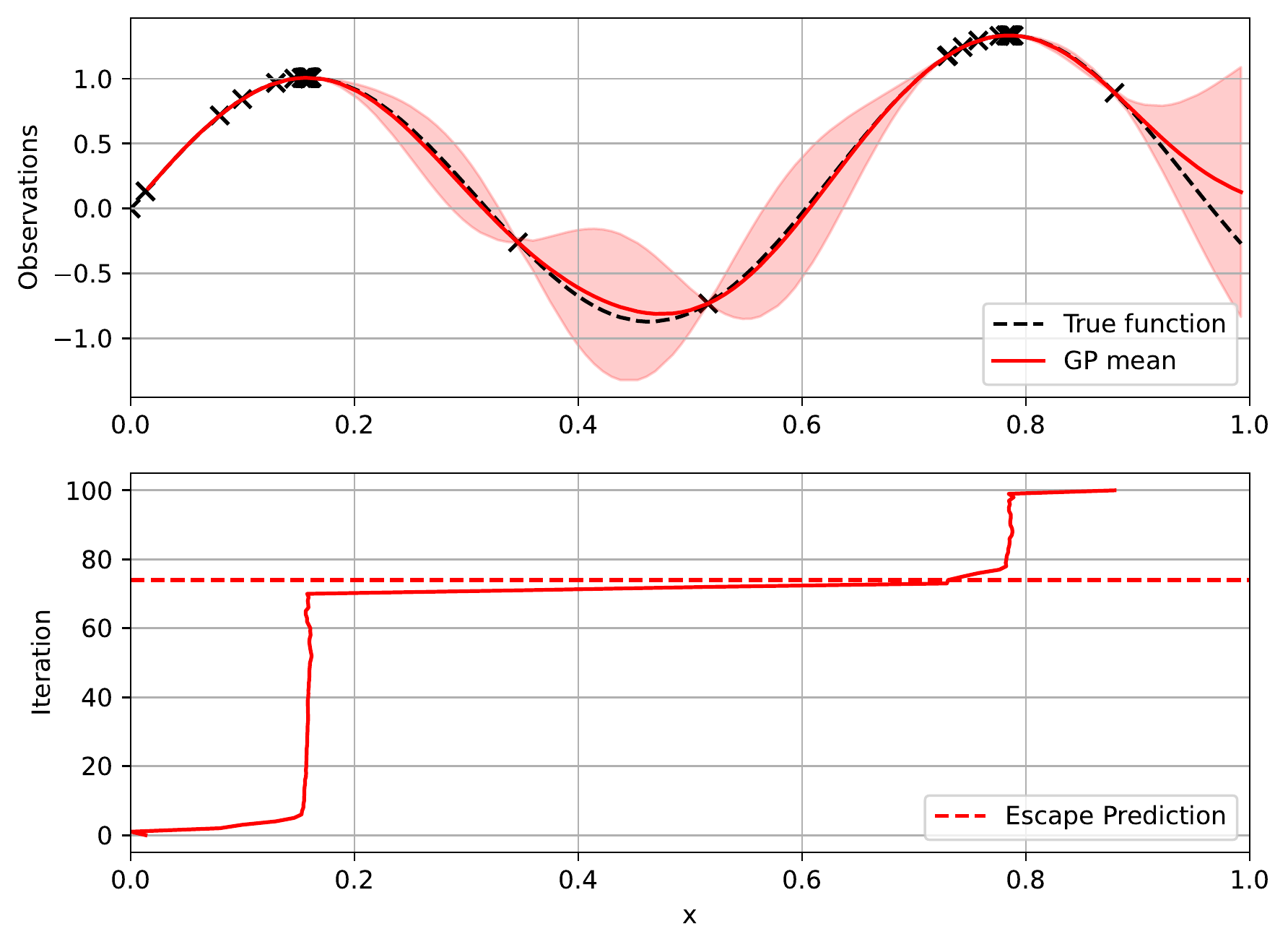}	
	\caption{Optimization with 0.1-Point Deletion. $T = 100$.}
	\end{subfigure}
	\hfill
	\begin{subfigure}[t]{0.475\textwidth}
	\includegraphics[width = \textwidth]{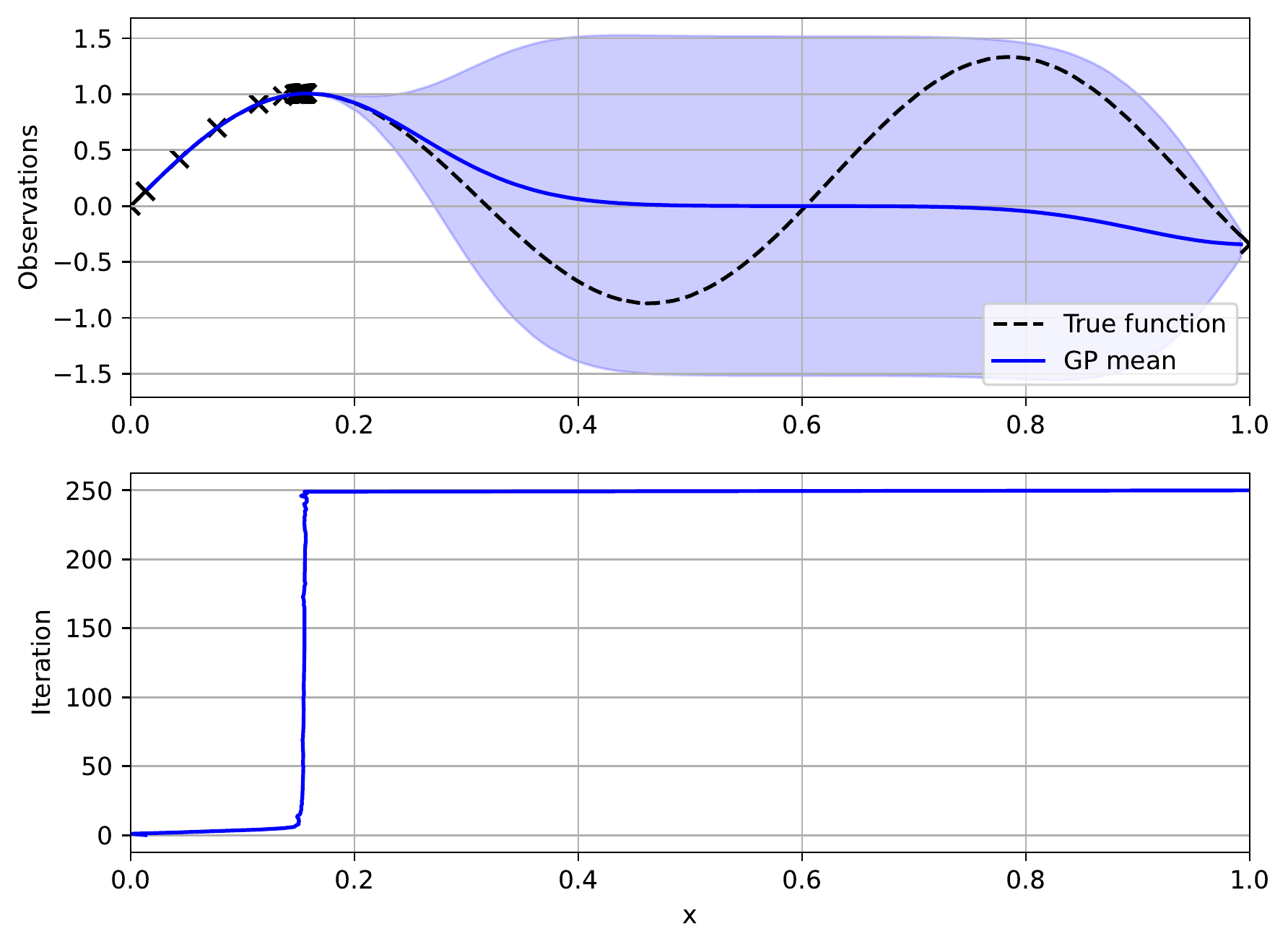}	
	\caption{Optimization with Resampling. $T = 250$.}
	\end{subfigure}
	\hfill
	\begin{subfigure}[t]{0.475\textwidth}
	\includegraphics[width = \textwidth]{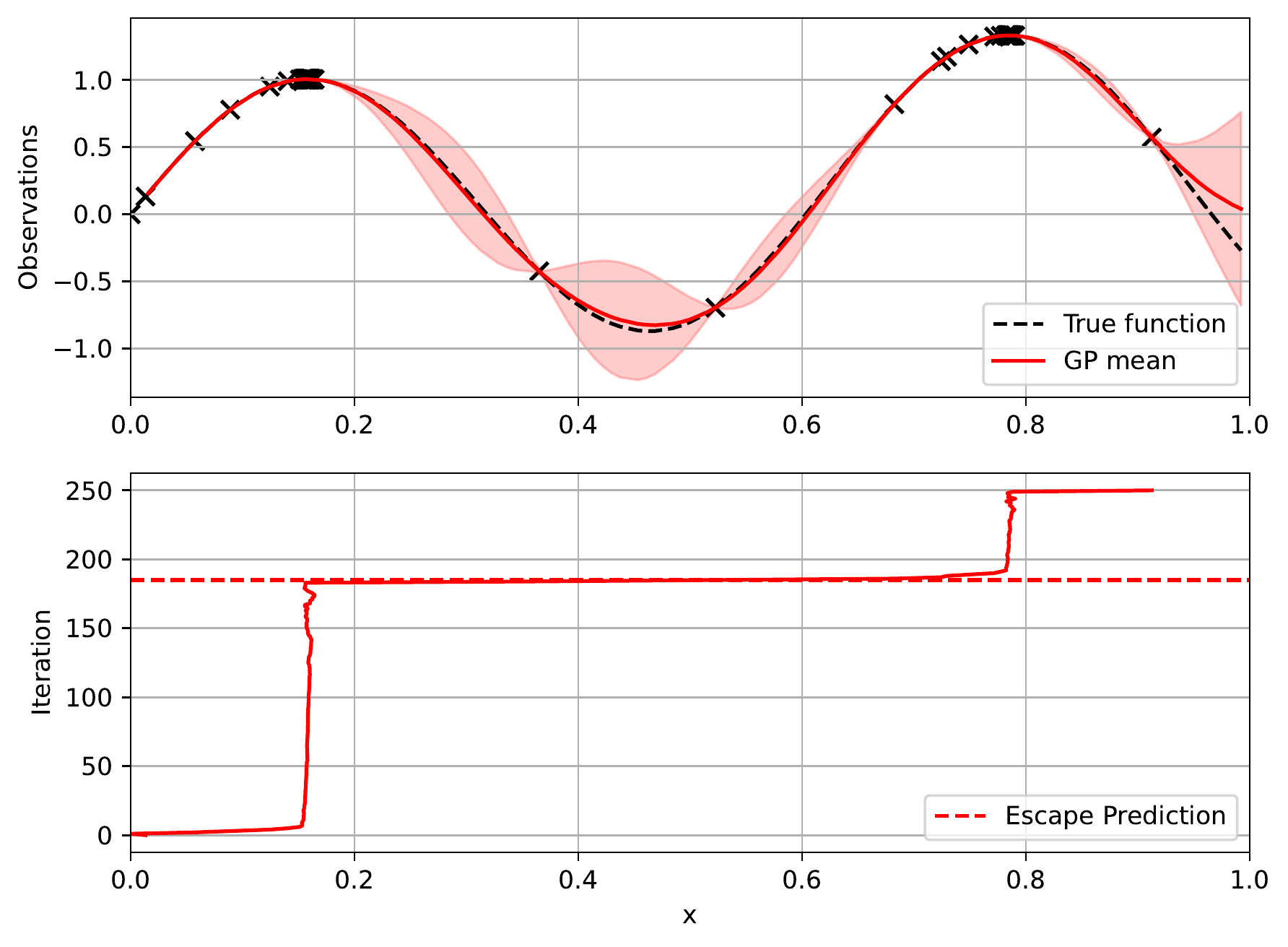}	
	\caption{Optimization with 0.1-Point Deletion. $T = 250$.}
	\end{subfigure}
	\caption{We investigate the effect of Point Deletion, and empirically confirm our analysis from Section \ref{sec: importance_point_deletion}. For Point Deletion, we calculate the escape prediction as $p T$, using $\hat{p} \approx 0.74$, which we estimated in Section \ref{sec: appendix_empirical_stationary}. We can see that without Point Deletion, the escape happens until \textit{until the very last} iteration, independently of the budget (see Remark \ref{remark: escape_prob} for an explanation). With Point Deletion, we can see that our escape predictions are accurate, and the exploration of the actual optimum increases with the budget.}
	\label{fig: PDvsRS}
\end{figure}

Figure \ref{fig: PDvsRS} empirically confirms the analysis of Section \ref{sec: theory_epsilon_point_deletion} showing the effectiveness of $\epsilon$-Point Deletion, and displaying the effect of increasing the budget from 100 to 250 iterations.

\section{Implementation Details} \label{sec: appendix_implement_details}
This section outlines the implementation details and hyper-parameter choices for all the methods compared in the paper. The code used in the paper is available at the following link: \url{https://github.com/cog-imperial/SnAKe}.

\subsection{Computational Considerations (Extended)} \label{sec: appendix_computational}

Unfortunately, Algorithm \ref{algo: EaS} (SnAKe) is computationally expensive in two aspects. First, for large budgets, we may struggle to train and sample the GPs. For our experiments, training was not an issue and we were able to use full model GPs. However, we could use Sparse GPs \citep{snelson2006sparsegps} if needed.
The \citet{wilson2020efficiently} approach  allows us to create GP samples efficiently, and possibly optimize them using gradient methods. The sampling can be done in linear time (\textit{after} the GP has been trained). We use the \citet{wilson2020efficiently} method to create our samples, and then optimize the samples using Adam \citep{kingma2014adam}.

The second bottleneck is solving the TSP, which is NP-hard, and we may need to solve it almost at every iteration (for small values of $t_{delay}$). There are heuristic solutions that give approximate solutions quickly. We use Simulated Annealing \citep{kirkpatrick1983simmulated} which grows linearly with the budget size, $T$. For small budgets, simulated annealing should find good solutions, but it could struggle as the budget size increases. To solve this, note that we do not actually require a super-specific solution to the problem: we are only expecting to query the first few points on a path before replacing it by an entirely new path.

We build an adaptive grid (at each iteration) consisting of two separate parts. A very coarse grid, $\xi_{global}$, covers most of the search space, and a very fine grid, $\xi_{local, t}$, consisting of the $N_l$ samples closest to $x_t$. This allows us to define the grid $\xi_t = \xi_{global} \cup \xi_{local, t}$. The remaining $T - N_l$ samples will be assigned to the closest point in $\xi_t$ (using the Euclidean distance).

The adaptive grid means we expect to have multiple samples assigned to the same point, specifically in the coarse areas of our grid. But this is not important, because we expect our immediate attention will be in the area around our current input where there should be little to no repetition. This will allow the algorithm to focus on testing solutions which are relevant to our problem.

The adaptive grid introduces two hyper-parameters: the size of the global and local grids, respectively $N_g$ and $N_l$. Using this method, we run the TSP heuristics on a graph with at most $\min(N_g + N_l, T)$ nodes. For the experiments, we create the global grid using a simple Sobol grid \citep{sobol1967distribution}.

\subsection{Gaussian Processes}

For every GP, we use the RBF Kernel with an output-scale, $\theta_0$:
\begin{equation*}
k_{RBF}(x_1, x_2) = \theta_0\exp\left({-\frac{1}{2}(x_1 - x_2)^T \Theta^{-2}(x_1 - x_2)}\right)
\end{equation*}
where $\Theta = \text{diag}(\ell_1, ..., \ell_d)$ and $\ell_i$ denotes the length-scale of the $i$th variable. For the prior mean, we used a constant function with trainable value, $\mu_0$. We implemented them all using the package GPyTorch \citep{gardner2018gpytorch}.

\subsection{Training the hyper-parameters of the Gaussian Processes}

Our method is well suited for physical systems. Hence, we assume that there is good prior knowledge of the hyper-parameters. In particular, we found it reasonable that each hyper-parameter would be given a lower and upper bound. Normally, we would simply have a large initialization sample. However, we believe this goes against the nature of the problem because we want to explore the space slowly to avoid large input costs. So any type of initialization would be costly.

We simulate this in the following way: we first randomly sample $\max(T / 5, 10d)$ points from the $d$-dimensional input space, and train a GP on these data-points. The hyper-parameters of the GP are optimized by maximizing the marginal log-likelihood \citep{rasmussen2005gps} over 500 epochs using Adam \citep{kingma2014adam} with a learning rate of 0.01. The resulting hyper-parameters will correspond to the `educated guess'. We then set the following bounds:
\begin{itemize}
\item[a)] For the length-scale, the lower bound is half the educated guess, and the upper bound double the `educated guess'.
\item[b)] For the output-scale, $\theta_0$, the lower bound is half the educated guess, and the upper bound double the `educated guess'.
\item[c)] For the initial mean, $\mu_0$, the lower bound is the educated guess minus a third of the initial variance, the upper bound is the educated guess plus a third of the initial variance.
\item[d)] The noise parameter we simply set to be greater than $10^{-5}$.
\end{itemize}

We (partially) enforce the constraints by setting a SmoothedBoxPrior on each parameter, with a variance of 0.001. Finally, under the constraints defined above, we re-estimate the hyper-parameters every time we obtain 25 new observations.

To make sure all models receive fair initializations, we set the same seed for each run and function pair. 

\subsection{SnAKe} \label{sec: appendix_implement_eas}

We used Simulated Annealing \citep{kirkpatrick1983simmulated} to solve the Travelling Salesman Problem. We implemented it using the NetworkX package \citep{hagberg2008networkx}. We initalized the cycle with the `greedy' sub-algorithm and used all default options.

We generated the Thompson Samples using the method introduced in \citet{wilson2020efficiently} which we implemented ourselves. To optimize the samples we used Adam \citep{kingma2014adam} and PyTorch \citep{paszke2019pytorch} over $10d$ epochs, with a learning rate of 0.01. We used $10d$ multi-starts for each sample. To create the samples, we used $\ell = 1024$ Fourier bases.

For $\ell$-SnAKe, we define an adaptive deletion constant, $\epsilon_t = \min(\ell_{1, t}, ..., \ell_{d, t})$, where $\ell_{i, t}$ denotes the length scale of the $i$th variable at time $t$ (recall we are re-training the hyper-parameters every new 25 observations, so the length scales change with time). 

For the adaptive grid, we use $N_l = 25$ local samples, and a corse global Sobol grid \citep{sobol1967distribution} of $N_g = 100$ points.

\subsection{Classical Bayesian Optimization}
We used BoTorch \citep{balandat2020botorch} to implement all methods in this section. We optimized the acquisition functions across 150 epochs using Adam \citep{kingma2014adam} with a learning rate of 0.0001 using 7500 random multi-starts.

\subsubsection{Expected Improvement}
Expected Improvement \citep{mockus2014EI} optimizes the acquisition function:
\begin{equation*}
EI(x) = \mathbb{E}\left[\max(y - y_{best}, 0) \right], \quad y \sim f(x)
\end{equation*}
where $y_{best}$ is our best observation so far.

Expected Improvement per Unit Cost \cite{snoek2012practical} optimizes the acquisition function:
\begin{equation*}
    EI(x, x_{t-1}) = \frac{EI(x))}{C(x, x_{t-1}) + \gamma}
\end{equation*}
We set $\gamma = 1$ in all experiments to address the fact that not moving incurs zero cost. Note that as $\gamma \rightarrow \infty$, the method will behave closer to normal Expected Improvement, and as $\gamma \rightarrow 0$ the method will tend to stay closer and closer to the current input.

Truncated Expected Improvement \cite{samaniego2021bayesian} first maximizes the normal Expected Improvement acquisition function. It then travels a distance of at most $\ell$ towards the proposed point, where $\ell$ is the GPs length-scale. We exclude this method from synchronous SnAr, as it is not suited for general cost functions.

\subsubsection{Upper Confidence Bound}
Upper Confidence Bound \citep{srinivas2009gaussian} optimizes the acquisition function:
\begin{equation*}
UCB(x) = \mu_t(x) + \beta_t \sigma_t	(x)
\end{equation*}
We set $\beta_t = 0.2 d \log(2t)$ following \citet{kandasamy2019multi}.

\subsubsection{Probability of Improvement}
Probability of Improvement \citep{kushner1964PI} optimizes the acquisition function:
\begin{equation*}
PI(x) = \mathbb{P}(y \geq y_{best})	, \quad y \sim f(x)
\end{equation*}

\subsubsection{Truncated Expected Improvement}
Truncated Expected Improvement was developed by \cite{samaniego2021bayesian} trying to solve the task of using an automated boat to monitor the water quality of Ypacari Lake in Paraguay. The method seeks to take distance travelled into account when doing Bayesian Optimization. It does by selecting a point using Expected Improvement, setting a direct path from our current input to the new point, and then \textit{truncating} the path one length-scale away. That is:
\begin{align*}
    p_t &= \argmax_{x \in \mathcal{X}}{EI(x \ | \ D_t)} \\
    x_{t+1} &= x_{t} + \frac{p_t - x_t}{|| p_t - x_t ||} \min(\ell, ||p_t - x_t ||) 
\end{align*}

\subsection{Asynchronous Bayesian Optimization}

\subsubsection{Local Penalization}

We use the penalization method as is described in \citet{gonzalez2016batch}. For the Lipschitz constant, we estimate it by calculating the gradient of $\mu_t$ (using auto-differentiation) in a Sobol grid \citep{sobol1967distribution} of 50$d$ points and selecting $L$ to be the maximum gradient in the grid.

\textbf{UCB with Local Penalization} We set $\beta_t = 0.2 d \log(2t)$ and $M = y_{best}$.

\textbf{EIpu with Local Penalization} We set $\gamma = 1$.

\subsubsection{Thompson Sampling}
We use the sample procedure as in \ref{sec: appendix_implement_eas}, except we only optimize a single sample at every iteration.

\subsection{Description of Benchmark Functions}

We chose the benchmark functions to observe the behavior of SnAKe in a variety of scenarios. More details of all the benchmark functions can be found in \citet{simulationlib}.

\subsubsection{Branin2D}

The two-dimensional Branin function is given by. The function has three global maximums:
\begin{equation*}
	f(x) = a(x_2 - b x_1^2 + c x_1 - r)^2 + s (1 - t) \cos(x_1) + s
\end{equation*}
where we optimize over $\mathcal{X} = [-5, 10] \times [0, 15]$. We set $a = -1$, $ = 5.1 / (4 \pi^2)$, $c = 5 / \pi $, $r = 6$, $s = -10$, and $t = 1 / (8 \pi)$.

\subsubsection{Ackley4D}
The four-dimensional Ackley function has a lot of local optimums, with the optima in the center of the search space. The function is given by:
\begin{equation*}
	f(x) = a \exp\left(-b \sqrt{\frac{1}{4}\sum_{i=1}^4 x_i^2}\right) + \exp \left(\frac{1}{d} \sum_{i=1}^4 \cos(c x_i) \right) - a - \exp(1)
\end{equation*}
where we slightly shift the search space and optimize over $\mathcal{X} = [-1.8, 2.2]^4$. This is to avoid having the optimum exactly at a point in the Sobol grid and giving SnAKe an unfair advantage. We set $a = 20$, $ = 0.2$, and $c = 2 \pi$.
\subsubsection{Michaelwicz2D}

The two-dimensional Michalewicz function is characterized by multiple local maxima and a lot of flat regions. The function is given by:
\begin{equation*}
	f(x) = \sum_{i = 1}^2 \sin(x_i) \sin^{2m} \left(\frac{i x_i^2}{\pi} \right)
\end{equation*}
where we set $m = 10$ and we optimize on the region $\mathcal{X} = [0, \pi]^2$.

\subsubsection{Hartmann}
We select this function to see how the algorithms behave in \textit{similar} functions as dimension increases. We do three versions of the Hartmann function, with dimensions $d = 3, 4,$ and $6$. The equation is given by:
\begin{equation*}
	f(x) = \sum_{i = 1}^4 \alpha_i \exp\left(- \sum_{j = 1}^d A_{ij}(x_j - P_{ij}) \right)
\end{equation*}
where $\alpha = (1, 1.2, 3, 3.2)^T$. 
For $d = 3$ we use:
\begin{equation*}
	A =
	\begin{pmatrix}
	3 & 10 & 30\\
	0.1 & 10 & 35 \\
	3 & 10 & 30 \\
	0.1 & 10 & 35
	\end{pmatrix}
\qquad 
P = 10^{-4}
\begin{pmatrix}
	3689 & 1170 & 2673\\
	4699 & 4387 & 7470 \\
	1091 & 8732 & 5547 \\
	381 & 5743 & 8828
	\end{pmatrix}
\end{equation*}
for $d = 4$ and $d = 6$ we use:
\begin{equation*}
	A =
	\begin{pmatrix}
	10 & 3 & 17 & 3.5 & 1.7 & 8\\
	0.05 & 10 & 17 & 0.1 & 8 & 14\\
	3 & 3.5 & 1.7 & 10 & 17 & 8\\
	17 & 8 & 0.05 & 10 & 0.1 & 14
	\end{pmatrix}
\qquad 
P = 10^{-4}
\begin{pmatrix}
	1312 & 1696 & 5569 & 124 & 8283 & 5886\\
	2329 & 4135 & 8307 & 3736 & 1004 & 9991\\
	2348 & 1451 & 3522 & 2883 & 3047 & 6650\\
	4047 & 8828 & 8732 & 5743 & 1091 & 381
	\end{pmatrix}
\end{equation*}
They are all evaluated on the unit cube $[0, 1]^d$.

\subsubsection{Perm10D}
We select the 10-dimensional version of the Perm benchmark to test the capabilities of the algorithms in a very high-dimensional setting (by BO standards). The equation is given by:
\begin{equation*}
	f(x) = - 10^{-21} \sum_{i = 1}^{10} \left( \sum_{j = 1}^{10}(j^i + \beta)\left(\left(\frac{x_j}{j}\right)^i - 1 \right)\right)^2
\end{equation*}
where we set $\beta = 10$. We evaluate it on $\mathcal{X} = [-10, 10]^d$.

\subsubsection{SnAr4D} \label{sec: appendix_implement_details_snar}

We implement the simulation using Summit \cite{felton2021summit}. We control temperature between 40 and 120 degrees, concentration from 0.1 to 0.5 moles per liter, and residence time between 0.5 and 2 minutes. We set the cost parameters for temperature $\alpha_{temp} = 5, \beta_{temp} = 1$, $\gamma_{temp} = 1$, for concentration $\alpha_{conc} = 2, \beta_{conc} = 0.01$, $\gamma_{conc} = 1$, and for residence time $\alpha_{residence} = 3$, $\beta_{residence} = 0.05$ and $\gamma_{residence} = 1$. We further optimize over the equivalents of pyrrolidine between 1 and 5 units, but we assume changing it incurs no input costs.

Since the SnAr benchmark is a multi-objective problem, we optimize a weighted sum of the two objectives:
\begin{equation*}
	\text{SnAr}(x) = \omega_1 \times \text{yield} - \omega_2 \times \text{e-factor}
\end{equation*}
where we set $\omega_1 = 10^{-4}$ and $\omega_2 = 0.1$. We optimize over $\mathcal{X} = [40, 120] \times [0.1, 0.5] \times [0.5, 2] \times [1, 5]$.

\subsection{Ypacarai Implementation Details}

As per \cite{samaniego2021bayesian} we model the Lake using the Shekel function which is given as:
\begin{equation*}
    f(x) = \sum_{i = 1}^m \left(\sum_{j = 1}^2(10 x_j - C_{ji})^2 + \beta_i \right)^{-1}
\end{equation*}

For the objective we use $m = 2$, $m = 3$, and $m = 2$. The other parameters are:
\begin{equation*}
	C^{(1)} =
	\begin{pmatrix}
	2 & 6.7 \\
	9 & 2 \\
	\end{pmatrix}
\qquad 
    C^{(2)} =
	\begin{pmatrix}
	7 & 6 \\
	3.8 & 9.9 \\
	9 & 0.1 \\
	\end{pmatrix}
\qquad 
    C^{(3)} =
	\begin{pmatrix}
	4 & 3 \\
	8.5 & 4 \\
	\end{pmatrix}
\end{equation*}

\begin{equation*}
	\beta^{(1)} =
	\begin{pmatrix}
	9 & 9 \\
	\end{pmatrix}
\qquad 
    \beta^{(2)} =
	\begin{pmatrix}
	10 & 8 & 8\\
	\end{pmatrix}
\qquad 
    \beta^{(3)} =
	\begin{pmatrix}
	7 & 9 \\
	\end{pmatrix}
\end{equation*}

Where $C^{(i)}$ and $\beta^{(i)}$ represent the parameters of the $i$th objective. All objectives are optimized across on a subset of $\mathcal{X} = [0, 1]^2$. The subset is defined by creating grid of points mapping a high-resolution black and white image of Lake Ypacarai onto $\mathcal{X}$. For simplicity, we assume the cost of moving from one point to another is simply the distance between the points, even though in practice we might need to take a longer route to avoid land. For Truncated Expected Improvement, we simply project the truncation into the closest point grid, to avoid sampling points outside of the Lake.

\section{Full Experiment Results} \label{sec: appedix_full_results}

\subsection{Tables of Results}

\subsubsection{Synchronous Experiments} \label{sec: appendix_synch_experiments}

This section includes the full tables of results of the synthetic synchronous experiments. The results are shown in Table \ref{tab: cost_comparison_exp_sec_1_small_budgets} and \ref{tab: reg_comparison_exp_1_small_budgets}.

\begin{table}[ht]
	\caption {Comparison of 2-norm cost for different BO benchmark functions. The best three performances are shown in bold, and the best one in italics. We can see that SnAKe constantly achieves low cost, especially for larger budgets. The best cost performance is achieved by 0.0-SnAKe, however, we do this at the expense of worse regret. The only function for which SnAKe struggles is the very high dimensional Perm10D.}
	\label{tab: cost_comparison_exp_sec_1_small_budgets}
	\resizebox{\columnwidth}{!}{
		\begin{tabular}{cccccccccccc}
			\toprule   
			{\bf Method} & {\bf Budget} & {\bf 0.0-SnAKe} & {\bf 0.1-SnAKe} & {\bf 1.0-SnAKe} & {\bf $\ell$-SnAKe} & {\bf EI} & {\bf EIpu} & {\bf TrEI} & {\bf UCB} & {\bf PI} & {\bf Random}\\
			\midrule
			\multirow{4}{*}{Branin2D}
			& 15&$4.5 \pm 1.8$&$5.4 \pm 1.6$&$5.8 \pm 1.7$&$5.7 \pm 1.5$&$6.7 \pm 1.5$&$5.3 \pm 1.2$&$5.1 \pm 1.0$&$\textbf{4.4} \pm \textbf{1.6}$&$\emph{\textbf{1.0}} \pm \emph{\textbf{0.9}}$&$\textbf{4.1} \pm \textbf{0.4}$\\
			
			& 50&$\textbf{5.8} \pm \textbf{3.0}$&$9.3 \pm 3.1$&$10 \pm 4$&$9.8 \pm 3.2$&$17 \pm 6$&$\textbf{7.3} \pm \textbf{1.7}$&$13.4 \pm 2.3$&$15 \pm 7$&$\emph{\textbf{4.0}} \pm \emph{\textbf{3.2}}$&$7.5 \pm 0.4$\\
			
			& 100&$\emph{\textbf{5.4}} \pm \emph{\textbf{2.9}}$&$10 \pm 4$&$10 \pm 4$&$11 \pm 4$&$37 \pm 13$&$\textbf{9.1} \pm \textbf{1.7}$&$25 \pm 4$&$33 \pm 17$&$12 \pm 7$&$\textbf{10.2} \pm \textbf{0.5}$\\
			
			& 250&$\emph{\textbf{7.1}} \pm \emph{\textbf{2.3}}$&$15.4 \pm 3.4$&$16 \pm 4$&$\textbf{15.3} \pm \textbf{2.8}$&$112 \pm 32$&$\textbf{12.2} \pm \textbf{2.0}$&$59 \pm 13$&$\left(9 \pm 5\right) \times 10^{1}$&$32 \pm 22$&$16.5 \pm 0.7$\\
			
			\midrule
			\multirow{4}{*}{Ackley4D}
			& 15&$\textbf{14.7} \pm \textbf{1.2}$&$\textbf{14.9} \pm \textbf{1.3}$&$16.0 \pm 1.0$&$16.4 \pm 0.9$&$21.1 \pm 0.8$&$15.4 \pm 2.6$&$21.4 \pm 0.8$&$19.5 \pm 2.1$&$18.5 
			\pm 3.4$&$\emph{\textbf{8.0}} \pm \emph{\textbf{0.4}}$\\
			
			& 50&$\textbf{25} \pm \textbf{5}$&$\textbf{28} \pm \textbf{4}$&$31 \pm 6$&$30 \pm 5$&$66 \pm 6$&$60 \pm 8$&$69.4 \pm 1.9$&$60 \pm 9$&$52 \pm 9$&$\emph{\textbf{19.5}} \pm \emph{\textbf{0.6}}$\\
			
			& 100&$\emph{\textbf{23}} \pm \emph{\textbf{5}}$&$27 \pm 6$&$\textbf{24} \pm \textbf{6}$&$\textbf{23} \pm \textbf{5}$&$128 \pm 9$&$123 \pm 10$&$135.8 \pm 3.2$&$107 \pm 16$&$93 \pm 10$&$32.6 \pm 0.9$\\
			
			& 250&$\emph{\textbf{25}} \pm \emph{\textbf{8}}$&$\textbf{32} \pm \textbf{9}$&$\textbf{33} \pm \textbf{8}$&$33 \pm 6$&$302 \pm 28$&$311 \pm 17$&$330 \pm 6$&$221 \pm 30$&$210 \pm 10$&$59.9 \pm 1.1$\\
			
			\midrule
			\multirow{4}{*}{Michaelwicz2D}
			& 15&$\textbf{1.7} \pm \textbf{1.1}$&$\textbf{1.9} \pm \textbf{0.6}$&$1.9 \pm 0.8$&$2.1 \pm 1.1$&$11.8 \pm 2.6$&$\emph{\textbf{0.8}} \pm \emph{\textbf{0.4}}$&$2.3 \pm 
			0.4$&$6.7 \pm 1.6$&$4.2 \pm 3.3$&$3.9 \pm 0.4$\\
			
			& 50&$\textbf{2.1} \pm \textbf{0.8}$&$3.0 \pm 0.9$&$\textbf{2.7} \pm \textbf{0.8}$&$3.1 \pm 1.1$&$23 \pm 4$&$\emph{\textbf{1.58}} \pm \emph{\textbf{0.33}}$&$7.7 \pm 0.9$&$24 \pm 7$&$8 \pm 12$&$7.5 \pm 0.4$\\
			
			& 100&$\textbf{2.4} \pm \textbf{1.2}$&$3.9 \pm 0.8$&$\textbf{3.6} \pm \textbf{1.0}$&$3.7 \pm 1.0$&$30 \pm 11$&$\emph{\textbf{2.1}} \pm \emph{\textbf{0.5}}$&$15.6 \pm 2.0$&$36 \pm 4$&$17 \pm 23$&$10.5 \pm 0.4$\\
			
			& 250&$\emph{\textbf{2.8}} \pm \emph{\textbf{1.4}}$&$5.8 \pm 1.6$&$\textbf{4.7} \pm \textbf{1.6}$&$5.4 \pm 1.0$&$\left(6 \pm 4\right) \times 10^{1}$&$\textbf{3.1} \pm 
			\textbf{0.4}$&$40.3 \pm 2.6$&$107 \pm 22$&$\left(6 \pm 5\right) \times 10^{1}$&$16.4 \pm 0.6$\\
			
			\midrule
			\multirow{4}{*}{Hartmann3D}
			& 15&$\emph{\textbf{2.5}} \pm \emph{\textbf{1.0}}$&$3.4 \pm 1.2$&$4.2 \pm 1.3$&$3.6 \pm 1.1$&$6.7 \pm 3.0$&$\textbf{2.6} \pm \textbf{0.8}$&$\textbf{3.1} \pm \textbf{0.6}$&$5.9 \pm 2.7$&$4.4 \pm 3.1$&$6.24 \pm 0.32$\\
			
			& 50&$\emph{\textbf{4.3}} \pm \emph{\textbf{2.1}}$&$\textbf{5.7} \pm \textbf{2.2}$&$6.9 \pm 3.4$&$7.0 \pm 3.3$&$15.1 \pm 3.4$&$\textbf{4.9} \pm \textbf{2.6}$&$5.8 \pm 
			1.7$&$14 \pm 4$&$9 \pm 4$&$13.7 \pm 0.5$\\
			
			& 100&$\emph{\textbf{4.9}} \pm \emph{\textbf{2.6}}$&$9 \pm 5$&$\textbf{8} \pm \textbf{4}$&$9 \pm 5$&$46 \pm 8$&$\textbf{7.0} \pm \textbf{2.8}$&$9.4 \pm 2.8$&$26 \pm 10$&$29 \pm 12$&$21.5 \pm 0.6$\\
			
			& 250&$\emph{\textbf{4.9}} \pm \emph{\textbf{2.3}}$&$\textbf{10} \pm \textbf{4}$&$\textbf{8.5} \pm \textbf{3.5}$&$9.8 \pm 3.4$&$96 \pm 4$&$13.1 \pm 3.5$&$20 \pm 4$&$53 \pm 31$&$\left(9 \pm 4\right) \times 10^{1}$&$38.5 \pm 1.3$\\
			
			\midrule
			\multirow{4}{*}{Hartmann6D}
			& 15&$\emph{\textbf{6}} \pm \emph{\textbf{4}}$&$\textbf{6} \pm \textbf{4}$&$8 \pm 4$&$8 \pm 4$&$18 \pm 4$&$\textbf{6.9} \pm \textbf{3.1}$&$9 \pm 4$&$17 \pm 4$&$17 \pm 
			5$&$10.5 \pm 0.5$\\
			
			& 50&$\emph{\textbf{11}} \pm \emph{\textbf{5}}$&$\textbf{11} \pm \textbf{4}$&$12 \pm 6$&$\textbf{12} \pm \textbf{4}$&$61 \pm 11$&$39 \pm 9$&$32 \pm 14$&$54 \pm 9$&$49 
			\pm 14$&$29.6 \pm 0.8$\\
			
			& 100&$\emph{\textbf{11}} \pm \emph{\textbf{5}}$&$\textbf{13} \pm \textbf{6}$&$15 \pm 9$&$\textbf{12} \pm \textbf{6}$&$117 \pm 21$&$94 \pm 19$&$65 \pm 29$&$102 \pm 11$&$91 \pm 28$&$51.8 \pm 1.0$\\
			
			& 250&$\emph{\textbf{13}} \pm \emph{\textbf{6}}$&$\textbf{15} \pm \textbf{7}$&$15 \pm 8$&$\textbf{15} \pm \textbf{9}$&$\left(2.7 \pm 0.6\right) \times 10^{2}$&$\left(2.5 \pm 0.7\right) \times 10^{2}$&$\left(1.6 \pm 0.7\right) \times 10^{2}$&$224 \pm 24$&$\left(2.1 \pm 0.7\right) \times 10^{2}$&$107.6 \pm 1.3$\\
			
			\midrule
			\multirow{4}{*}{Perm10D}
			& 15&$22.2 \pm 1.0$&$22.2 \pm 1.1$&$22.3 \pm 1.1$&$22.2 \pm 1.2$&$23.7 \pm 2.8$&$\textbf{6.3} \pm \textbf{0.5}$&$\emph{\textbf{2.0}} \pm \emph{\textbf{0.6}}$&$21.6 \pm 2.8$&$\textbf{14} \pm \textbf{4}$&$15.05 \pm 0.27$\\
			
			& 50&$65.7 \pm 2.6$&$65.7 \pm 2.3$&$67.6 \pm 2.3$&$66.0 \pm 2.7$&$85 \pm 11$&$\textbf{29.4} \pm \textbf{3.5}$&$\emph{\textbf{6.8}} \pm \emph{\textbf{2.0}}$&$76 \pm 7$&$48 \pm 14$&$\textbf{45.2} \pm \textbf{0.7}$\\
			
			& 100&$118.5 \pm 3.3$&$118.0 \pm 3.2$&$129 \pm 9$&$118 \pm 4$&$173 \pm 24$&$\textbf{67} \pm \textbf{7}$&$\emph{\textbf{14}} \pm \emph{\textbf{4}}$&$155 \pm 12$&$98 \pm 32$&$\textbf{82.5} \pm \textbf{0.9}$\\
			
			& 250&$254 \pm 12$&$254 \pm 11$&$282 \pm 11$&$251 \pm 9$&$\left(4.3 \pm 0.5\right) \times 10^{2}$&$\textbf{202} \pm \textbf{24}$&$\emph{\textbf{35}} \pm \emph{\textbf{10}}$&$400 \pm 32$&$\left(2.6 \pm 0.8\right) \times 10^{2}$&$\textbf{183.2} \pm \textbf{1.2}$\\
			\bottomrule  
	\end{tabular}}
\end{table}

\begin{table}[htp]
	\caption {Comparison of final $\log(\text{regret})$ for different BO benchmark functions. The best three performances are shown in bold, and the best one in italics. We can see that SnAKe constantly achieves regret comparable with classical Bayesian Optimization methods. The worst performance happens when $\epsilon = 0$, this could be explained by the method getting stuck in local optimums.}
	\label{tab: reg_comparison_exp_1_small_budgets}
	\resizebox{\columnwidth}{!}{
		\begin{tabular}{cccccccccccc}
			\toprule   
			{\bf Method} & {\bf Budget} & {\bf 0.0-SnAKe} & {\bf 0.1-SnAKe} & {\bf 1.0-SnAKe} & {\bf $\ell$-SnAKe} & {\bf EI} & {\bf EIpu} & {\bf TrEI} & {\bf UCB} & {\bf PI} & {\bf Random}\\
			\midrule
			\multirow{4}{*}{Branin2D}
			& 15&$-3.9 \pm 1.7$&$-3.5 \pm 0.7$&$-3.7 \pm 1.6$&$-3.4 \pm 0.9$&$-\textbf{4.7} \pm \textbf{1.5}$&$-\textbf{4.5} \pm \textbf{1.4}$&$-4.4 \pm 1.3$&$-\emph{\textbf{5.0}} \pm \emph{\textbf{1.5}}$&$-3.2 \pm 1.5$&$-3.3 \pm 0.8$\\
			
			& 50&$-6.2 \pm 2.5$&$-8.1 \pm 2.6$&$-7.9 \pm 2.4$&$-\textbf{8.3} \pm \textbf{2.3}$&$-\emph{\textbf{8.7}} \pm \emph{\textbf{1.7}}$&$-7.0 \pm 1.9$&$-6.1 \pm 1.6$&$-\textbf{8.5} \pm \textbf{2.6}$&$-6.2 \pm 2.5$&$-4.4 \pm 1.2$\\
			
			& 100&$-9.1 \pm 3.2$&$-\textbf{11.2} \pm \textbf{2.2}$&$-\textbf{11.4} \pm \textbf{2.7}$&$-10.7 \pm 2.2$&$-\emph{\textbf{13}} \pm \emph{\textbf{5}}$&$-7.4 \pm 2.0$&$-7.0 \pm 2.0$&$-10.7 \pm 2.5$&$-10.8 \pm 1.7$&$-5.1 \pm 1.1$\\
			
			& 250&$-12.3 \pm 1.6$&$-13.2 \pm 1.3$&$-\textbf{13.6} \pm \textbf{1.4}$&$-\textbf{13.5} \pm \textbf{1.4}$&$-\emph{\textbf{15}} \pm \emph{\textbf{7}}$&$-7.9 \pm 1.7$&$-8.1 \pm 1.5$&$-11.9 \pm 2.5$&$-12 \pm 5$&$-6.3 \pm 1.2$\\
			
			\midrule
			\multirow{4}{*}{Ackley4D}
			& 15&$1.62 \pm 0.25$&$1.62 \pm 0.25$&$\textbf{1.62} \pm \textbf{0.25}$&$1.67 \pm 0.18$&$1.71 \pm 0.15$&$1.69 \pm 0.20$&$1.73 \pm 0.19$&$1.71 \pm 0.15$&$\textbf{1.59} \pm \textbf{0.22}$&$\emph{\textbf{1.40}} \pm \emph{\textbf{0.11}}$\\
			
			& 50&$1.64 \pm 0.19$&$1.64 \pm 0.19$&$1.64 \pm 0.19$&$1.65 \pm 0.19$&$\textbf{1.59} \pm \textbf{0.22}$&$1.69 \pm 0.20$&$1.73 \pm 0.19$&$1.70 \pm 0.14$&$\textbf{1.34} \pm \textbf{0.21}$&$\emph{\textbf{1.21}} \pm \emph{\textbf{0.14}}$\\
			
			& 100&$1.70 \pm 0.15$&$1.69 \pm 0.15$&$1.69 \pm 0.15$&$1.72 \pm 0.11$&$\textbf{1.54} \pm \textbf{0.27}$&$1.69 \pm 0.20$&$1.73 \pm 0.19$&$1.70 \pm 0.14$&$\textbf{1.21} 
			\pm \textbf{0.11}$&$\emph{\textbf{1.06}} \pm \emph{\textbf{0.17}}$\\
			
			& 250&$1.52 \pm 0.22$&$\textbf{0.9} \pm \textbf{0.8}$&$1.2 \pm 0.5$&$1.0 \pm 0.5$&$1.4 \pm 0.5$&$1.69 \pm 0.20$&$1.73 \pm 0.19$&$1.5 \pm 0.4$&$\emph{\textbf{0.85}} \pm \emph{\textbf{0.32}}$&$\textbf{0.89} \pm \textbf{0.20}$\\
			
			\midrule
			\multirow{4}{*}{Michaelwicz2D}
			& 15&$-4.2 \pm 1.7$&$-4.8 \pm 1.6$&$-5.1 \pm 1.8$&$-4.5 \pm 1.9$&$-\textbf{5.4} \pm \textbf{1.1}$&$-\textbf{5.8} \pm \textbf{1.1}$&$-5.4 \pm 1.0$&$-\emph{\textbf{5.9}} \pm \emph{\textbf{1.3}}$&$-4.3 \pm 1.6$&$-4.5 \pm 1.0$\\
			
			& 50&$-6.4 \pm 1.5$&$-\textbf{7.3} \pm \textbf{1.4}$&$-6.8 \pm 1.3$&$-6.3 \pm 1.4$&$-6.0 \pm 1.3$&$-6.4 \pm 1.1$&$-\textbf{7.0} \pm \textbf{1.1}$&$-\emph{\textbf{7.8}} \pm \emph{\textbf{1.9}}$&$-6.2 \pm 1.7$&$-5.4 \pm 0.4$\\
			
			& 100&$-7.0 \pm 2.2$&$-7.2 \pm 1.7$&$-\textbf{7.8} \pm \textbf{1.8}$&$-\textbf{7.6} \pm \textbf{1.9}$&$-6.2 \pm 0.8$&$-6.6 \pm 1.2$&$-7.5 \pm 1.1$&$-\emph{\textbf{8.2}} \pm \emph{\textbf{2.0}}$&$-6.6 \pm 1.9$&$-6.0 \pm 0.4$\\
			
			& 250&$-6.6 \pm 1.6$&$-\textbf{8.1} \pm \textbf{2.2}$&$-\textbf{8.5} \pm \textbf{2.2}$&$-8.0 \pm 2.2$&$-6.5 \pm 0.7$&$-6.7 \pm 1.3$&$-7.7 \pm 1.0$&$-\emph{\textbf{8.7}} \pm \emph{\textbf{2.5}}$&$-7.0 \pm 1.4$&$-6.5 \pm 0.7$\\
			
			\midrule
			\multirow{4}{*}{Hartmann3D}
			& 15&$-1.2 \pm 1.5$&$-2.0 \pm 1.9$&$-1.9 \pm 1.7$&$-1.8 \pm 1.6$&$-\textbf{2.2} \pm \textbf{1.6}$&$-\textbf{2.7} \pm \textbf{1.5}$&$-\emph{\textbf{3.2}} \pm \emph{\textbf{1.5}}$&$-2.0 \pm 1.7$&$-0.6 \pm 1.1$&$-0.3 \pm 0.6$\\
			
			& 50&$-2.9 \pm 2.6$&$-4.8 \pm 3.0$&$-4.8 \pm 2.4$&$-4.8 \pm 2.3$&$-\emph{\textbf{7.4}} \pm \emph{\textbf{1.6}}$&$-5.0 \pm 1.5$&$-5.1 \pm 1.1$&$-\textbf{6.1} \pm \textbf{2.5}$&$-\textbf{5.2} \pm \textbf{2.0}$&$-1.4 \pm 0.8$\\
			
			& 100&$-5.4 \pm 2.9$&$-\textbf{8.3} \pm \textbf{1.5}$&$-7.9 \pm 2.4$&$-8.2 \pm 2.1$&$-\emph{\textbf{10.9}} \pm \emph{\textbf{1.3}}$&$-5.8 \pm 1.5$&$-5.6 \pm 1.0$&$-8.1 \pm 3.2$&$-\textbf{9.9} \pm \textbf{2.6}$&$-1.5 \pm 0.5$\\
			
			& 250&$-6 \pm 4$&$-9.8 \pm 2.7$&$-9.2 \pm 2.5$&$-9.4 \pm 2.0$&$-\emph{\textbf{12.4}} \pm \emph{\textbf{2.0}}$&$-6.8 \pm 1.8$&$-6.7 \pm 1.2$&$-\textbf{10} \pm \textbf{4}$&$-\textbf{12.0} \pm \textbf{2.1}$&$-2.4 \pm 0.7$\\
			
			\midrule
			\multirow{4}{*}{Hartmann6D}
			& 15&$0.6 \pm 0.5$&$\textbf{0.6} \pm \textbf{0.5}$&$0.6 \pm 0.5$&$\emph{\textbf{0.5}} \pm \emph{\textbf{0.6}}$&$0.88 \pm 0.33$&$0.8 \pm 0.5$&$\textbf{0.5} \pm \textbf{0.5}$&$0.77 \pm 0.32$&$0.8 \pm 0.4$&$0.7 \pm 0.4$\\
			
			& 50&$0.3 \pm 0.6$&$\textbf{0.1} \pm \textbf{0.7}$&$-\emph{\textbf{0.0}} \pm \emph{\textbf{0.8}}$&$\textbf{0.0} \pm \textbf{0.7}$&$0.6 \pm 0.4$&$0.3 \pm 0.6$&$0.1 \pm 
			0.6$&$0.4 \pm 0.5$&$0.5 \pm 0.6$&$0.49 \pm 0.26$\\
			
			& 100&$-\textbf{0.2} \pm \textbf{0.8}$&$-\textbf{0.3} \pm \textbf{1.0}$&$-0.1 \pm 0.9$&$-\emph{\textbf{0.6}} \pm \emph{\textbf{0.8}}$&$0.2 \pm 0.8$&$-0.1 \pm 0.9$&$-0.1 \pm 0.5$&$0.0 \pm 0.6$&$0.0 \pm 0.7$&$0.1 \pm 0.5$\\
			
			& 250&$-0.6 \pm 0.8$&$-0.7 \pm 1.5$&$-0.7 \pm 1.0$&$-\textbf{0.9} \pm \textbf{1.0}$&$-0.7 \pm 1.0$&$-0.5 \pm 1.0$&$-0.4 \pm 0.6$&$-\emph{\textbf{0.9}} \pm \emph{\textbf{0.8}}$&$-\textbf{0.8} \pm \textbf{0.9}$&$-0.04 \pm 0.34$\\
			
			\midrule
			\multirow{4}{*}{Perm10D}
			& 15&$-2.0 \pm 1.7$&$-2.0 \pm 1.7$&$-2.1 \pm 1.7$&$-2.3 \pm 1.7$&$-5.4 \pm 1.8$&$-4.9 \pm 1.0$&$-4.3 \pm 2.6$&$-\emph{\textbf{6.6}} \pm \emph{\textbf{1.6}}$&$-\textbf{6.1} \pm \textbf{2.0}$&$-\textbf{5.7} \pm \textbf{1.1}$\\
			
			& 50&$-3.2 \pm 1.9$&$-3.2 \pm 1.9$&$-3.6 \pm 2.0$&$-3.3 \pm 1.8$&$-6.0 \pm 2.2$&$-\textbf{7.5} \pm \textbf{1.9}$&$-5.1 \pm 2.5$&$-\textbf{7.5} \pm \textbf{1.7}$&$-7.1 
			\pm 1.9$&$-\emph{\textbf{7.5}} \pm \emph{\textbf{1.7}}$\\
			
			& 100&$-3.4 \pm 1.4$&$-3.4 \pm 1.4$&$-4.7 \pm 1.7$&$-3.7 \pm 1.3$&$-6.4 \pm 2.2$&$-\textbf{8.4} \pm \textbf{1.7}$&$-6.4 \pm 2.0$&$-8.0 \pm 1.7$&$-\emph{\textbf{8.5}} \pm \emph{\textbf{1.8}}$&$-\textbf{8.2} \pm \textbf{1.2}$\\
			
			& 250&$-5.0 \pm 2.3$&$-5.0 \pm 2.3$&$-6.4 \pm 2.3$&$-4.1 \pm 2.1$&$-6.8 \pm 2.4$&$-\textbf{9.3} \pm \textbf{1.2}$&$-8.1 \pm 1.0$&$-8.5 \pm 1.9$&$-\emph{\textbf{9.6}} \pm \emph{\textbf{1.9}}$&$-\textbf{9.5} \pm \textbf{1.6}$\\
			\bottomrule  
	\end{tabular}}
\end{table}

\subsubsection{Asynchronous Experiments}

In this section we include the full table results of all asynchronous experiments. The results are shown in Table \ref{tab: cost_comparison_exp_aysync} and \ref{tab: reg_comparison_exp_aysync}.

\begin{table}[htp]
	\caption {Comparison of 2-norm cost for different benchmark functions in the \textit{asynchronous setting}. The best three performances are shown in bold, and the best one in italics. SnAKe achieves considerable lower cost with respect to other methods, achieving the top 3 lowest costs all but one time.} 
	\label{tab: cost_comparison_exp_aysync}
	\resizebox{\columnwidth}{!}{
		\begin{tabular}{ccccccccccc}
			\toprule   
			{\bf Method} & {\bf Budget} & {\bf Delay} & {\bf 0.0-SnAKe} & {\bf 0.1-SnAKe} & {\bf 1.0-SnAKe} & {\bf $\ell$-SnAKe} & {\bf Random} & {\bf TS} & {\bf UCBwLP} & {\bf EIpuLP}\\
			
			\midrule
			\multirow{4}{*}{Branin2D}
			& 100 & 10 &$\emph{\textbf{7.0}} \pm \emph{\textbf{1.9}}$&$\textbf{10.0} \pm \textbf{2.2}$&$10.6 \pm 3.2$&$\textbf{9.8} \pm \textbf{2.6}$&$10.4 \pm 0.4$&$49 \pm 5$&$27.5 \pm 2.5$&$22 \pm 5$\\
			
			& 100 & 25 &$\emph{\textbf{7.8}} \pm \emph{\textbf{1.6}}$&$11.2 \pm 2.7$&$\textbf{9.6} \pm \textbf{2.6}$&$10.6 \pm 2.4$&$\textbf{10.2} \pm \textbf{0.5}$&$52 \pm 6$&$51 \pm 4$&$25 \pm 7$\\
			
			& 250 & 10 &$\emph{\textbf{7.6}} \pm \emph{\textbf{1.7}}$&$\textbf{14.5} \pm \textbf{1.9}$&$15.9 \pm 3.3$&$\textbf{14.9} \pm \textbf{2.9}$&$16.6 \pm 0.8$&$120 \pm 12$&$37 \pm 5$&$37 \pm 14$\\
			
			& 250 & 25 &$\emph{\textbf{8.8}} \pm \emph{\textbf{2.1}}$&$\textbf{13.4} \pm \textbf{1.5}$&$14.7 \pm 3.3$&$\textbf{13.1} \pm \textbf{2.7}$&$16.6 \pm 0.6$&$122 \pm 10$&$56 \pm 4$&$48 \pm 14$\\
			
			\midrule
			\multirow{4}{*}{Ackley4D}
			& 100 & 10 &$\emph{\textbf{22.2}} \pm \emph{\textbf{2.3}}$&$23.7 \pm 1.9$&$\textbf{22.5} \pm \textbf{2.0}$&$\textbf{22.4} \pm \textbf{2.2}$&$32.5 \pm 0.8$&$110 \pm 6$&$96 \pm 8$&$100 \pm 26$\\
			
			& 100 & 25 &$\emph{\textbf{19.6}} \pm \emph{\textbf{2.4}}$&$24.0 \pm 2.5$&$\textbf{22.9} \pm \textbf{3.2}$&$\textbf{23.3} \pm \textbf{3.5}$&$32.6 \pm 0.9$&$101 \pm 10$&$100 \pm 4$&$59 \pm 4$\\
			
			& 250 & 10 &$\textbf{27} \pm \textbf{4}$&$30.1 \pm 2.9$&$\textbf{26.3} \pm \textbf{2.6}$&$\emph{\textbf{26.2}} \pm \emph{\textbf{1.5}}$&$59.6 \pm 1.1$&$219 \pm 23$&$243 \pm 25$&$\left(2.5 \pm 0.7\right) \times 10^{2}$\\
			
			& 250 & 25 &$\textbf{28.5} \pm \textbf{3.3}$&$32 \pm 4$&$\emph{\textbf{25.4}} \pm \emph{\textbf{2.3}}$&$\textbf{25.6} \pm \textbf{2.3}$&$59.5 \pm 0.9$&$\left(2.2 \pm 0.5\right) \times 10^{2}$&$240 \pm 20$&$175 \pm 5$\\
			
			\midrule
			\multirow{4}{*}{Michaelwicz2D}
			& 100 & 10 &$\emph{\textbf{3.4}} \pm \emph{\textbf{1.4}}$&$5.0 \pm 1.5$&$\textbf{4.5} \pm \textbf{1.0}$&$\textbf{4.7} \pm \textbf{1.0}$&$10.5 \pm 0.4$&$17.8 \pm 2.4$&$23.0 \pm 2.6$&$54 \pm 7$\\
			
			& 100 & 25 &$\emph{\textbf{4.2}} \pm \emph{\textbf{1.1}}$&$6.0 \pm 1.2$&$\textbf{5.7} \pm \textbf{1.3}$&$\textbf{5.9} \pm \textbf{1.2}$&$10.5 \pm 0.5$&$25.8 \pm 2.6$&$33.5 \pm 2.9$&$47.4 \pm 3.3$\\
			
			& 250 & 10 &$\emph{\textbf{3.5}} \pm \emph{\textbf{1.7}}$&$\textbf{5.9} \pm \textbf{0.8}$&$\textbf{5.7} \pm \textbf{1.3}$&$6.4 \pm 1.6$&$16.3 \pm 0.9$&$27.2 \pm 3.4$&$37 \pm 7$&$67 \pm 9$\\
			
			& 250 & 25 &$\emph{\textbf{3.7}} \pm \emph{\textbf{1.1}}$&$\textbf{6.6} \pm \textbf{1.1}$&$6.7 \pm 1.2$&$\textbf{6.6} \pm \textbf{1.0}$&$16.4 \pm 0.7$&$35.3 \pm 2.6$&$36.0 \pm 2.8$&$96 \pm 9$\\
			
			\midrule
			\multirow{4}{*}{Hartmann3D}
			& 100 & 10 &$\emph{\textbf{7.7}} \pm \emph{\textbf{3.4}}$&$11 \pm 4$&$\textbf{9.7} \pm \textbf{3.2}$&$\textbf{10} \pm \textbf{4}$&$21.7 \pm 0.6$&$20.6 \pm 3.3$&$34 \pm 6$&$28 \pm 4$\\
			
			& 100 & 25 &$\emph{\textbf{9.6}} \pm \emph{\textbf{2.9}}$&$\textbf{13} \pm \textbf{4}$&$\textbf{12.8} \pm \textbf{3.2}$&$14 \pm 5$&$21.3 \pm 0.5$&$32 \pm 4$&$55 \pm 5$&$49 \pm 5$\\
			
			& 250 & 10 &$\emph{\textbf{5.8}} \pm \emph{\textbf{2.4}}$&$\textbf{11} \pm \textbf{4}$&$\textbf{9.8} \pm \textbf{3.1}$&$11 \pm 4$&$38.6 \pm 0.7$&$27 \pm 4$&$42 \pm 5$&$36 \pm 5$\\
			
			& 250 & 25 &$\emph{\textbf{7.6}} \pm \emph{\textbf{2.3}}$&$\textbf{12} \pm \textbf{4}$&$\textbf{11.2} \pm \textbf{3.1}$&$12 \pm 4$&$38.7 \pm 0.9$&$38 \pm 4$&$64 \pm 9$&$57 \pm 5$\\
			
			\midrule
			\multirow{4}{*}{Hartmann4D}
			& 100 & 10 &$\emph{\textbf{12}} \pm \emph{\textbf{5}}$&$\textbf{18} \pm \textbf{6}$&$\textbf{18} \pm \textbf{5}$&$21 \pm 8$&$32.6 \pm 0.7$&$39 \pm 13$&$56 \pm 9$&$47 \pm 13$\\
			
			& 100 & 25 &$\emph{\textbf{15}} \pm \emph{\textbf{4}}$&$\textbf{20} \pm \textbf{4}$&$\textbf{21} \pm \textbf{5}$&$23 \pm 5$&$32.5 \pm 0.8$&$42 \pm 8$&$71 \pm 9$&$58 \pm 7$\\
			
			& 250 & 10 &$\emph{\textbf{13}} \pm \emph{\textbf{7}}$&$\textbf{22} \pm \textbf{12}$&$\textbf{20} \pm \textbf{10}$&$22 \pm 13$&$59.3 \pm 1.0$&$\left(8 \pm 5\right) \times 10^{1}$&$103 \pm 28$&$\left(8 \pm 4\right) \times 10^{1}$\\
			
			& 250 & 25 &$\emph{\textbf{17}} \pm \emph{\textbf{11}}$&$\textbf{28} \pm \textbf{14}$&$\textbf{24} \pm \textbf{11}$&$31 \pm 15$&$59.4 \pm 1.1$&$\left(9 \pm 4\right) \times 10^{1}$&$113 \pm 20$&$96 \pm 34$\\
			
			\midrule
			\multirow{4}{*}{Hartmann6D}
			& 100 & 10 &$\emph{\textbf{14.9}} \pm \emph{\textbf{3.5}}$&$\textbf{17} \pm \textbf{5}$&$17.9 \pm 3.5$&$\textbf{18} \pm \textbf{4}$&$52.1 \pm 1.2$&$39 \pm 10$&$120 \pm 19$&$124 \pm 11$\\
			
			& 100 & 25 &$\emph{\textbf{20.9}} \pm \emph{\textbf{2.7}}$&$\textbf{22.4} \pm \textbf{2.7}$&$23.9 \pm 3.2$&$\textbf{24} \pm \textbf{4}$&$51.7 \pm 1.0$&$56 \pm 10$&$109 \pm 11$&$114 \pm 9$\\
			
			& 250 & 10 &$\emph{\textbf{18}} \pm \emph{\textbf{5}}$&$\textbf{18} \pm \textbf{5}$&$\textbf{18} \pm \textbf{4}$&$20 \pm 7$&$107.1 \pm 1.6$&$53 \pm 19$&$\left(2.6 \pm 
			0.6\right) \times 10^{2}$&$\left(3.1 \pm 0.5\right) \times 10^{2}$\\
			
			& 250 & 25 &$\emph{\textbf{21}} \pm \emph{\textbf{5}}$&$\textbf{23} \pm \textbf{6}$&$24 \pm 6$&$\textbf{23} \pm \textbf{5}$&$107.0 \pm 1.8$&$74 \pm 18$&$\left(2.7 \pm 
			0.5\right) \times 10^{2}$&$306 \pm 30$\\
			
			\bottomrule  
	\end{tabular}}
\end{table}

\begin{table}
	\caption {Comparison of $\log(\text{regret})$ for different benchmark functions in the \textit{asynchronous setting}. The best three performances are shown in bold, and the best one in italics. SnAKe achieves regret comparable with other Bayesian Optimization methods.}
	\label{tab: reg_comparison_exp_aysync}
	\resizebox{\columnwidth}{!}{
	\begin{tabular}{ccccccccccc}
		\toprule   
		{\bf Method} & {\bf Budget} & {\bf Delay} & {\bf 0.0-SnAKe} & {\bf 0.1-SnAKe} & {\bf 1.0-SnAKe} & {\bf $\ell$-SnAKe} & {\bf Random} & {\bf TS} & {\bf UCBwLP} & {\bf EIpuLP}\\
			
		\midrule
		\multirow{4}{*}{Branin2D}
		& 100 & 10 &$-9.6 \pm 2.4$&$-9.7 \pm 2.2$&$-\textbf{10.3} \pm \textbf{2.4}$&$-10.1 \pm 2.4$&$-5.7 \pm 1.8$&$-\textbf{11.7} \pm \textbf{1.1}$&$-\emph{\textbf{12.1}} \pm \emph{\textbf{1.5}}$&$-7.3 \pm 2.5$\\
		
		& 100 & 25 &$-7.1 \pm 2.9$&$-\textbf{8.3} \pm \textbf{3.1}$&$-7.5 \pm 2.6$&$-7.1 \pm 2.2$&$-5.4 \pm 1.0$&$-\emph{\textbf{11.7}} \pm \emph{\textbf{1.6}}$&$-\textbf{8.2} \pm \textbf{2.1}$&$-5.4 \pm 1.7$\\
		
		& 250 & 10 &$-11.9 \pm 1.9$&$-12.9 \pm 0.8$&$-12.8 \pm 1.1$&$-\textbf{13.3} \pm \textbf{1.2}$&$-5.9 \pm 1.1$&$-\textbf{13.8} \pm \textbf{1.0}$&$-\emph{\textbf{14.2}} \pm \emph{\textbf{1.6}}$&$-9.3 \pm 2.9$\\
		
		& 250 & 25 &$-11.6 \pm 2.0$&$-11.6 \pm 0.8$&$-\textbf{12.1} \pm \textbf{0.8}$&$-12.1 \pm 1.1$&$-6.2 \pm 1.1$&$-\emph{\textbf{15}} \pm \emph{\textbf{6}}$&$-\textbf{14.8} \pm \textbf{1.3}$&$-8.9 \pm 2.7$\\
		
		\midrule
		\multirow{4}{*}{Ackley4D}
		& 100 & 10 &$1.0 \pm 0.5$&$\textbf{0.9} \pm \textbf{0.7}$&$\textbf{0.9} \pm \textbf{0.7}$&$\emph{\textbf{0.8}} \pm \emph{\textbf{0.7}}$&$1.05 \pm 0.18$&$1.46 \pm 0.15$&$1.08 \pm 0.21$&$1.4 \pm 0.4$\\
		
		& 100 & 25 &$1.21 \pm 0.19$&$\textbf{1.1} \pm \textbf{0.4}$&$1.15 \pm 0.33$&$1.20 \pm 0.23$&$\textbf{1.03} \pm \textbf{0.17}$&$1.32 \pm 0.16$&$\emph{\textbf{1.02}} \pm \emph{\textbf{0.23}}$&$1.16 \pm 0.17$\\
		
		& 250 & 10 &$-\emph{\textbf{0.4}} \pm \emph{\textbf{0.8}}$&$-\textbf{0.4} \pm \textbf{1.0}$&$0.0 \pm 0.8$&$-\textbf{0.2} \pm \textbf{0.7}$&$0.8 \pm 0.4$&$0.7 \pm 0.8$&$0.1 \pm 0.8$&$1.3 \pm 0.4$\\
		
		& 250 & 25 &$-0.3 \pm 0.5$&$-\emph{\textbf{0.9}} \pm \emph{\textbf{0.5}}$&$-\textbf{0.6} \pm \textbf{0.5}$&$-\textbf{0.6} \pm \textbf{0.5}$&$0.95 \pm 0.13$&$0.4 \pm 0.9$&$0.74 \pm 0.29$&$1.02 \pm 0.18$\\
		
		\midrule
		\multirow{4}{*}{Michaelwicz2D}
		& 100 & 10 &$-7.1 \pm 1.8$&$-7.1 \pm 1.4$&$-6.8 \pm 1.7$&$-7.4 \pm 1.7$&$-6.2 \pm 0.7$&$-\textbf{7.9} \pm \textbf{2.5}$&$-\emph{\textbf{9.0}} \pm \emph{\textbf{1.1}}$&$-\textbf{7.9} \pm \textbf{1.5}$\\
		
		& 100 & 25 &$-7.1 \pm 2.1$&$-\textbf{7.4} \pm \textbf{2.0}$&$-7.0 \pm 1.4$&$-6.8 \pm 1.1$&$-5.86 \pm 0.21$&$-7.4 \pm 1.6$&$-\emph{\textbf{11.2}} \pm \emph{\textbf{1.5}}$&$-\textbf{9.2} \pm \textbf{0.9}$\\
		
		& 250 & 10 &$-6.8 \pm 1.7$&$-8.3 \pm 2.1$&$-8.5 \pm 2.2$&$-8.4 \pm 2.4$&$-6.5 \pm 0.8$&$-\textbf{8.5} \pm \textbf{3.2}$&$-\textbf{9.0} \pm \textbf{1.1}$&$-\emph{\textbf{9.9}} \pm \emph{\textbf{1.7}}$\\
		
		& 250 & 25 &$-6.8 \pm 1.3$&$-8.1 \pm 2.0$&$-\textbf{8.6} \pm \textbf{2.5}$&$-8.4 \pm 2.3$&$-6.6 \pm 0.9$&$-7.9 \pm 2.4$&$-\emph{\textbf{11.2}} \pm \emph{\textbf{1.5}}$&$-\textbf{10.3} \pm \textbf{1.4}$\\
		
		\midrule
		\multirow{4}{*}{Hartmann3D}
		& 100 & 10 &$-5.3 \pm 2.8$&$-\textbf{7.8} \pm \textbf{3.0}$&$-6.8 \pm 3.4$&$-7.4 \pm 2.7$&$-1.9 \pm 0.9$&$-\textbf{9.5} \pm \textbf{1.1}$&$-\emph{\textbf{10.0}} \pm \emph{\textbf{1.2}}$&$-5.5 \pm 1.3$\\
		
		& 100 & 25 &$-4.7 \pm 2.4$&$-6.0 \pm 2.1$&$-6.1 \pm 1.8$&$-\textbf{6.4} \pm \textbf{1.7}$&$-1.6 \pm 0.8$&$-\emph{\textbf{8.6}} \pm \emph{\textbf{1.3}}$&$-\textbf{6.3} 
		\pm \textbf{1.5}$&$-4.3 \pm 1.4$\\
		
		& 250 & 10 &$-5 \pm 4$&$-7.9 \pm 3.5$&$-8.0 \pm 3.5$&$-\textbf{8.6} \pm \textbf{3.0}$&$-2.6 \pm 0.8$&$-\textbf{10.7} \pm \textbf{1.1}$&$-\emph{\textbf{12.8}} \pm \emph{\textbf{1.2}}$&$-6.3 \pm 1.3$\\
		
		& 250 & 25 &$-6 \pm 4$&$-8.4 \pm 3.0$&$-8.8 \pm 3.1$&$-\textbf{8.9} \pm \textbf{2.6}$&$-2.4 \pm 0.5$&$-\textbf{10.5} \pm \textbf{0.7}$&$-\emph{\textbf{12.4}} \pm \emph{\textbf{0.9}}$&$-6.2 \pm 1.1$\\
		
		\midrule
		\multirow{4}{*}{Hartmann4D}
		& 100 & 10 &$-3.1 \pm 3.0$&$-2.9 \pm 2.6$&$-\textbf{3.5} \pm \textbf{2.4}$&$-2.8 \pm 2.1$&$-1.1 \pm 0.6$&$-\emph{\textbf{6.5}} \pm \emph{\textbf{2.8}}$&$-\textbf{4.5} 
		\pm \textbf{1.9}$&$-2.6 \pm 1.3$\\
		
		& 100 & 25 &$-\textbf{2.1} \pm \textbf{1.8}$&$-2.0 \pm 1.7$&$-2.0 \pm 1.9$&$-1.9 \pm 1.3$&$-1.0 \pm 0.5$&$-\emph{\textbf{4.2}} \pm \emph{\textbf{1.4}}$&$-2.1 \pm 1.0$&$-\textbf{2.2} \pm \textbf{1.0}$\\
		
		& 250 & 10 &$-6 \pm 4$&$-\textbf{6} \pm \textbf{4}$&$-6 \pm 4$&$-4.2 \pm 3.3$&$-1.2 \pm 0.4$&$-\emph{\textbf{8.7}} \pm \emph{\textbf{2.1}}$&$-\textbf{7.2} \pm \textbf{1.2}$&$-4.6 \pm 2.0$\\
		
		& 250 & 25 &$-6 \pm 4$&$-6 \pm 4$&$-\textbf{6} \pm \textbf{4}$&$-5.5 \pm 3.2$&$-1.4 \pm 0.5$&$-\emph{\textbf{8.7}} \pm \emph{\textbf{1.4}}$&$-\textbf{6.5} \pm \textbf{1.6}$&$-4.2 \pm 1.7$\\
		
		\midrule
		\multirow{4}{*}{Hartmann6D}
		& 100 & 10 &$-\textbf{0.2} \pm \textbf{0.6}$&$-\emph{\textbf{0.3}} \pm \emph{\textbf{0.9}}$&$-0.2 \pm 0.7$&$-\textbf{0.3} \pm \textbf{1.0}$&$0.1 \pm 0.4$&$-0.2 \pm 0.9$&$-0.0 \pm 0.7$&$0.3 \pm 0.5$\\
		
		& 100 & 25 &$-\textbf{0.3} \pm \textbf{0.7}$&$-\textbf{0.2} \pm \textbf{0.5}$&$-0.2 \pm 0.6$&$-0.2 \pm 0.6$&$0.07 \pm 0.29$&$-\emph{\textbf{0.4}} \pm \emph{\textbf{0.6}}$&$0.0 \pm 0.4$&$0.0 \pm 0.5$\\
		
		& 250 & 10 &$-0.5 \pm 1.1$&$-0.7 \pm 1.0$&$-\textbf{0.8} \pm \textbf{0.7}$&$-\textbf{0.9} \pm \textbf{1.6}$&$-0.2 \pm 0.5$&$-0.5 \pm 1.0$&$-\emph{\textbf{1.1}} \pm \emph{\textbf{0.7}}$&$-0.2 \pm 0.8$\\
		
		& 250 & 25 &$-0.5 \pm 0.8$&$-0.8 \pm 0.9$&$-0.7 \pm 1.1$&$-\emph{\textbf{1.1}} \pm \emph{\textbf{1.5}}$&$-0.05 \pm 0.30$&$-\textbf{0.9} \pm \textbf{0.9}$&$-\textbf{1.0} \pm \textbf{0.6}$&$-0.2 \pm 0.5$\\
			
		\bottomrule  
	\end{tabular}}
\end{table}

\subsubsection{SnAr Benchmark}

We ran additional experiments on the SnAr benchmark. For the first one (which includes the example looked at in the main paper) we tested on a budget of $T = 100$ iterations for different values of $t_{delay}$. The results are included in Tables \ref{tab: comparison_cost_snar_async} and \ref{tab: comparison_log_regret_snar_async}.

We also carried out synchronous results on the benchmark, for different budgets. The results are included in Tables \ref{tab: comparison_cost_snar} and \ref{tab: comparison_log_regret_snar}.

\begin{table}
	\caption {Comparison of cost on  SnAr benchmark (asynchronous) for $T = 100$ and different values of $t_{delay}$. The best three performances are shown in bold, and the best one in italics. SnAKe consistently achieves lower cost than BO methods. EIpuLP achieves lower cost for small delays, suggesting over-exploration in the early stages, and under-exploration once observations arrive.} 
	\label{tab: comparison_cost_snar_async}
	\resizebox{\textwidth}{8.5mm}{
		\begin{tabular}{ccccccccccc}
			\toprule   
			{\bf delay} &  {\bf 0.0-SnAKe} & {\bf 0.1-SnAKe} & {\bf 1.0-SnAKe} & {\bf $\ell$-SnAKe} & {\bf Random} & {\bf TS} & {\bf UCBwLP} & {\bf EIpuLP}\\
			
			\midrule
			5&$\left(\textbf{5.0} \pm \textbf{1.1}\right) \times \textbf{10}^{\textbf{2}}$&$\left(5.7 \pm 0.9\right) \times 10^{2}$&$\left(\textbf{5.5} \pm \textbf{0.8}\right) \times \textbf{10}^{\textbf{2}}$&$\left(6.2 \pm 0.6\right) \times 10^{2}$&$596 \pm 26$&$\left(1.12 \pm 0.05\right) \times 10^{3}$&$\left(9.2 \pm 0.6\right) \times 10^{2}$&$\left(\emph{\textbf{1.6}} \pm \emph{\textbf{0.5}}\right) \times \emph{\textbf{10}}^{\emph{\textbf{2}}}$\\
			
			10&$\left(\textbf{4.5} \pm \textbf{0.8}\right) \times \textbf{10}^{\textbf{2}}$&$\left(\textbf{5.3} \pm \textbf{0.7}\right) \times \textbf{10}^{\textbf{2}}$&$\left(5.3 \pm 0.7\right) \times 10^{2}$&$\left(5.8 \pm 0.8\right) \times 10^{2}$&$600 \pm 20$&$\left(1.11 \pm 0.06\right) \times 10^{3}$&$\left(8.9 \pm 0.8\right) \times 10^{2}$&$\left(\emph{\textbf{1.9}} \pm \emph{\textbf{0.4}}\right) \times \emph{\textbf{10}}^{\emph{\textbf{2}}}$\\
			
			25&$\left(\textbf{4.0} \pm \textbf{0.7}\right) \times \textbf{10}^{\textbf{2}}$&$\left(4.8 \pm 0.6\right) \times 10^{2}$&$\left(\textbf{4.5} \pm \textbf{0.8}\right) \times \textbf{10}^{\textbf{2}}$&$\left(5.1 \pm 0.6\right) \times 10^{2}$&$603 \pm 25$&$\left(1.09 \pm 0.06\right) \times 10^{3}$&$\left(9.3 \pm 0.9\right) \times 10^{2}$&$\emph{\textbf{307}} \pm \emph{\textbf{31}}$\\
			
			50&$\left(\emph{\textbf{4.0}} \pm \emph{\textbf{0.5}}\right) \times \emph{\textbf{10}}^{\emph{\textbf{2}}}$&$\left(\textbf{4.5} \pm \textbf{0.4}\right) \times \textbf{10}^{\textbf{2}}$&$\left(\textbf{4.2} \pm \textbf{0.6}\right) \times \textbf{10}^{\textbf{2}}$&$\left(4.6 \pm 0.5\right) \times 10^{2}$&$606 \pm 24$&$\left(1.12 \pm 0.05\right) \times 10^{3}$&$\left(1.07 \pm 0.06\right) \times 10^{3}$&$568 \pm 27$\\
			
			\bottomrule  
	\end{tabular}}
\end{table}

\begin{table}[ht]
	\centering
	\caption {Comparison of regret on  SnAr benchmark (asynchronous) for $T = 100$ and different values of $t_{delay}$. The best three performances are shown in bold, and the best one in italics. SnAKe consistently achieves regret comparable with BO methods. EIpuLP performs well for small delays, but poorly as the delay is increased.} 
	\label{tab: comparison_log_regret_snar_async}
	\resizebox{0.9\textwidth}{9.55mm}{
		\begin{tabular}{ccccccccccc}
			\toprule   
			{\bf delay} &  {\bf 0.0-SnAKe} & {\bf 0.1-SnAKe} & {\bf 1.0-SnAKe} & {\bf $\ell$-SnAKe} & {\bf Random} & {\bf TS} & {\bf UCBwLP} & {\bf EIpuLP}\\
			
			\midrule
			5&$-\textbf{4.5} \pm \textbf{1.1}$&$-4.5 \pm 1.2$&$-4.3 \pm 1.5$&$-3.5 \pm 1.1$&$-0.93 \pm 0.25$&$-\textbf{4.6} \pm \textbf{1.4}$&$-\emph{\textbf{4.9}} \pm \emph{\textbf{1.4}}$&$-4.1 \pm 0.8$\\
			
			10&$-3.5 \pm 1.2$&$-3.9 \pm 1.5$&$-3.7 \pm 1.1$&$-4.0 \pm 1.4$&$-1.1 \pm 0.4$&$-\emph{\textbf{4.3}} \pm \emph{\textbf{1.3}}$&$-\textbf{4.0} \pm \textbf{1.2}$&$-\textbf{4.2} \pm \textbf{0.8}$\\
			
			25&$-\textbf{3.8} \pm \textbf{1.0}$&$-\textbf{3.7} \pm \textbf{1.4}$&$-3.3 \pm 0.9$&$-3.6 \pm 1.3$&$-0.9 \pm 0.4$&$-\emph{\textbf{4.2}} \pm \emph{\textbf{1.3}}$&$-2.9 
			\pm 0.6$&$-3.1 \pm 1.1$\\
			
			50&$-3.2 \pm 1.2$&$-\textbf{3.7} \pm \textbf{1.3}$&$-3.1 \pm 0.8$&$-\textbf{3.5} \pm \textbf{0.9}$&$-1.1 \pm 0.5$&$-\emph{\textbf{4.3}} \pm \emph{\textbf{1.2}}$&$-3.2 
			\pm 0.7$&$-2.3 \pm 1.0$\\
			
			\bottomrule  
	\end{tabular}}
\end{table}

\begin{table}
	\caption {Comparison of cost for SnAr benchmark (synchronous) for different budgets. The best three performances are shown in bold, and the best one in italics. SnAKe's performance is poor for small budgets, but improves considerably for the later ones. EIpu achieves the lowest cost in every instance, however Table \ref{tab: comparison_log_regret_snar} shows this is due to under-exploration.} 
	\label{tab: comparison_cost_snar}
	\resizebox{\columnwidth}{9mm}{
		\begin{tabular}{ccccccccccc}
			\toprule   
			{\bf Budget} &  {\bf 0.0-SnAKe} & {\bf 0.1-SnAKe} & {\bf 1.0-SnAKe} & {\bf $\ell$-SnAKe} & {\bf EI} & {\bf EIpu} &{\bf UCB} & {\bf PI} & {\bf Random}\\
			
			\midrule
			10&$98 \pm 13$&$101 \pm 16$&$104 \pm 12$&$103 \pm 11$&$\textbf{89} \pm \textbf{11}$&$\emph{\textbf{15}} \pm \emph{\textbf{8}}$&$103 \pm 10$&$\textbf{38} \pm \textbf{12}$&$91 \pm 7$\\
			
			25&$219 \pm 35$&$230 \pm 25$&$247 \pm 21$&$250 \pm 21$&$230 \pm 23$&$\emph{\textbf{46}} \pm \emph{\textbf{24}}$&$289 \pm 19$&$\textbf{108} \pm \textbf{24}$&$\textbf{202} \pm \textbf{10}$\\
			
			50&$\left(\textbf{3.7} \pm \textbf{0.7}\right) \times \textbf{10}^{\textbf{2}}$&$\left(4.0 \pm 0.5\right) \times 10^{2}$&$\left(4.3 \pm 0.6\right) \times 10^{2}$&$\left(4.1 \pm 0.7\right) \times 10^{2}$&$503 \pm 31$&$\left(\emph{\textbf{1.2}} \pm \emph{\textbf{0.4}}\right) \times \emph{\textbf{10}}^{\emph{\textbf{2}}}$&$\left(5.5 \pm 0.4\right) \times 10^{2}$&$\left(3.9 \pm 0.4\right) \times 10^{2}$&$\textbf{360} \pm \textbf{16}$\\
			
			100&$\left(\textbf{6.0} \pm \textbf{1.8}\right) \times \textbf{10}^{\textbf{2}}$&$\left(6.9 \pm 1.3\right) \times 10^{2}$&$\left(7.0 \pm 1.5\right) \times 10^{2}$&$\left(7.2 \pm 1.3\right) \times 10^{2}$&$\left(1.12 \pm 0.04\right) \times 10^{3}$&$\left(\emph{\textbf{3.0}} \pm \emph{\textbf{0.9}}\right) \times \emph{\textbf{10}}^{\emph{\textbf{2}}}$&$\left(1.11 \pm 0.06\right) \times 10^{3}$&$\left(1.00 \pm 0.05\right) \times 10^{3}$&$\textbf{605} \pm \textbf{26}$\\
			
			\bottomrule  
	\end{tabular}}
\end{table}

\begin{table}
	\caption {Comparison of $\log(\text{regret})$ for SnAr benchmark (synchronous) for different budgets. The best three performances are shown in bold, and the best one in italics. SnAKe achieves regret comparable with Bayesian Optimization methods. EIpu achieves the worst non-random performance in every instance.} 
	\label{tab: comparison_log_regret_snar}
	\resizebox{\columnwidth}{!}{
		\begin{tabular}{ccccccccccc}
			\toprule   
			{\bf Budget} &  {\bf 0.0-SnAKe} & {\bf 0.1-SnAKe} & {\bf 1.0-SnAKe} & {\bf $\ell$-SnAKe} & {\bf EI} & {\bf EIpu} & {\bf UCB} & {\bf PI} & {\bf Random}\\
			
			\midrule
			10&$-1.9 \pm 0.9$&$-2.0 \pm 0.9$&$-\textbf{2.2} \pm \textbf{1.1}$&$-1.6 \pm 0.8$&$-\emph{\textbf{2.2}} \pm \emph{\textbf{1.0}}$&$-1.0 \pm 0.8$&$-\textbf{2.0} \pm \textbf{1.0}$&$-1.0 \pm 1.0$&$-0.29 \pm 0.26$\\
			
			25&$-\textbf{3.4} \pm \textbf{0.8}$&$-3.3 \pm 0.7$&$-3.3 \pm 1.1$&$-2.9 \pm 0.8$&$-\textbf{3.6} \pm \textbf{0.7}$&$-2.9 \pm 0.9$&$-3.0 \pm 0.8$&$-\emph{\textbf{3.8}} \pm \emph{\textbf{1.3}}$&$-0.51 \pm 0.27$\\
			
			50&$-4.5 \pm 1.1$&$-\textbf{4.5} \pm \textbf{1.0}$&$-4.1 \pm 1.2$&$-4.3 \pm 1.0$&$-\emph{\textbf{4.9}} \pm \emph{\textbf{0.8}}$&$-3.7 \pm 1.0$&$-4.2 \pm 0.9$&$-\textbf{4.9} \pm \textbf{0.7}$&$-0.9 \pm 0.5$\\
			
			100&$-5.3 \pm 1.3$&$-5.8 \pm 1.1$&$-\textbf{5.9} \pm \textbf{1.2}$&$-5.6 \pm 1.0$&$-\textbf{6.0} \pm \textbf{0.8}$&$-4.9 \pm 1.0$&$-\emph{\textbf{6.0}} \pm \emph{\textbf{0.5}}$&$-5.8 \pm 0.7$&$-1.0 \pm 0.4$\\
			
			\bottomrule  
	\end{tabular}}
\end{table}

\subsection{Graphs for results of Section \ref{subsec: synthetic_experiments}}
\begin{figure}[ht]
	\centering
	\begin{subfigure}{\textwidth}
	\centering
	\includegraphics[width = 0.32\textwidth]{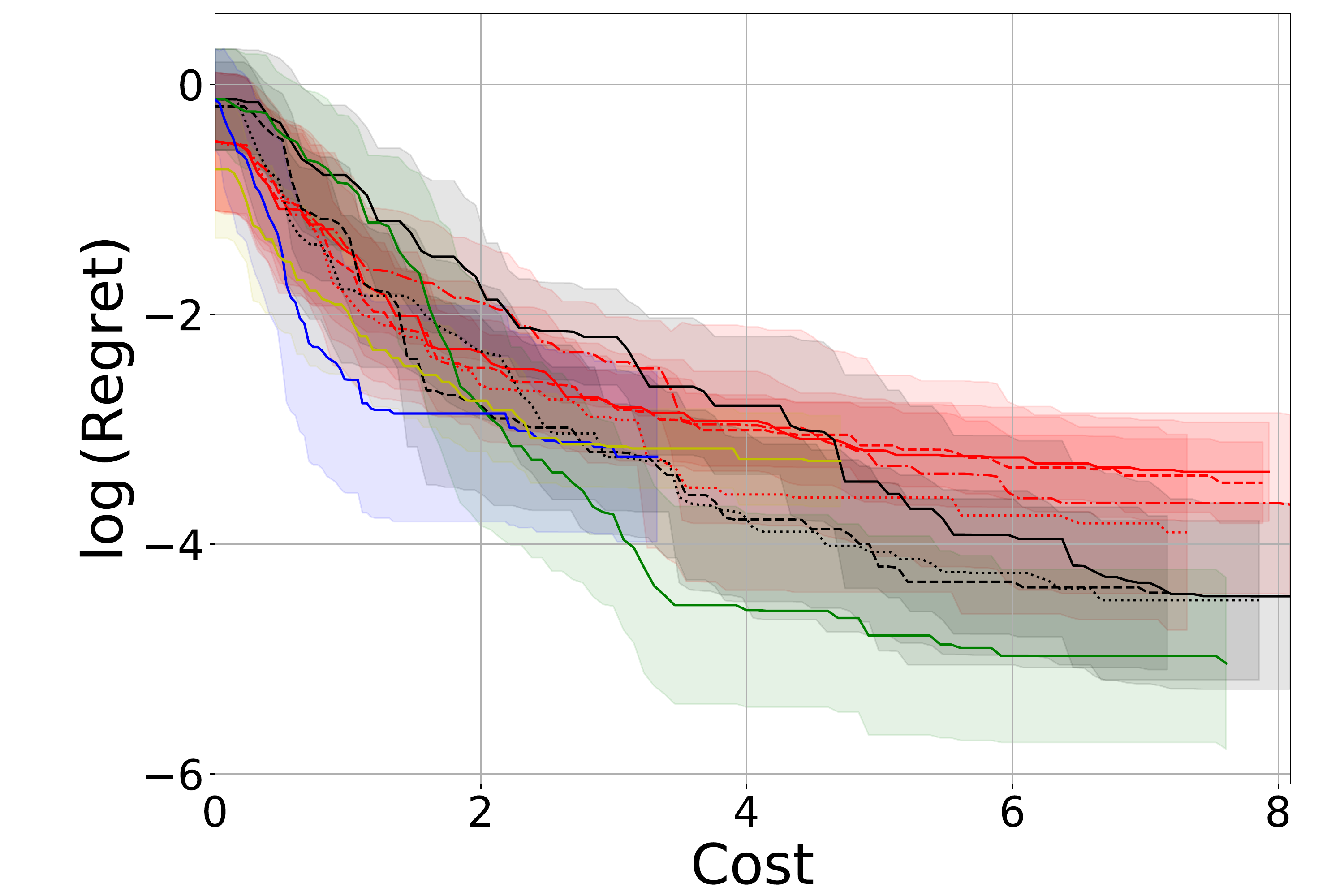}
	\includegraphics[width=0.32\textwidth]{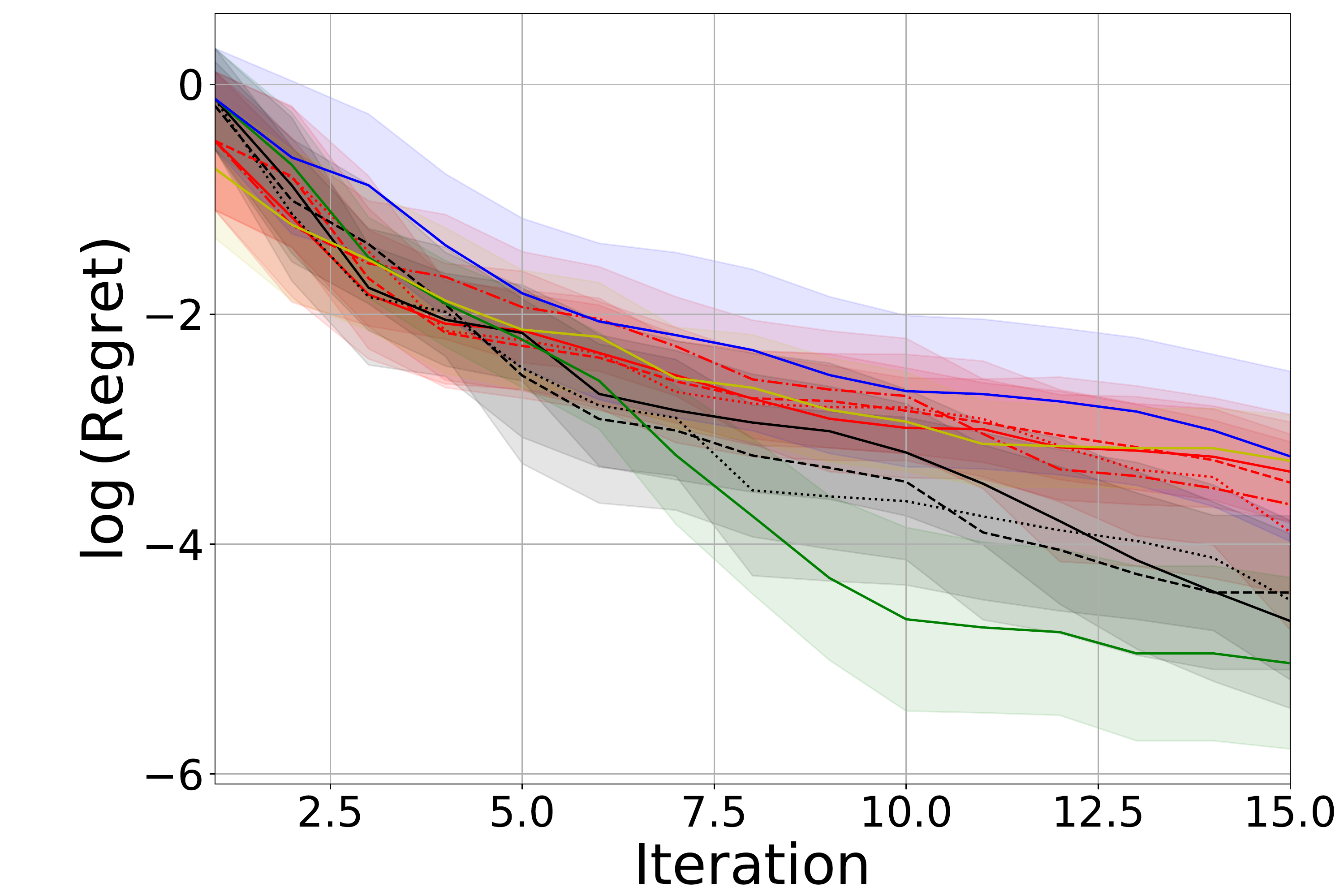}
	\includegraphics[width=0.32\textwidth]{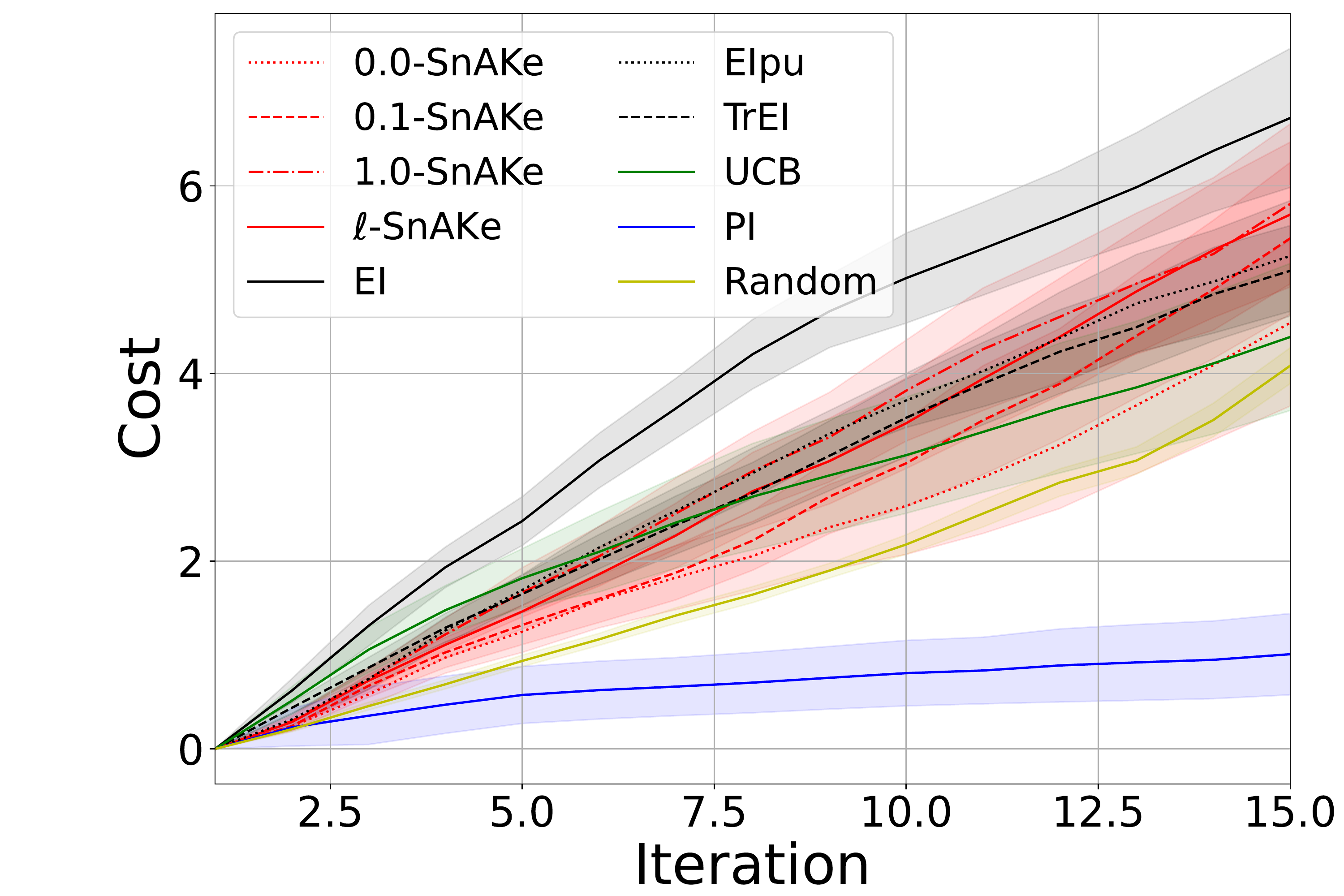}
	\caption{$T = 15$}
	\end{subfigure}
	\begin{subfigure}{\textwidth}
	\centering
	\includegraphics[width=0.32\textwidth]{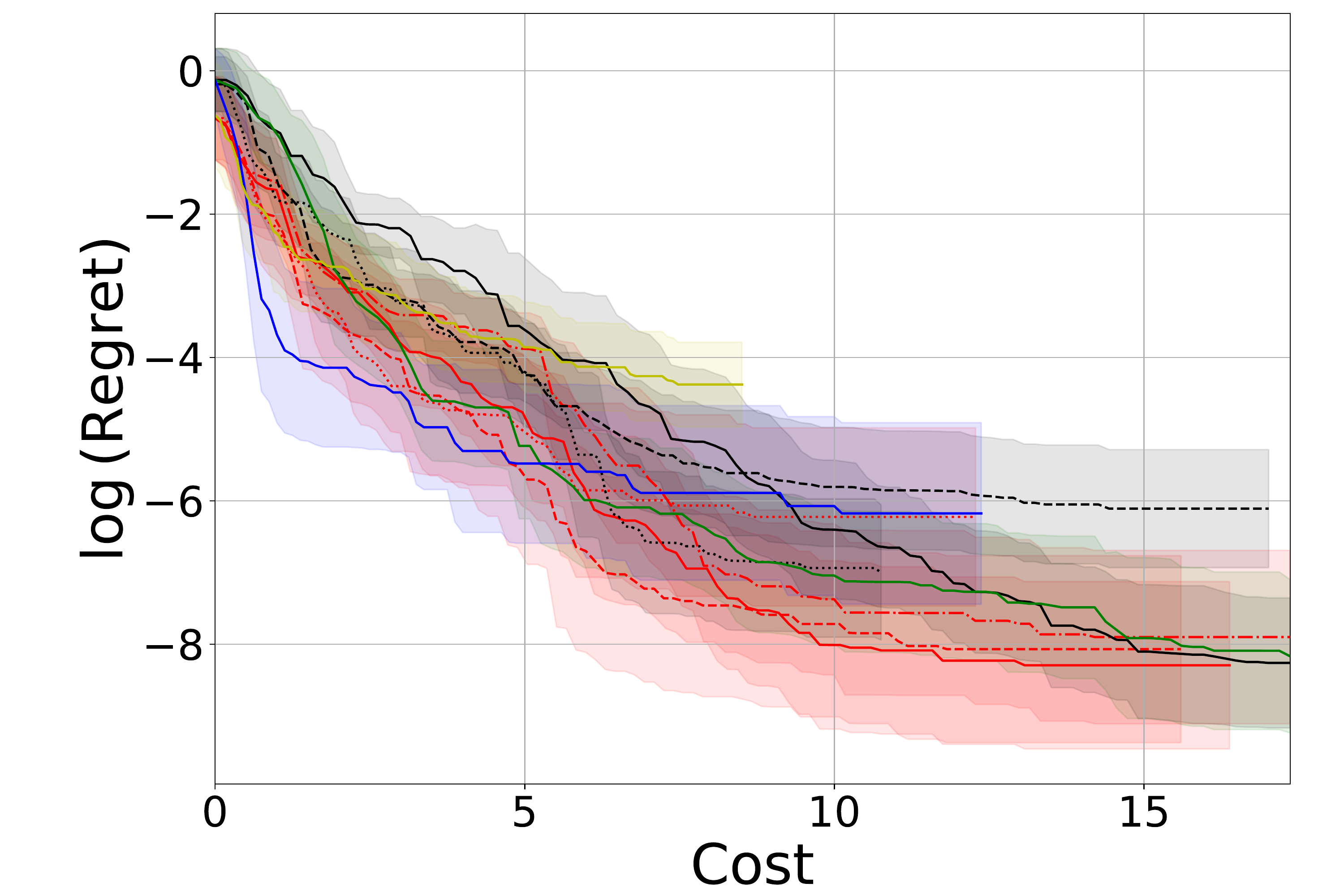}
	\includegraphics[width=0.32\textwidth]{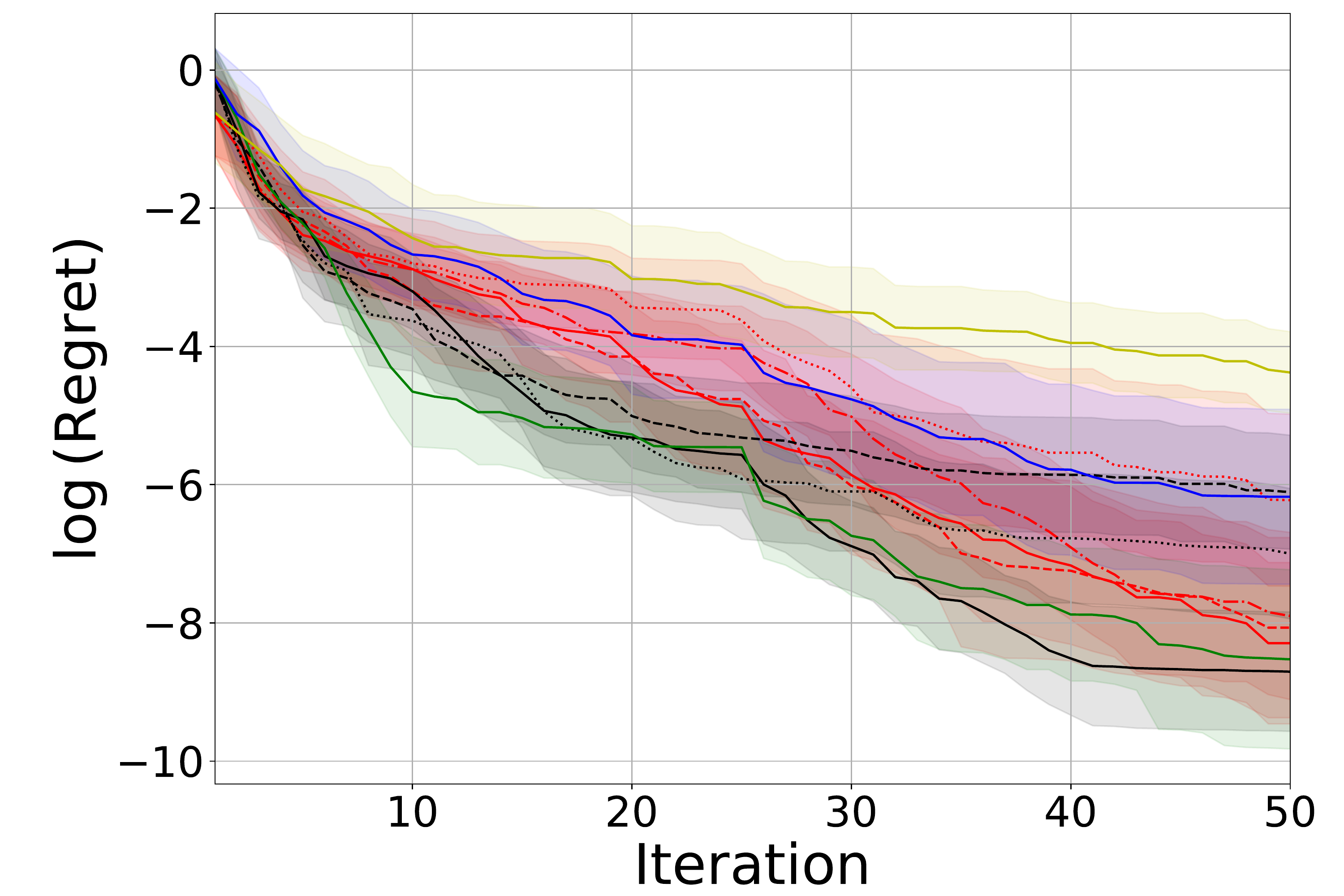}
	\includegraphics[width=0.32\textwidth]{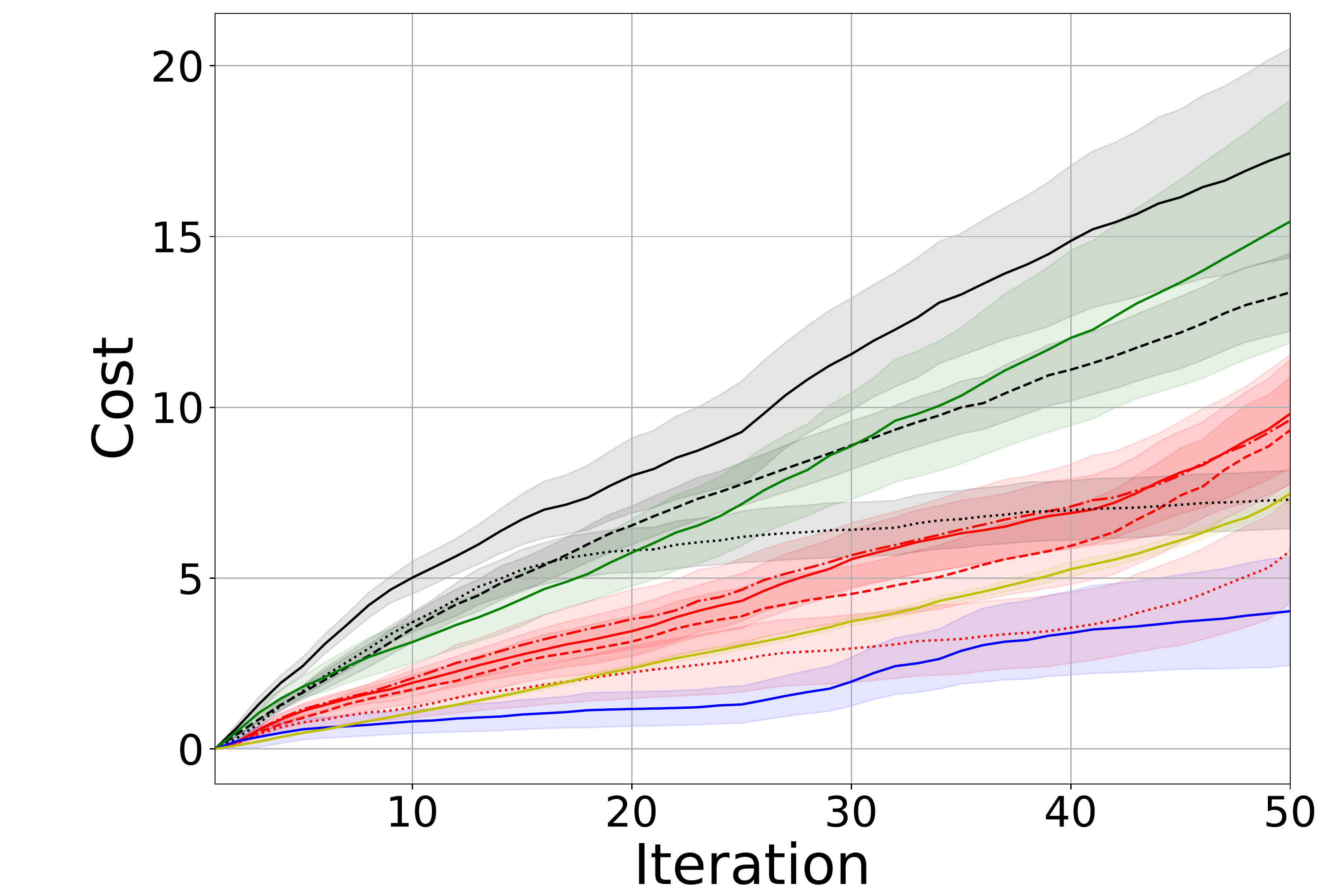}
	\caption{$T = 50$}
	\end{subfigure}
	\begin{subfigure}{\textwidth}
	\centering
	\includegraphics[width=0.32\textwidth]{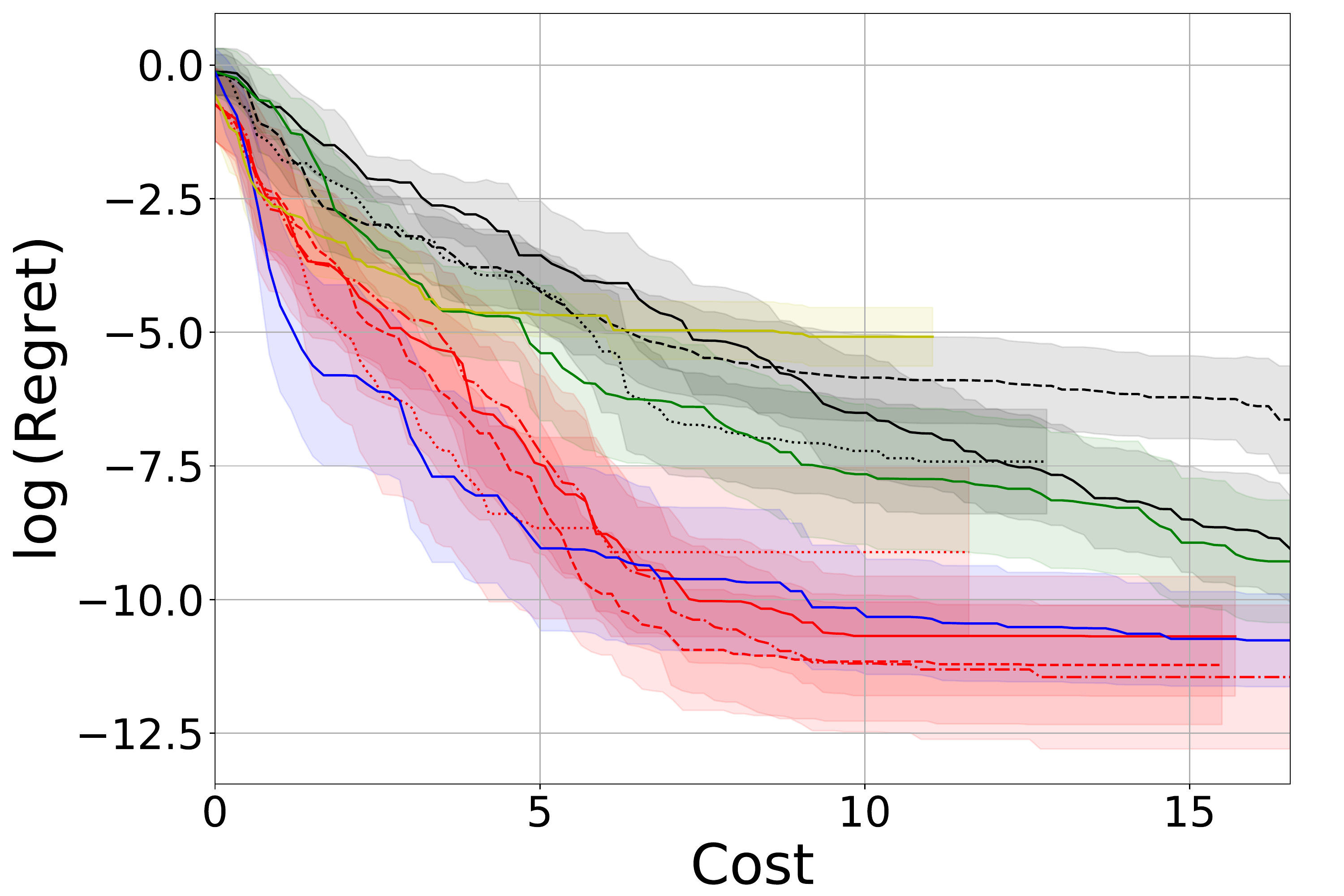}
	\includegraphics[width=0.32\textwidth]{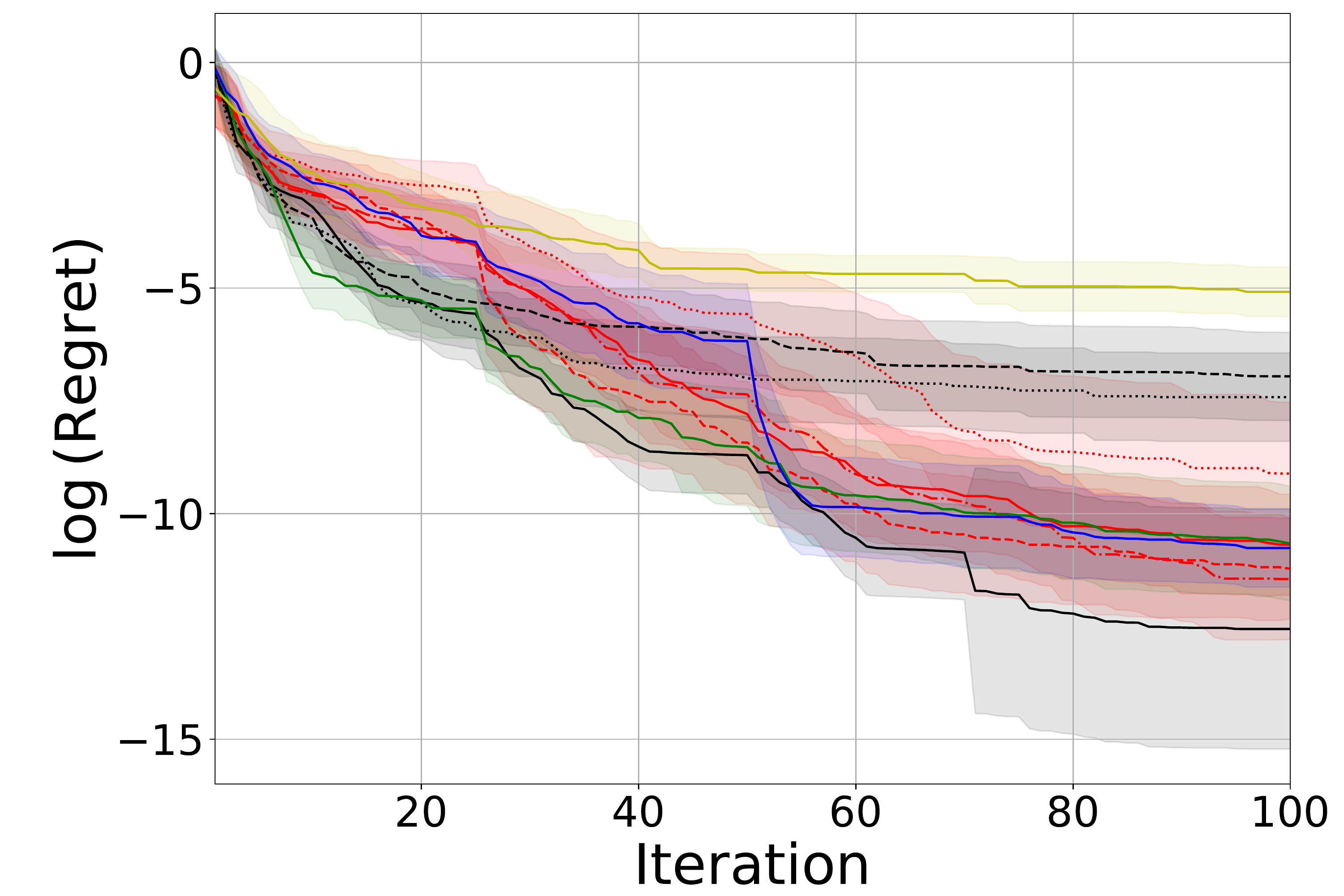}
	\includegraphics[width=0.32\textwidth]{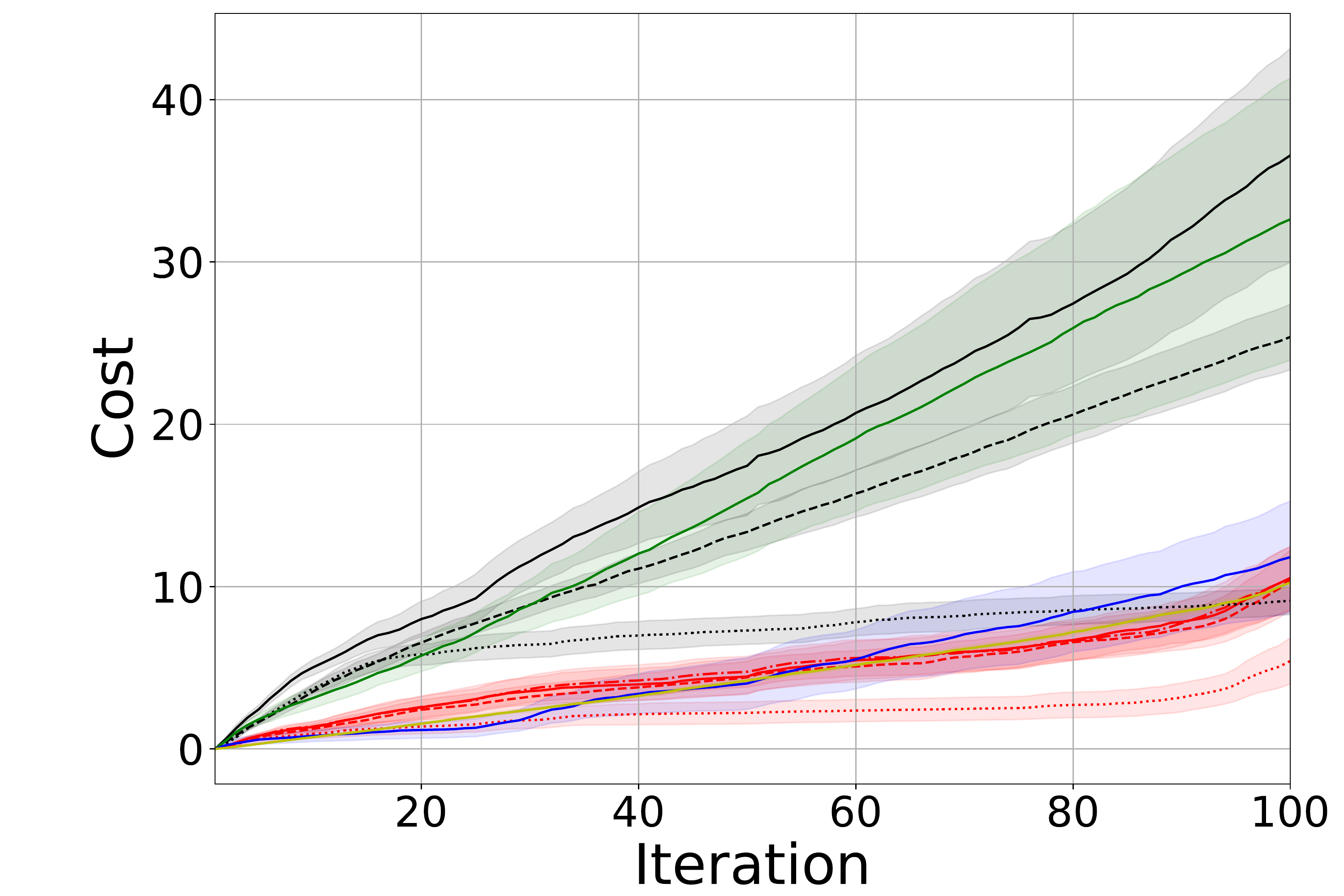}
	\caption{$T = 100$}
	\end{subfigure}
	\begin{subfigure}{\textwidth}
	\centering
	\includegraphics[width=0.32\textwidth]{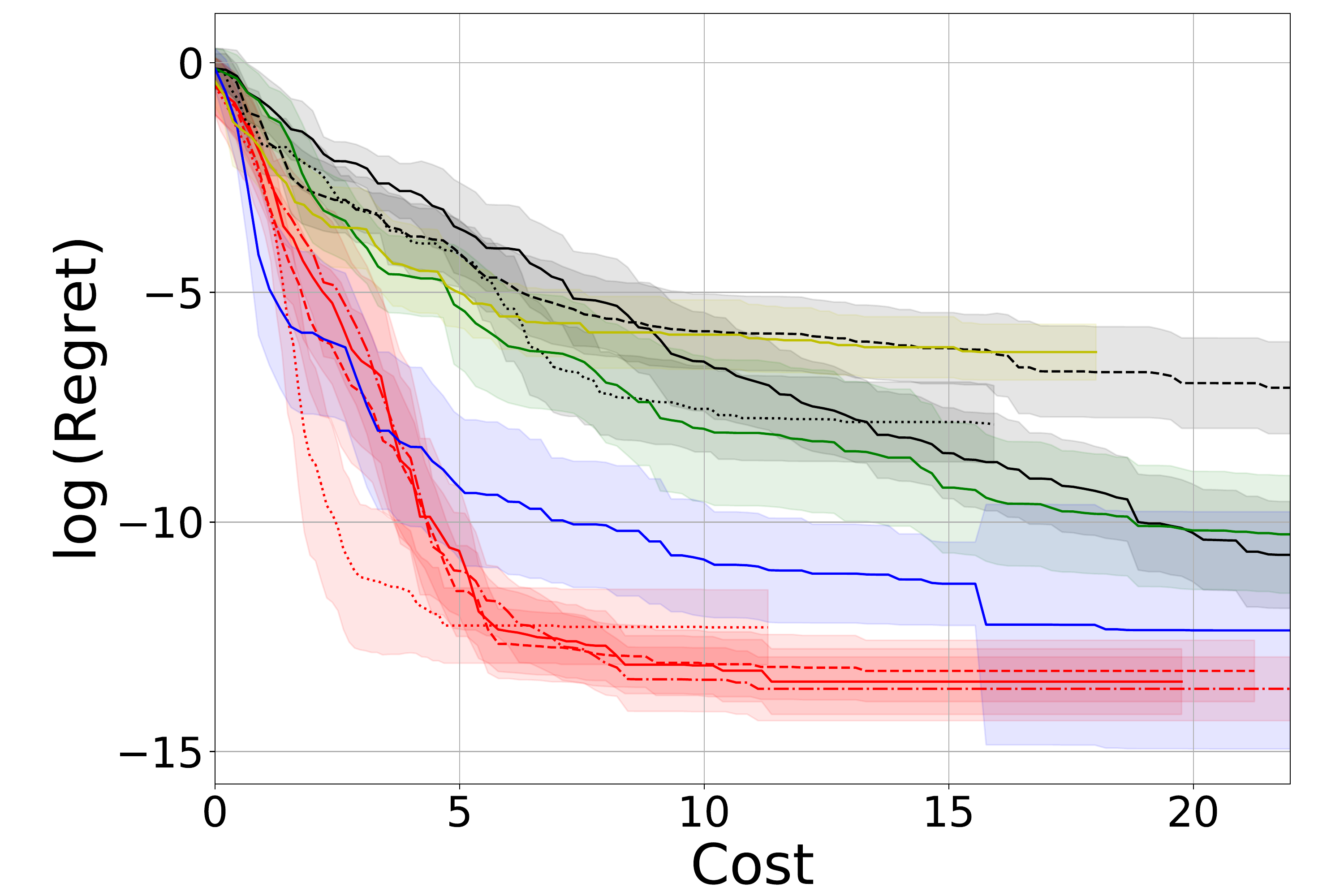}
	\includegraphics[width=0.32\textwidth]{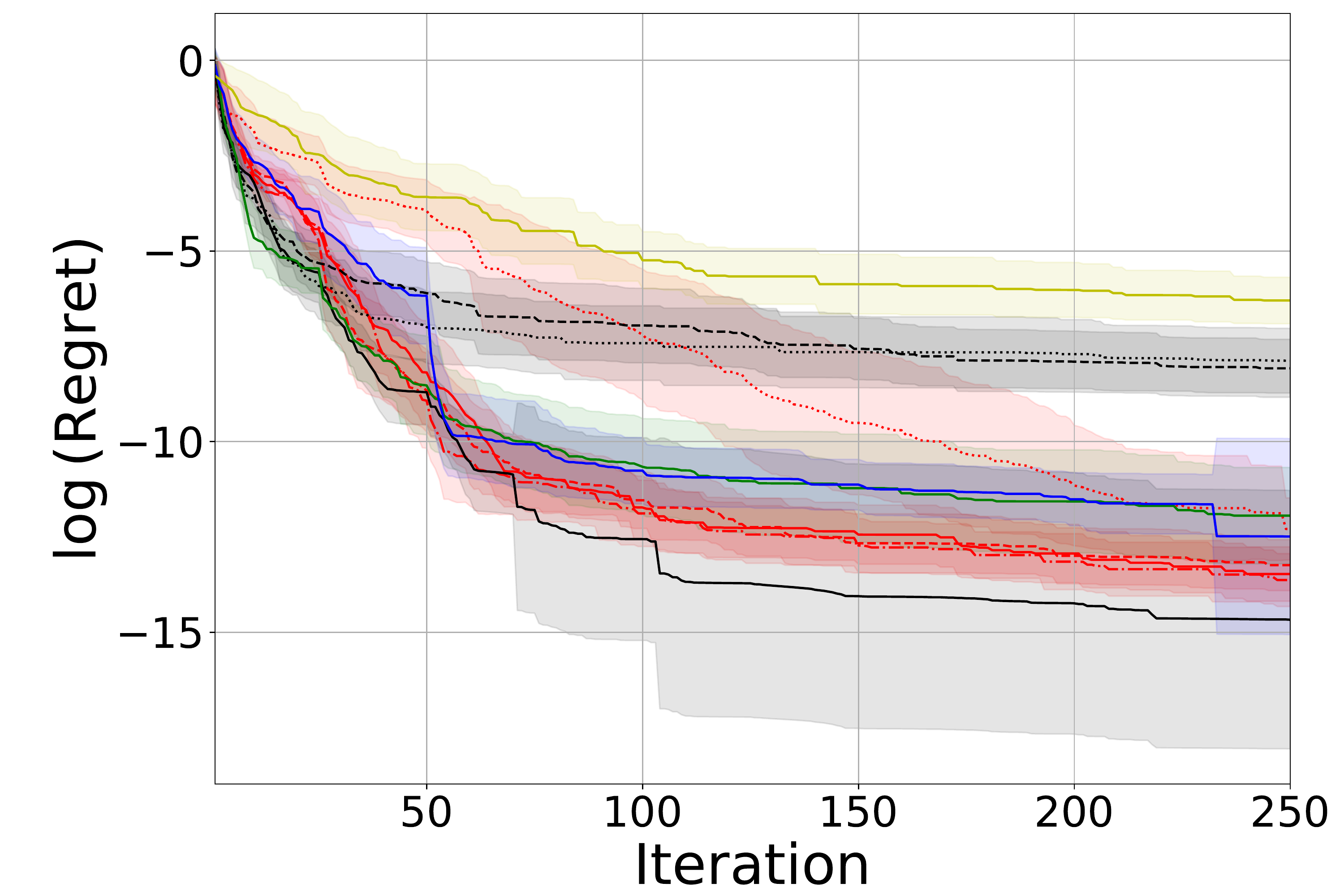}
	\includegraphics[width=0.32\textwidth]{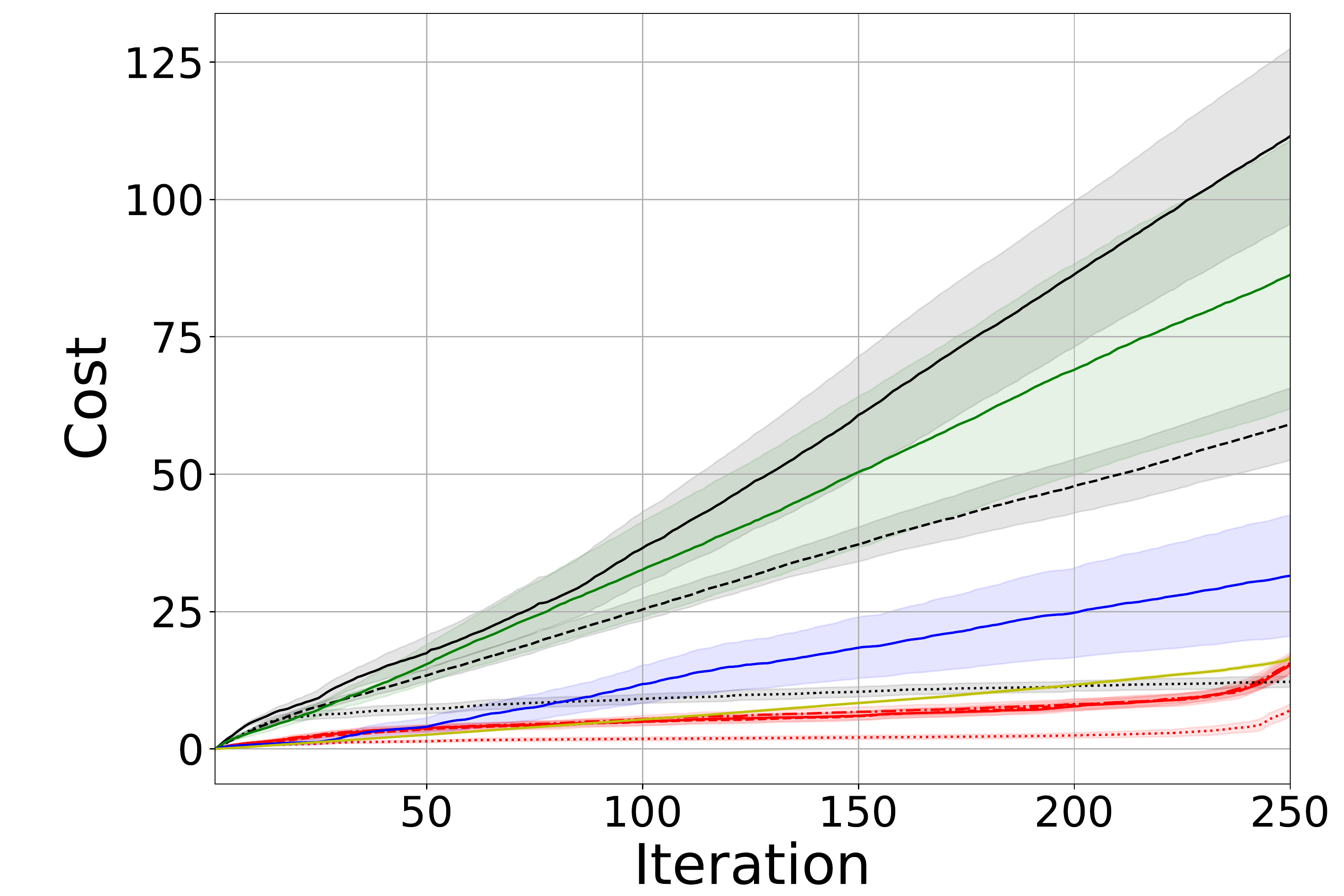}
	\caption{$T = 250$}
	\end{subfigure}
	\caption{Branin2D. Each row represents a different budget. The left column shows the evolution of regret against the cost used. The middle column shows the evolution of regret with iterations, and the right columns show the evolution of the 2-norm cost. As we increase the budget, SnAKe outperforms two BO methods in regret, and outperforms all methods in cost. $\epsilon = 0$ gives the smallest cost of all at the expense of some regret.}
	\label{fig: branin_2d_exp_1}
\end{figure}

\begin{figure}[ht]
	\centering
	\begin{subfigure}{\textwidth}
	\centering
	\includegraphics[width = 0.32\textwidth]{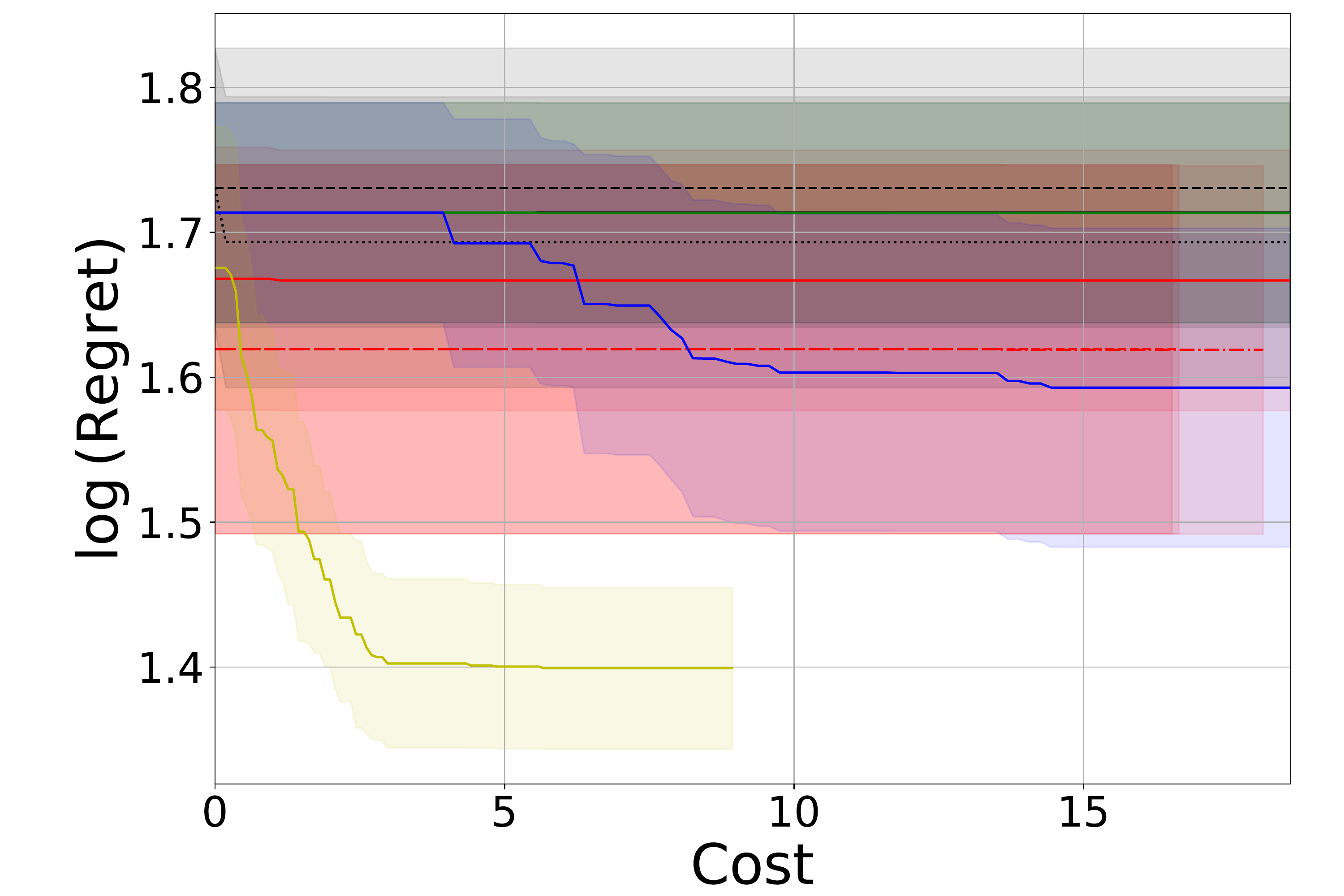}
	\includegraphics[width=0.32\textwidth]{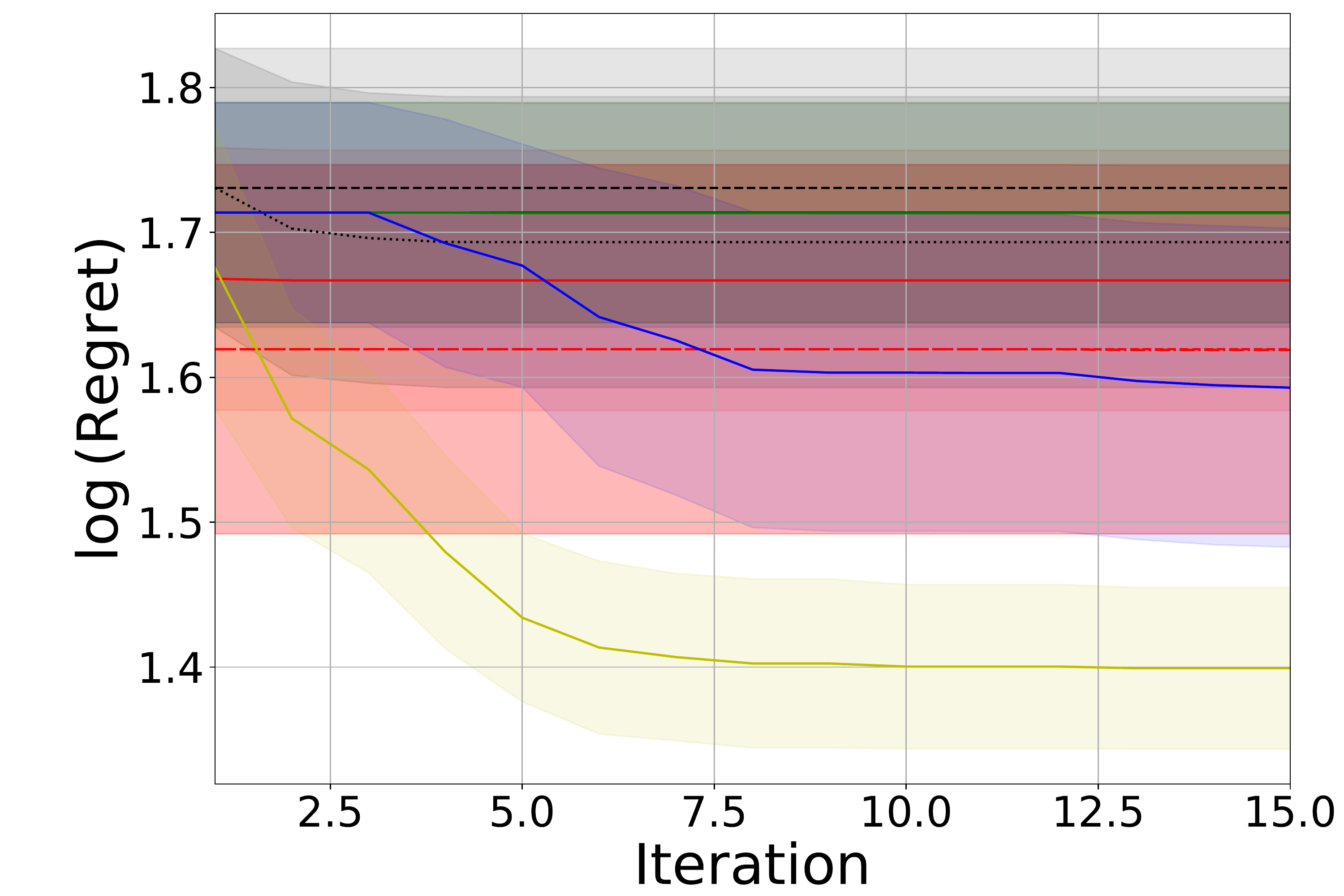}
	\includegraphics[width=0.32\textwidth]{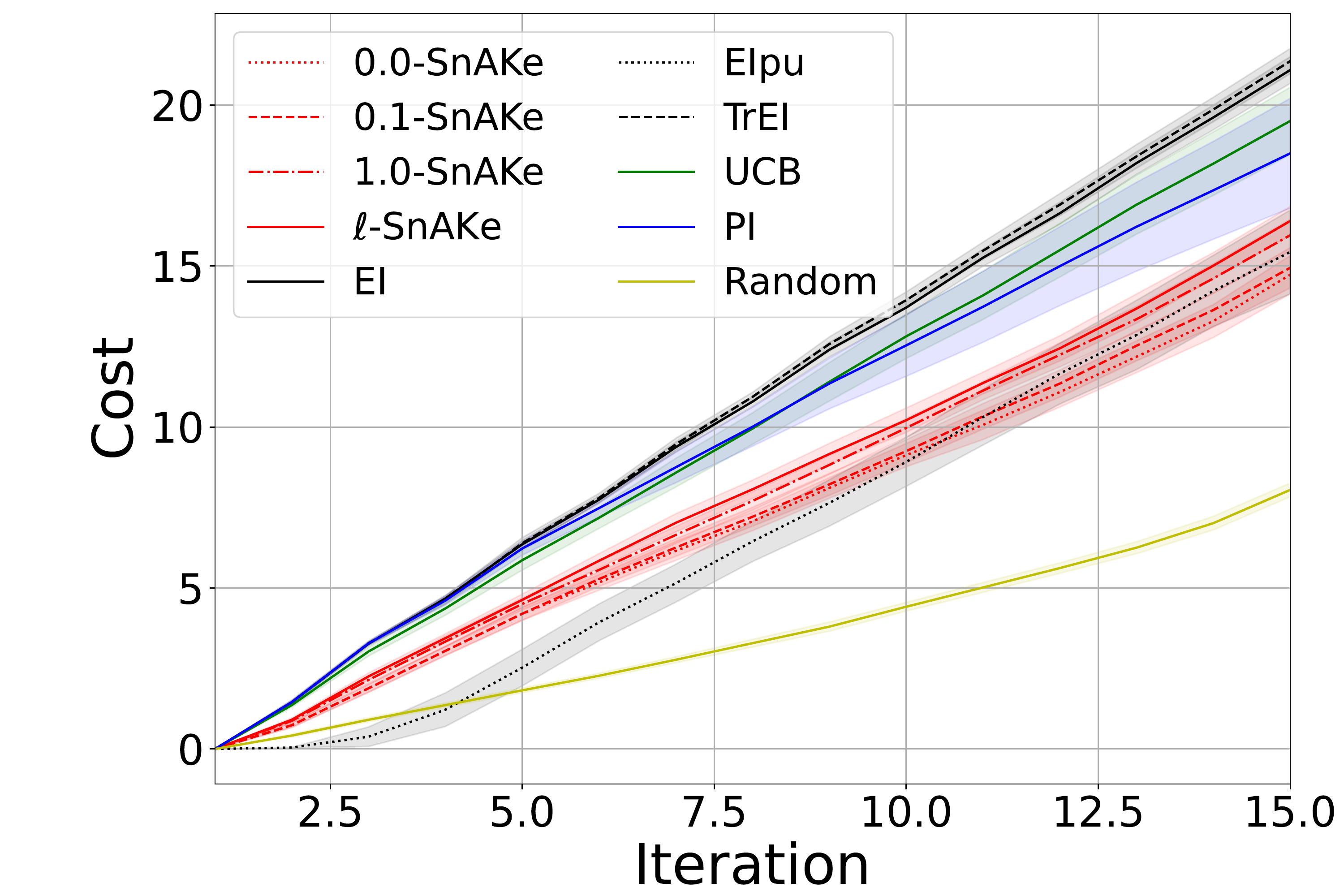}
	\caption{$T = 15$}
	\end{subfigure}
	\begin{subfigure}{\textwidth}
	\centering
	\includegraphics[width=0.32\textwidth]{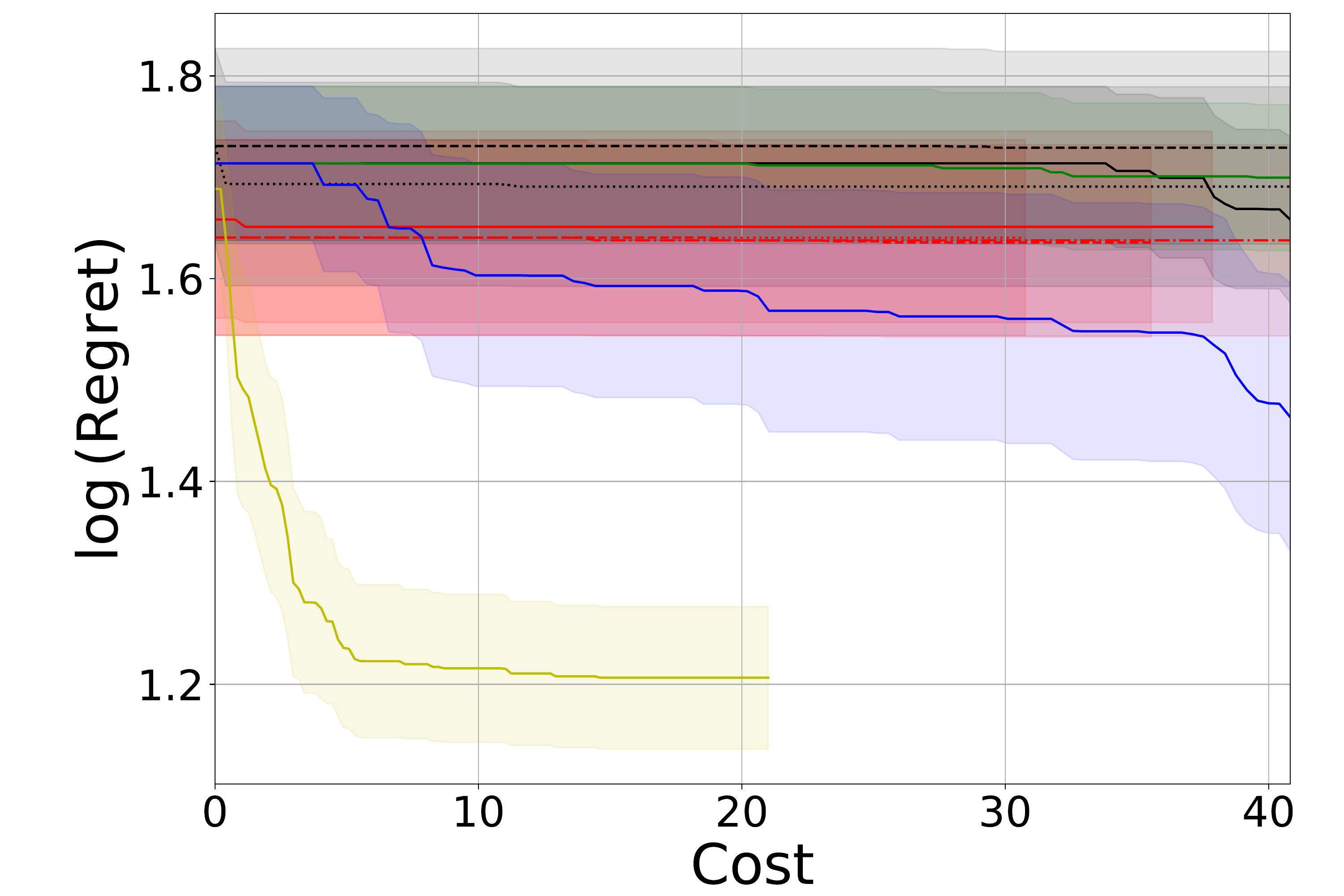}
	\includegraphics[width=0.32\textwidth]{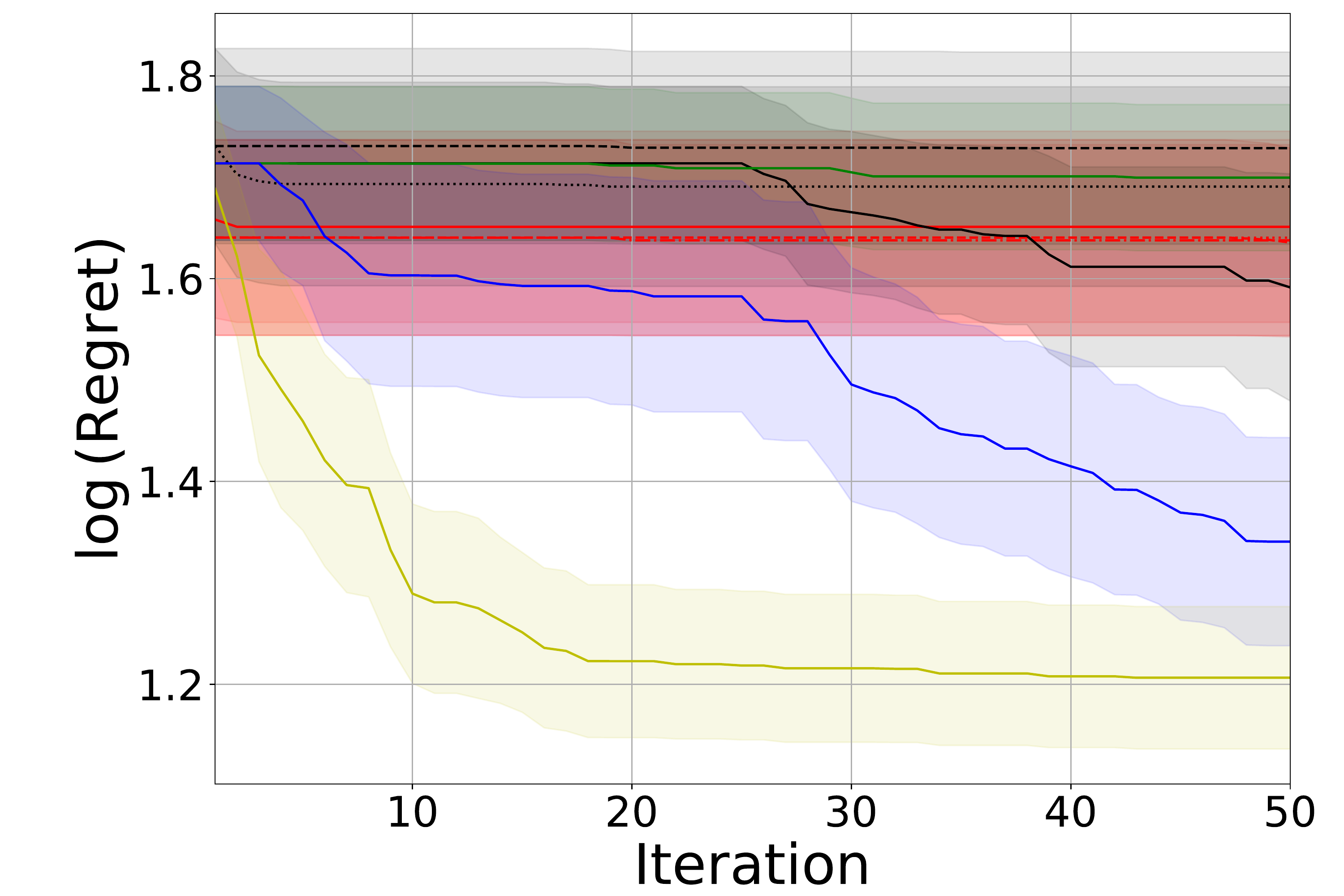}
	\includegraphics[width=0.32\textwidth]{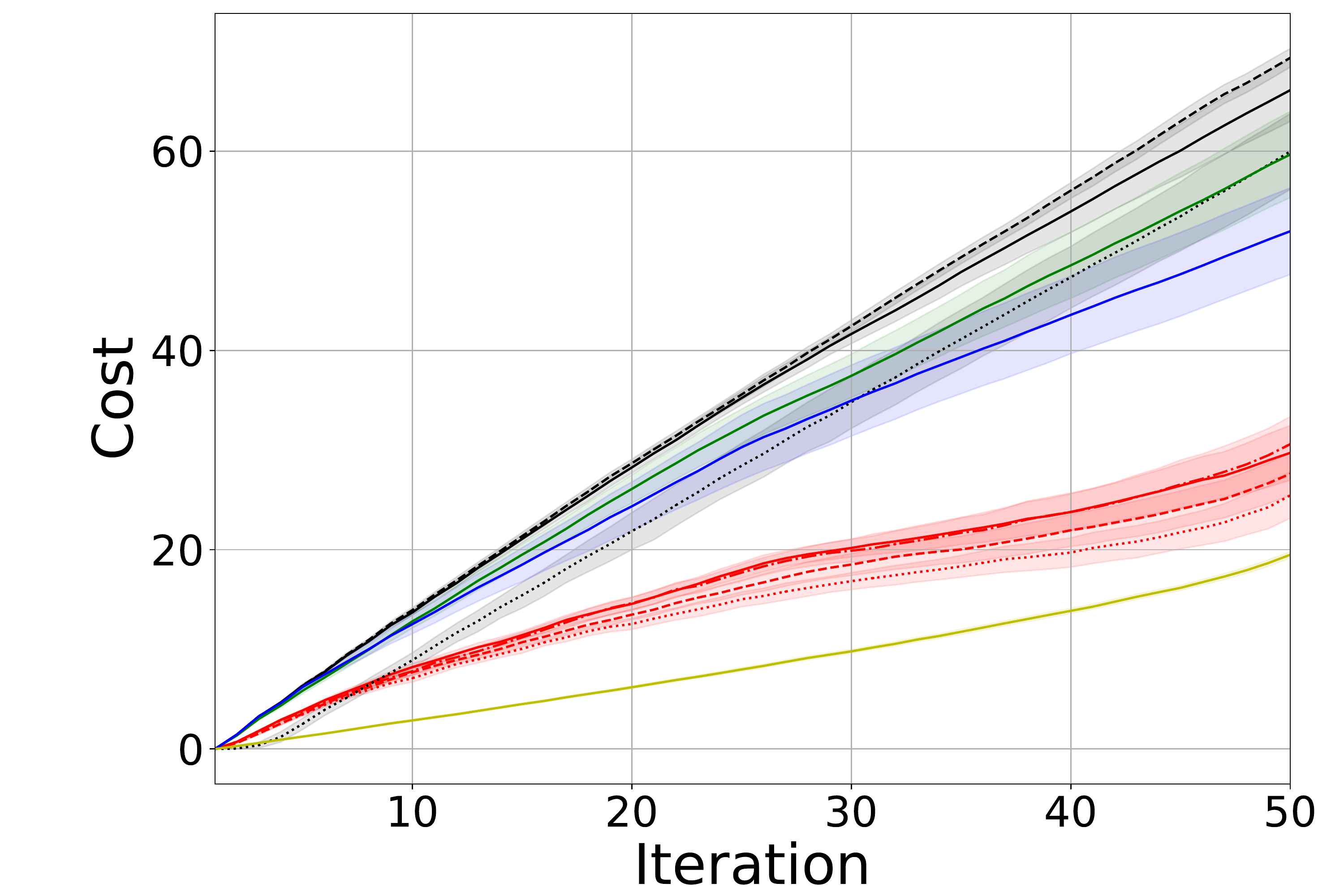}
	\caption{$T = 50$}
	\end{subfigure}
	\begin{subfigure}{\textwidth}
	\centering
	\includegraphics[width=0.32\textwidth]{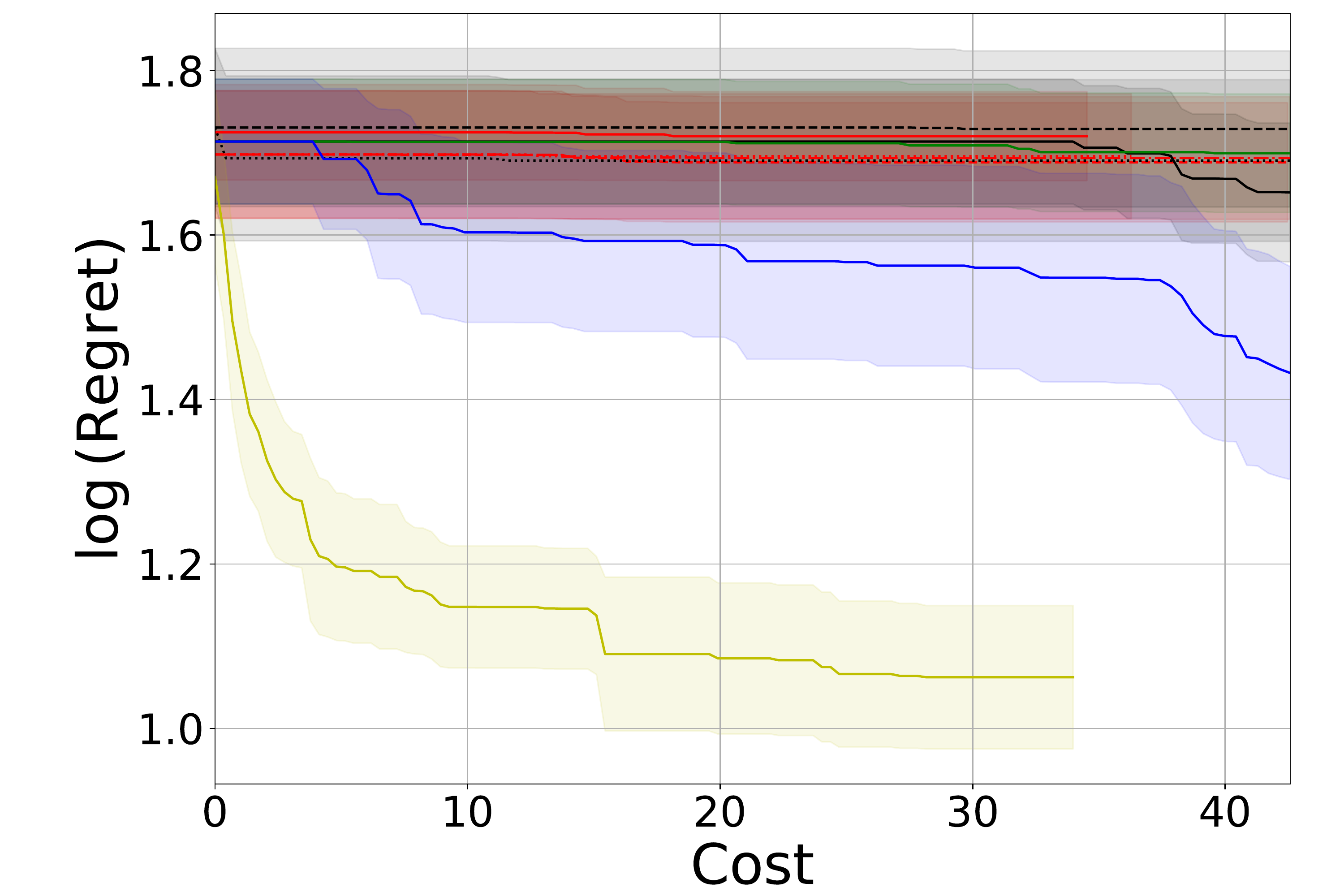}
	\includegraphics[width=0.32\textwidth]{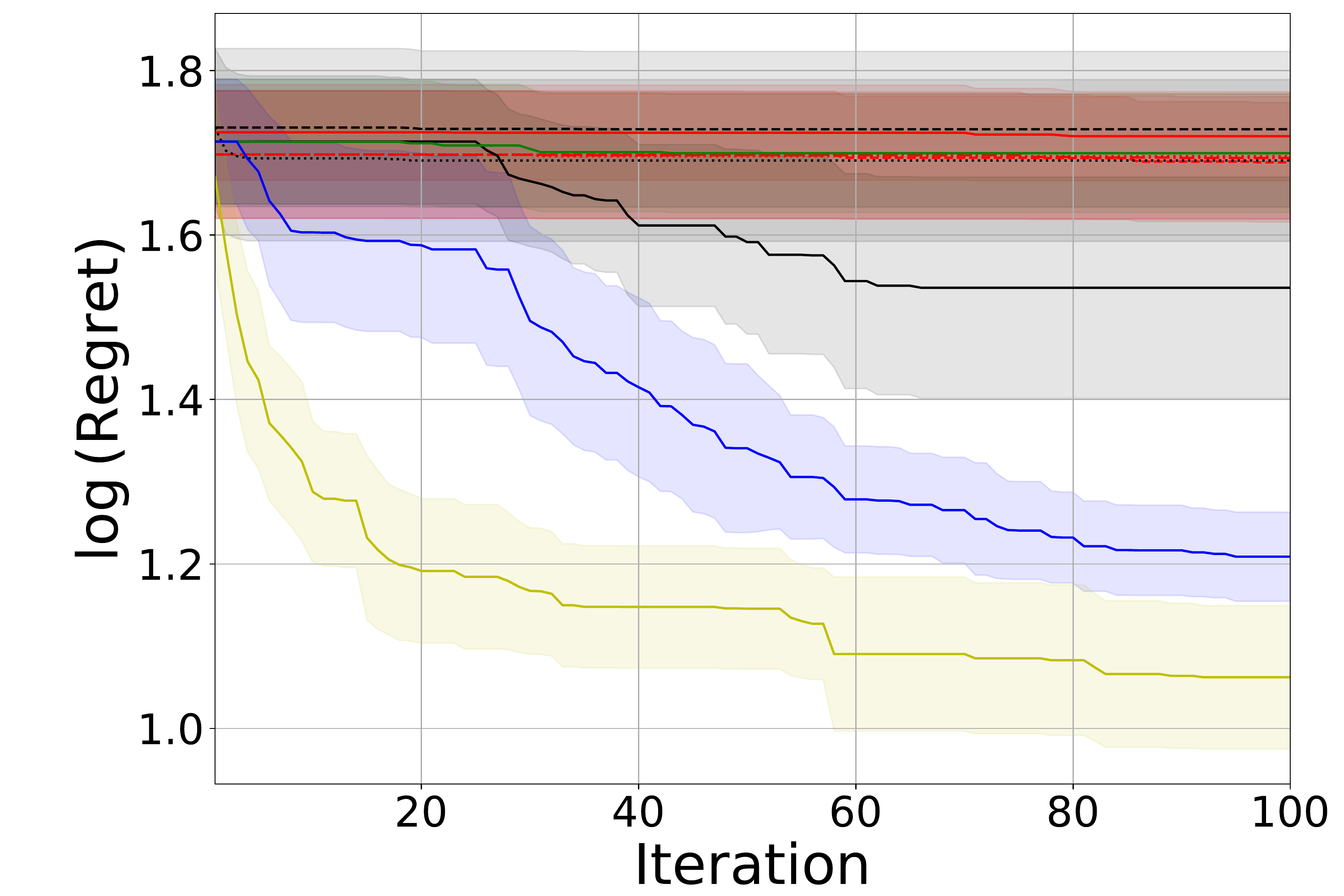}
	\includegraphics[width=0.32\textwidth]{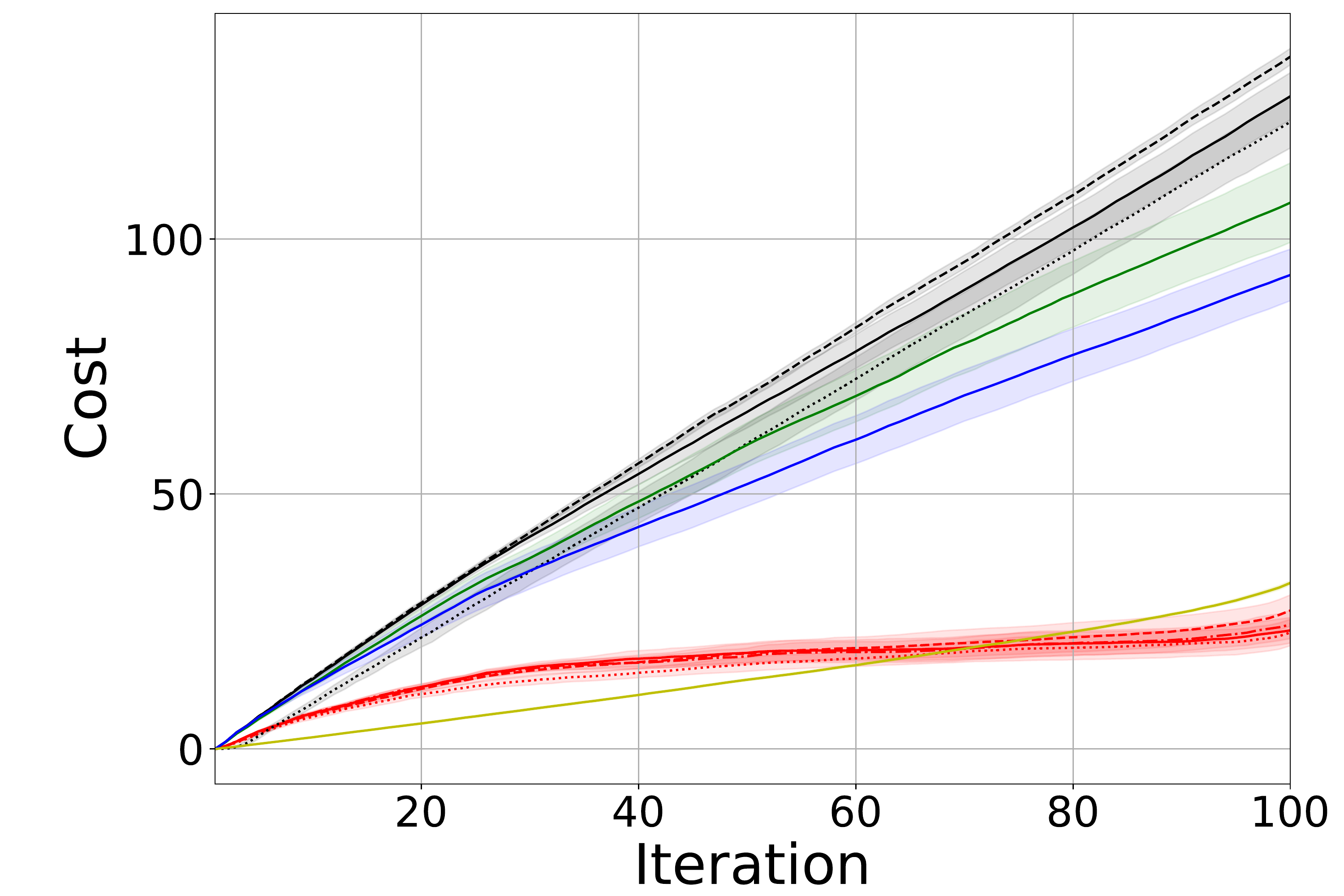}
	\caption{$T = 100$}
	\end{subfigure}
	\begin{subfigure}{\textwidth}
	\centering
	\includegraphics[width=0.32\textwidth]{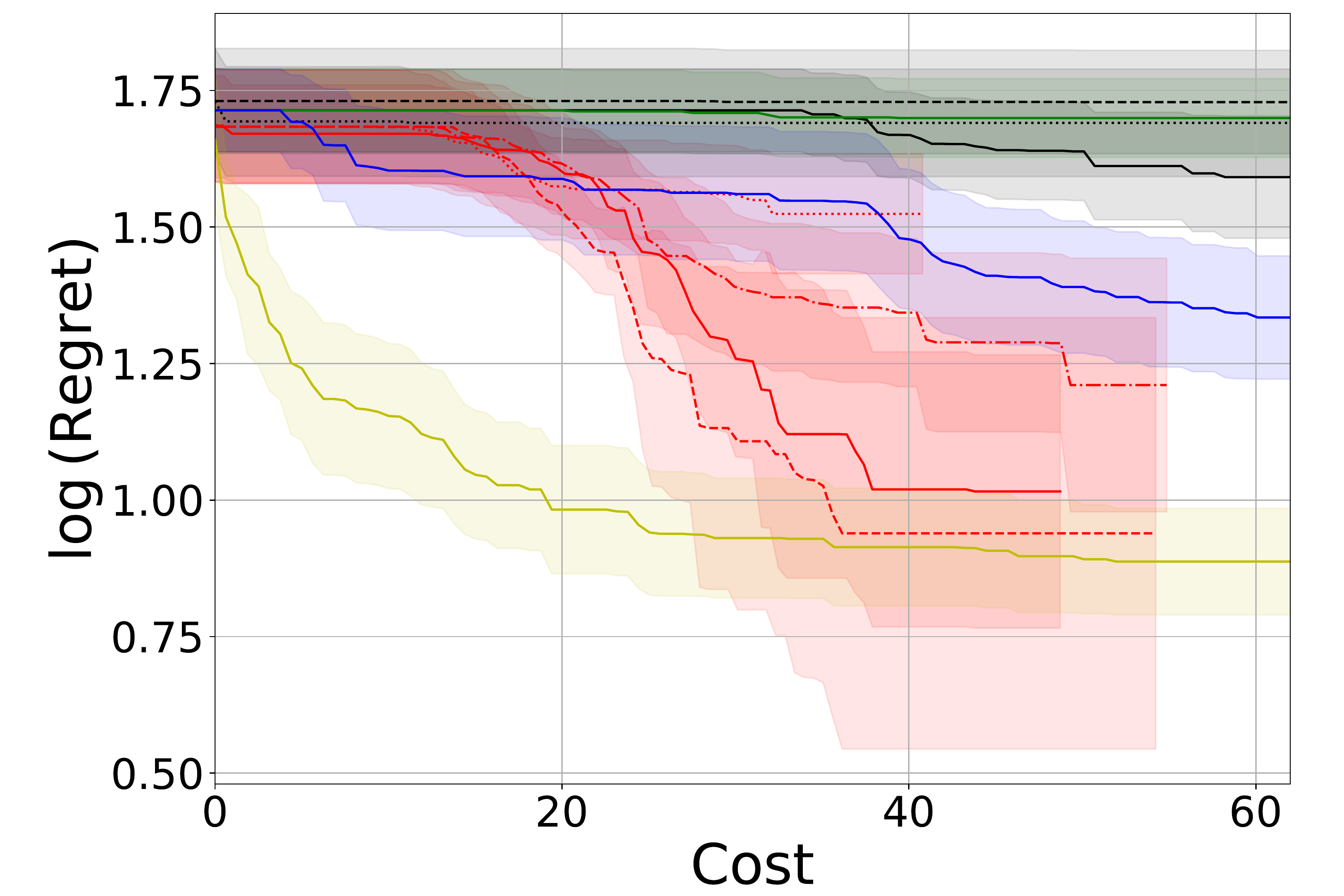}
	\includegraphics[width=0.32\textwidth]{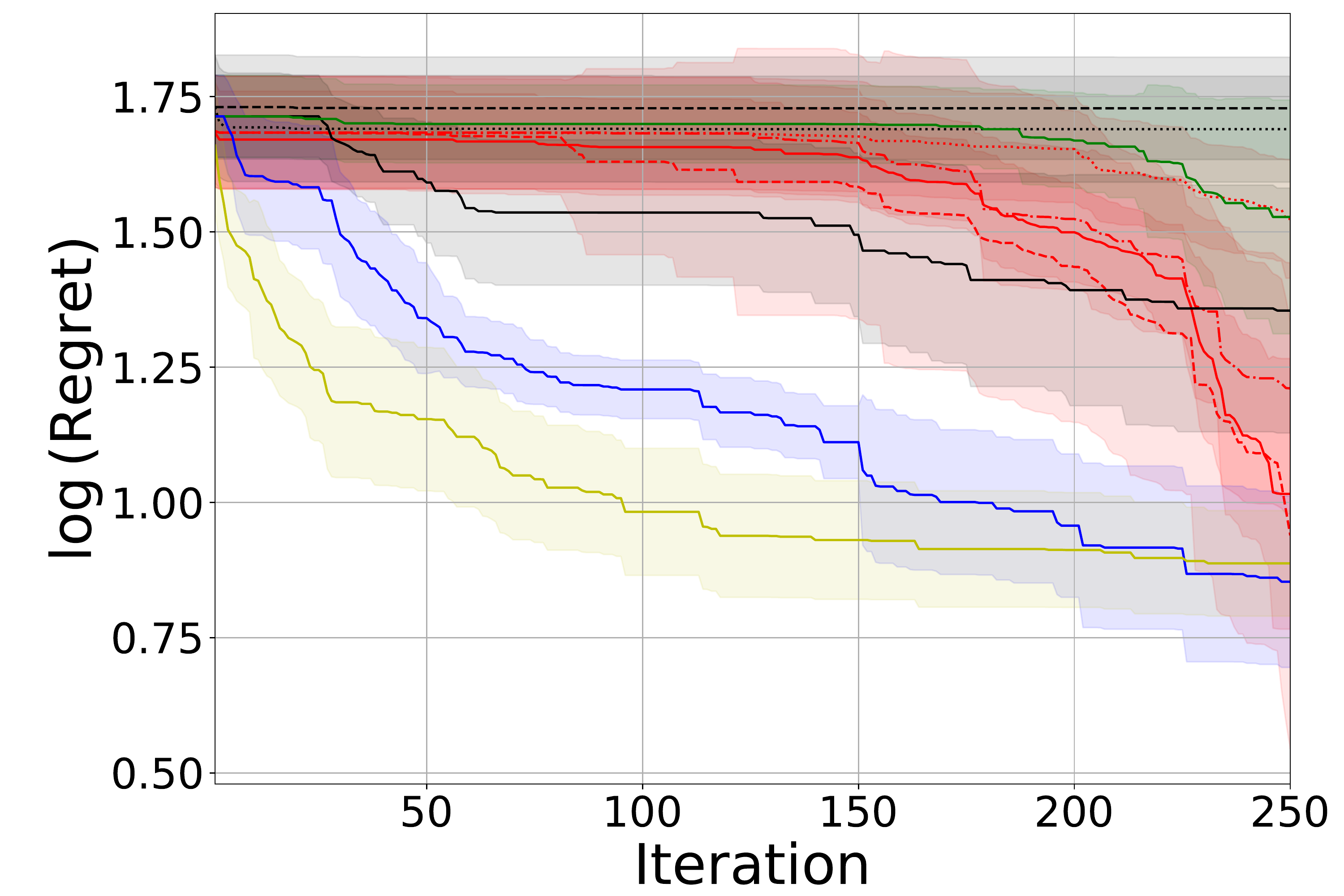}
	\includegraphics[width=0.32\textwidth]{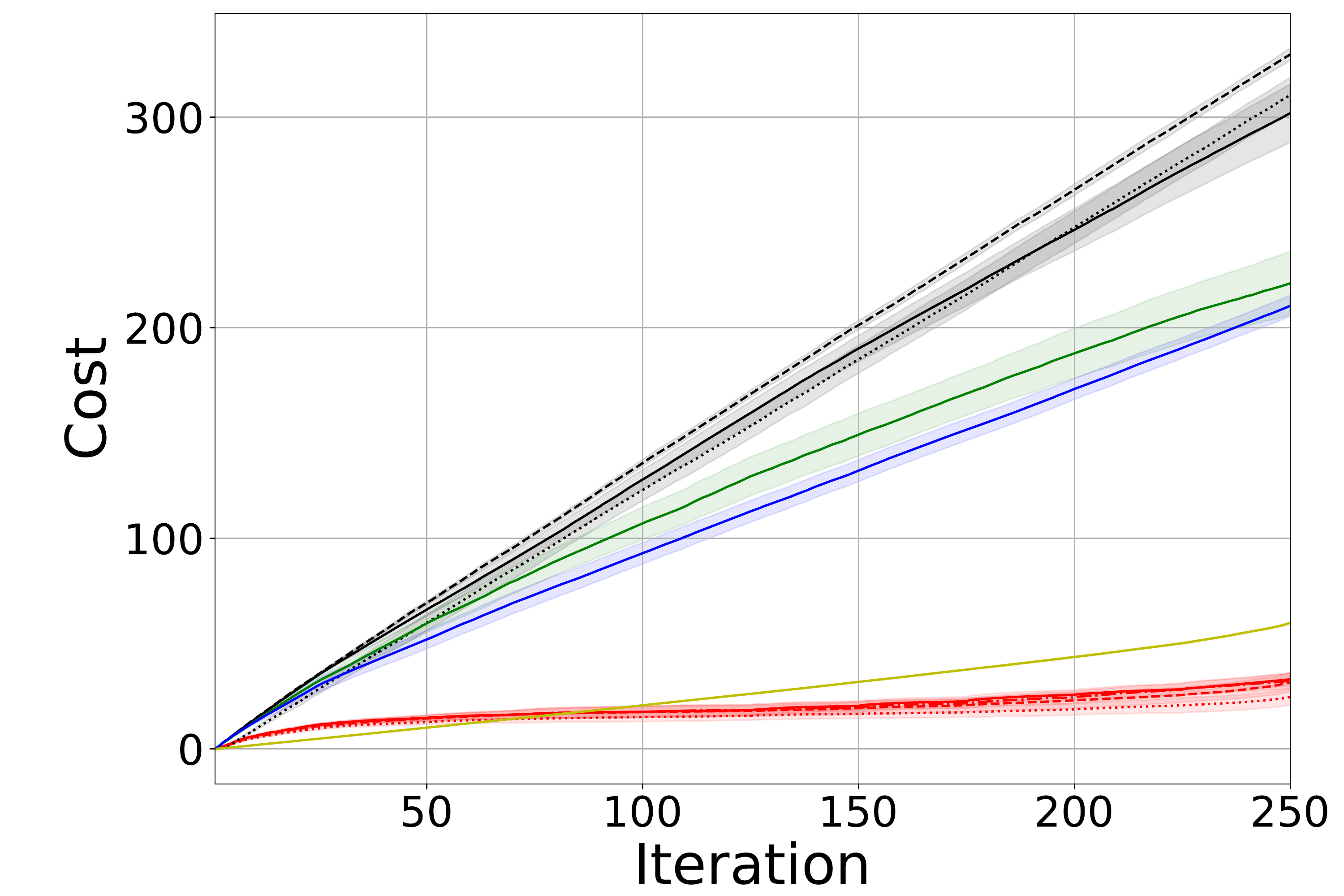}
	\caption{$T = 250$}
	\end{subfigure}
	\caption{Ackley4D. Each row represents a different budget. The left column shows the evolution of regret against the cost used. The middle column shows the evolution of regret with iterations, and the right columns show the evolution of the 2-norm cost. SnAKe performs badly for smaller budgets, this may be because of the Thompson Sampling (see Figure \ref{fig: ackley_async_4d_250}, TS performs very badly in asynchronous Ackely). For the largest budget SnAKe recovers and performs comparably with PI in terms of regret, but achieves low cost.}
	\label{fig: ackley_sync_4d_250}
\end{figure}

\begin{figure}[ht]
	\centering
	\begin{subfigure}{\textwidth}
	\centering
	\includegraphics[width = 0.32\textwidth]{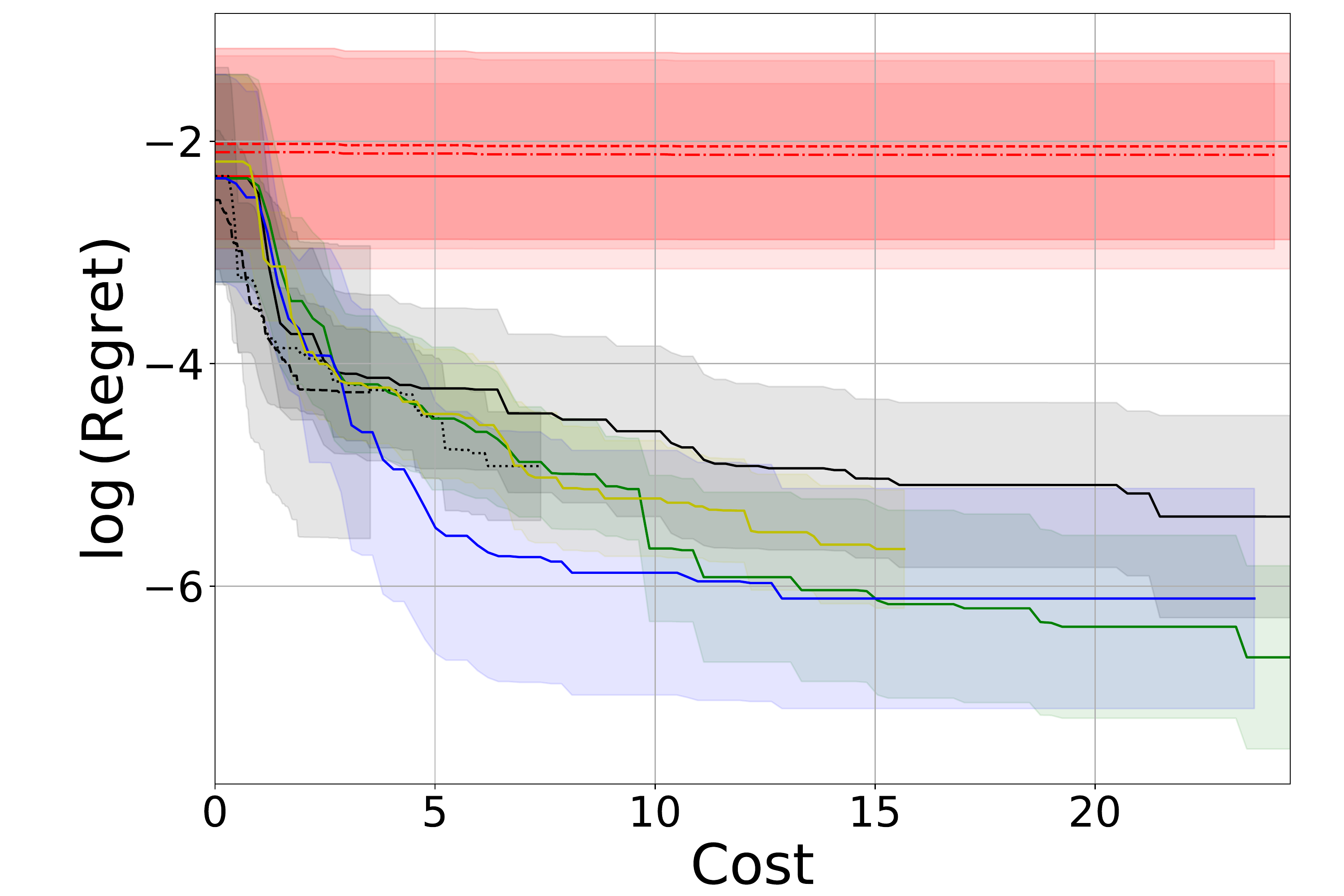}
	\includegraphics[width=0.32\textwidth]{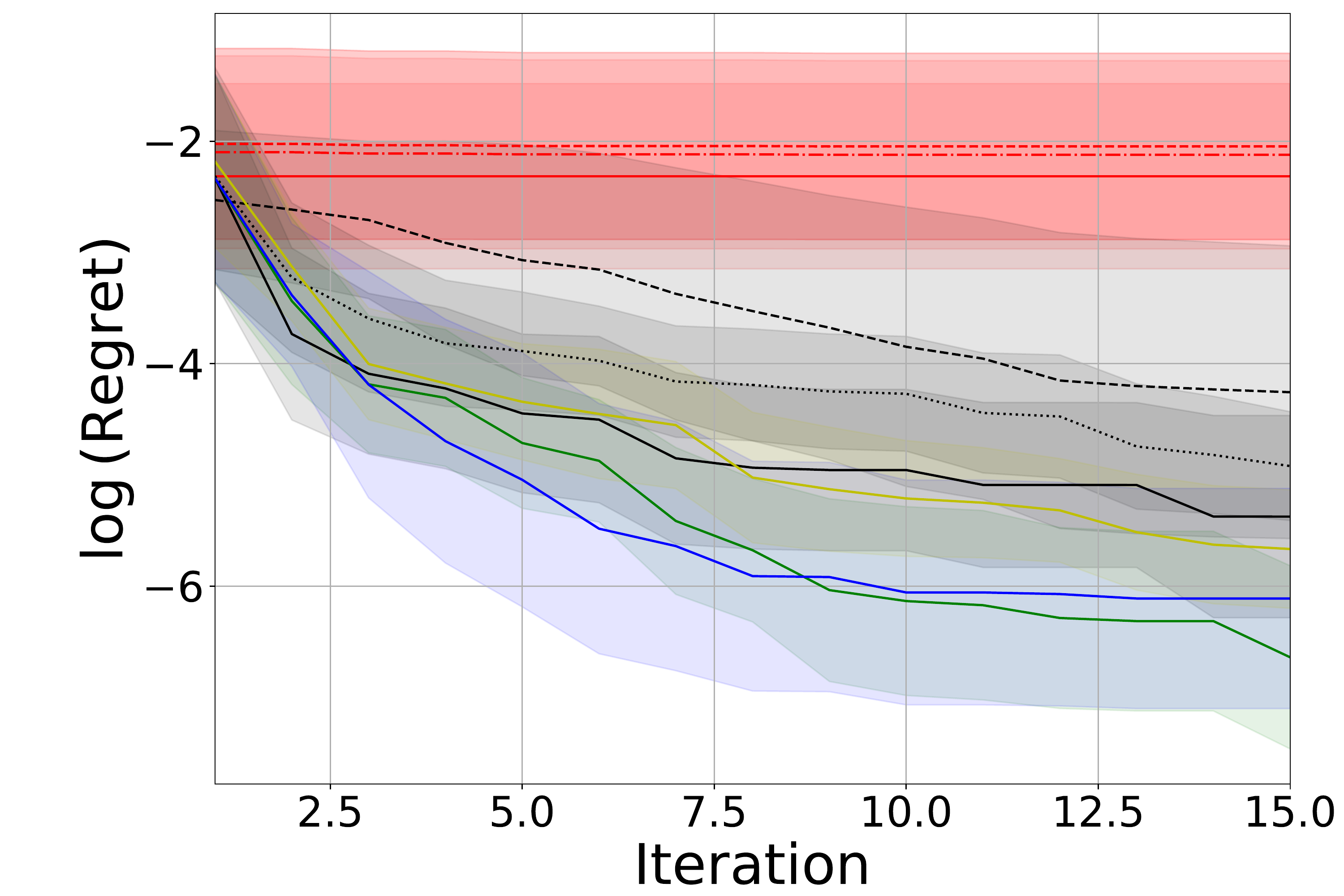}
	\includegraphics[width=0.32\textwidth]{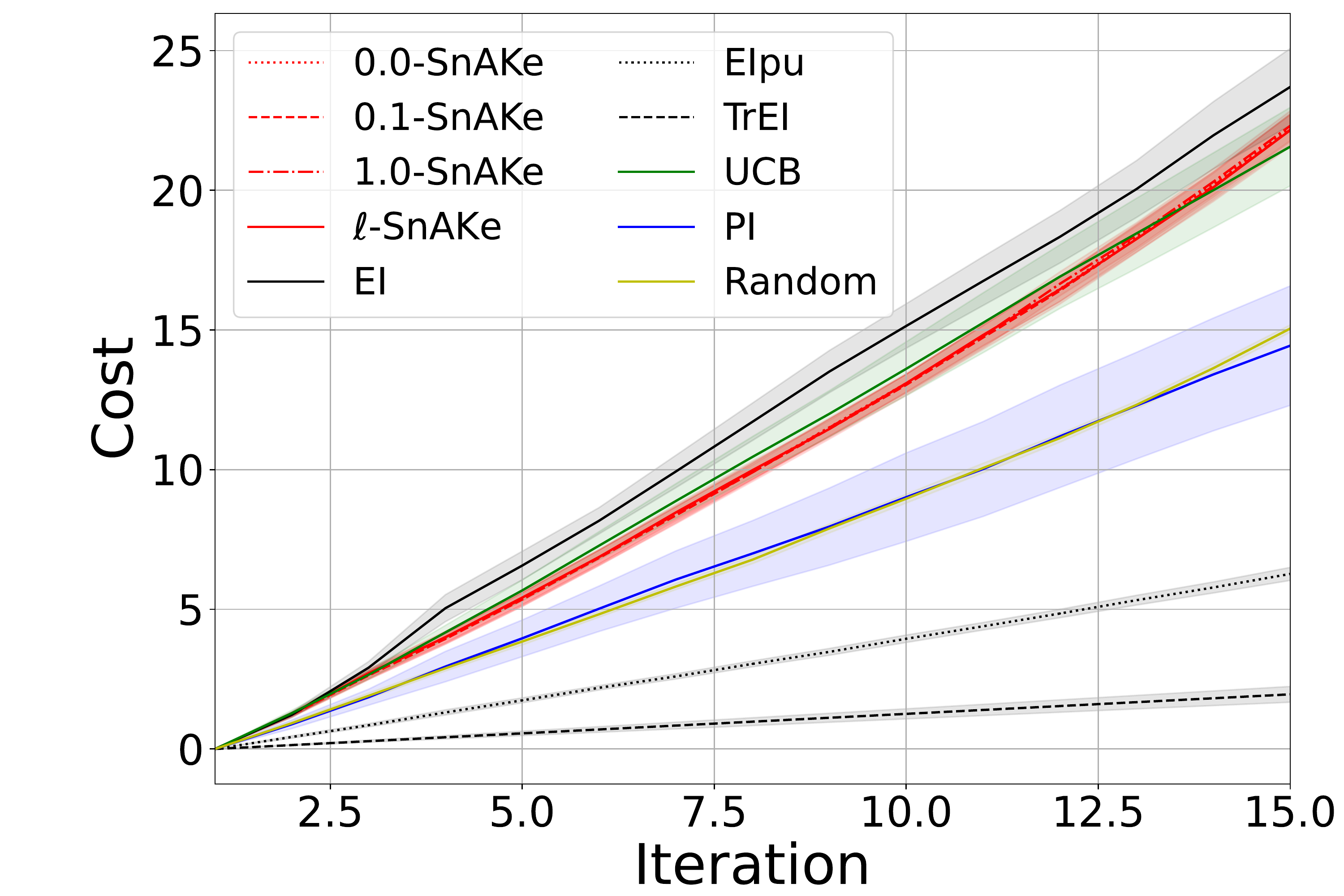}
	\caption{$T = 15$}
	\end{subfigure}
	\begin{subfigure}{\textwidth}
	\centering
	\includegraphics[width=0.32\textwidth]{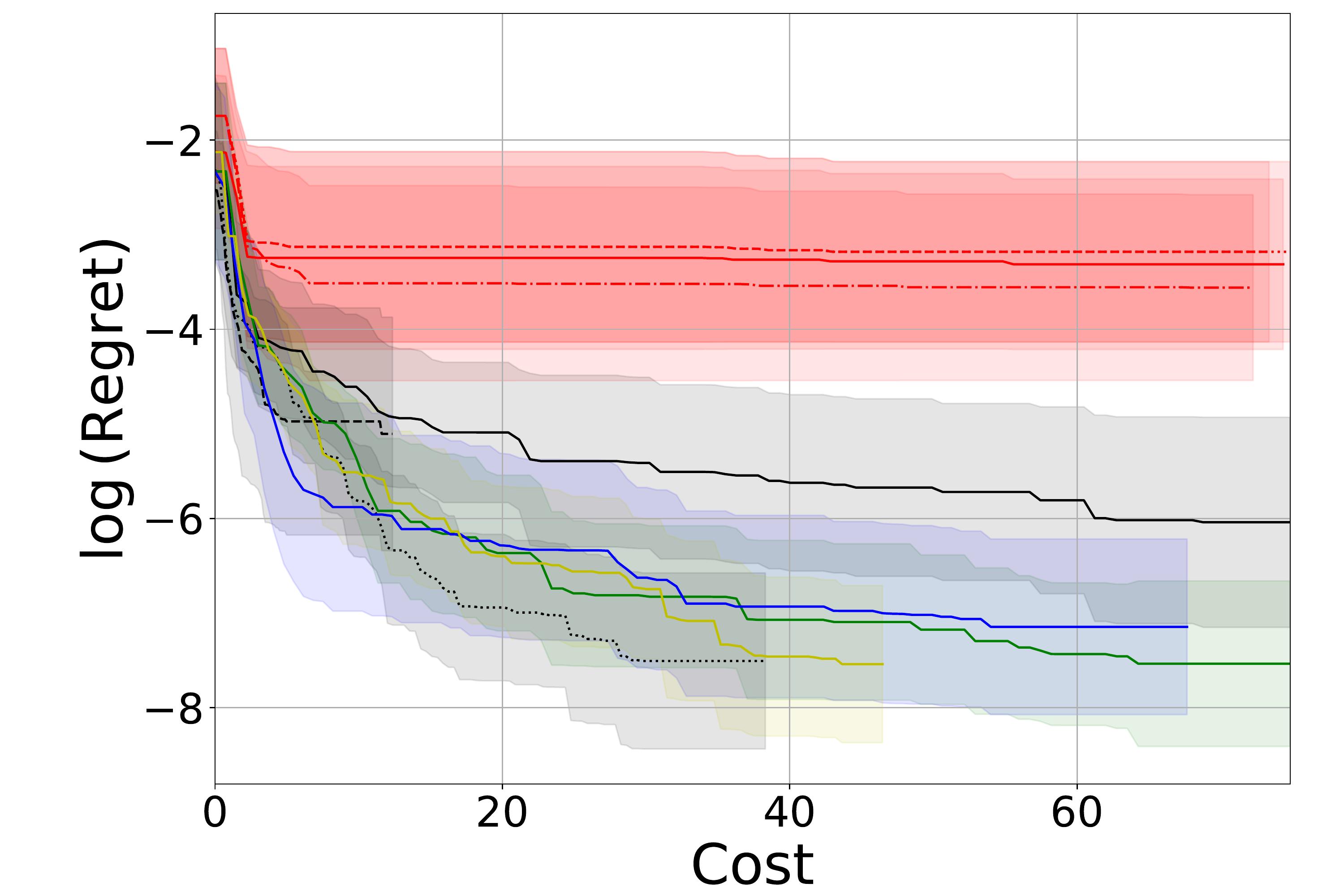}
	\includegraphics[width=0.32\textwidth]{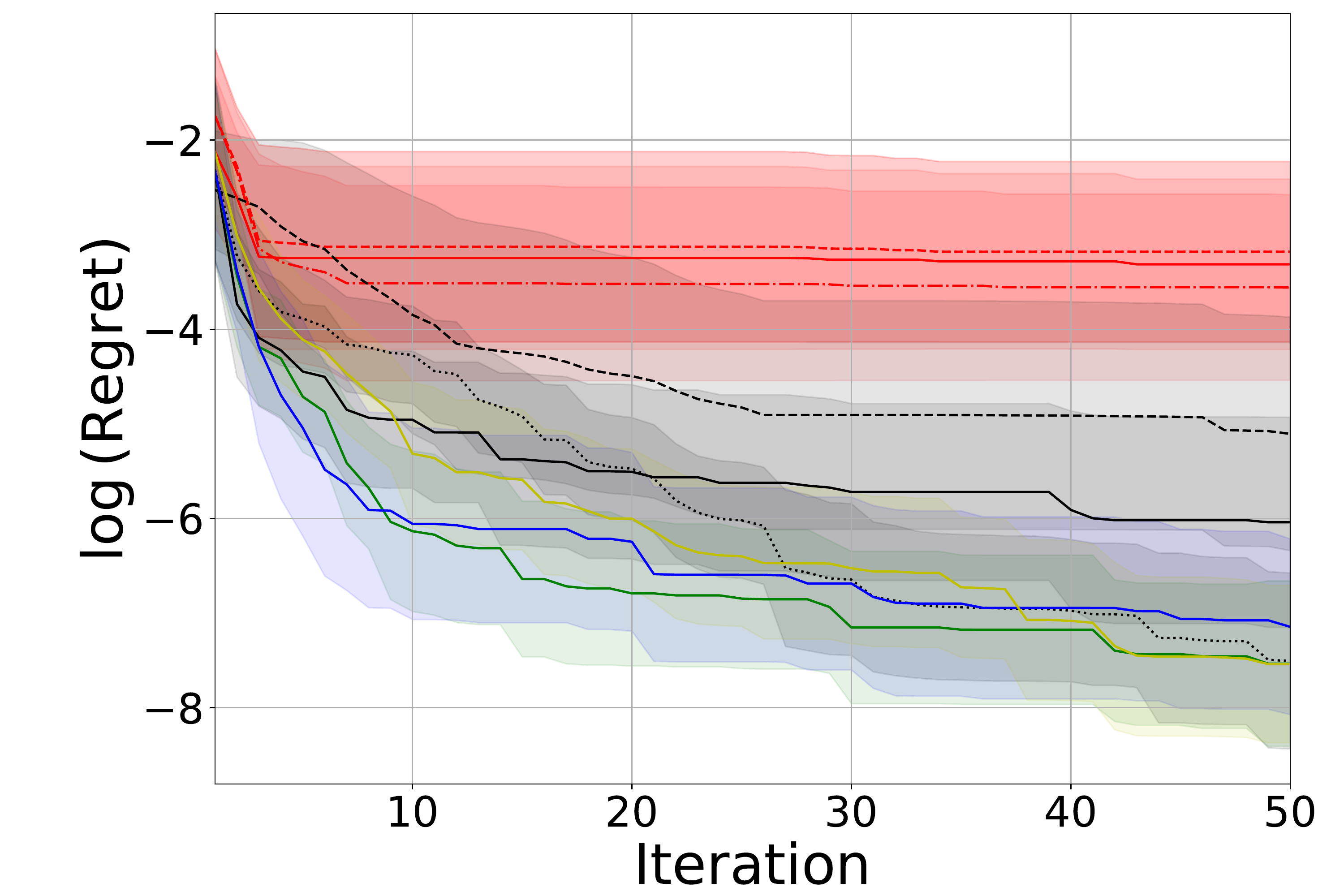}
	\includegraphics[width=0.32\textwidth]{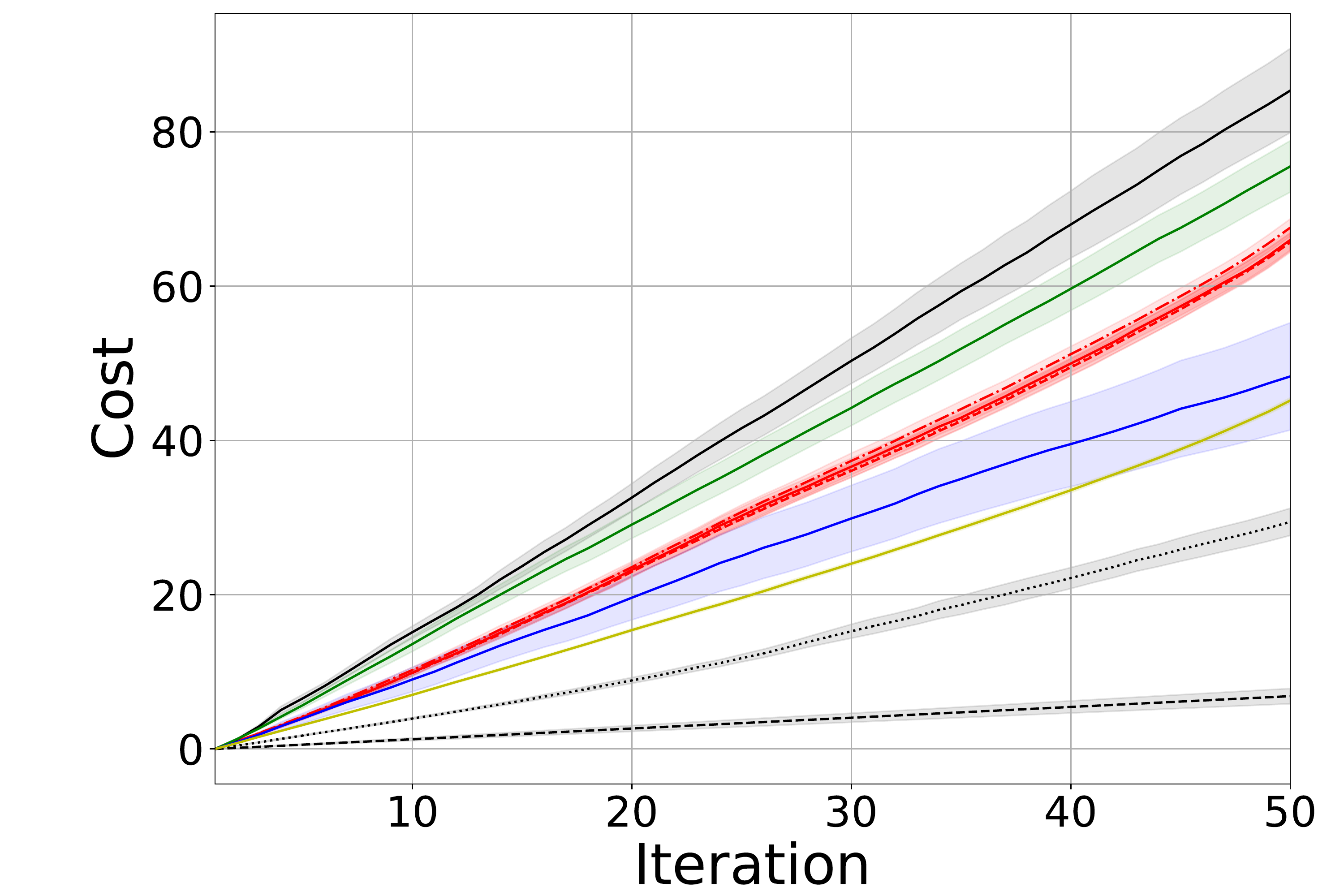}
	\caption{$T = 50$}
	\end{subfigure}
	\begin{subfigure}{\textwidth}
	\centering
	\includegraphics[width=0.32\textwidth]{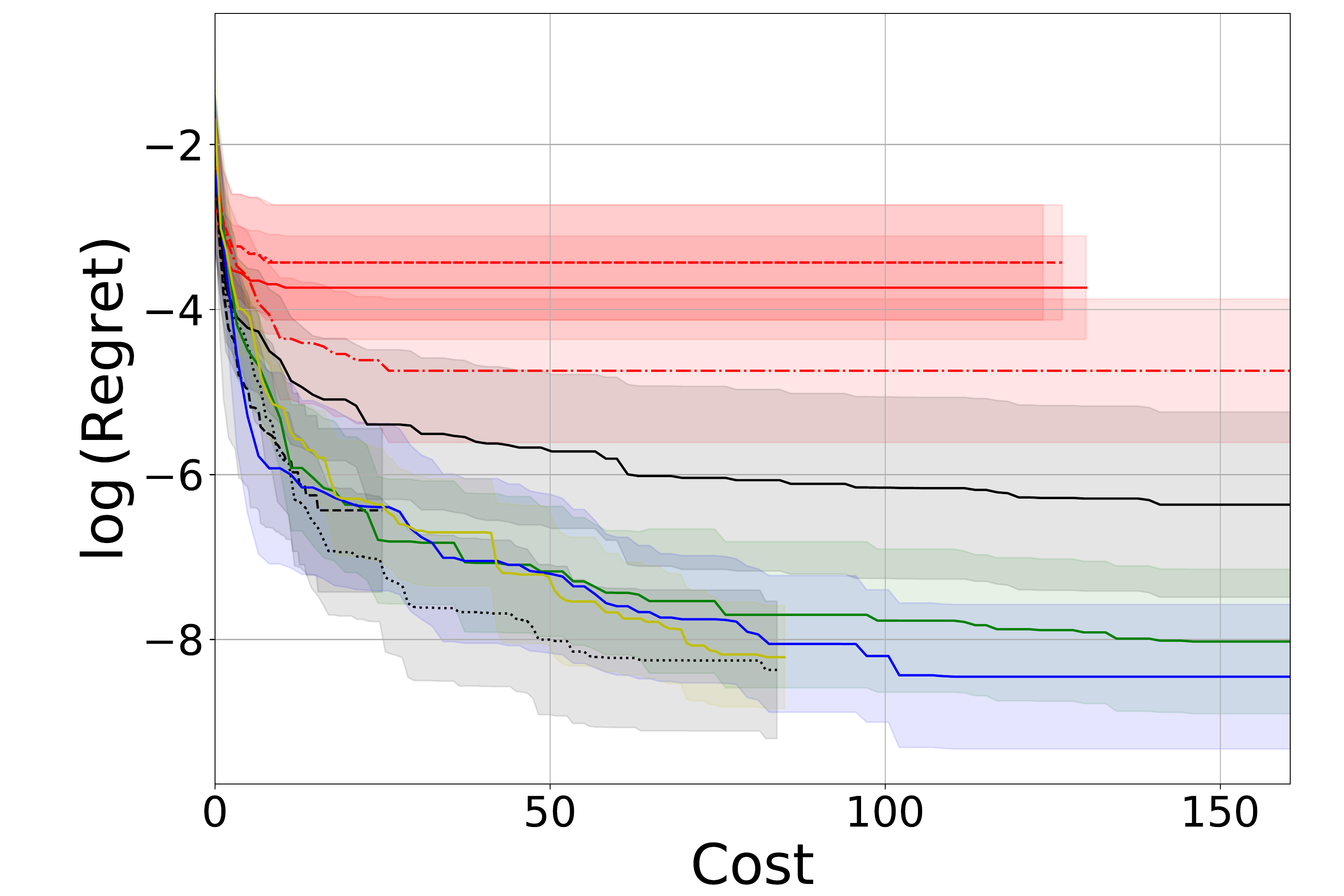}
	\includegraphics[width=0.32\textwidth]{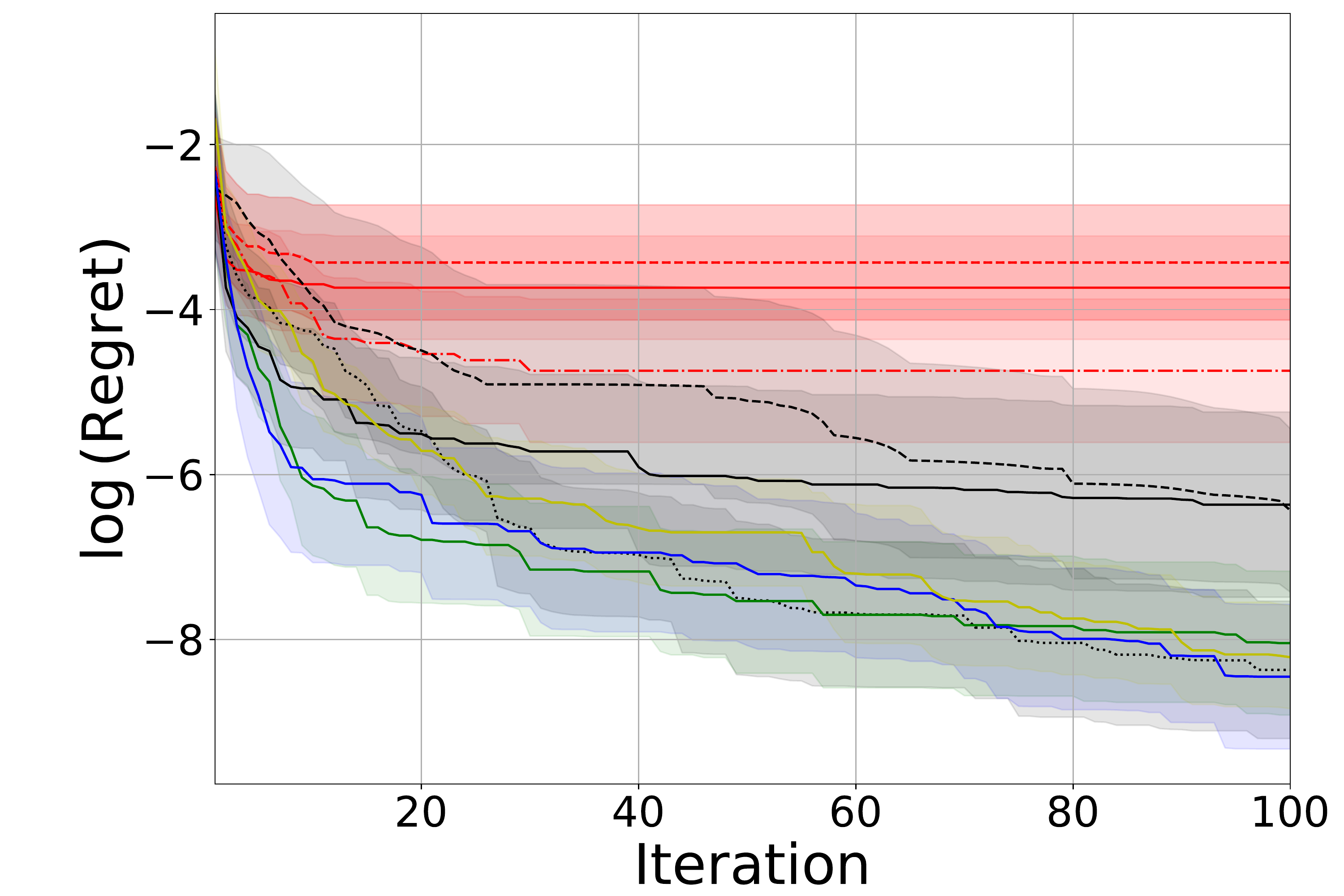}
	\includegraphics[width=0.32\textwidth]{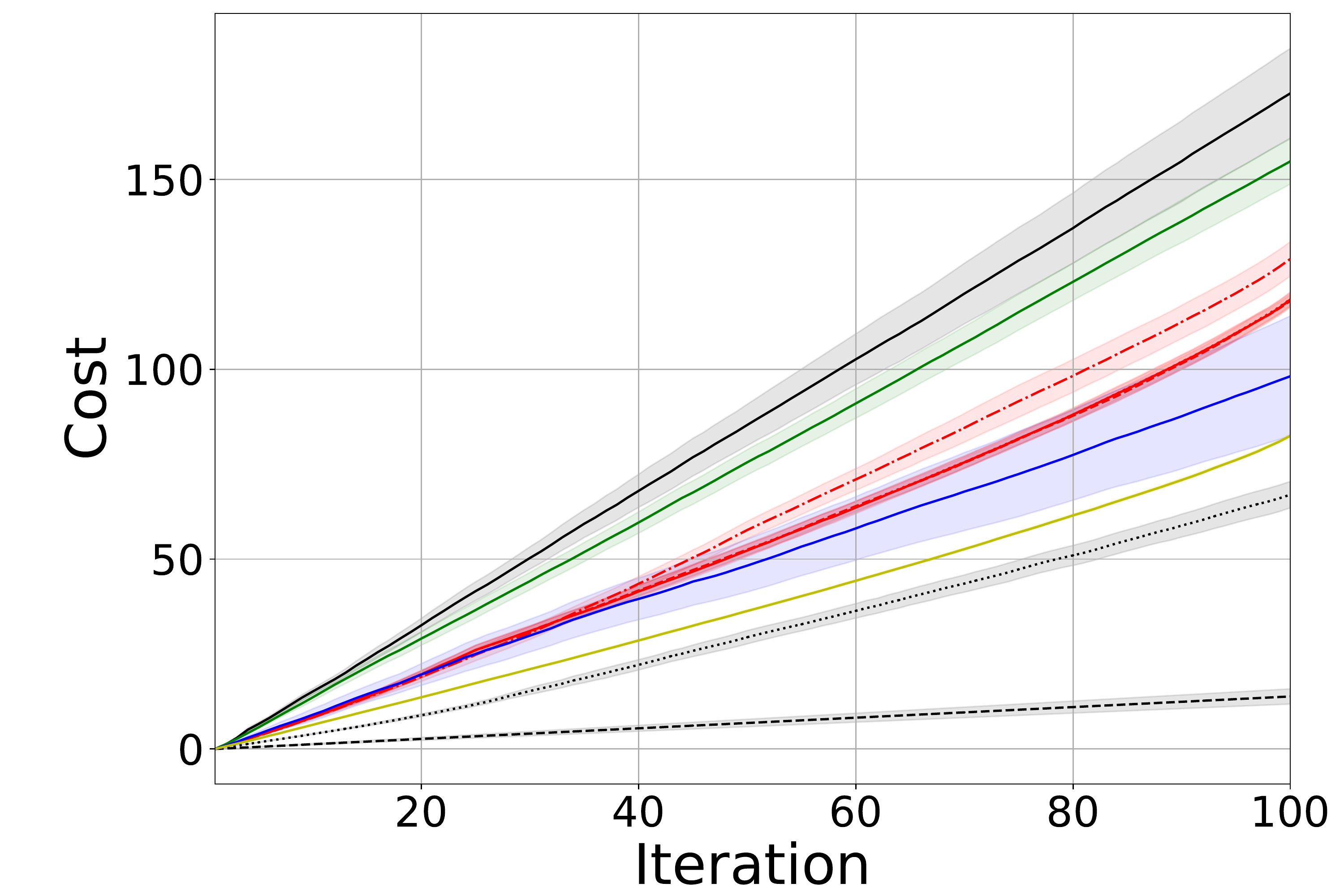}
	\caption{$T = 100$}
	\end{subfigure}
	\begin{subfigure}{\textwidth}
	\centering
	\includegraphics[width=0.32\textwidth]{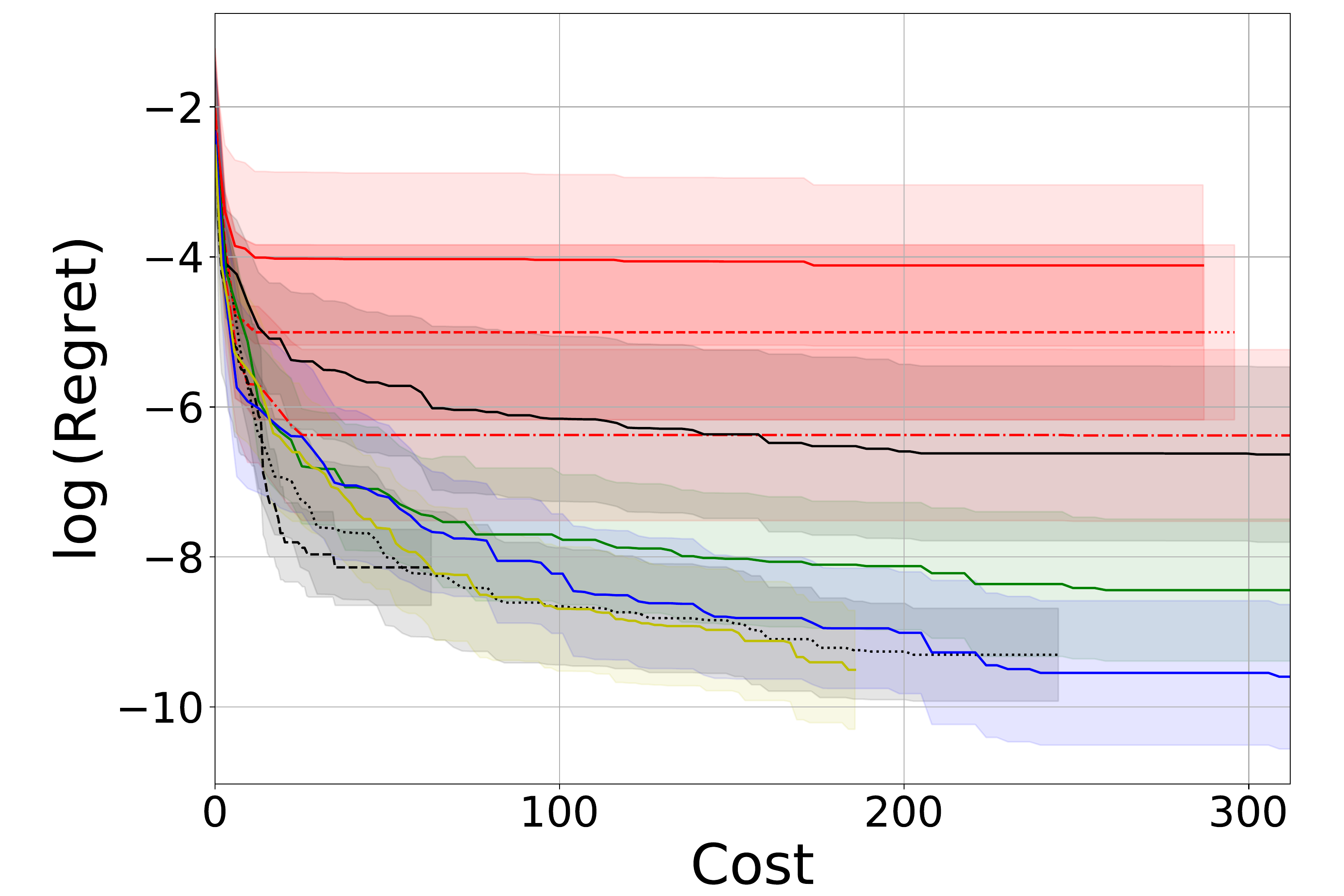}
	\includegraphics[width=0.32\textwidth]{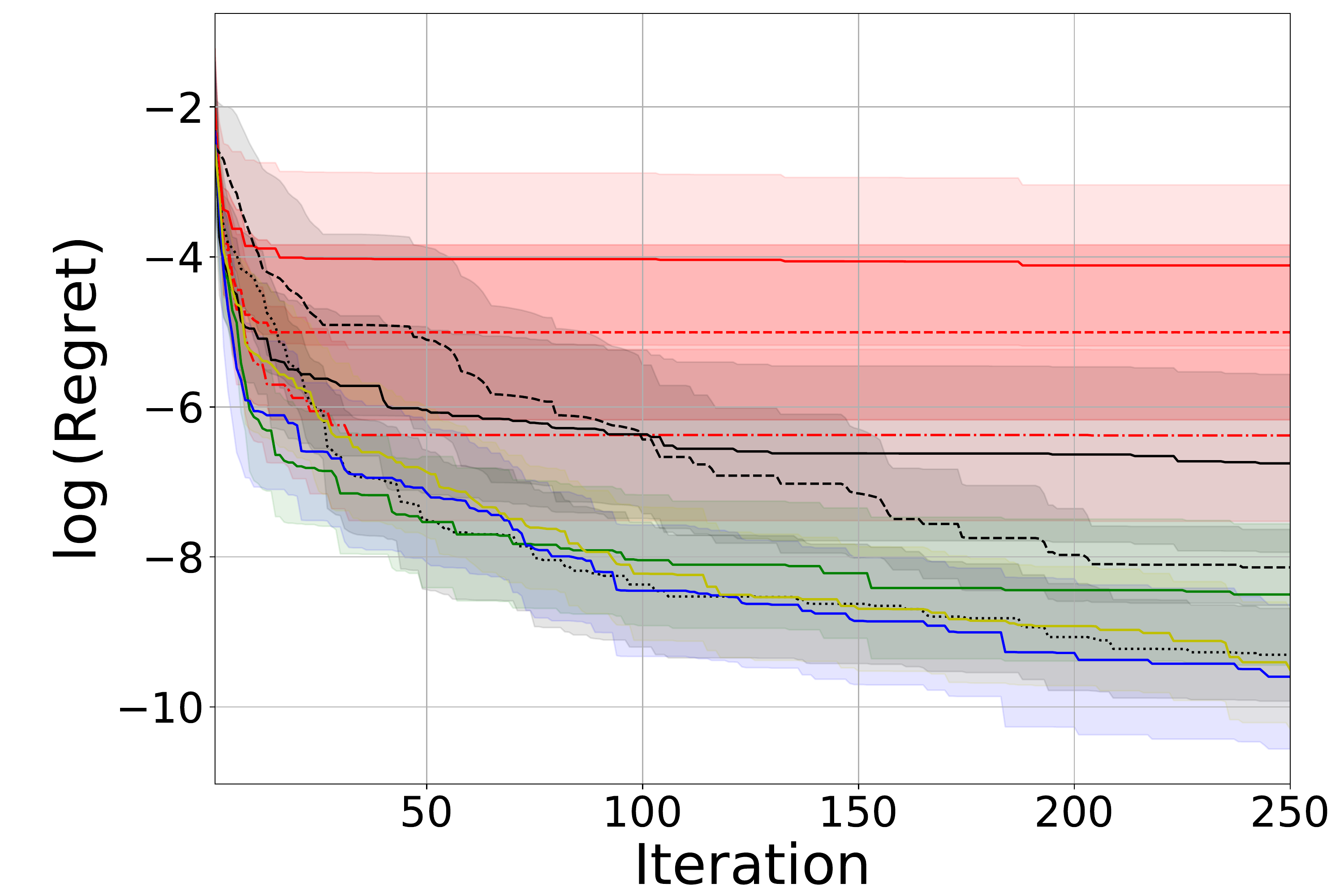}
	\includegraphics[width=0.32\textwidth]{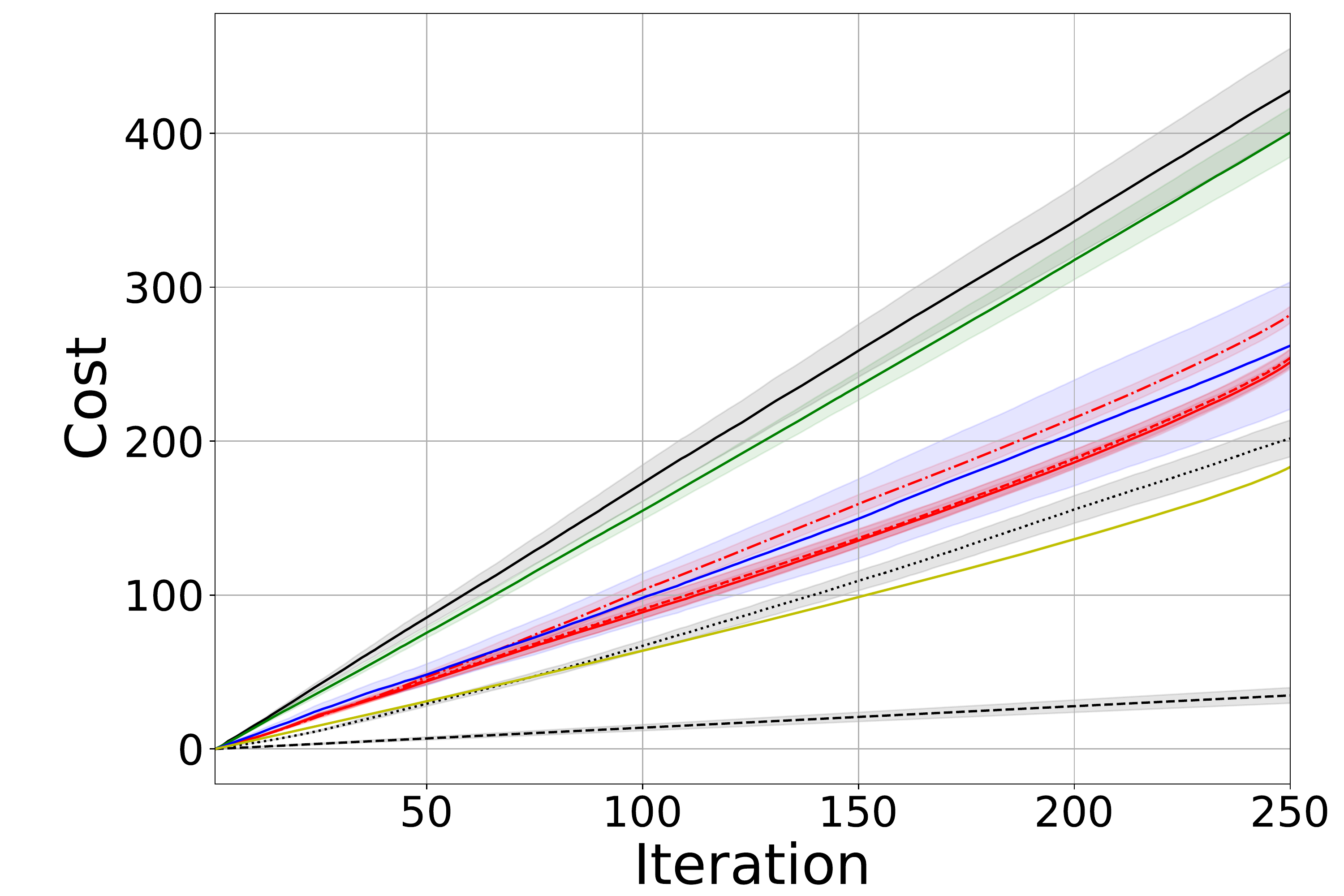}
	\caption{$T = 250$}
	\end{subfigure}
	\caption{Perm10D. Each row represents a different budget. The left column shows the evolution of regret against the cost used. The middle column shows the evolution of regret with iterations, and the right columns show the evolution of the 2-norm cost. SnAKe struggles in this benchmark, however, EI also struggles. As an interesting observation, if we \textit{did not} update the model, we would achieve a much better performance (as it would be equivalent to Random). We observe this behavior in the asynchronous case, where having a time-delay helps the method perform better (see asynchronous Ackley, Figure \ref{fig: ackley_async_4d_250}). EIpu and TrEI perform well in this example, we conjecture this is because they are doing far more localized searches while SnAKe is trying to cover all the space available (which is very difficult in higher dimensions).}
	\label{fig: perm_10d_exp_1}
\end{figure}

We include the full  graphs of the sequential Bayesian Optimization experiments. Each row represents a different budget. The left column shows the evolution of regret against the cost used, the middle column shows the evolution of regret with iterations, and the right column shows the evolution of the 2-norm cost. The results encompass Figures \ref{fig: branin_2d_exp_1} to \ref{fig: perm_10d_exp_1}. The caption in each figure tells us the benchmark function being evaluated. Each experiment is the mean $\pm$ half the standard deviation of 25 different runs.

\begin{figure}[ht]
	\centering
	\begin{subfigure}{\textwidth}
	\centering
	\includegraphics[width = 0.32\textwidth]{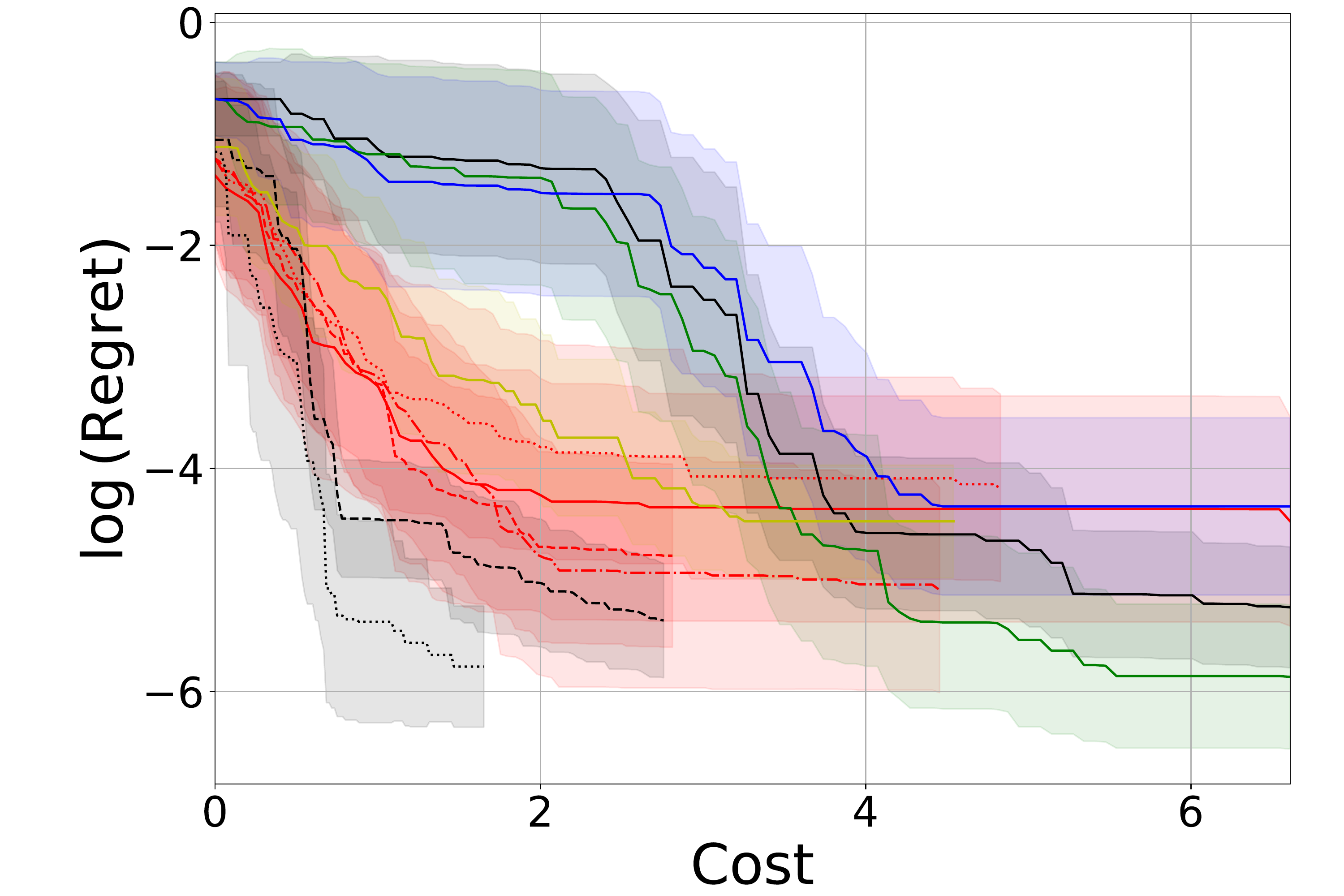}
	\includegraphics[width=0.32\textwidth]{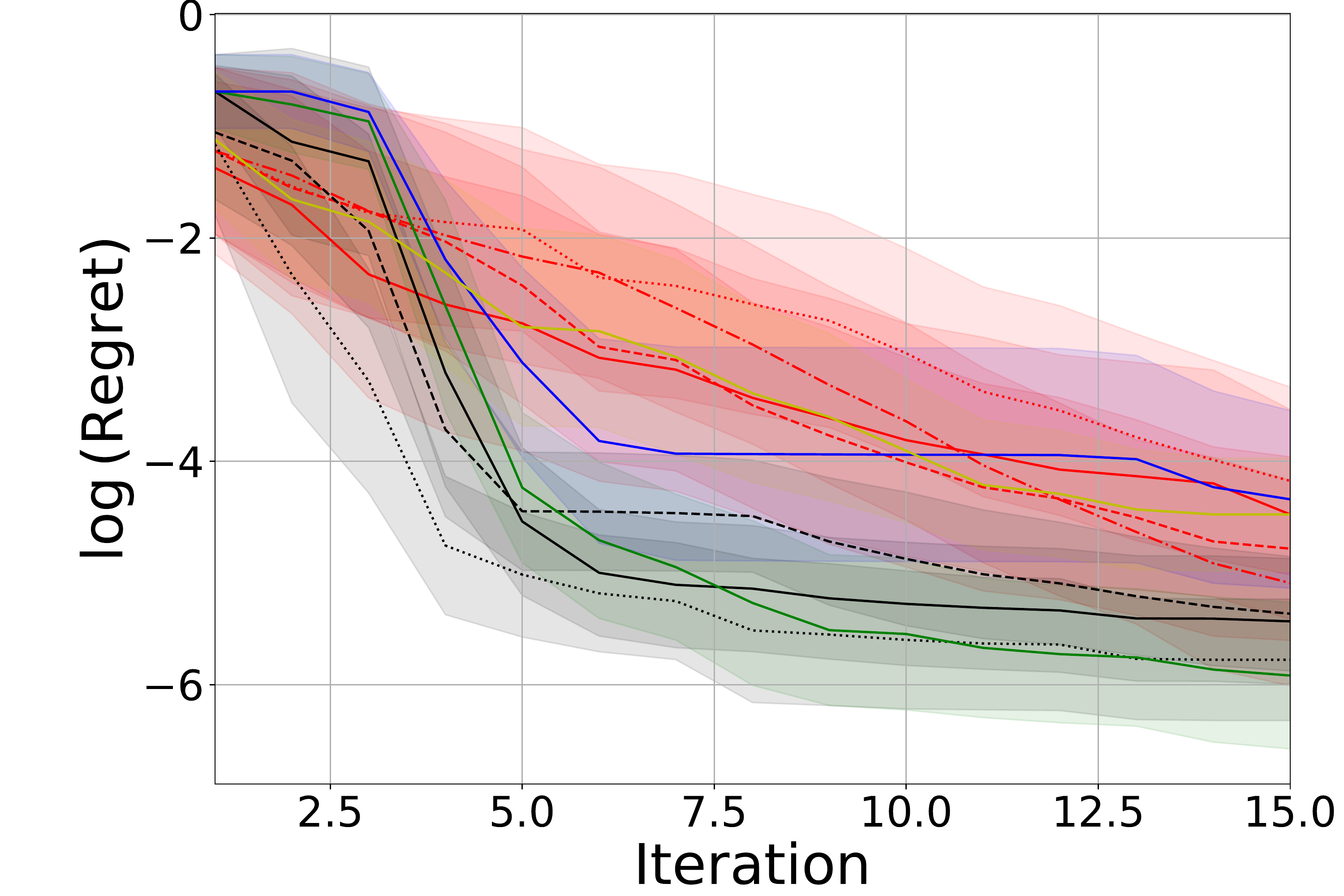}
	\includegraphics[width=0.32\textwidth]{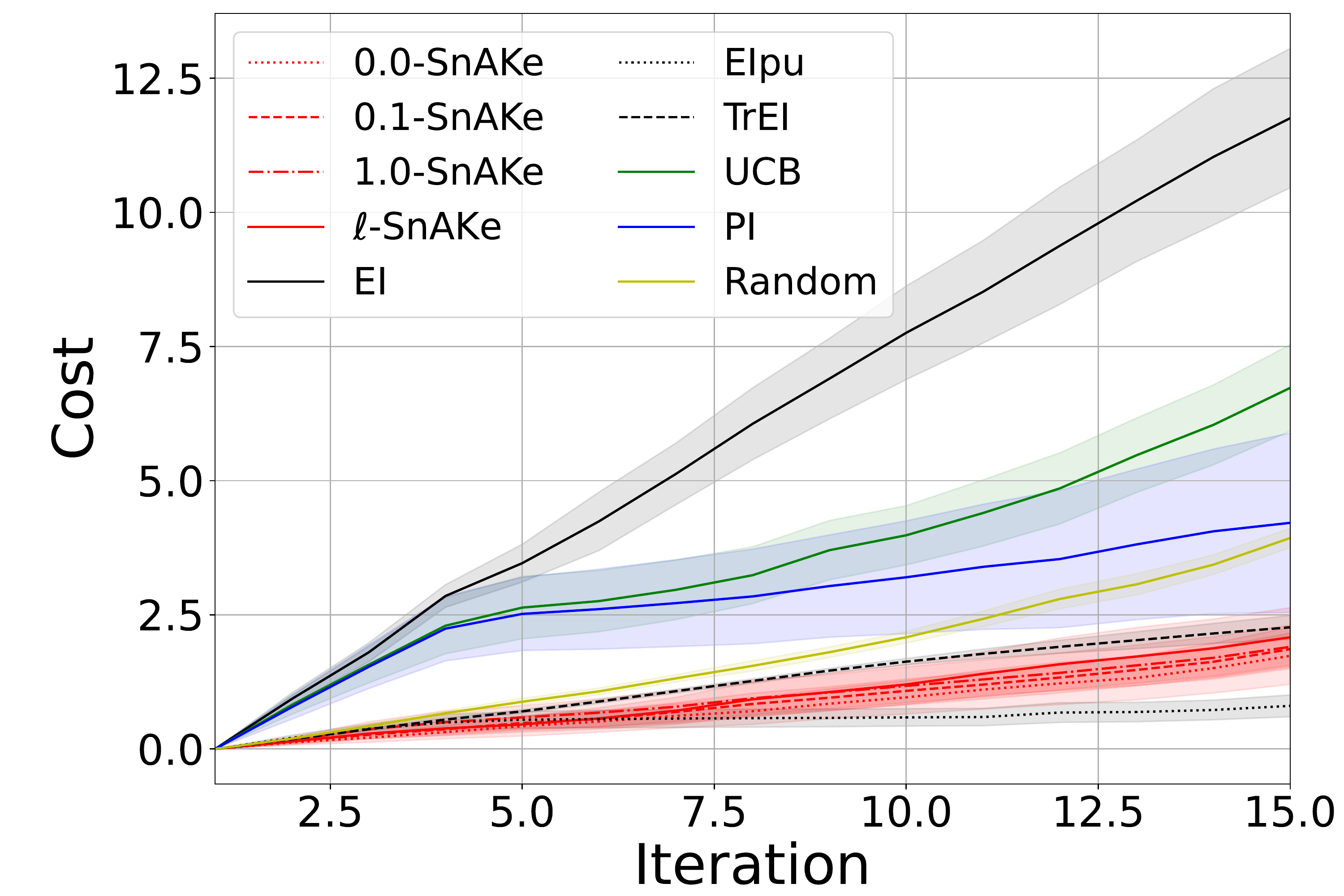}
	\caption{$T = 15$}
	\end{subfigure}
	\begin{subfigure}{\textwidth}
	\centering
	\includegraphics[width=0.32\textwidth]{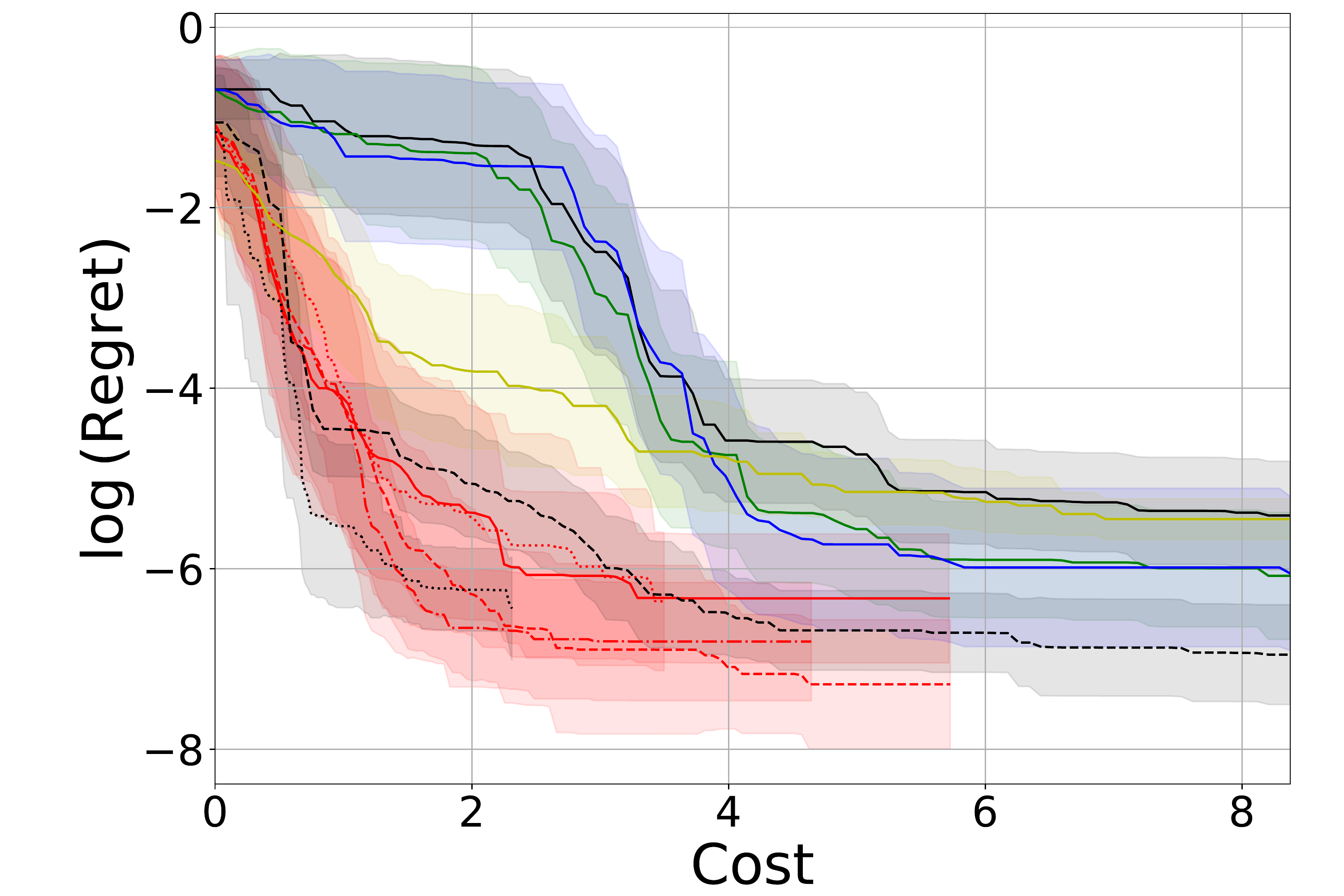}
	\includegraphics[width=0.32\textwidth]{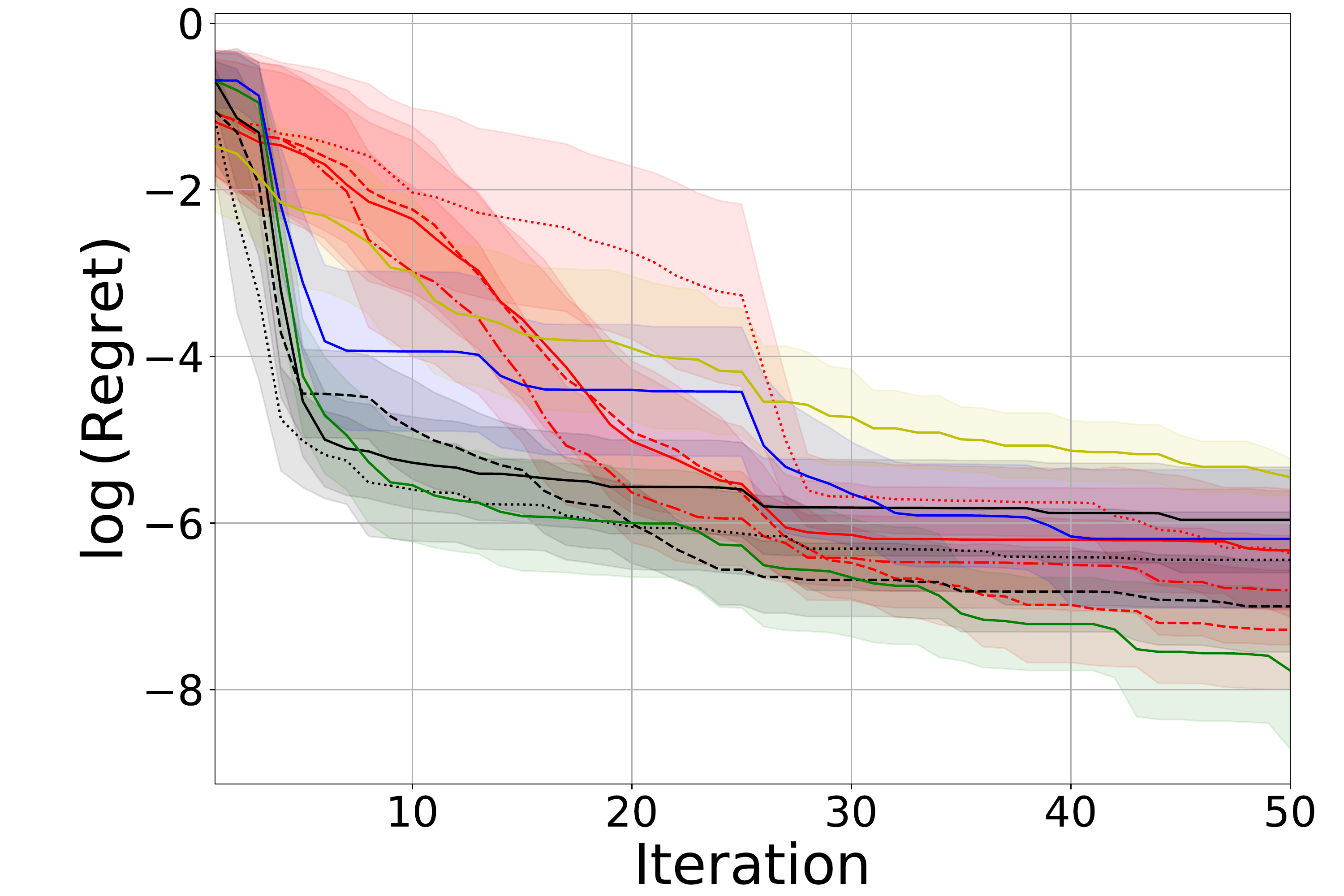}
	\includegraphics[width=0.32\textwidth]{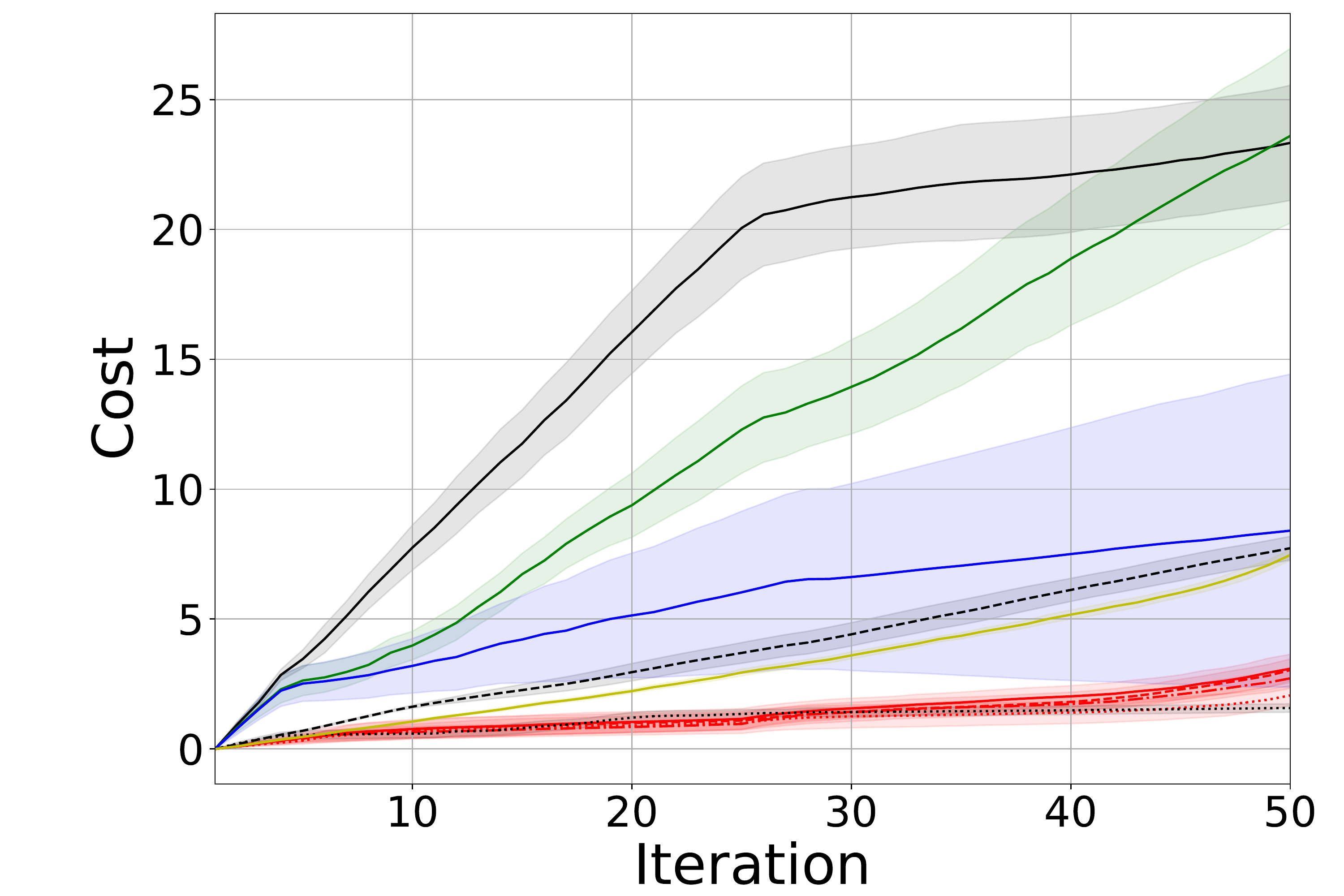}
	\caption{$T = 50$}
	\end{subfigure}
	\begin{subfigure}{\textwidth}
	\centering
	\includegraphics[width=0.32\textwidth]{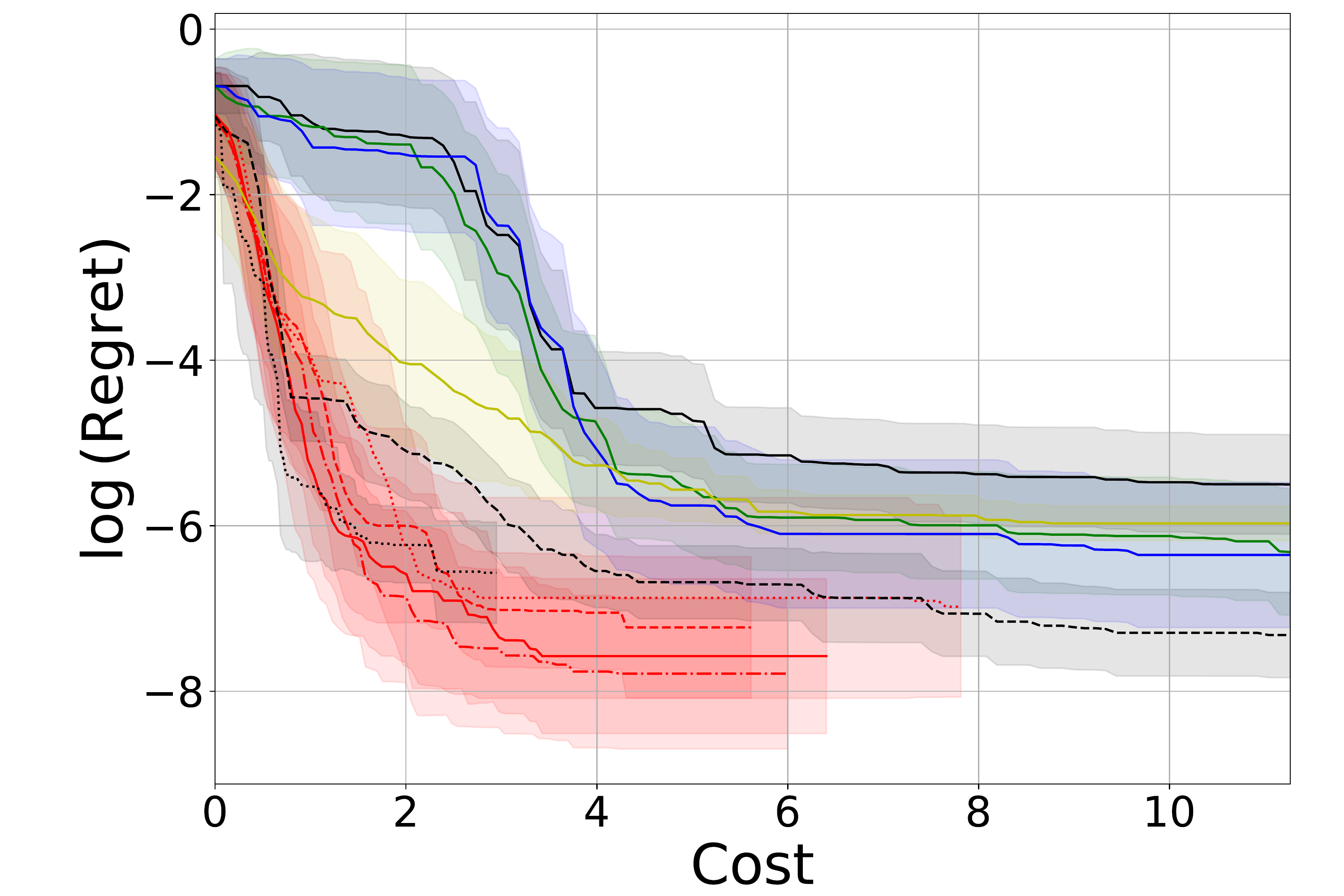}
	\includegraphics[width=0.32\textwidth]{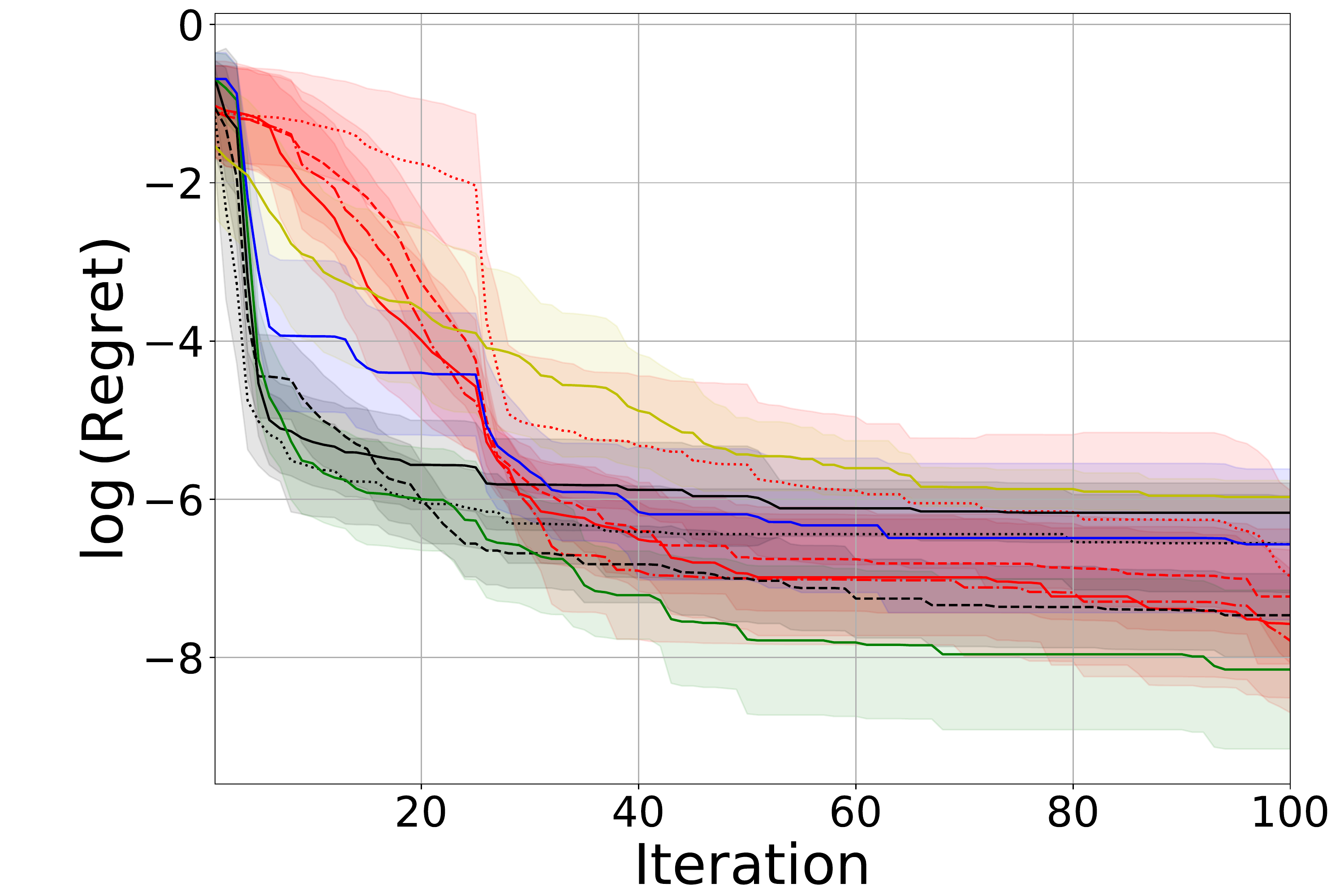}
	\includegraphics[width=0.32\textwidth]{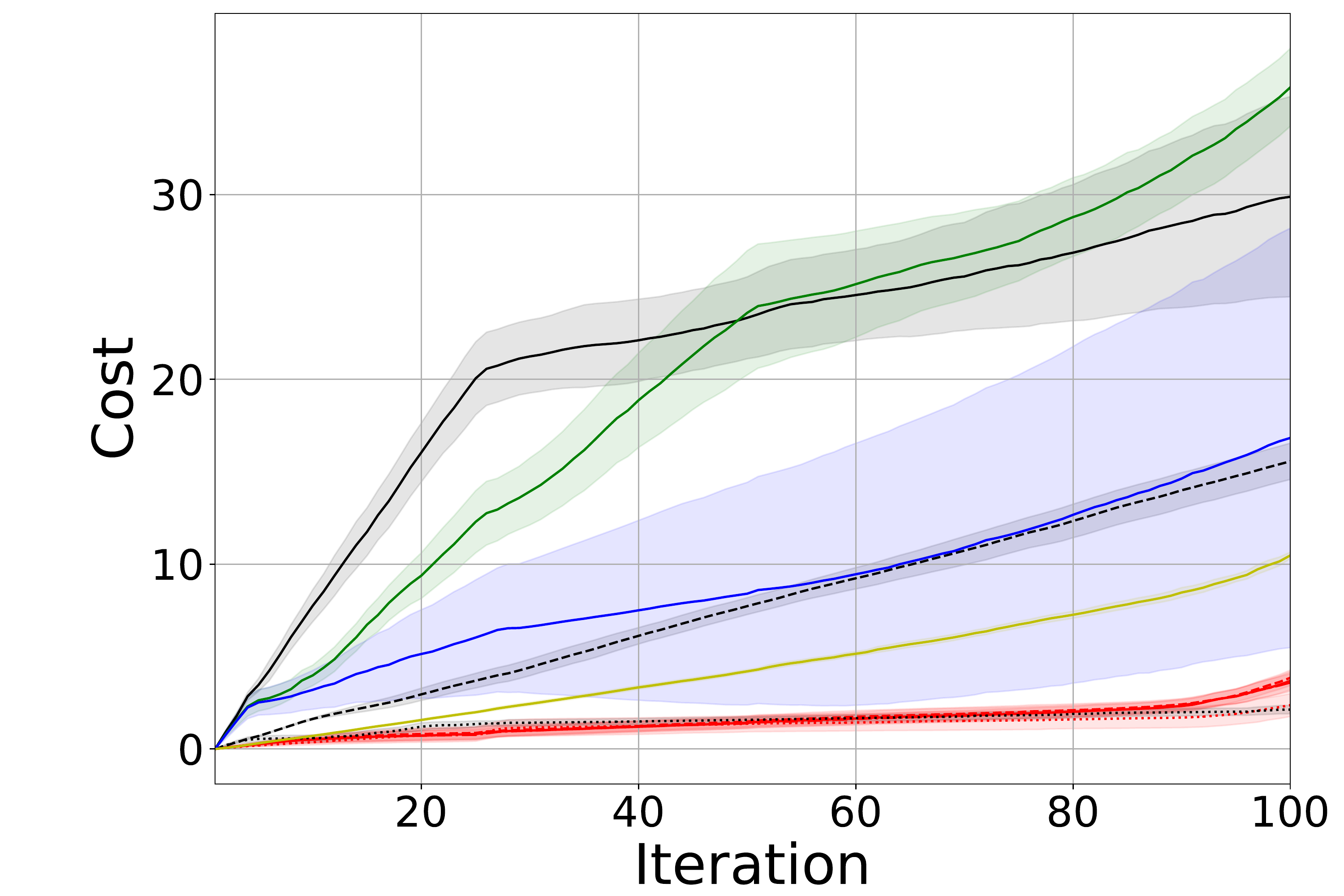}
	\caption{$T = 100$}
	\end{subfigure}
	\begin{subfigure}{\textwidth}
	\centering
	\includegraphics[width=0.32\textwidth]{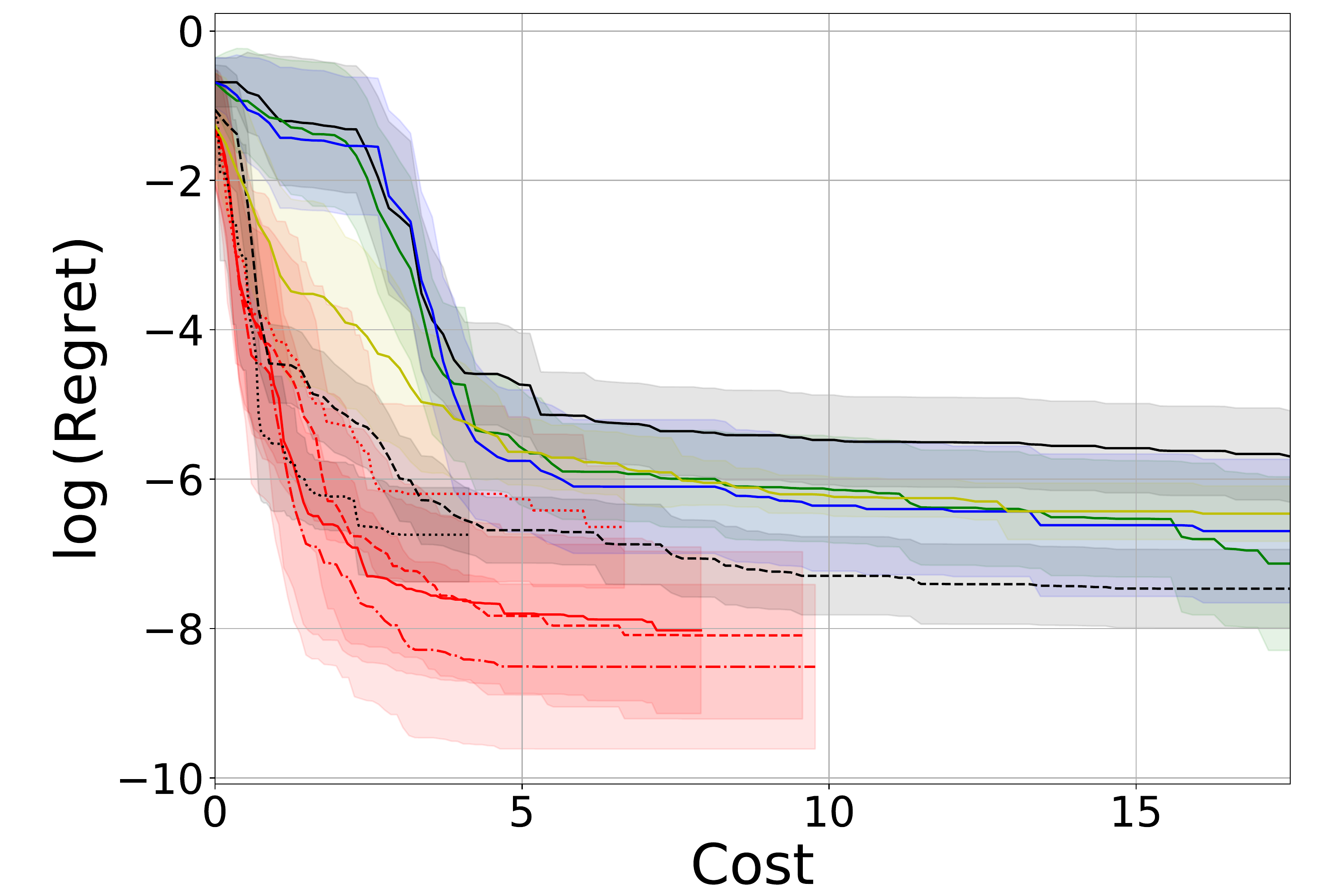}
	\includegraphics[width=0.32\textwidth]{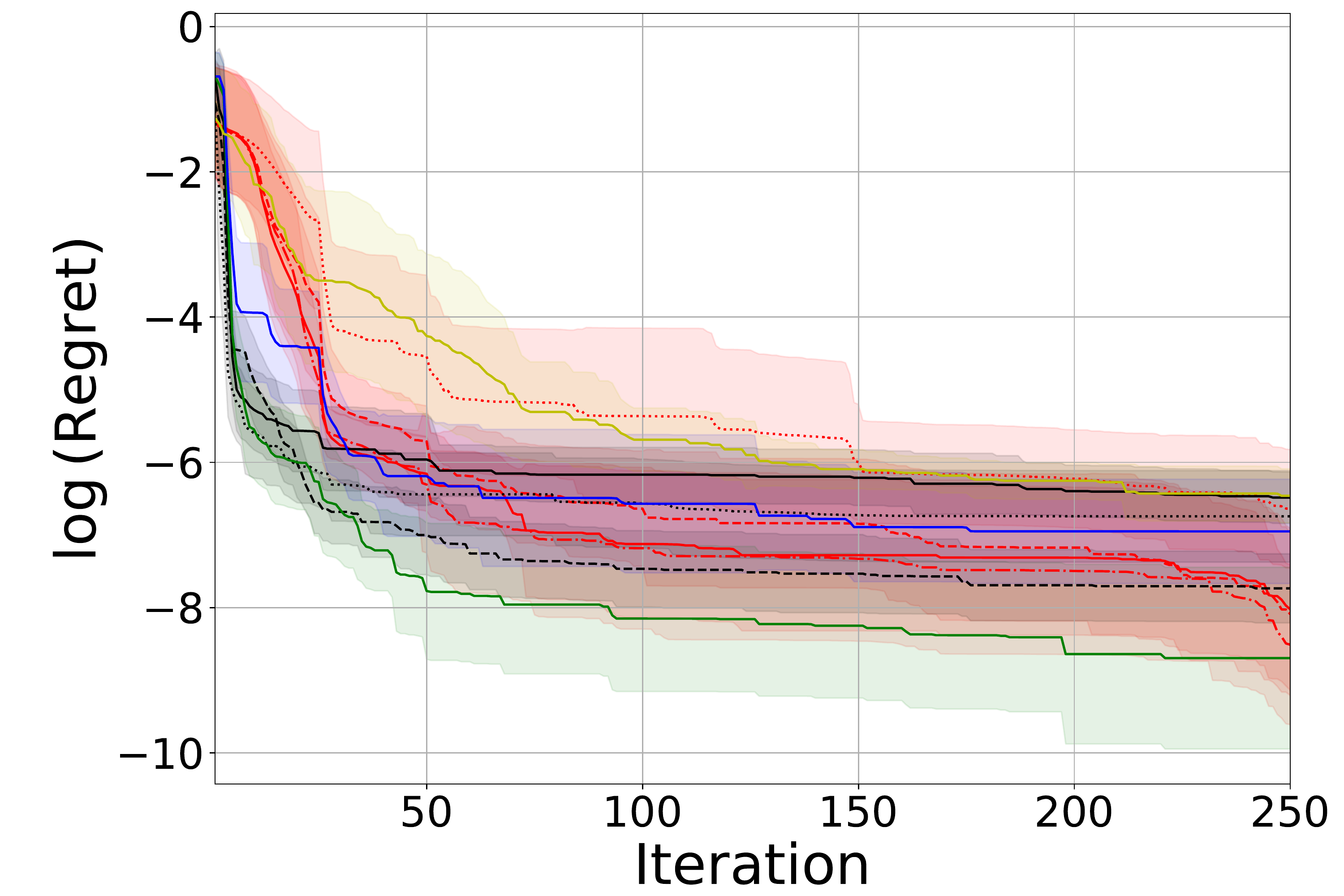}
	\includegraphics[width=0.32\textwidth]{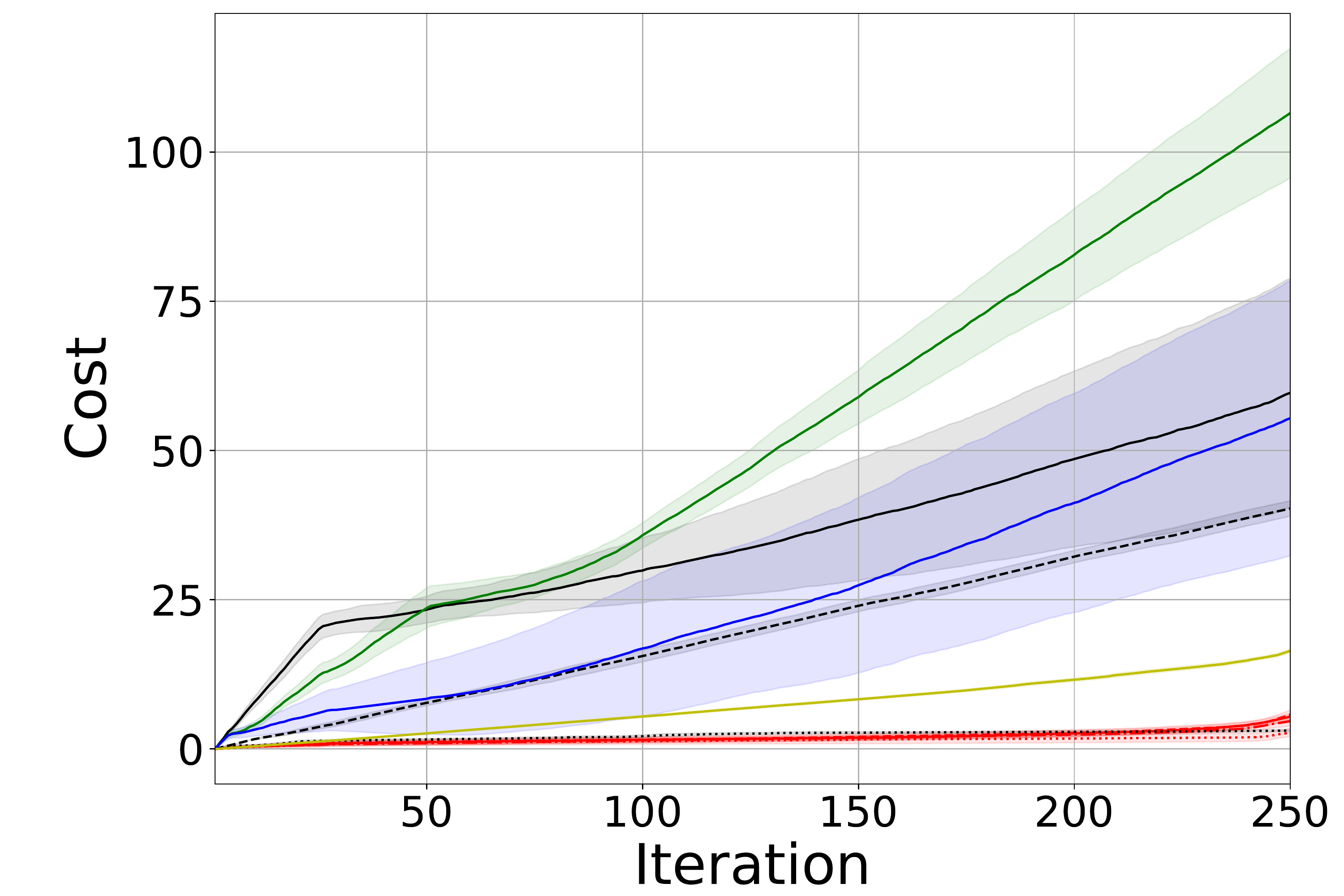}
	\caption{$T = 250$}
	\end{subfigure}
	\caption{Michaelwicz2D. Each row represents a different budget. The left column shows the evolution of regret against the cost used. The middle column shows the evolution of regret with iterations, and the right columns show the evolution of the 2-norm cost. SnAKe has regret comparable with other methods for all budgets (UCB outperforms the rest for larger ones). SnAKe achieves significantly less cost at all budgets, this may be due to SnAKe exploring the many local optimums carefully. The first column shows that SnAKe achieves by far the best regret for low cost. In this example, SnAKe and EIpu have similar performance.}
\end{figure}

\begin{figure}[ht]
	\centering
	\begin{subfigure}{\textwidth}
	\centering
	\includegraphics[width = 0.32\textwidth]{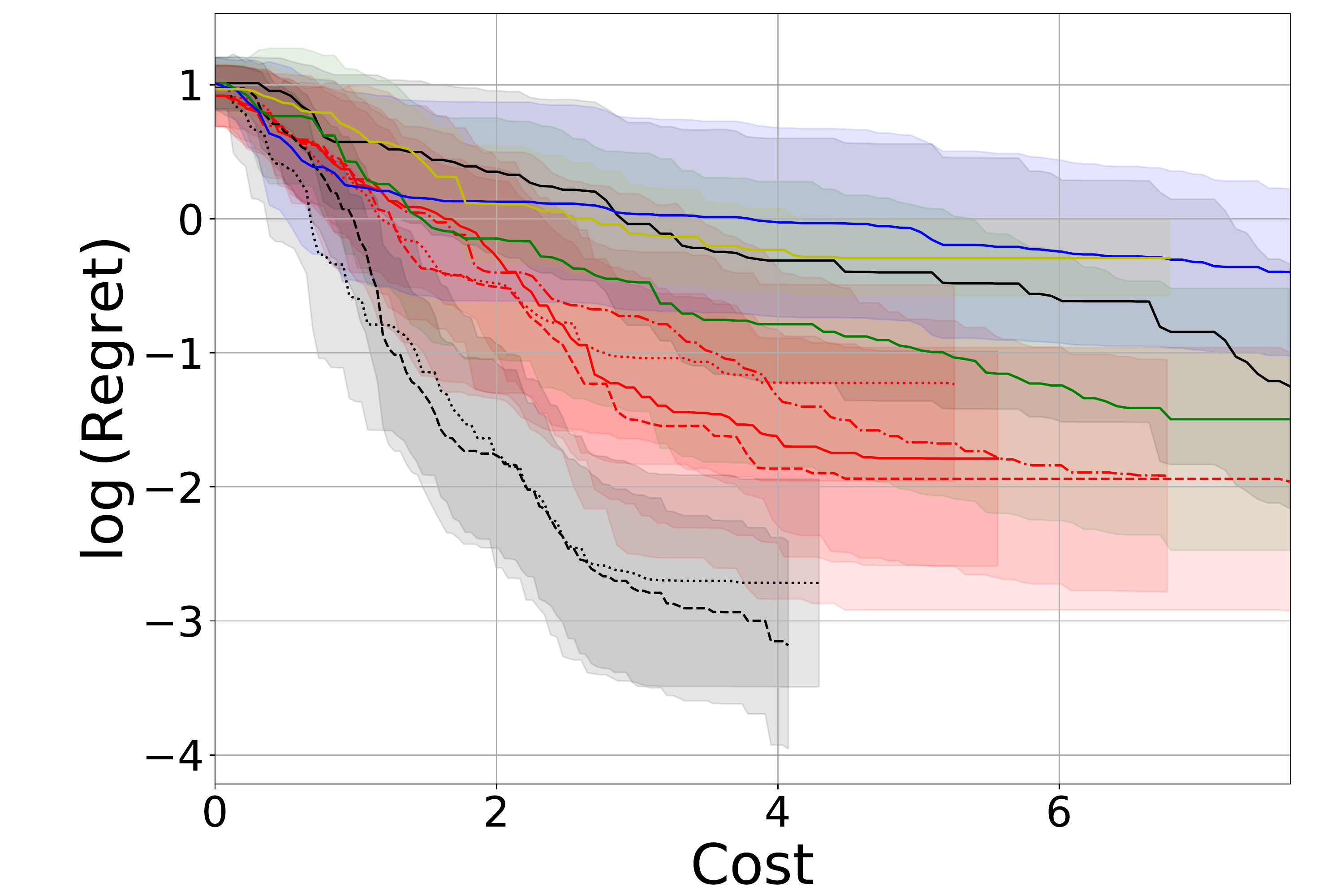}
	\includegraphics[width=0.32\textwidth]{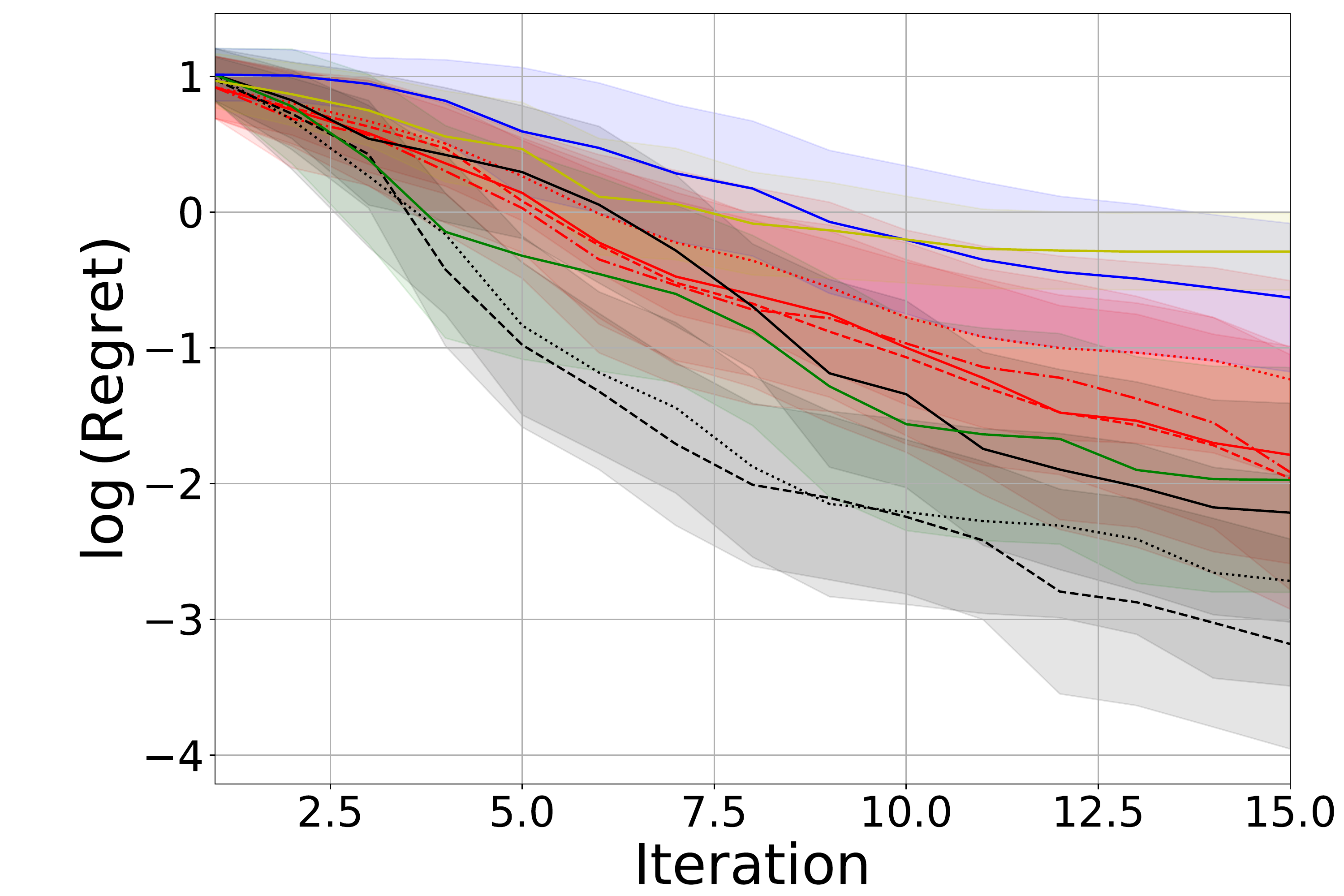}
	\includegraphics[width=0.32\textwidth]{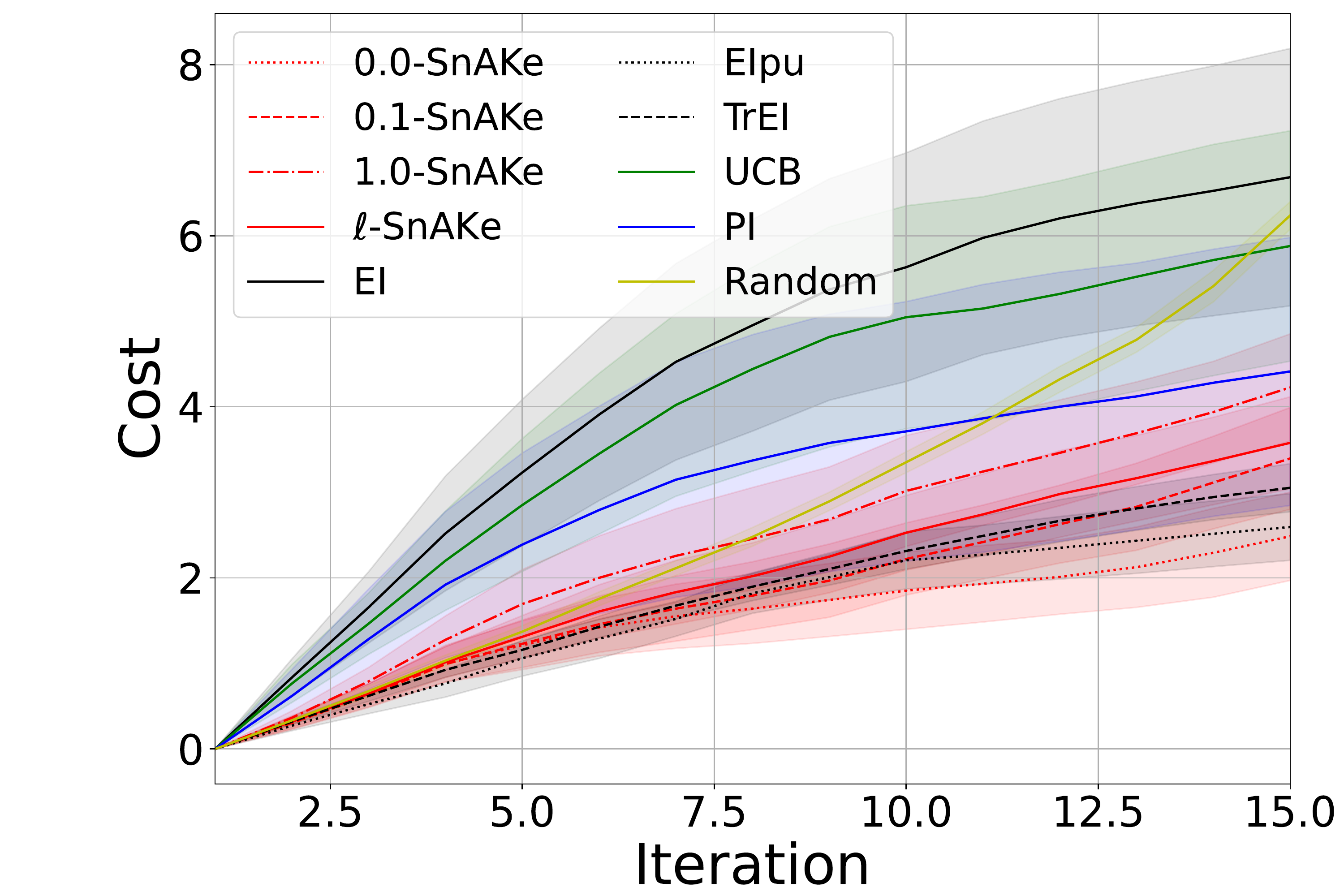}
	\caption{$T = 15$}
	\end{subfigure}
	\begin{subfigure}{\textwidth}
	\centering
	\includegraphics[width=0.32\textwidth]{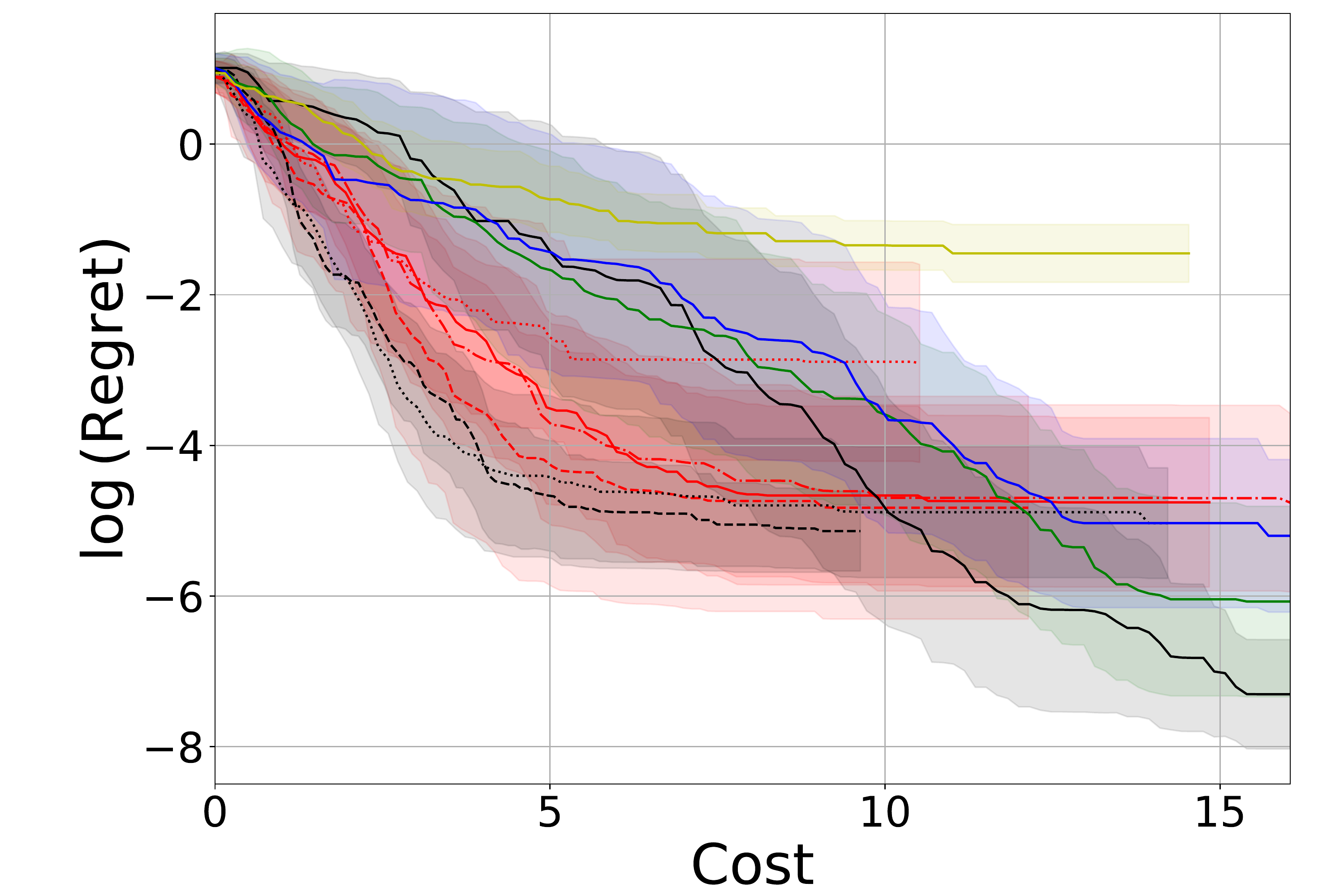}
	\includegraphics[width=0.32\textwidth]{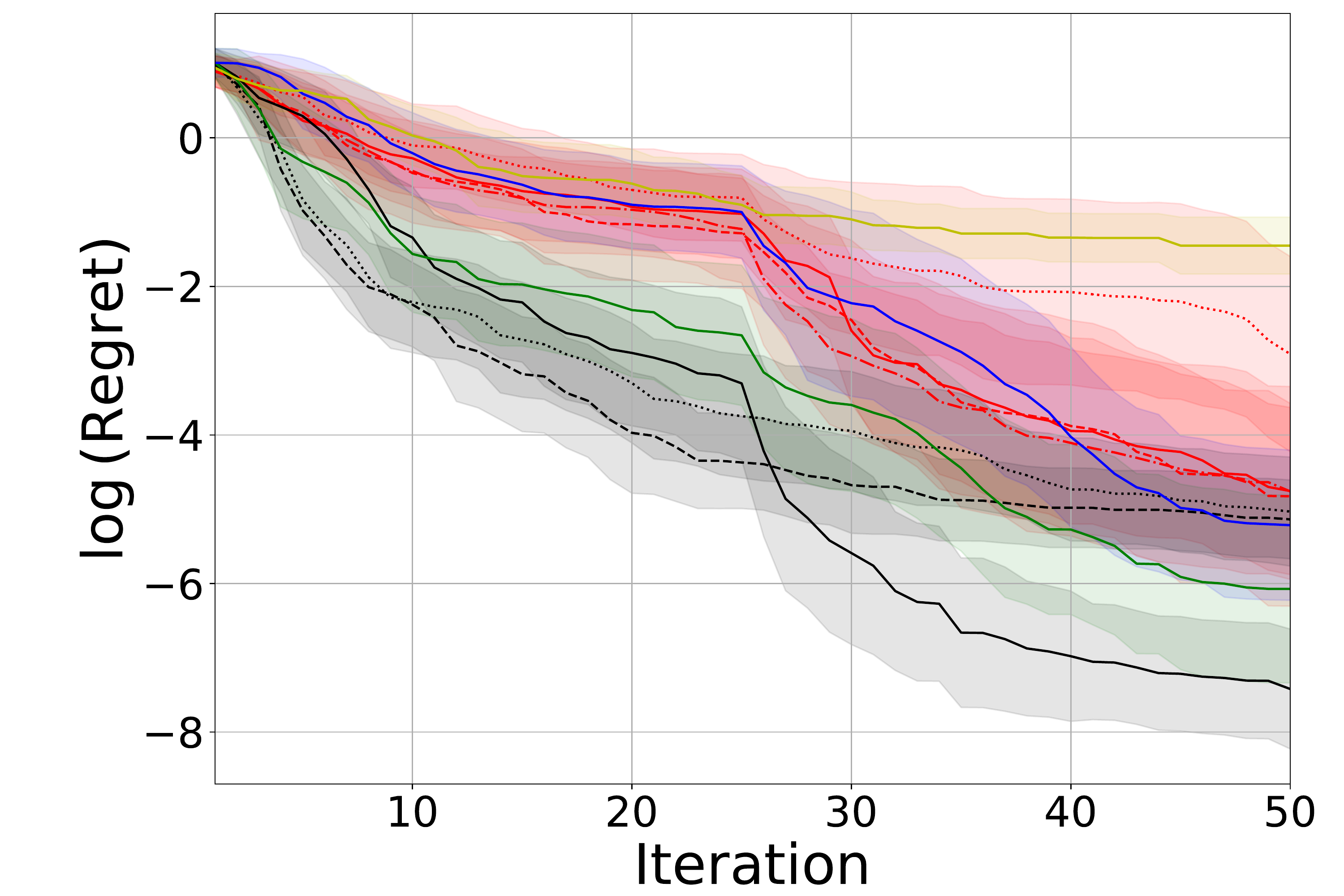}
	\includegraphics[width=0.32\textwidth]{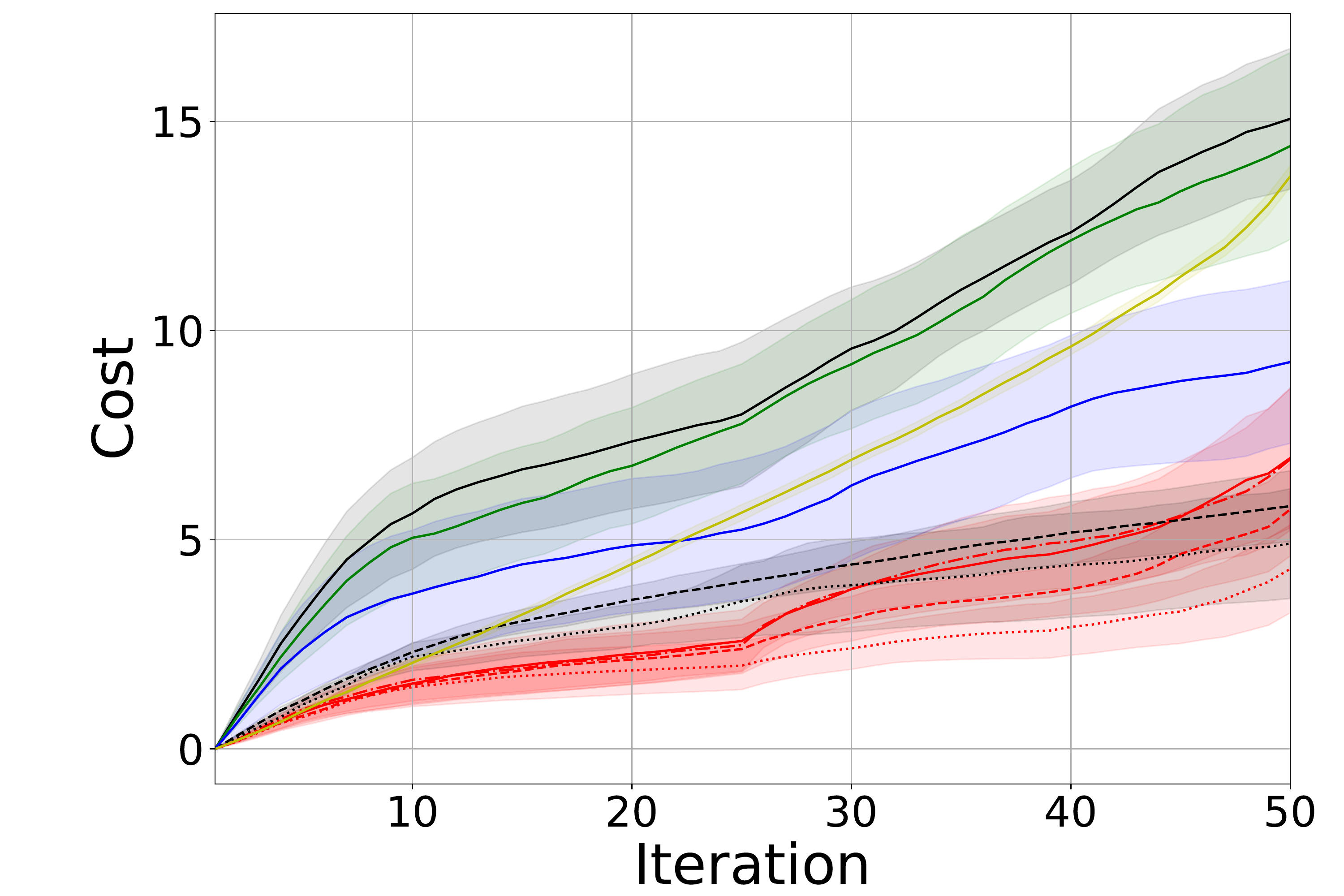}
	\caption{$T = 50$}
	\end{subfigure}
	\begin{subfigure}{\textwidth}
	\centering
	\includegraphics[width=0.32\textwidth]{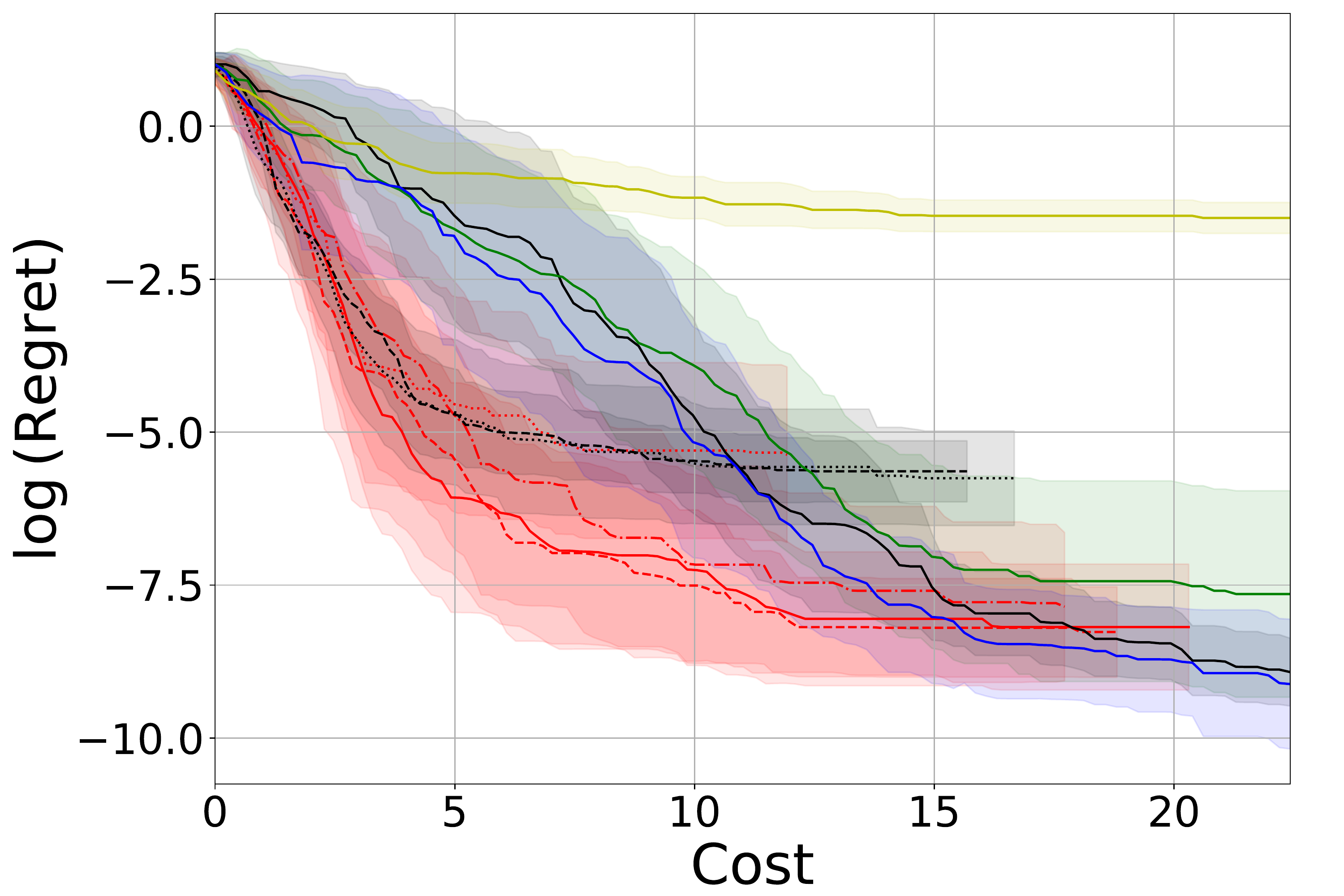}
	\includegraphics[width=0.32\textwidth]{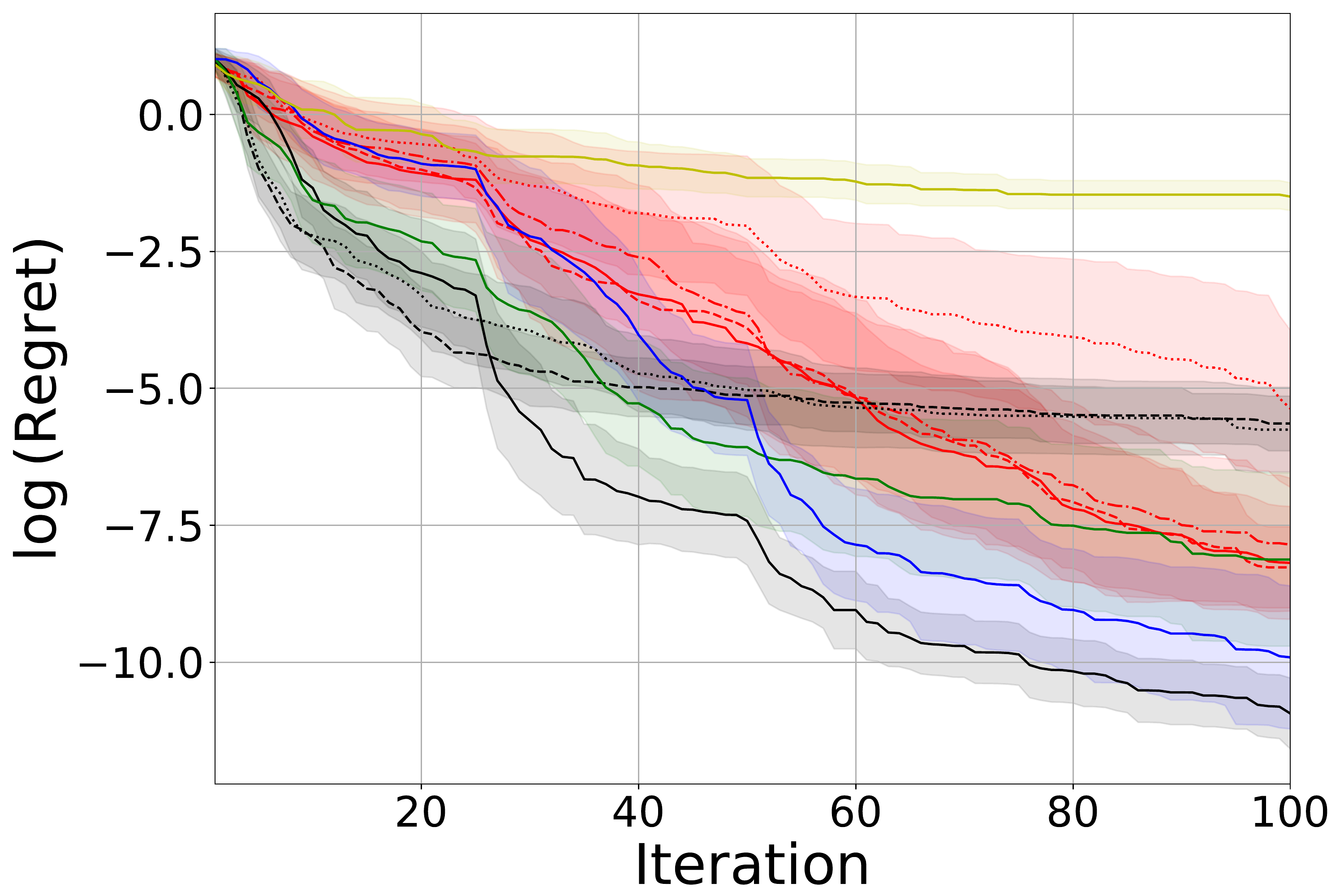}
	\includegraphics[width=0.32\textwidth]{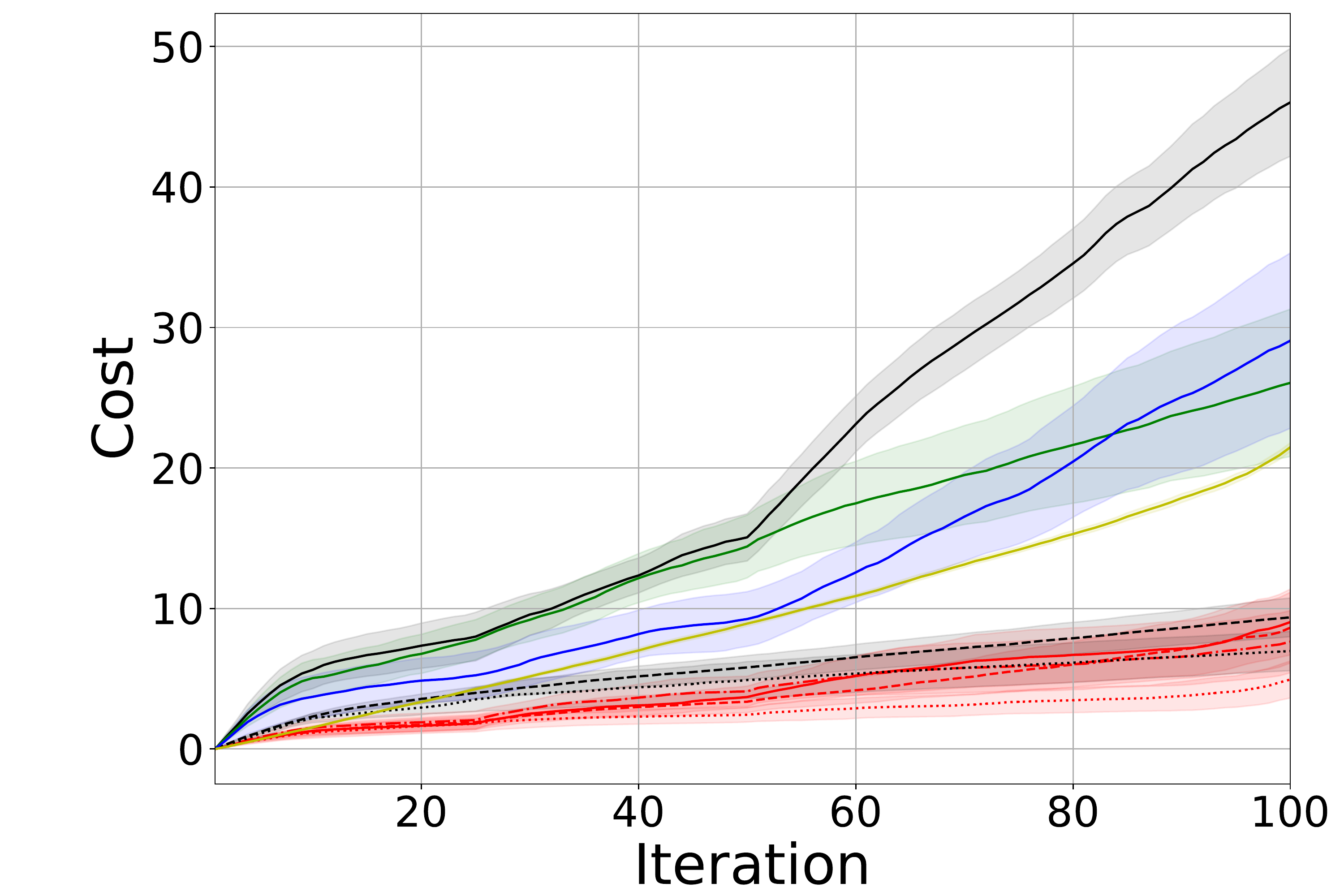}
	\caption{$T = 100$}
	\end{subfigure}
	\begin{subfigure}{\textwidth}
	\centering
	\includegraphics[width=0.32\textwidth]{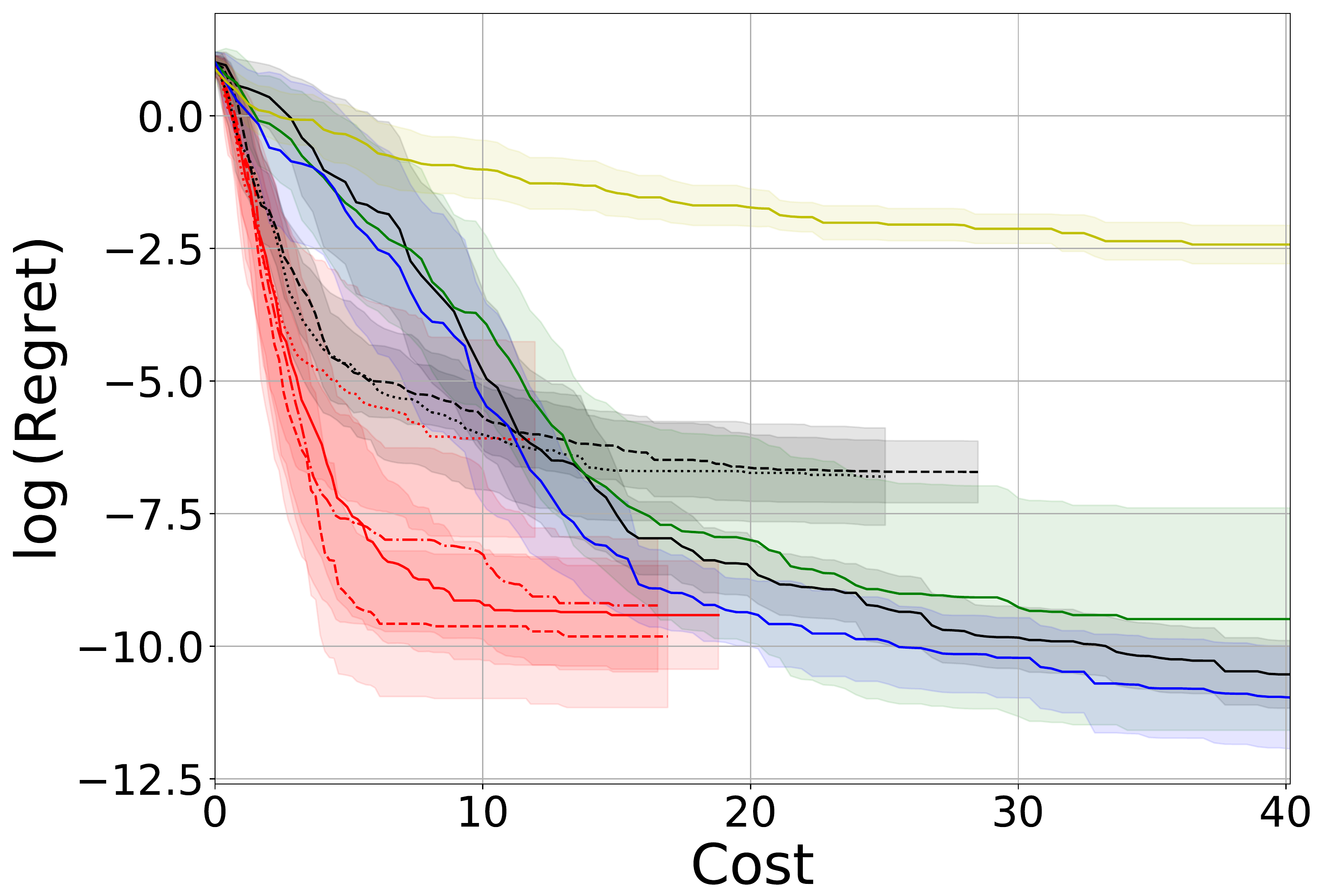}
	\includegraphics[width=0.32\textwidth]{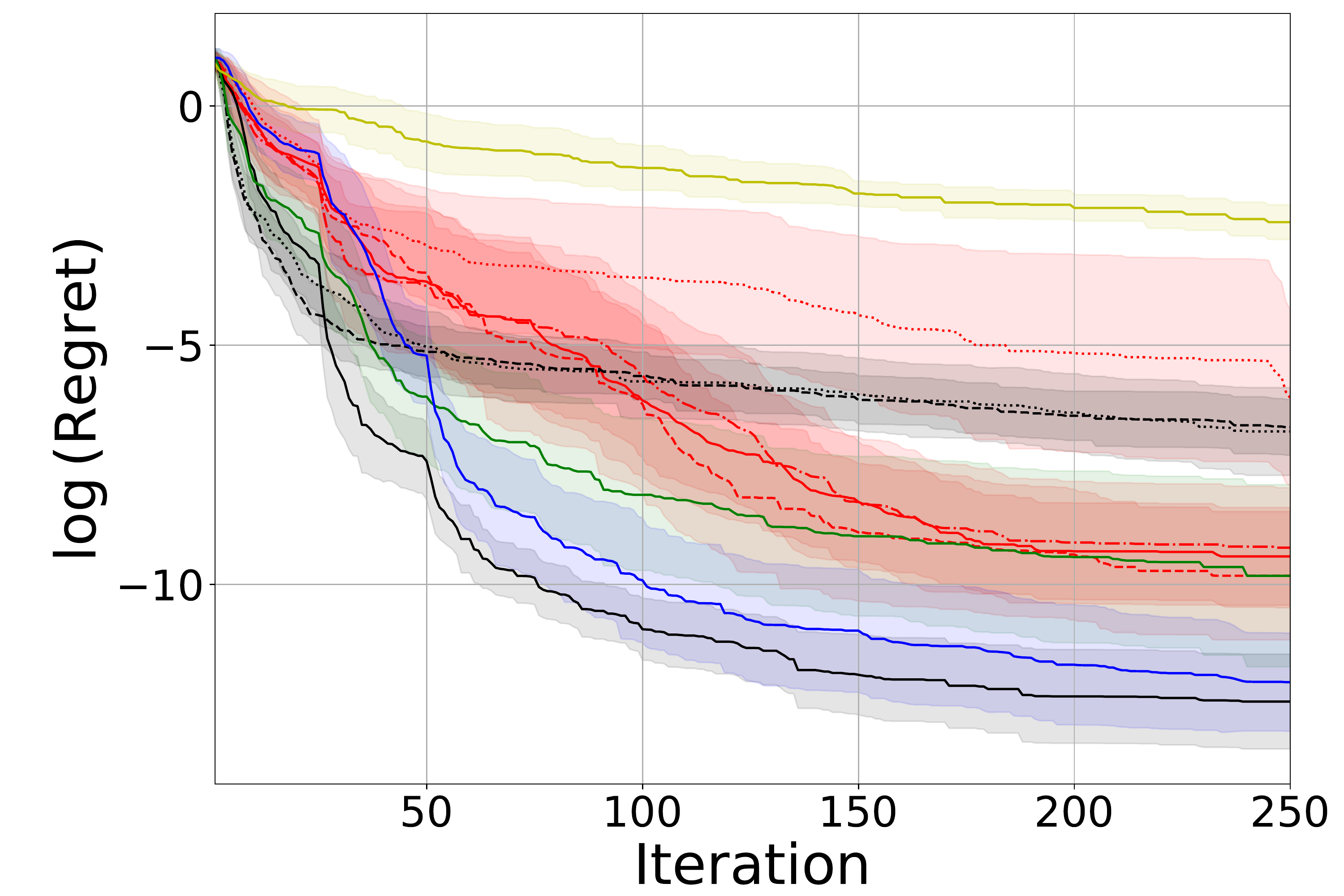}
	\includegraphics[width=0.32\textwidth]{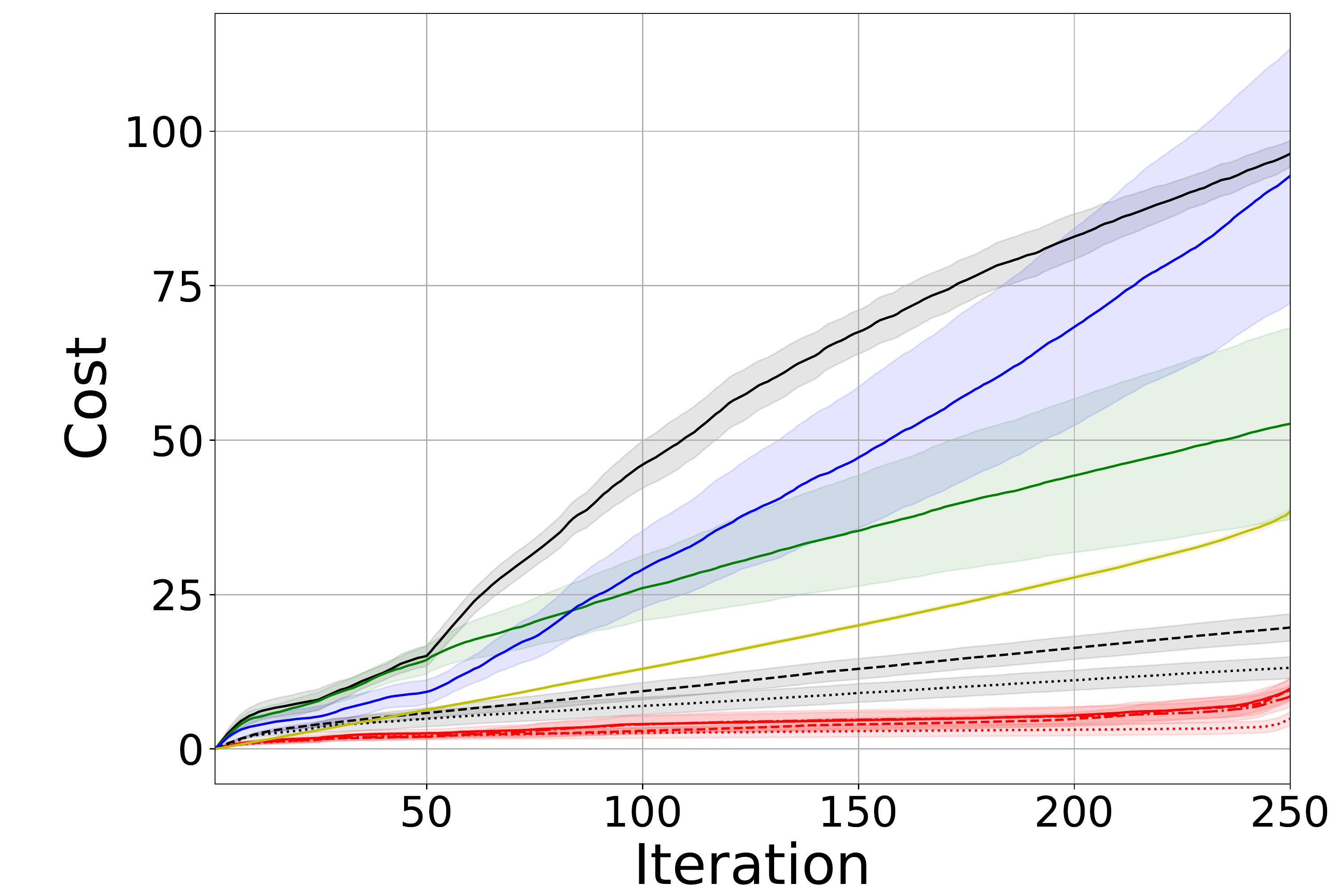}
	\caption{$T = 250$}
	\end{subfigure}
	\caption{Hartmann3D. Each row represents a different budget. The left column shows the evolution of regret against the cost used. The middle column shows the evolution of regret with iterations, and the right columns show the evolution of the 2-norm cost. Again, SnAKe achieves the best regret at low cost for all budgets. $\epsilon = 0$ struggles in this benchmark, showcasing the impact that Point Deletion can have. EIpu and TrEI achieve higher cost and worse regret than SnAKe.}
\end{figure}

\begin{figure}[ht]
	\centering
	\begin{subfigure}{\textwidth}
	\centering
	\includegraphics[width = 0.32\textwidth]{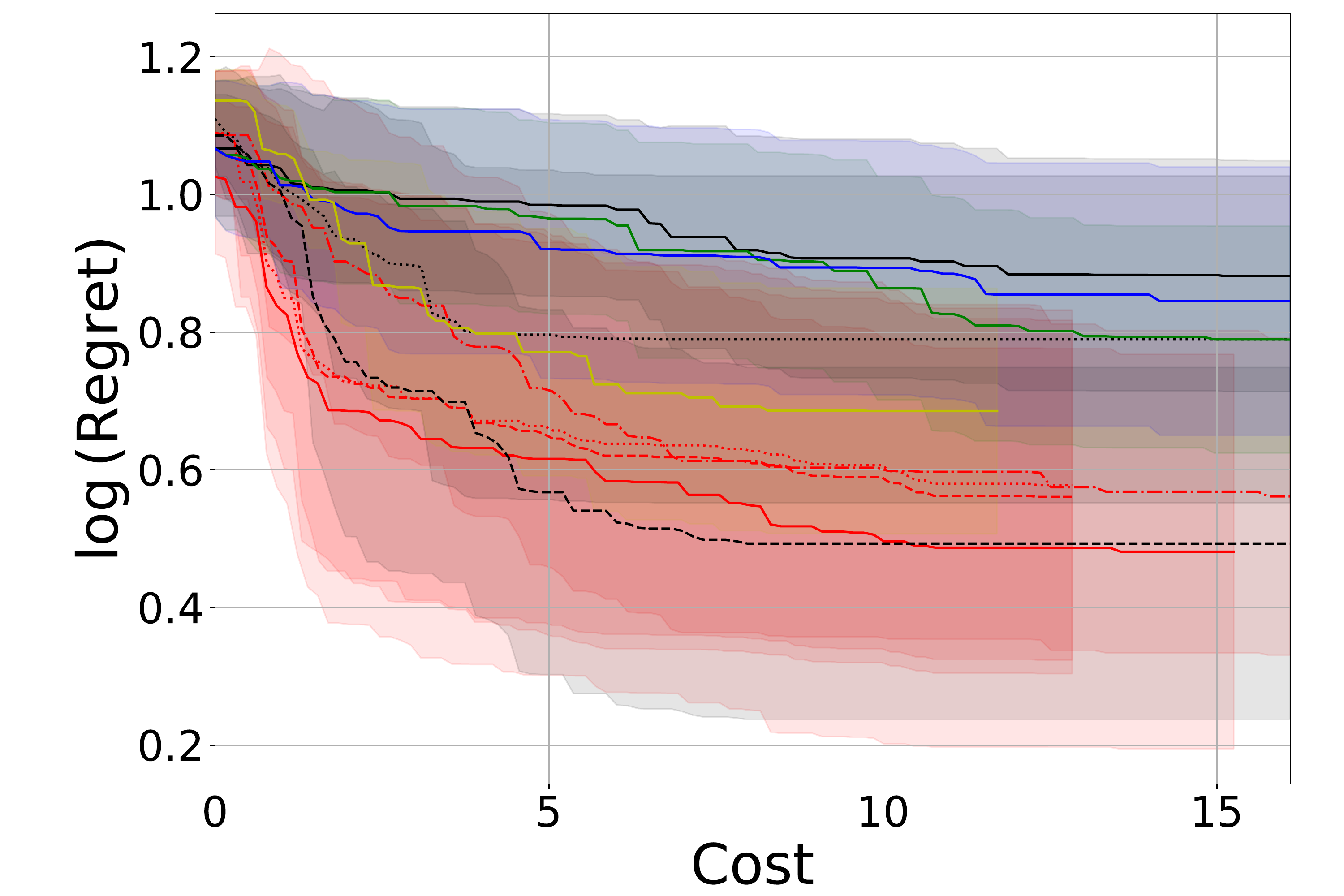}
	\includegraphics[width=0.32\textwidth]{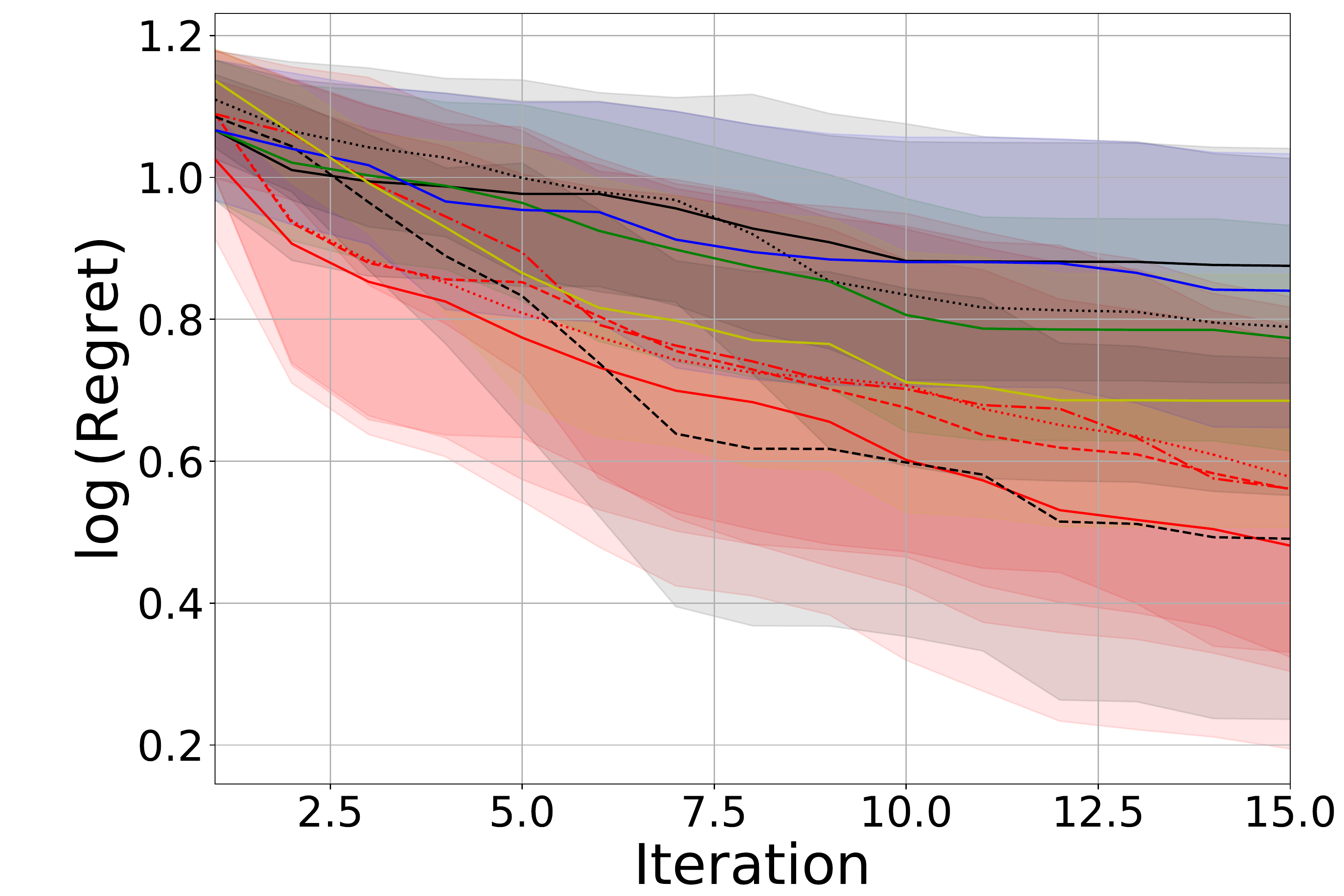}
	\includegraphics[width=0.32\textwidth]{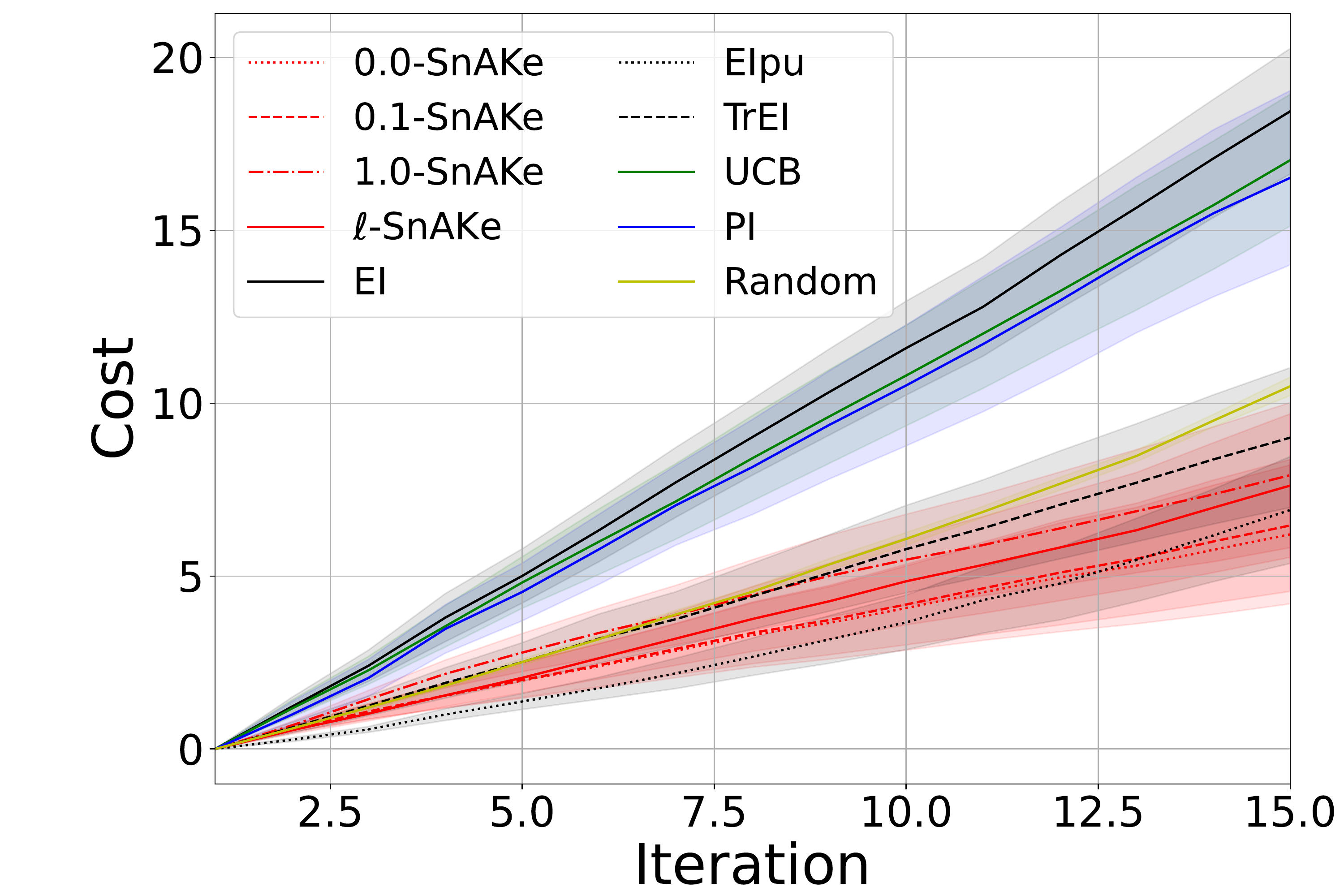}
	\caption{$T = 15$}
	\end{subfigure}
	\begin{subfigure}{\textwidth}
	\centering
	\includegraphics[width=0.32\textwidth]{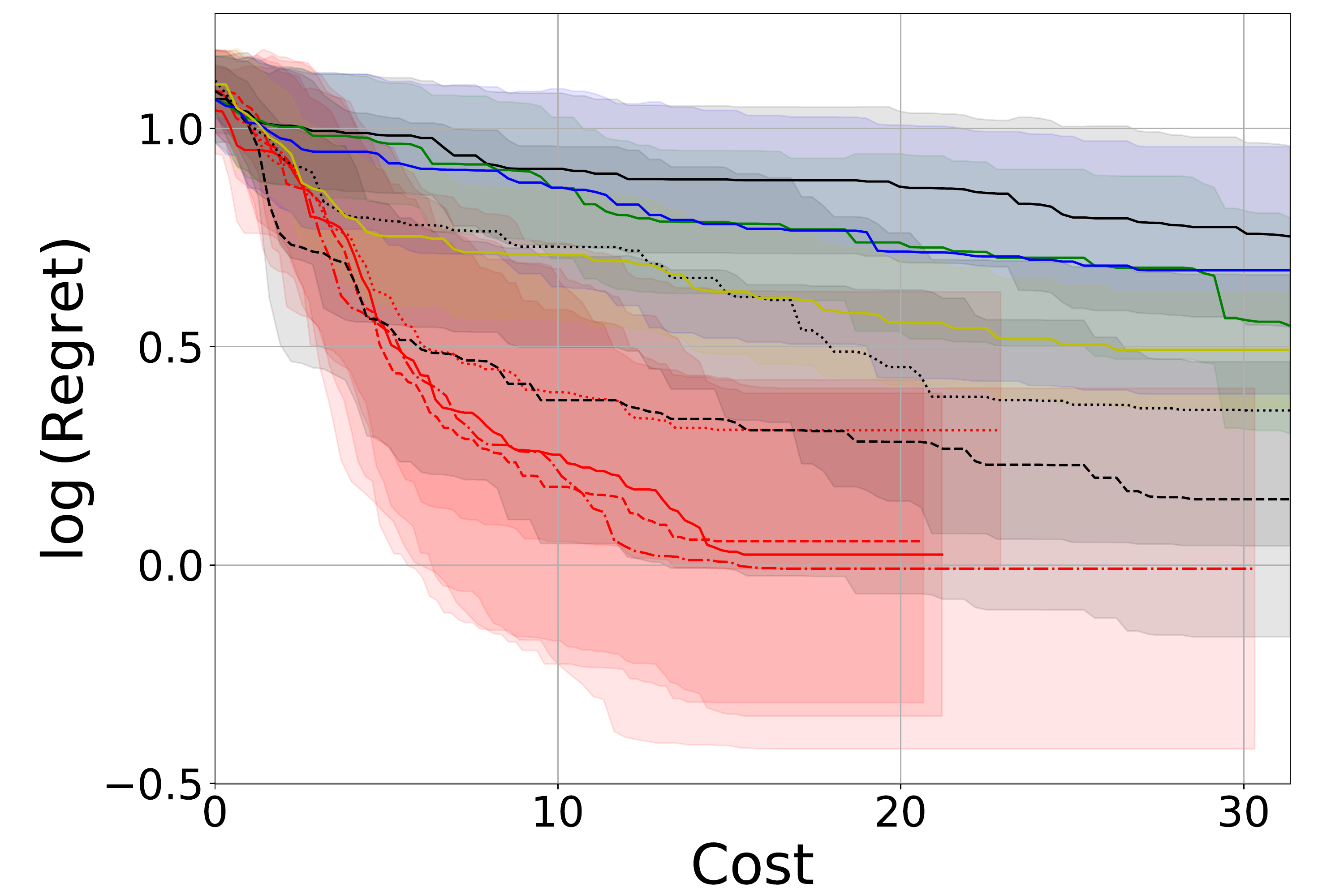}
	\includegraphics[width=0.32\textwidth]{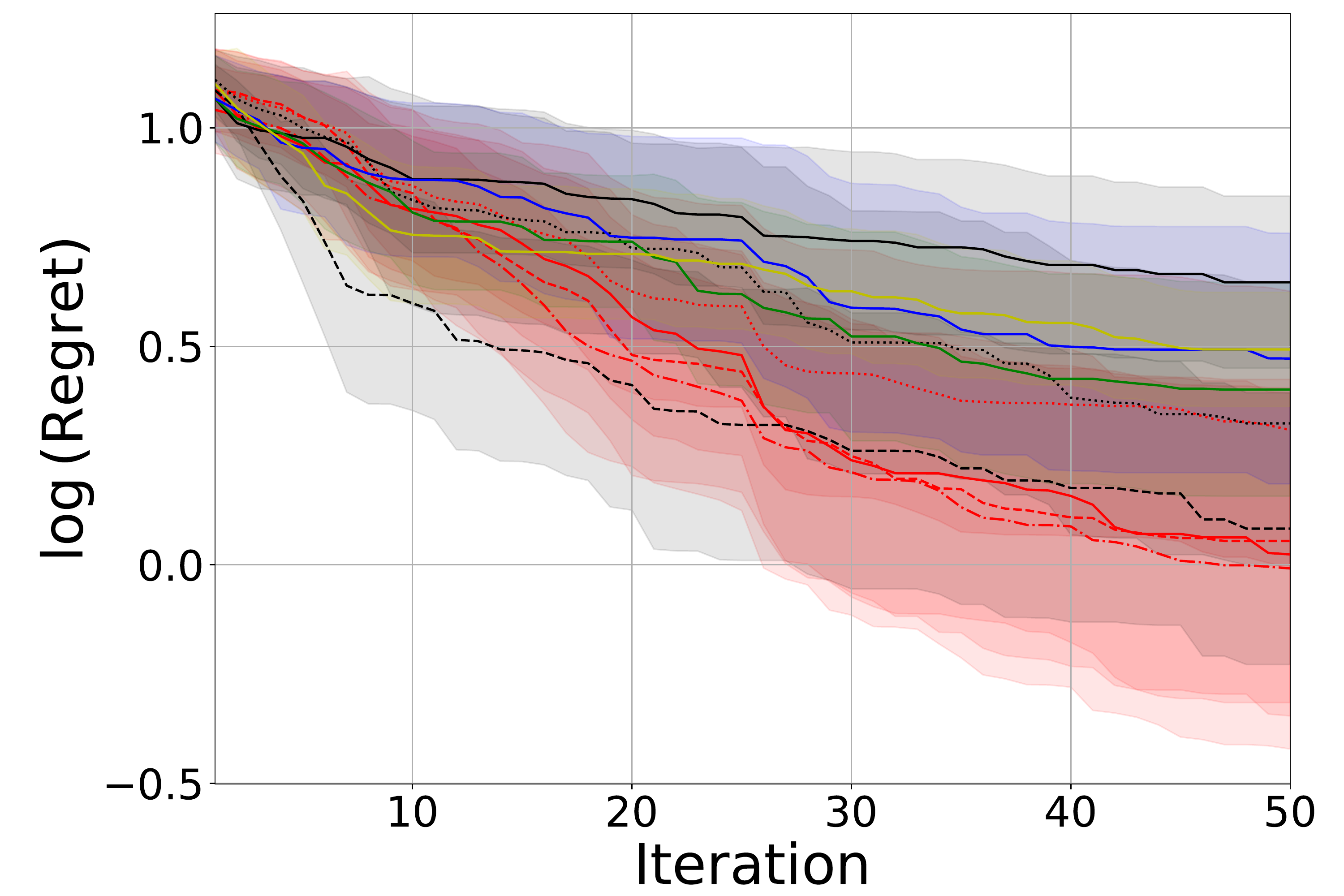}
	\includegraphics[width=0.32\textwidth]{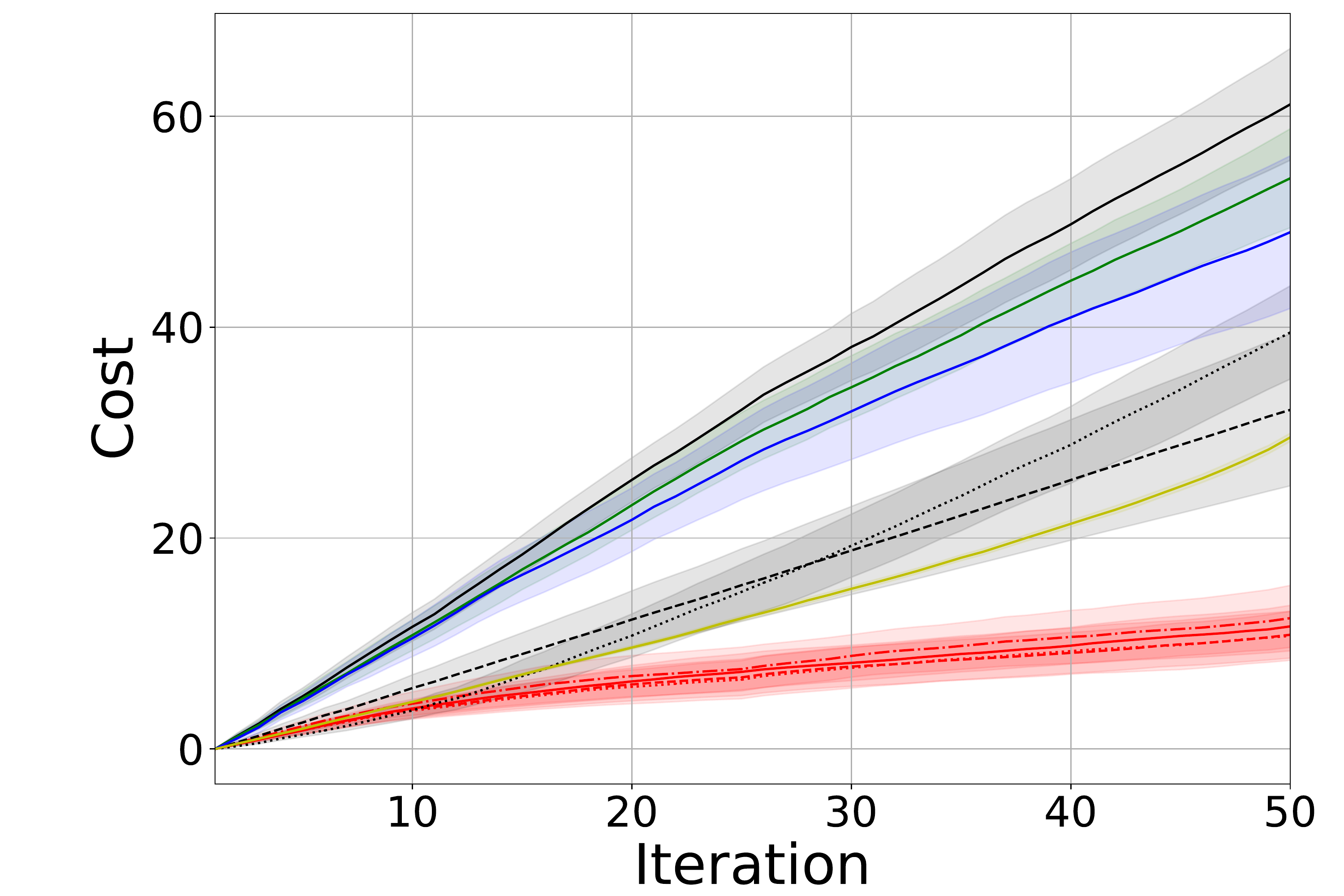}
	\caption{$T = 50$}
	\end{subfigure}
	\begin{subfigure}{\textwidth}
	\centering
	\includegraphics[width=0.32\textwidth]{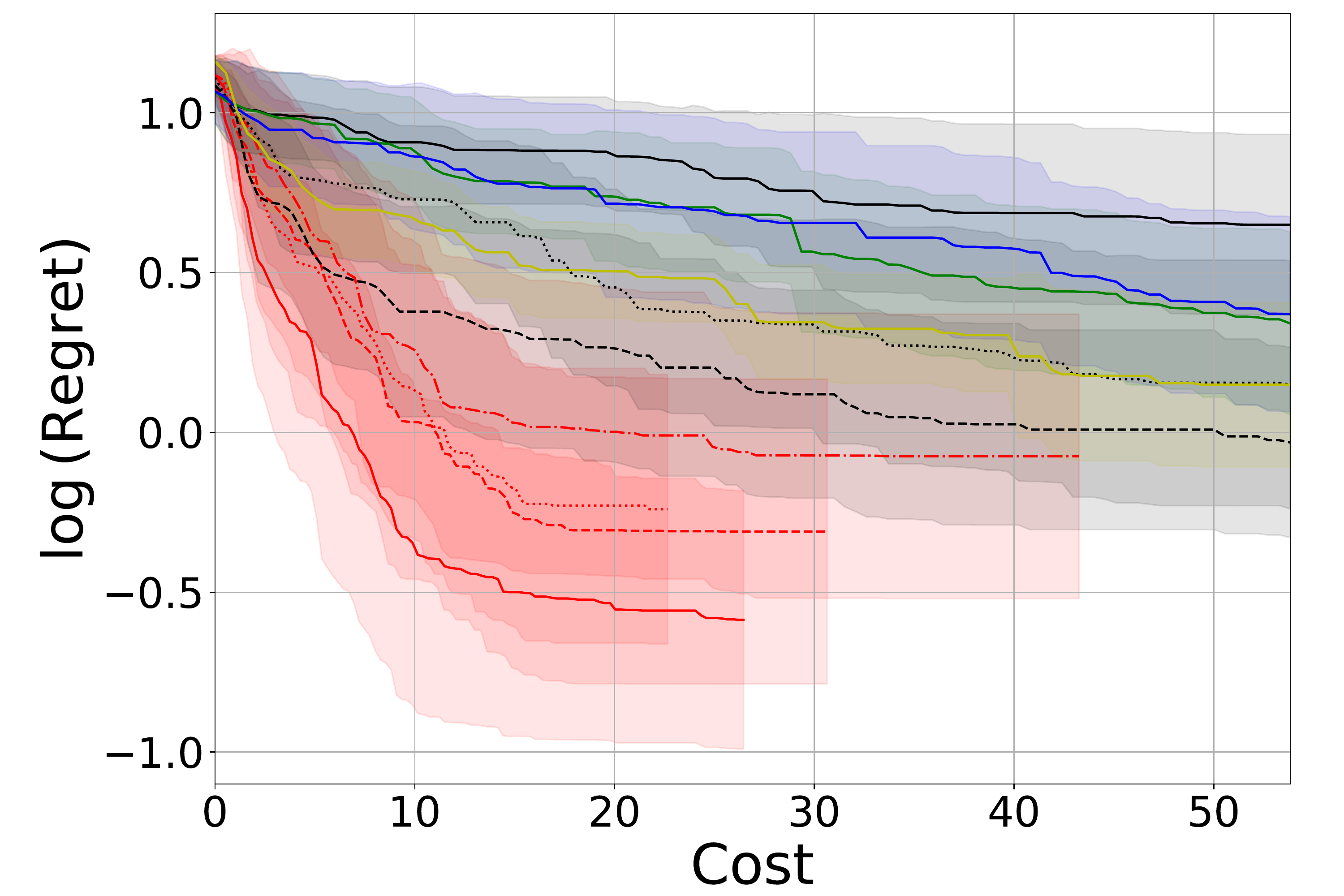}
	\includegraphics[width=0.32\textwidth]{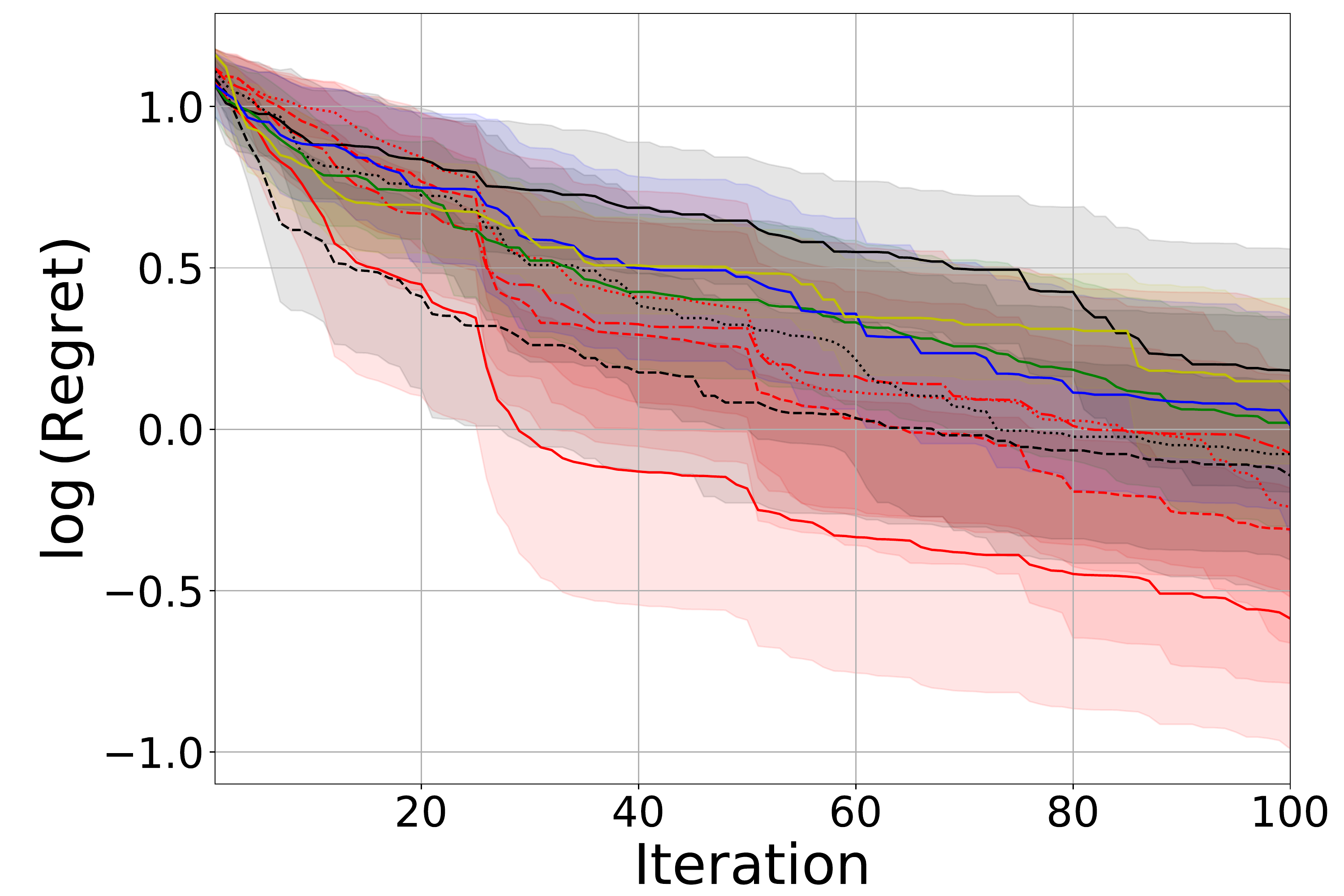}
	\includegraphics[width=0.32\textwidth]{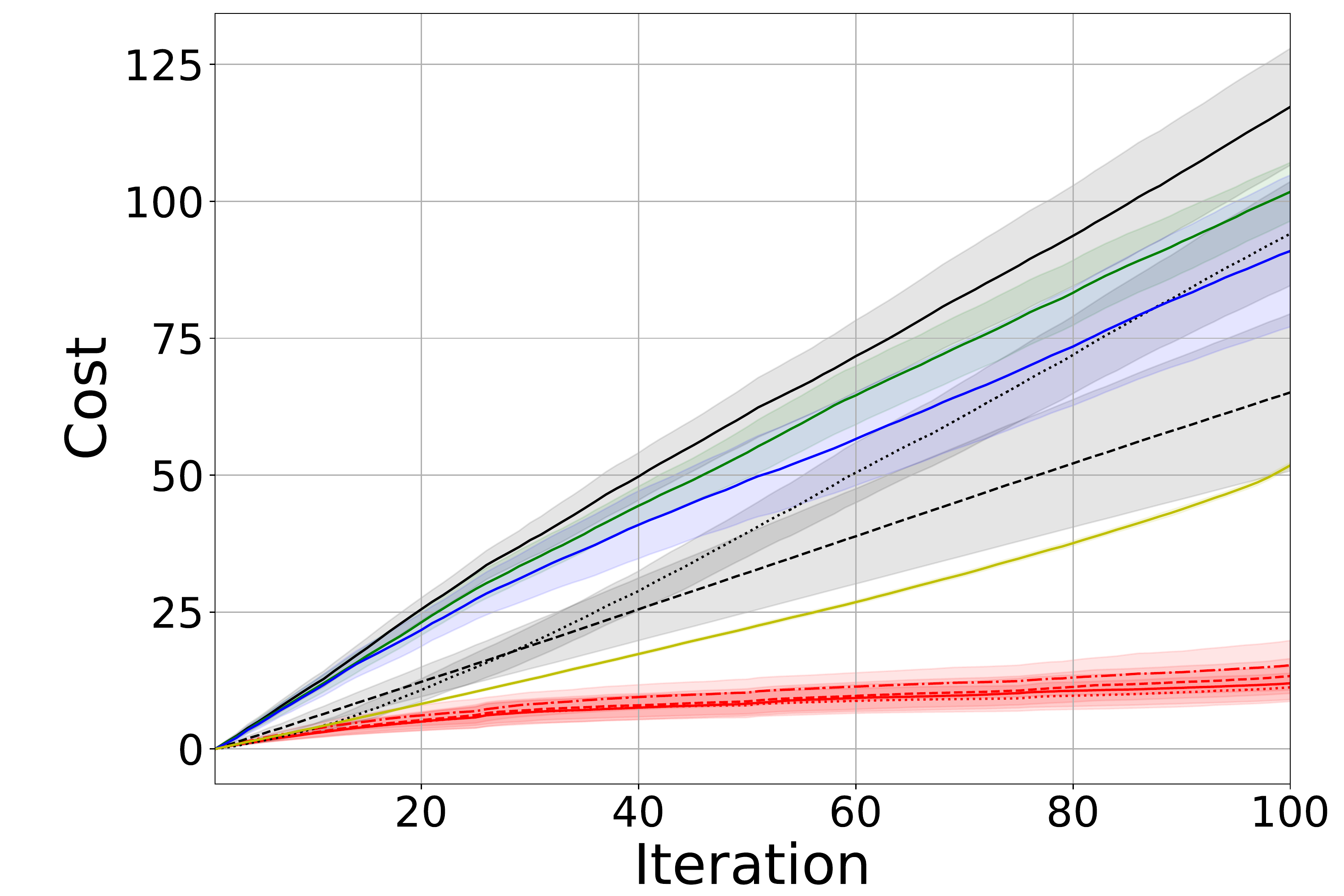}
	\caption{$T = 100$}
	\end{subfigure}
	\begin{subfigure}{\textwidth}
	\centering
	\includegraphics[width=0.32\textwidth]{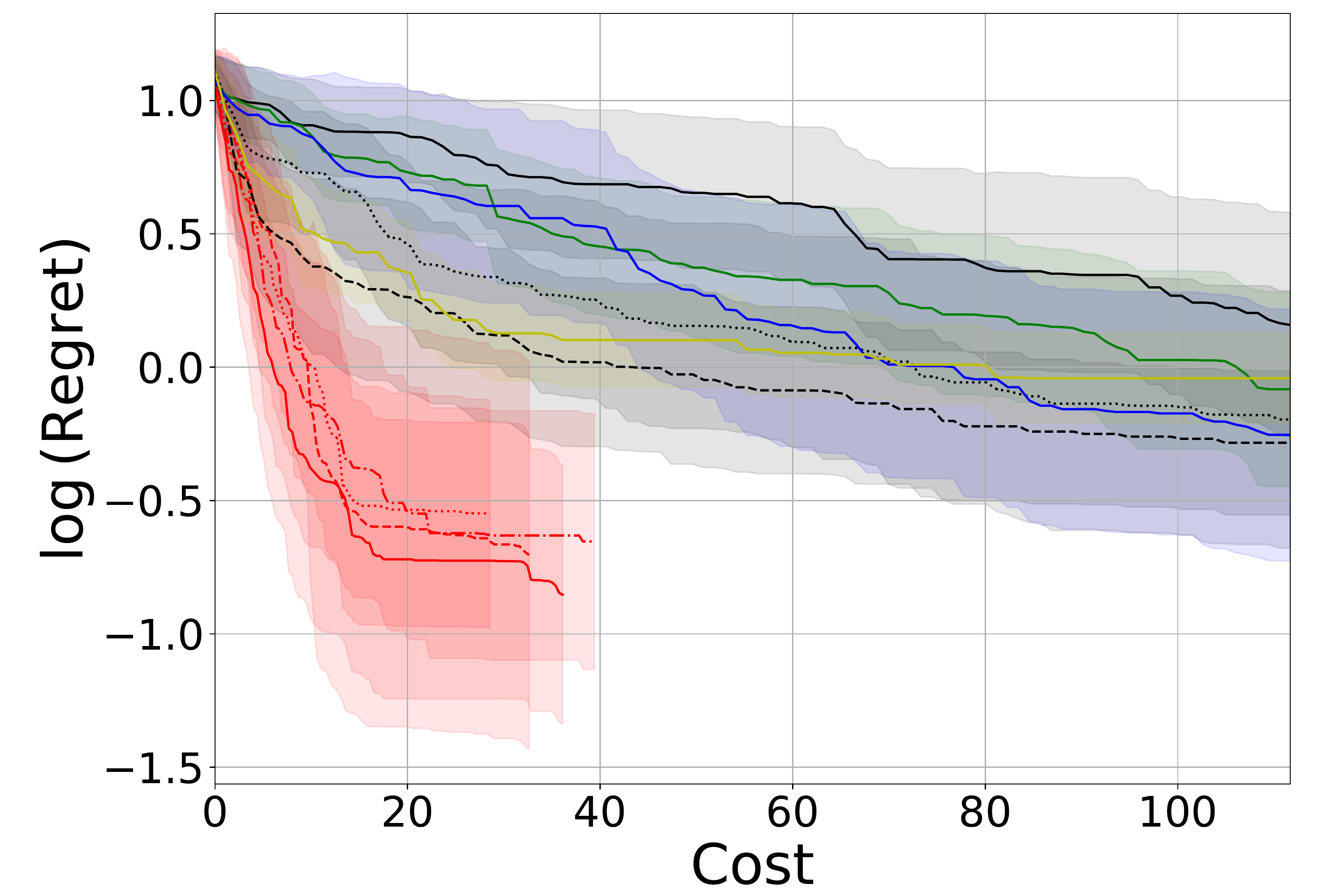}
	\includegraphics[width=0.32\textwidth]{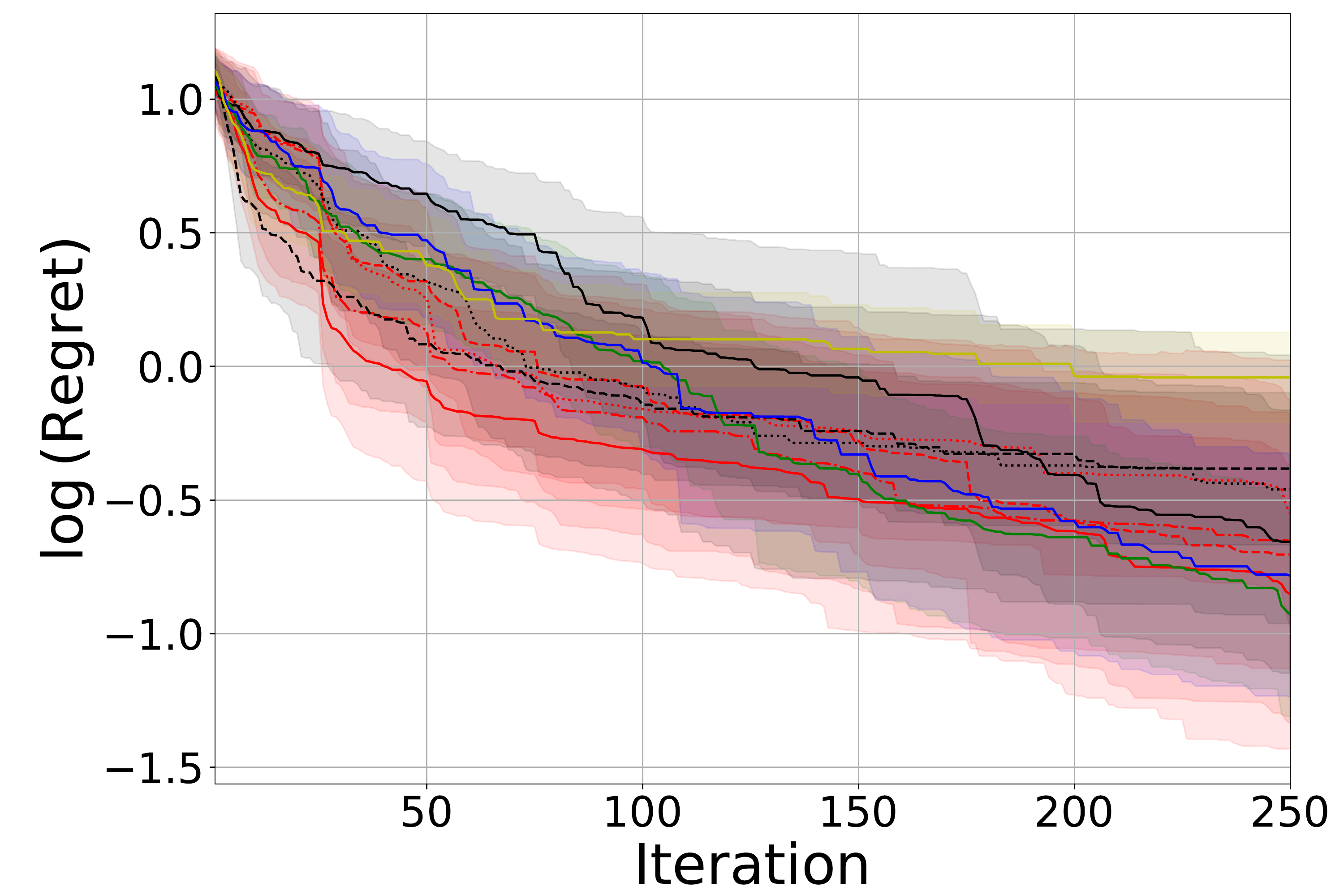}
	\includegraphics[width=0.32\textwidth]{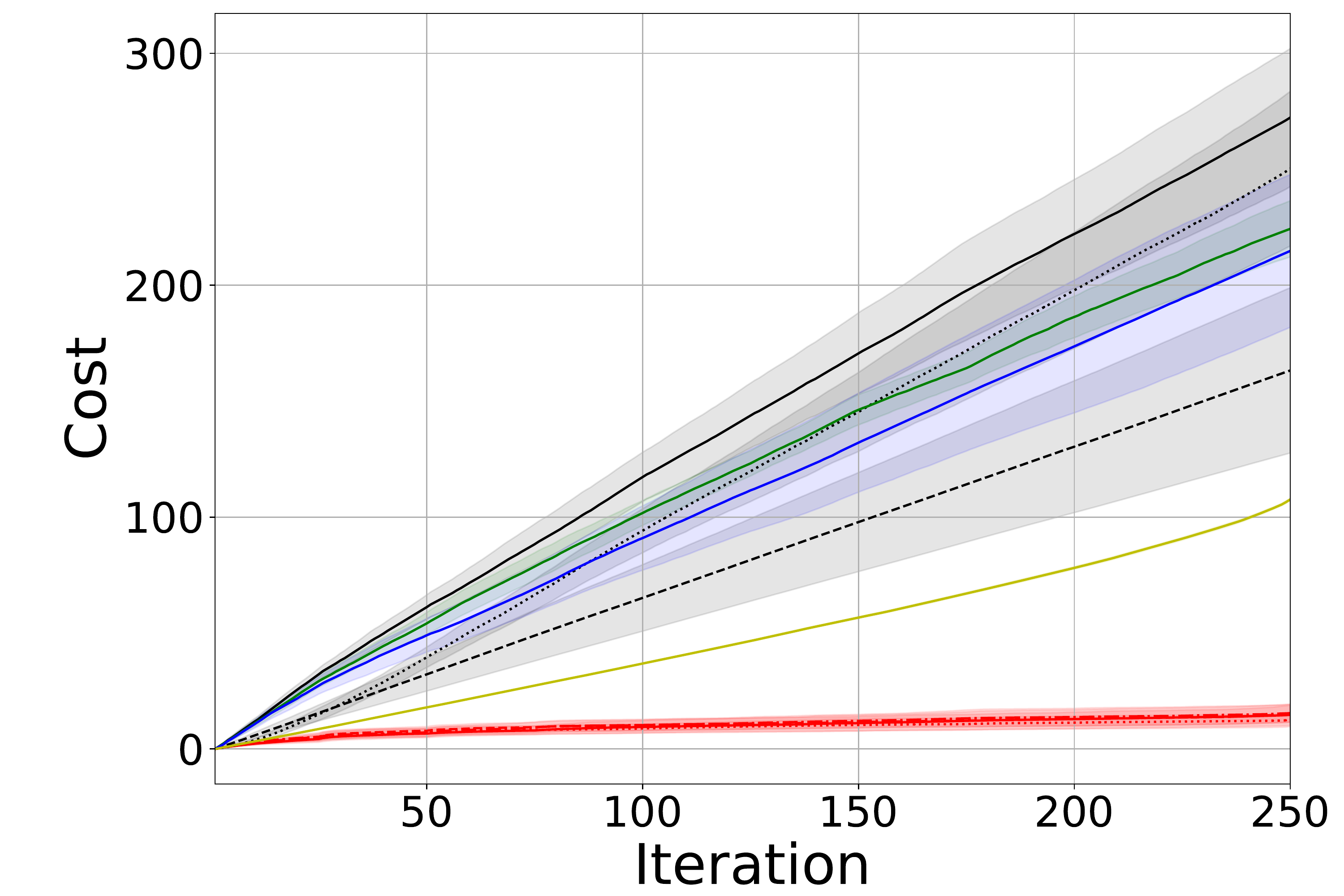}
	\caption{$T = 250$}
	\end{subfigure}
	\caption{Hartmann6D. Each row represents a different budget. The left column shows the evolution of regret against the cost used. The middle column shows the evolution of regret with iterations, and the right columns show the evolution of the 2-norm cost. A high-dimensional example where SnAKe performs exceedingly well, giving the best regret at low costs for all budgets except $T = 15$. The final cost is considerably lower for SnAKe than any other method.}
\end{figure}

\subsection{Graphs for results of Section \ref{subsec: synthetic_experiments}}
\begin{figure}[ht]
	\centering
	\begin{subfigure}[t]{\textwidth}
	\includegraphics[width = 0.32\textwidth]{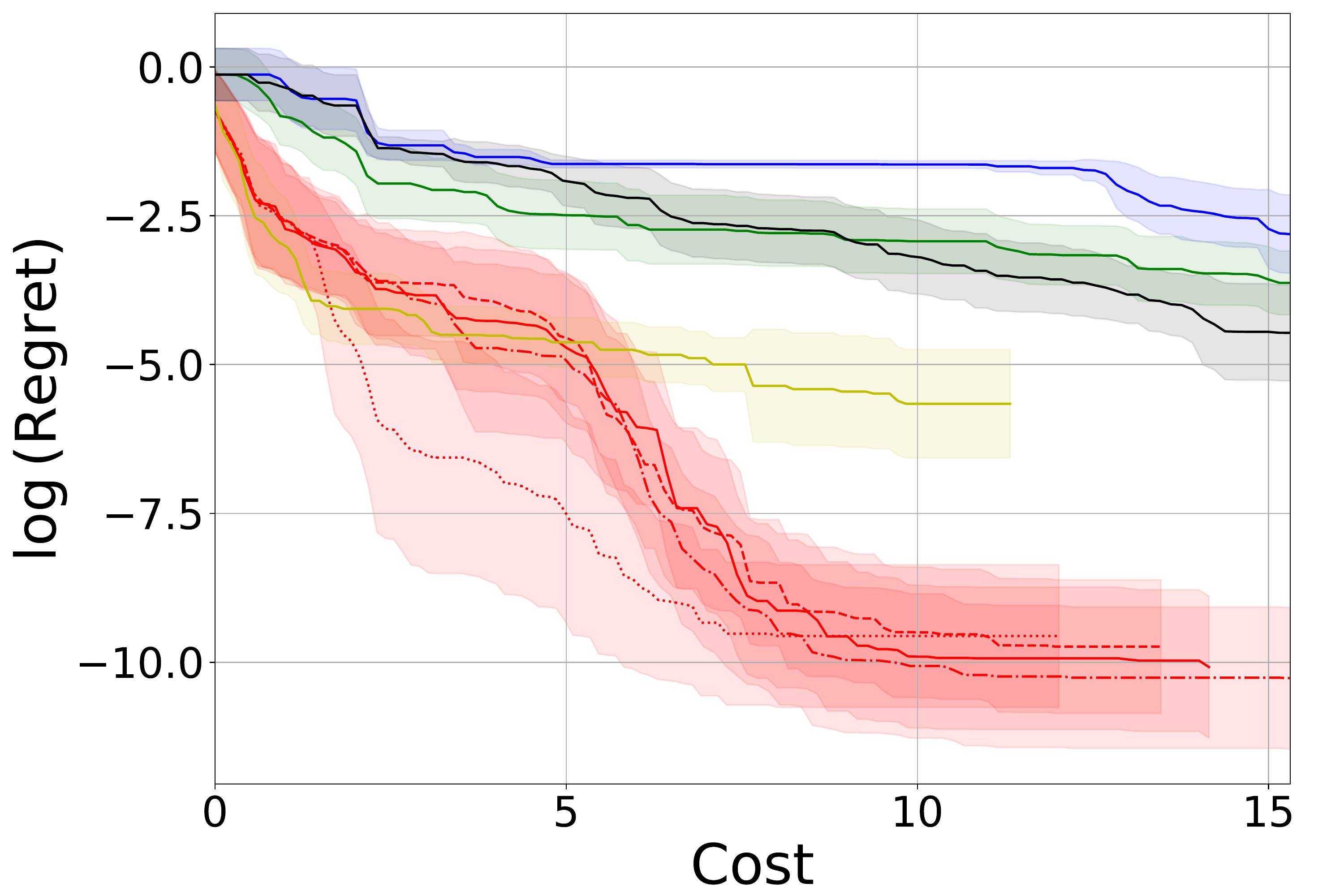}
	\includegraphics[width = 0.32\textwidth]{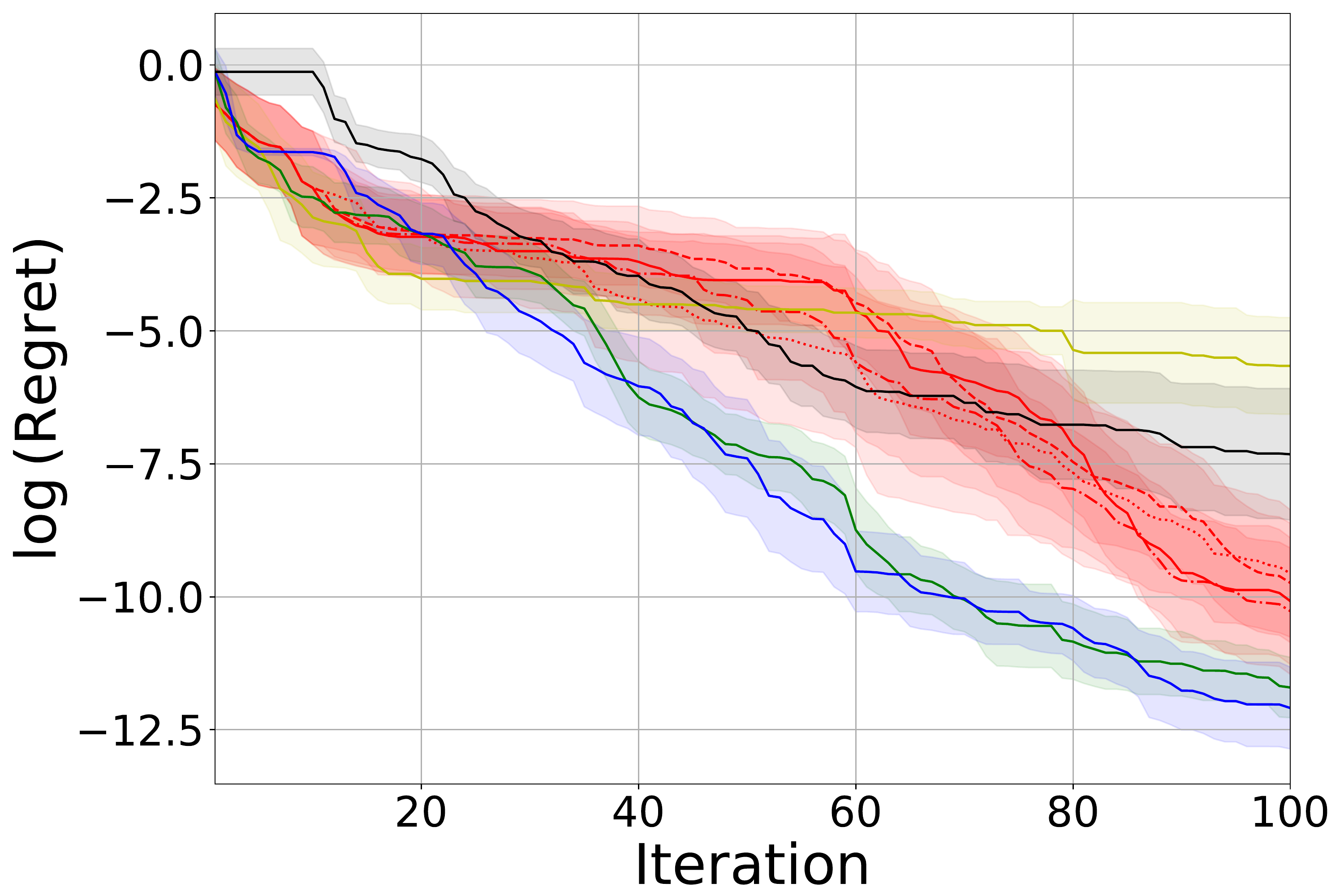}
	\includegraphics[width = 0.32\textwidth]{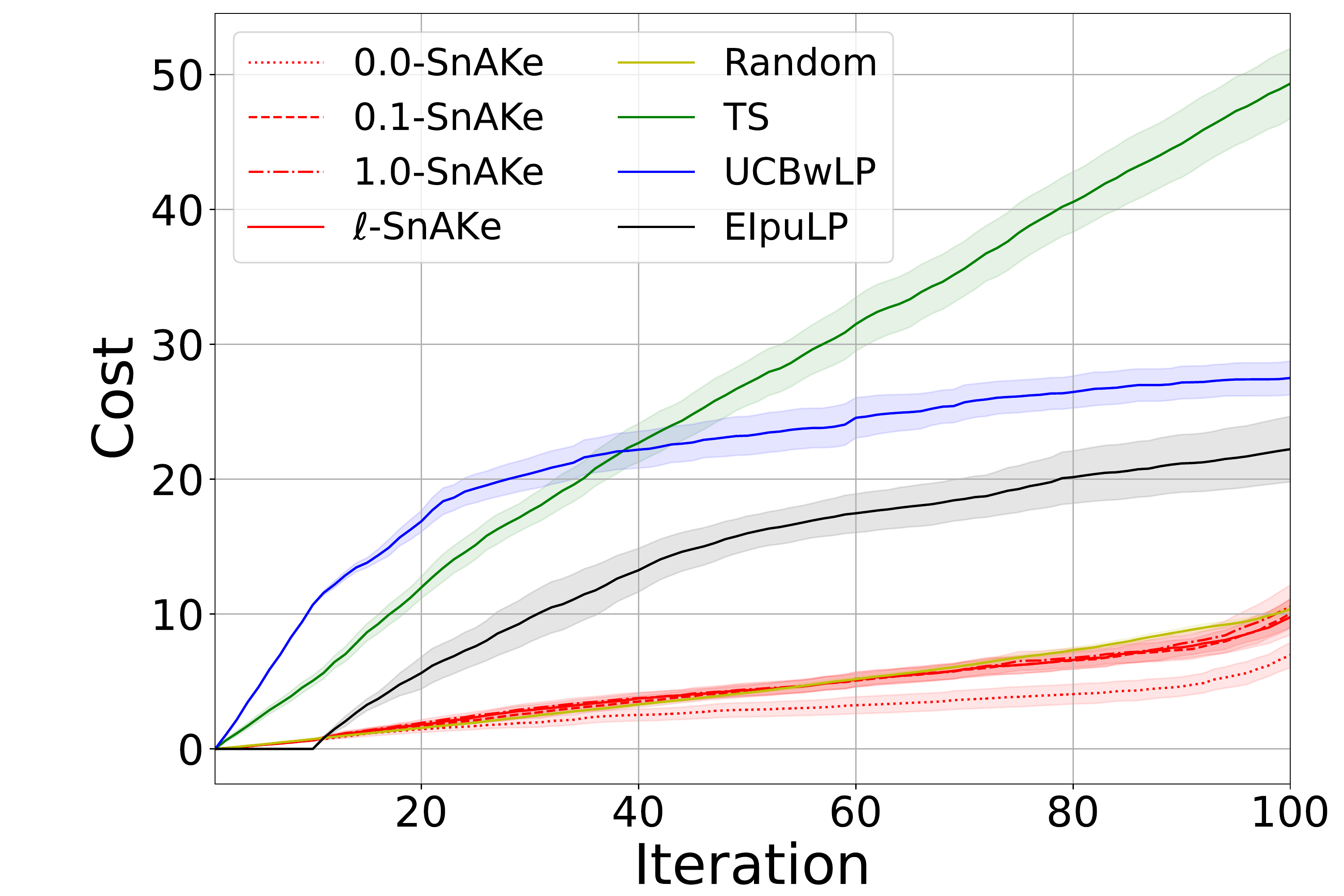}
	\caption{$T = 100$}
	\end{subfigure}
	\begin{subfigure}[t]{\textwidth}
	\includegraphics[width = 0.32\textwidth]{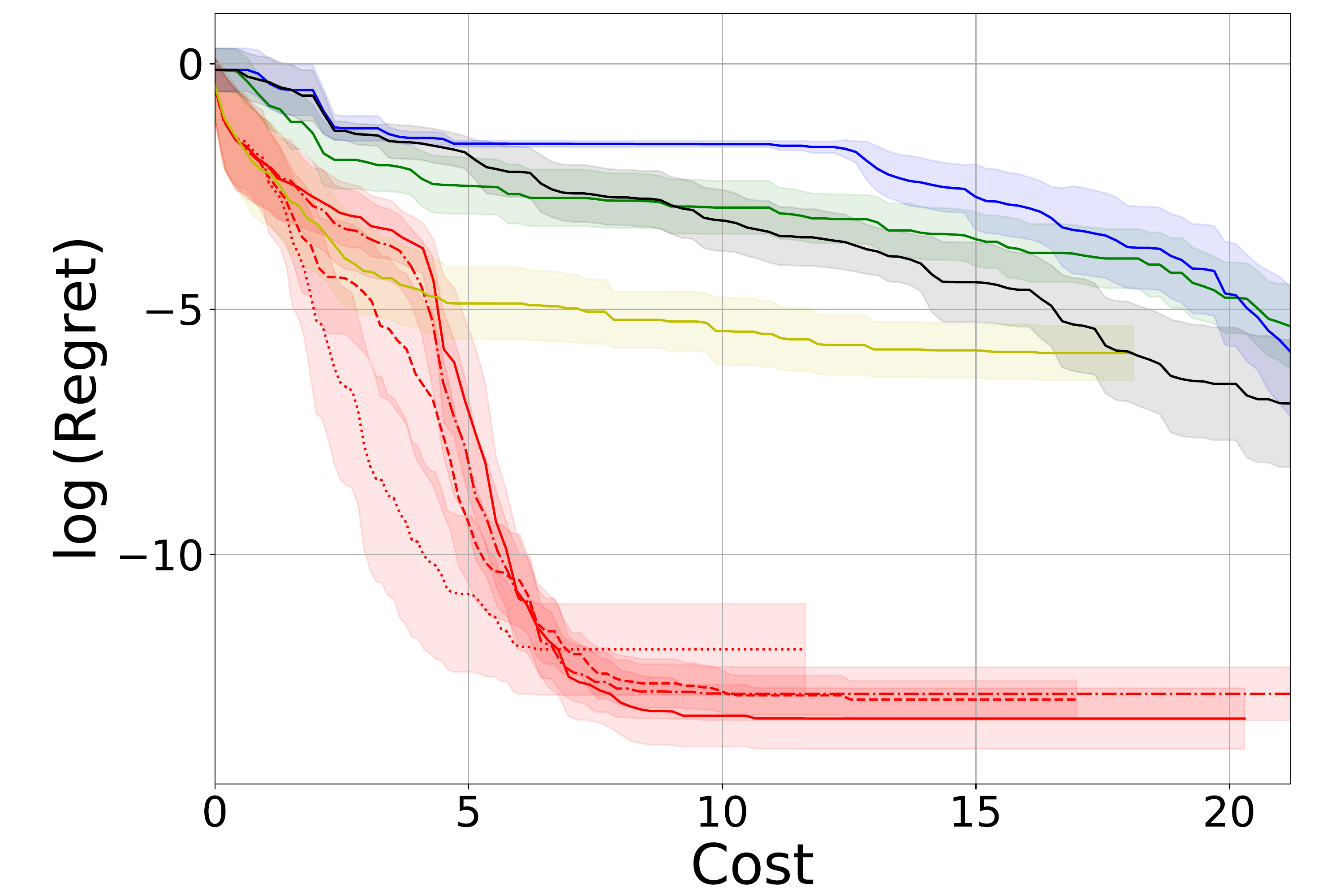}
	\includegraphics[width = 0.32\textwidth]{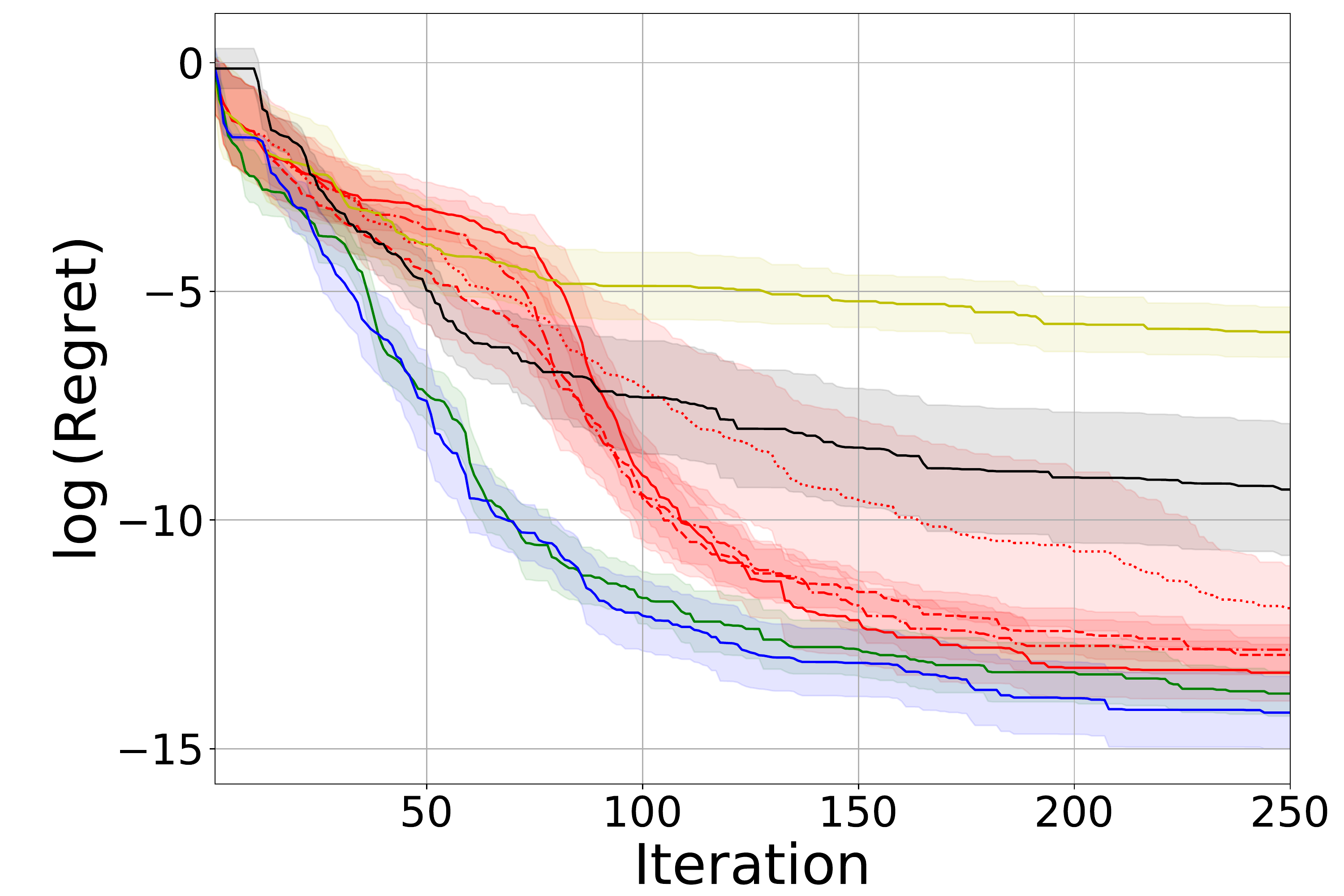}
	\includegraphics[width = 0.32\textwidth]{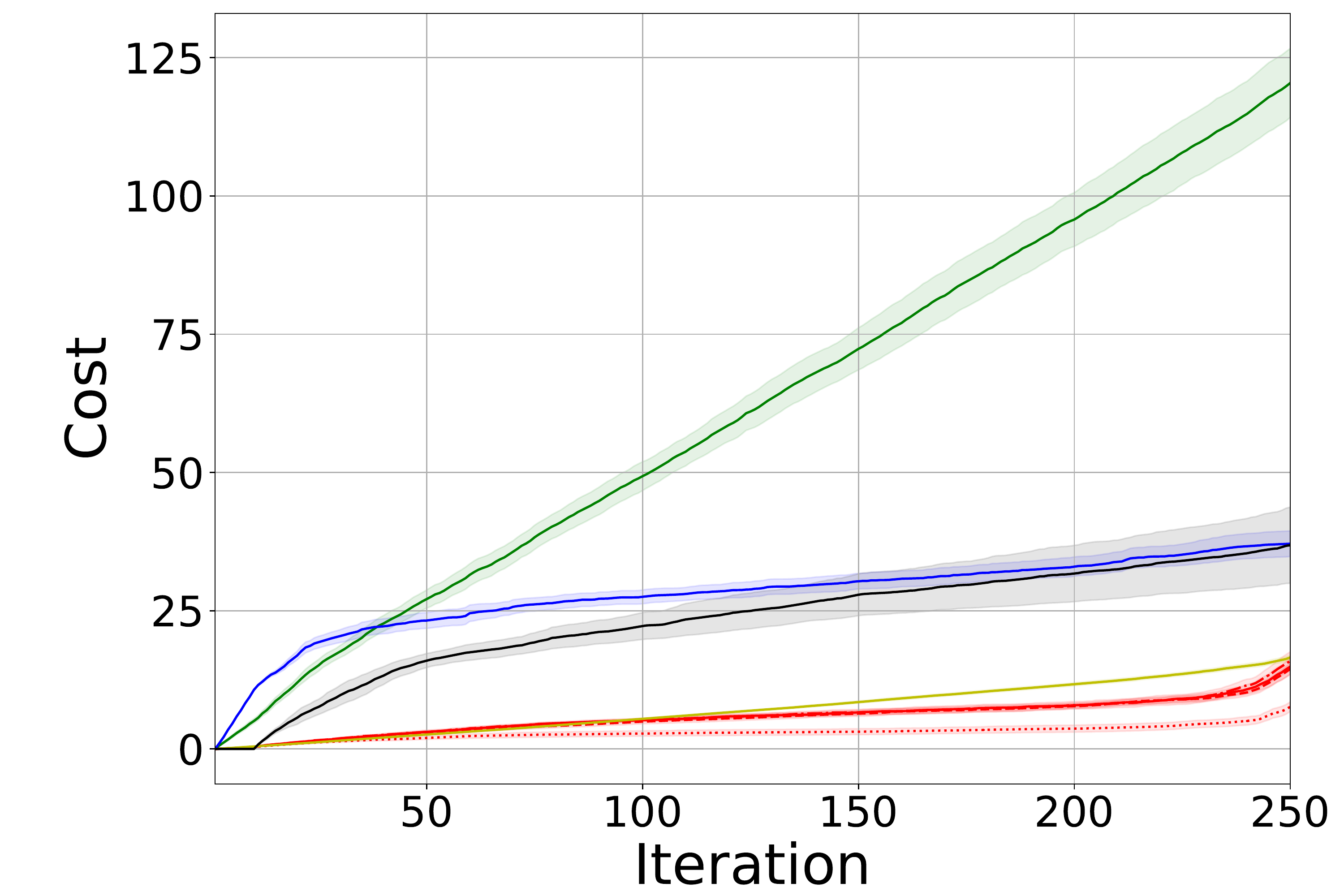}
	\caption{$T = 250$}
	\end{subfigure}
	\caption{Branin2D (Asynchronous),  $t_{delay}=10$. Each row represents a different budget. The left column shows the evolution of regret against the cost used. The middle column shows the evolution of regret with iterations, and the right columns show the evolution of the 2-norm cost. SnAKe achieves significantly better regret than all other methods at low costs. The final regret of other BO methods is slightly better, but this comes at the expense of much larger cost. EIpuLP performs poorly, as it seems Local Penalization is over-powering the cost term.}
	\label{fig: branin2d_10_async}
\end{figure}

\begin{figure}[ht]
	\centering
	\begin{subfigure}[t]{\textwidth}
	\includegraphics[width = 0.32\textwidth]{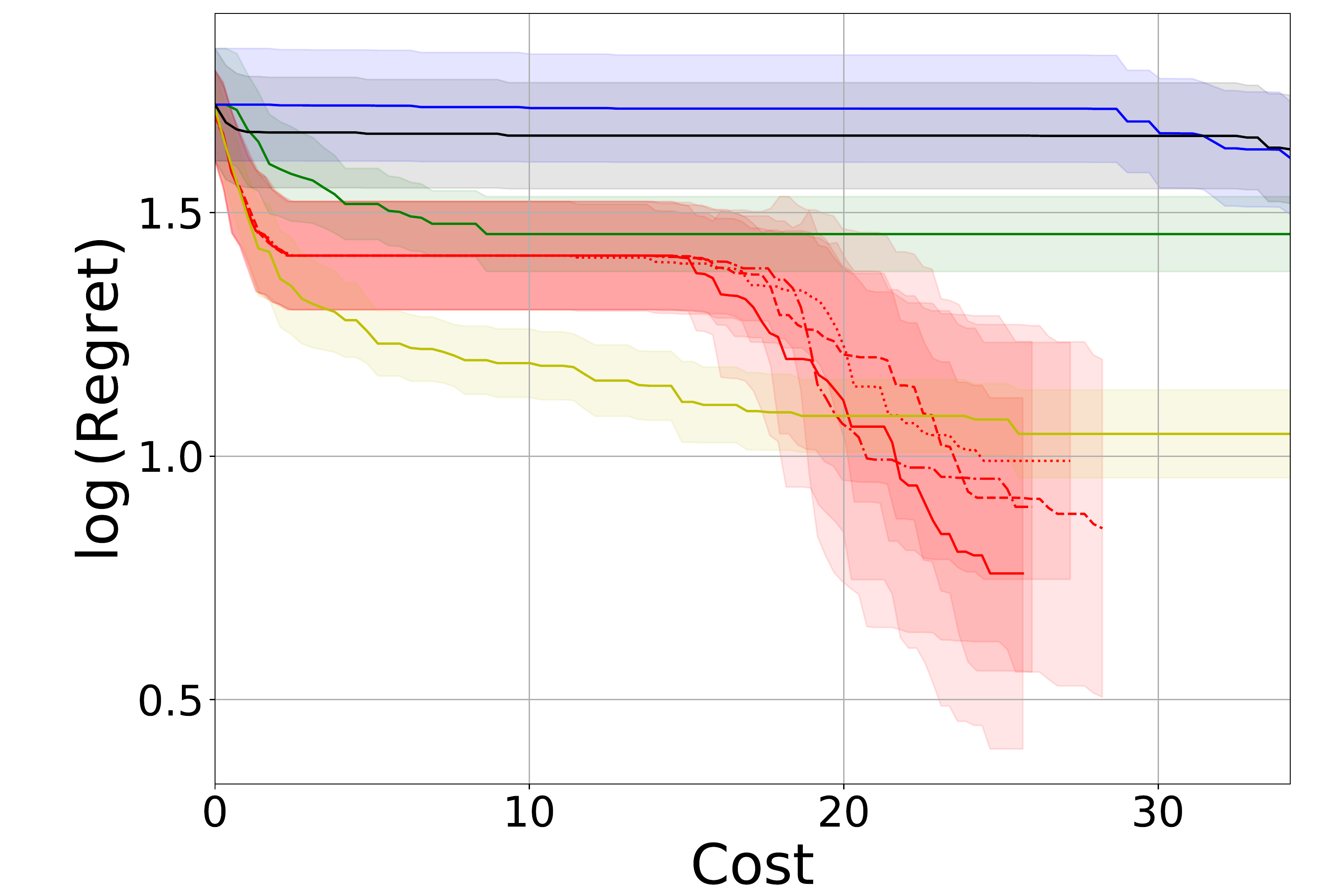}
	\includegraphics[width = 0.32\textwidth]{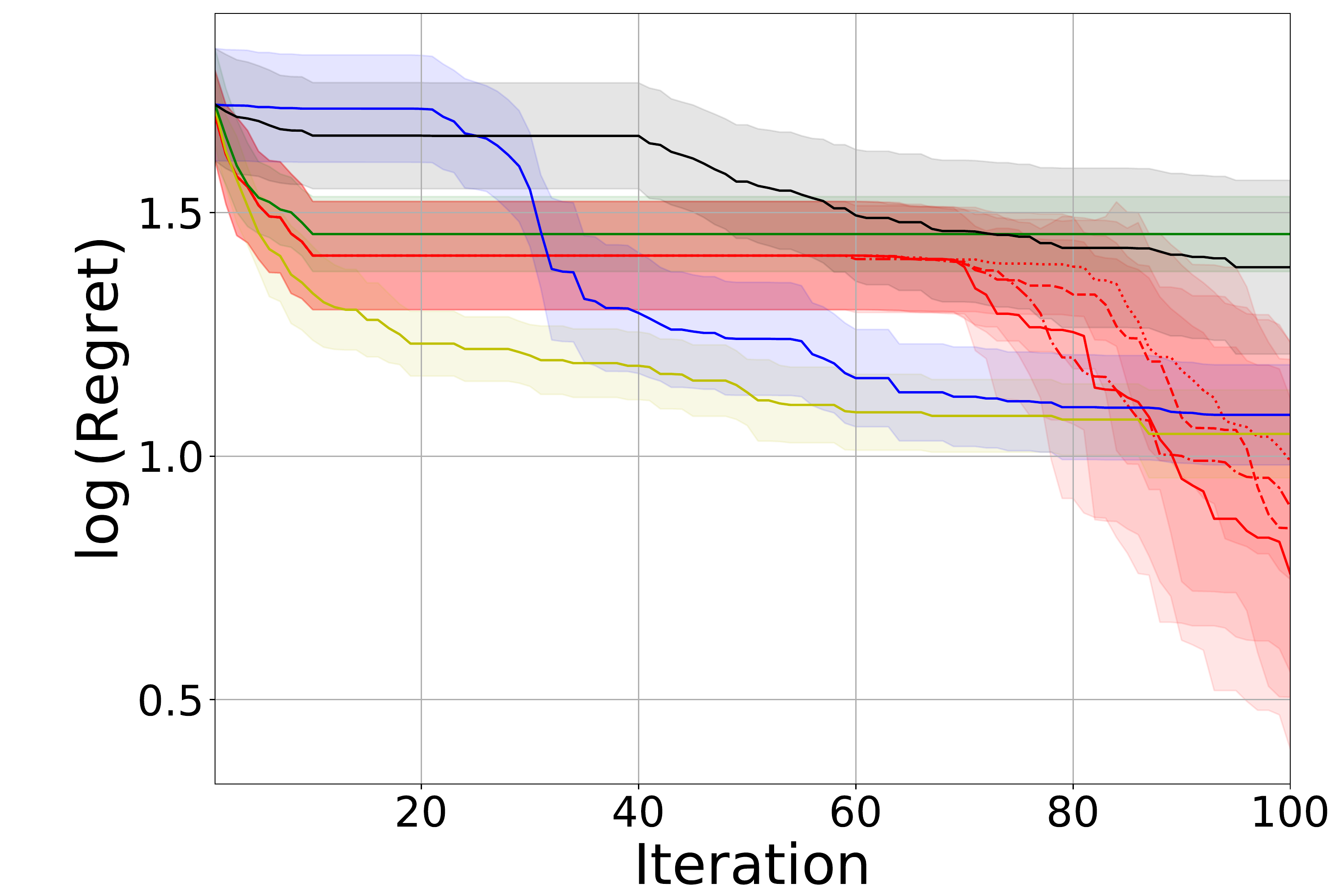}
	\includegraphics[width = 0.32\textwidth]{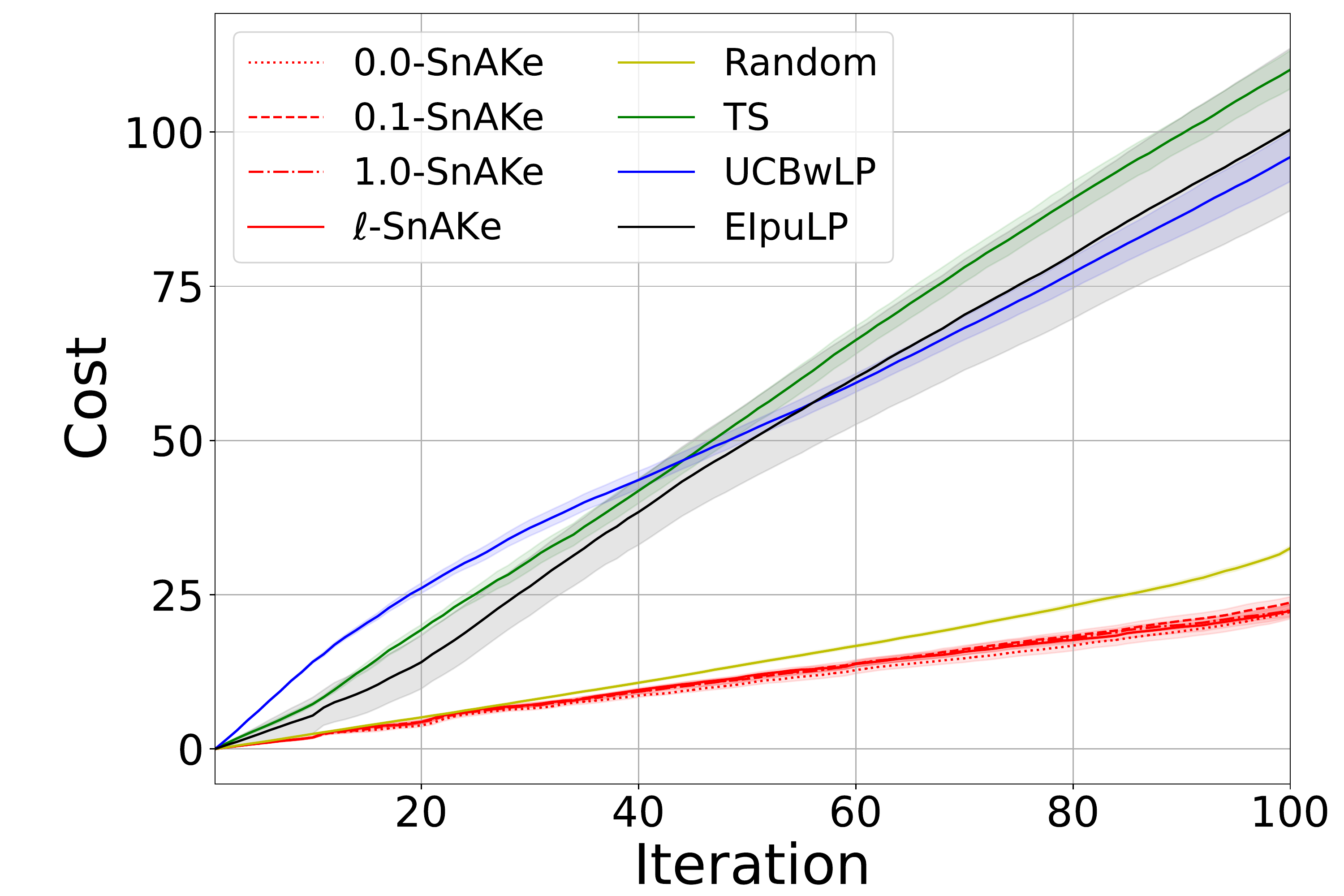}
	\caption{$T = 100$}
	\end{subfigure}
	\begin{subfigure}[t]{\textwidth}
	\includegraphics[width = 0.32\textwidth]{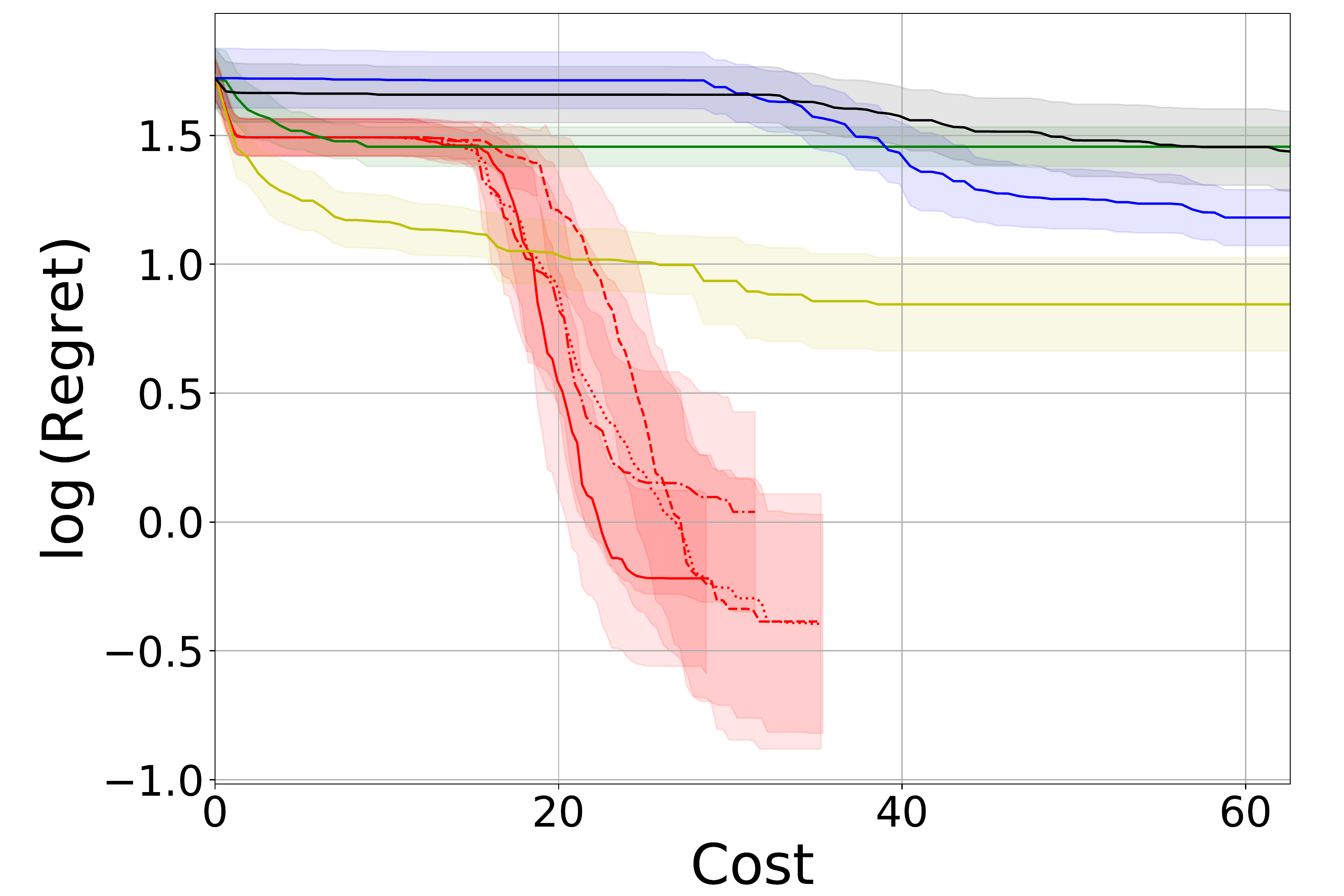}
	\includegraphics[width = 0.32\textwidth]{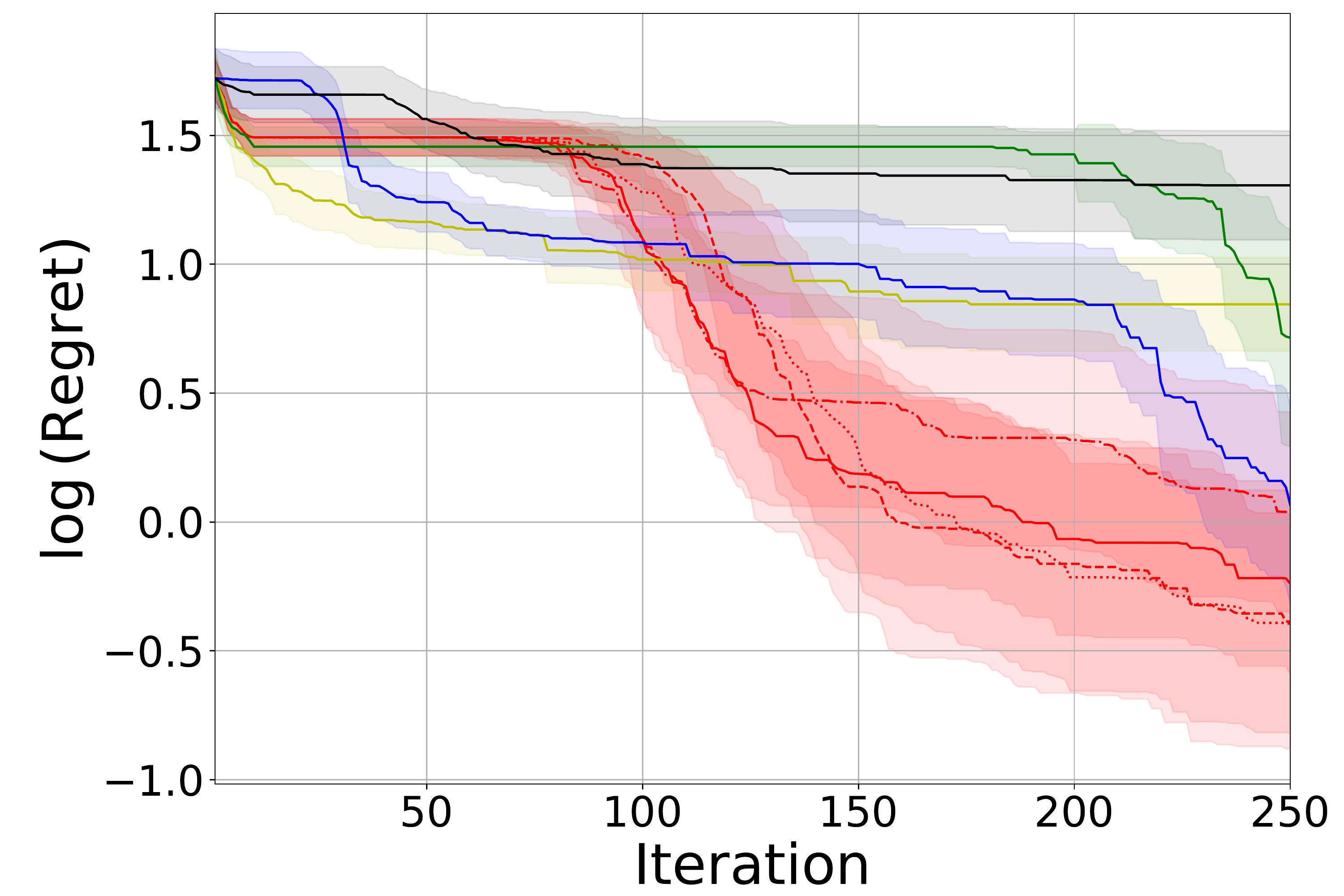}
	\includegraphics[width = 0.32\textwidth]{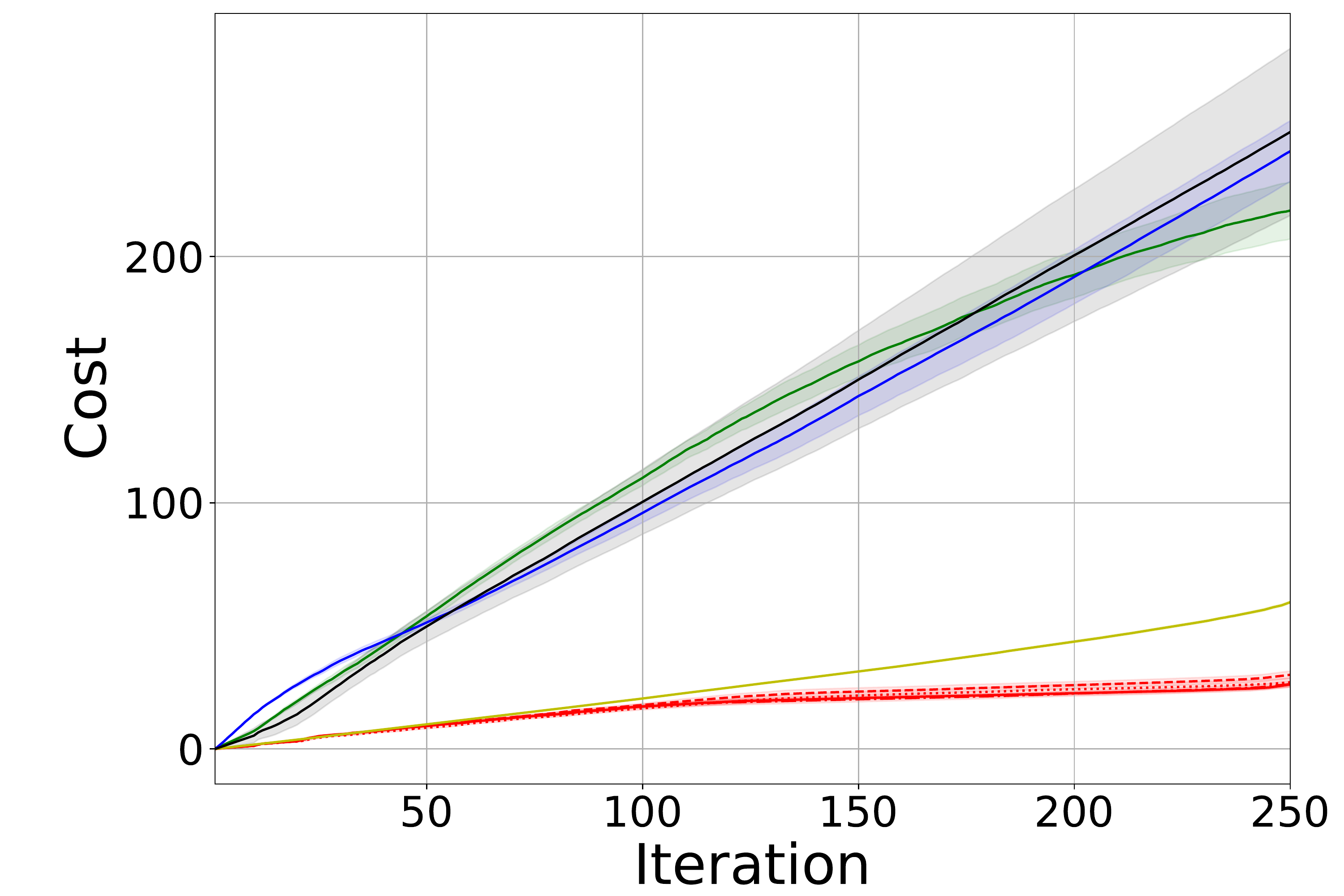}
	\caption{$T = 250$}
	\end{subfigure}
	\caption{Ackley4D (Asynchronous),  $t_{delay}=10$. Each row represents a different budget. The left column shows the evolution of regret against the cost used. The middle column shows the evolution of regret with iterations, and the right columns show the evolution of the 2-norm cost. For the larger budget, SnAKe outperforms all other methods in both regret and cost. Interestingly, the performance of SnAKe \textit{improves} when adding delay (see Figure \ref{fig: ackley_sync_4d_250} for synchronous results). EIpuLP performs poorly, as it seems Local Penalization is over-powering the cost term.}
	\label{fig: ackley_async_4d_10}
\end{figure}

\begin{figure}[ht]
	\centering
	\begin{subfigure}[t]{\textwidth}
	\includegraphics[width = 0.32\textwidth]{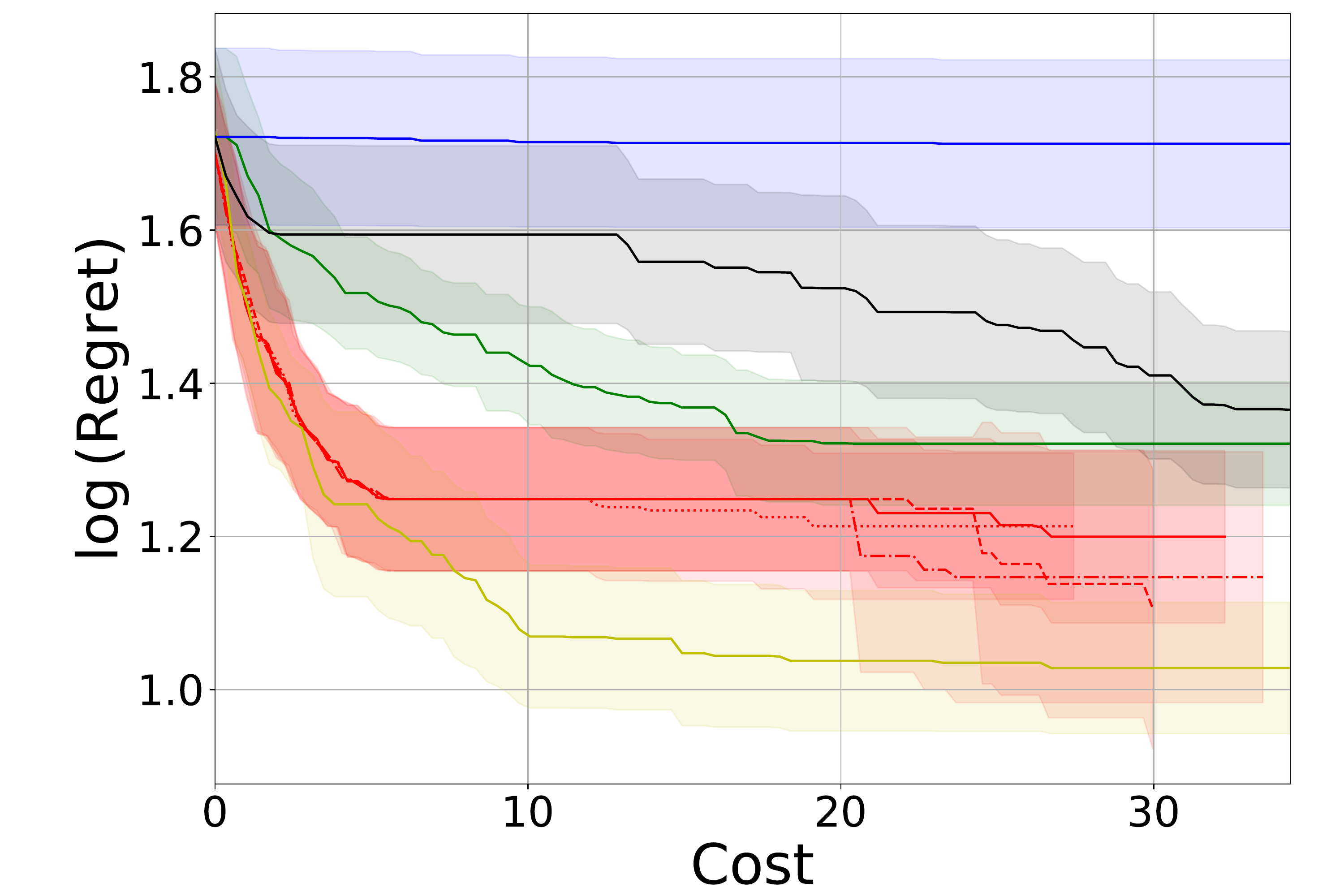}
	\includegraphics[width = 0.32\textwidth]{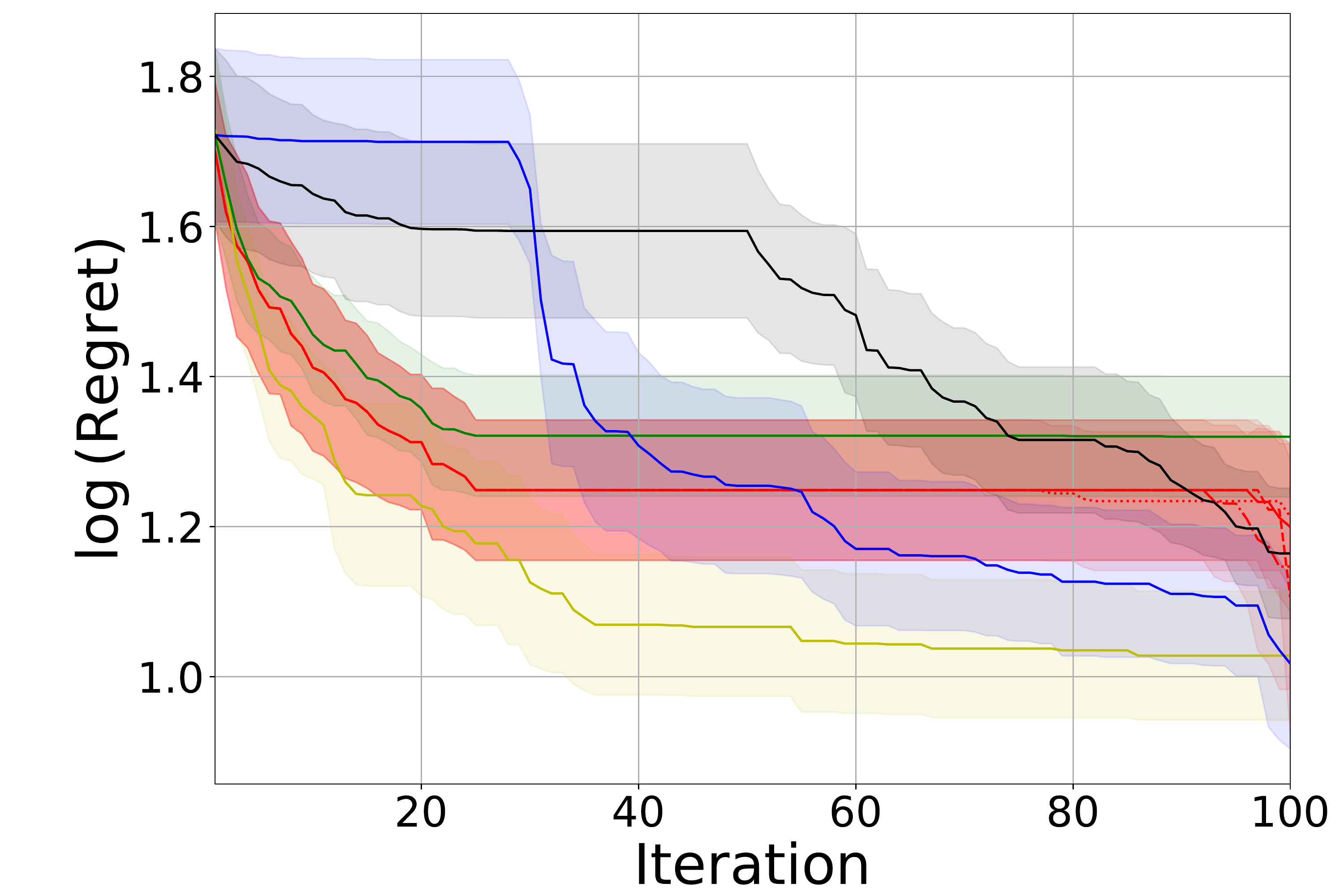}
	\includegraphics[width = 0.32\textwidth]{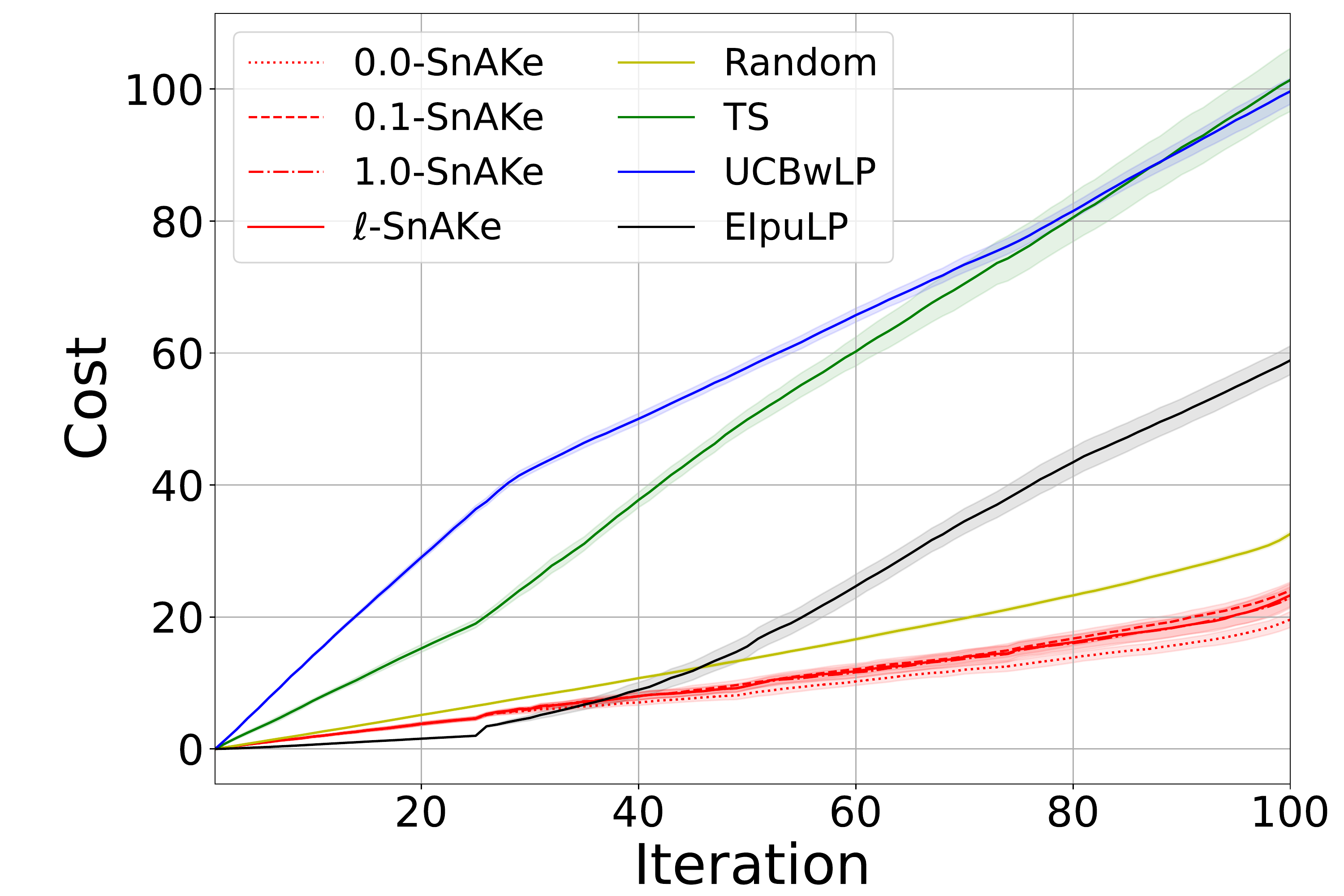}
	\caption{$T = 100$}
	\end{subfigure}
	\begin{subfigure}[t]{\textwidth}
	\includegraphics[width = 0.32\textwidth]{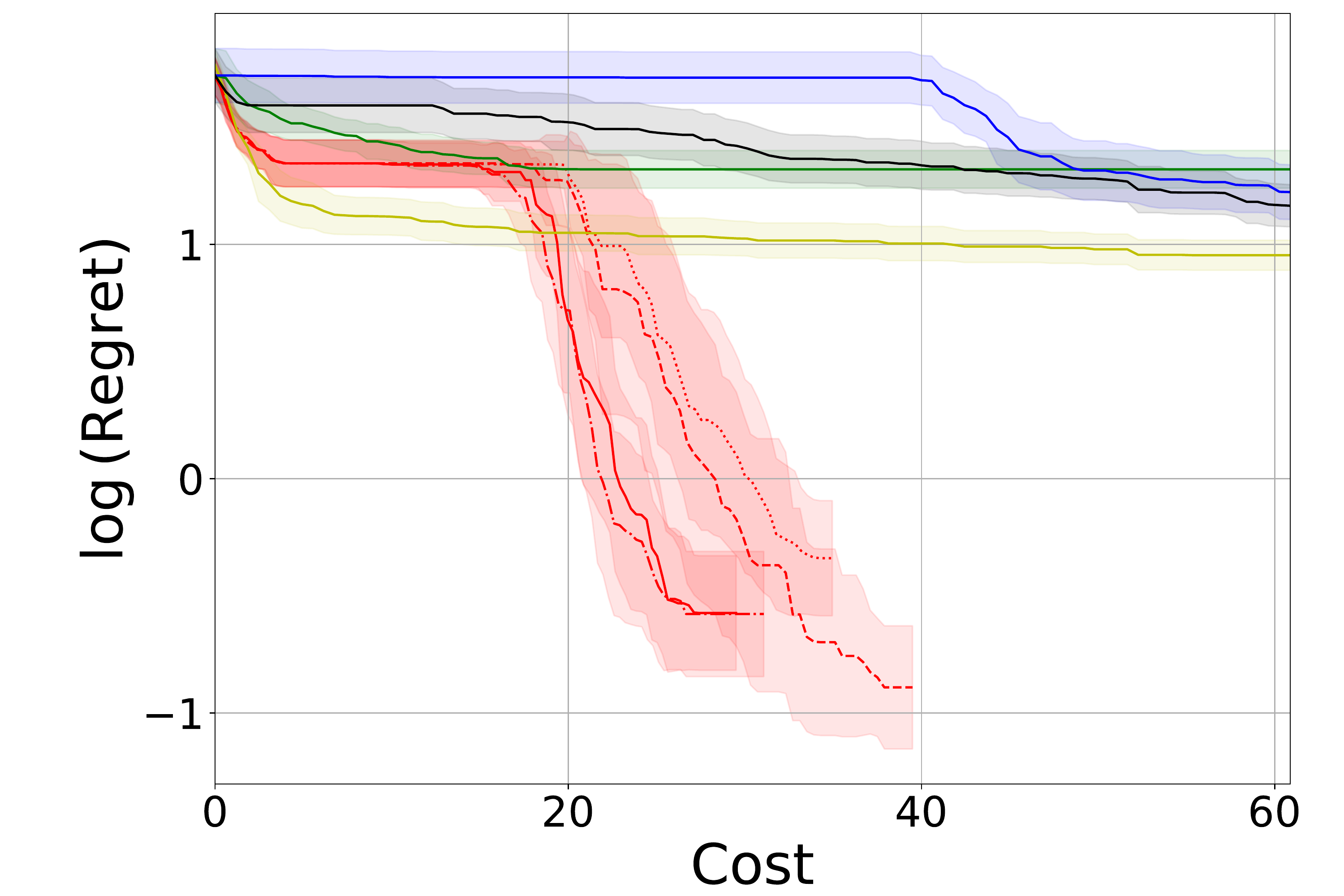}
	\includegraphics[width = 0.32\textwidth]{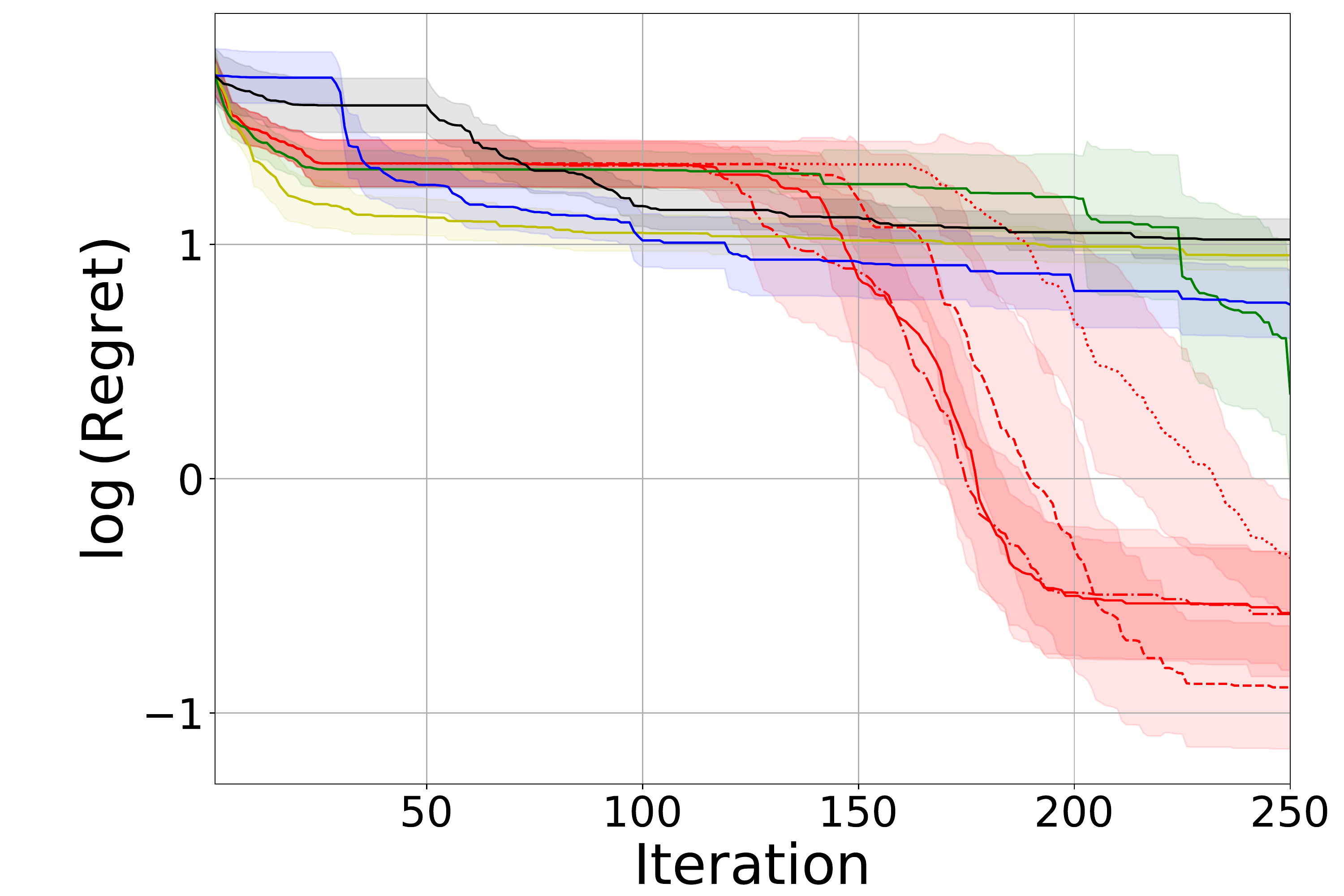}
	\includegraphics[width = 0.32\textwidth]{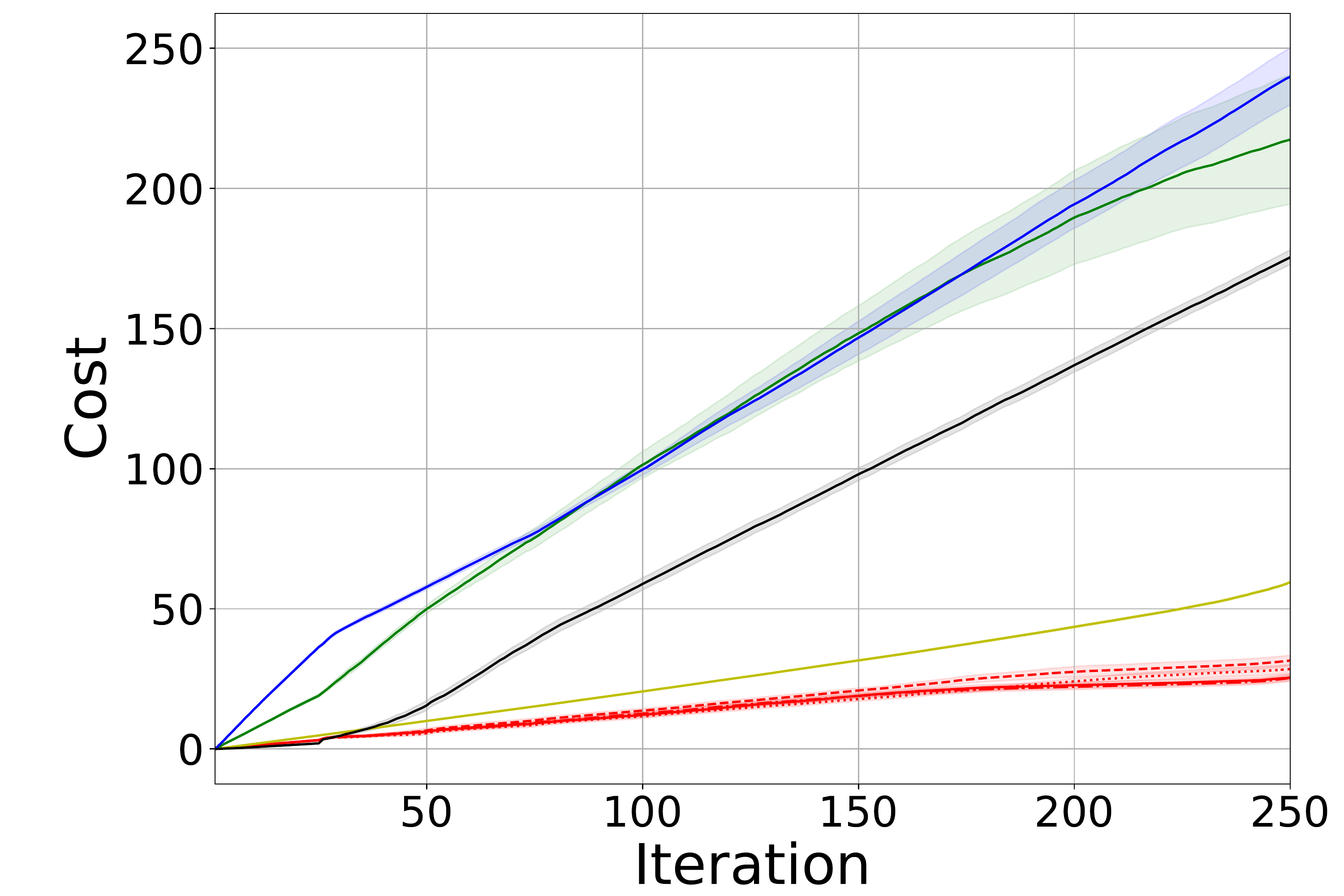}
	\caption{$T = 250$}
	\end{subfigure}
	\caption{Ackley4D (Asynchronous),  $t_{delay}=25$. Each row represents a different budget. The left column shows the evolution of regret against the cost used. The middle column shows the evolution of regret with iterations, and the right columns show the evolution of the 2-norm cost. Results are similar to the case when $t_{delay} = 10$, see Figure \ref{fig: ackley_async_4d_10}.}
	\label{fig: ackley_async_4d_250}
\end{figure}

\begin{figure}[ht]
	\centering
	\begin{subfigure}[t]{\textwidth}
	\includegraphics[width = 0.32\textwidth]{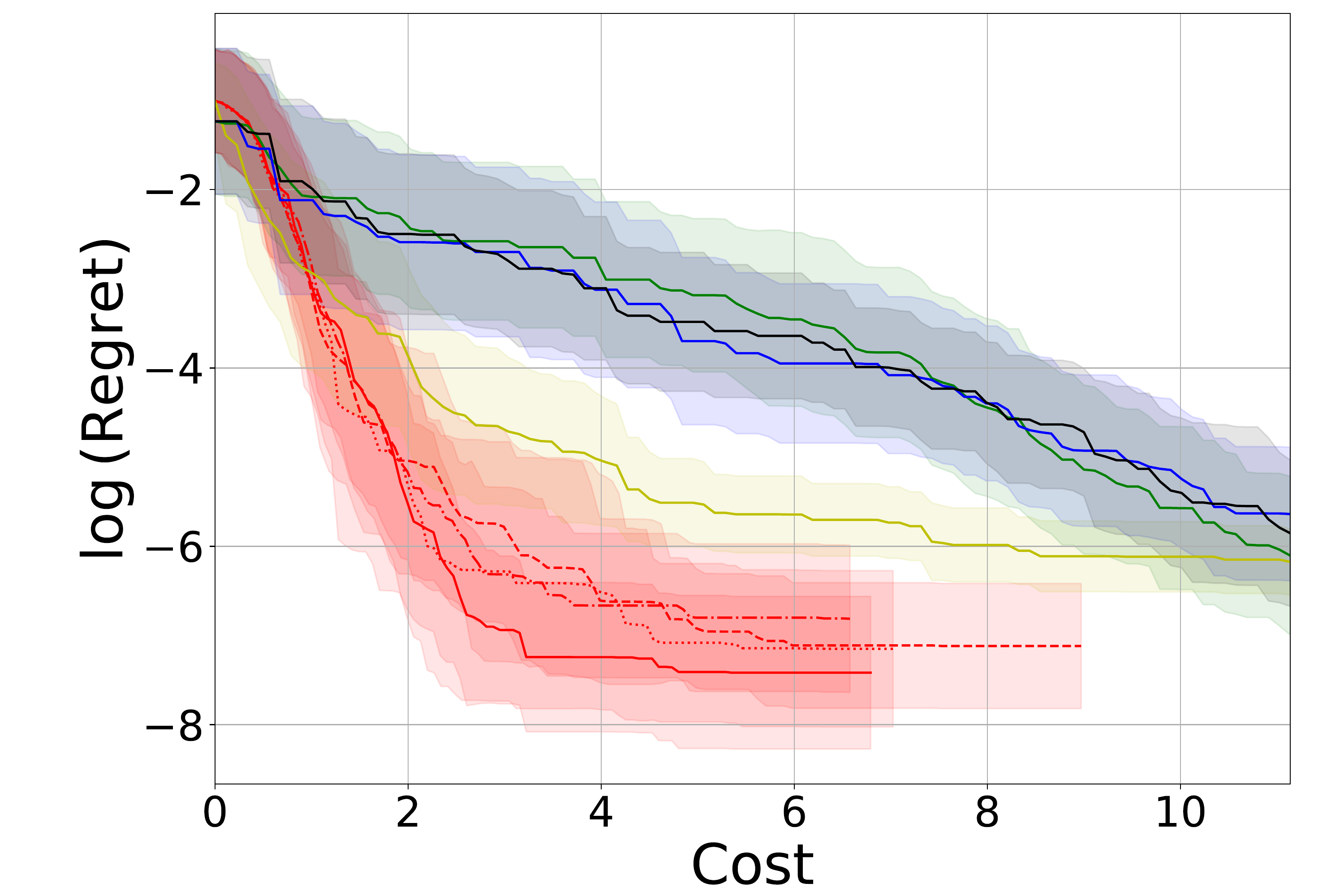}
	\includegraphics[width = 0.32\textwidth]{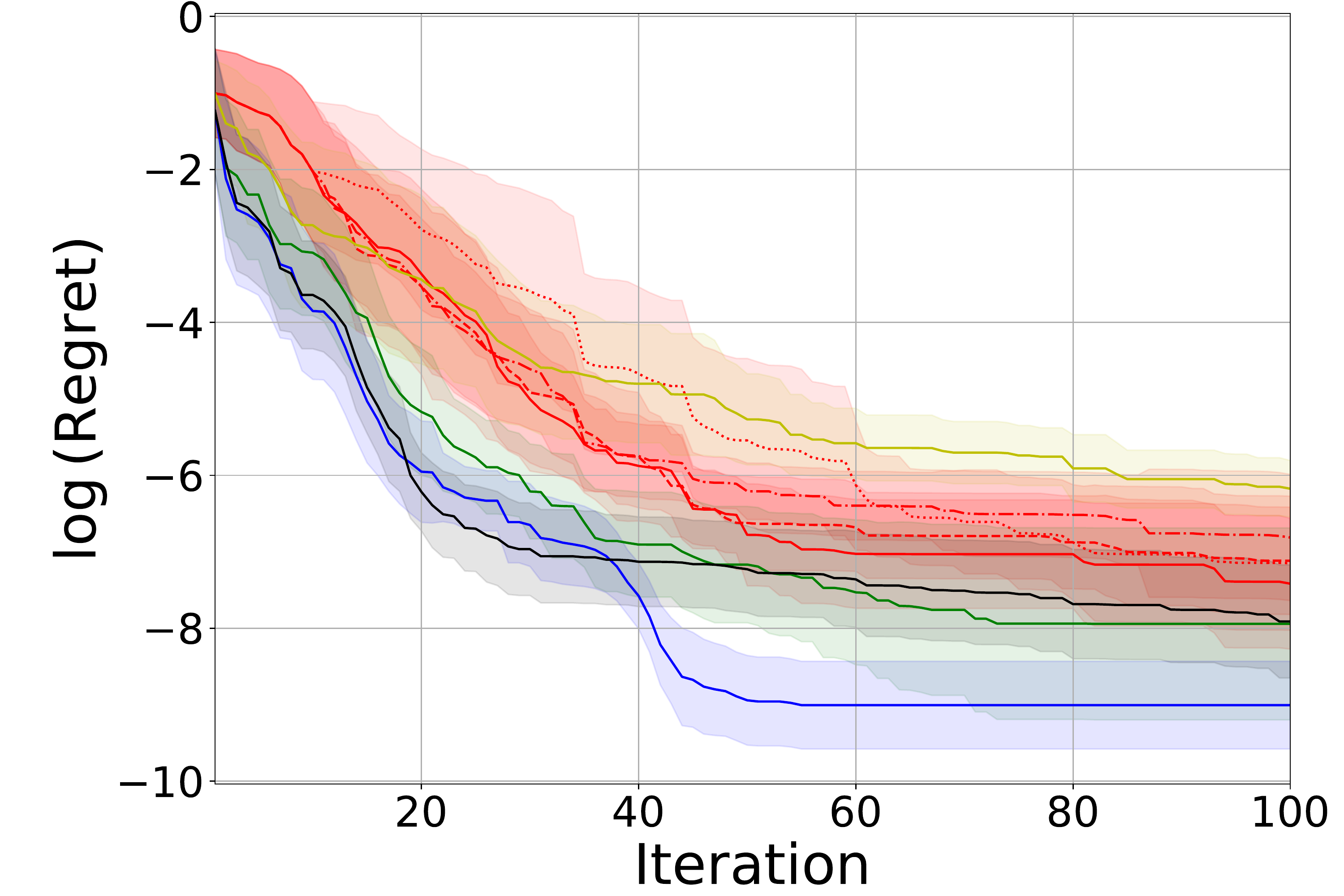}
	\includegraphics[width = 0.32\textwidth]{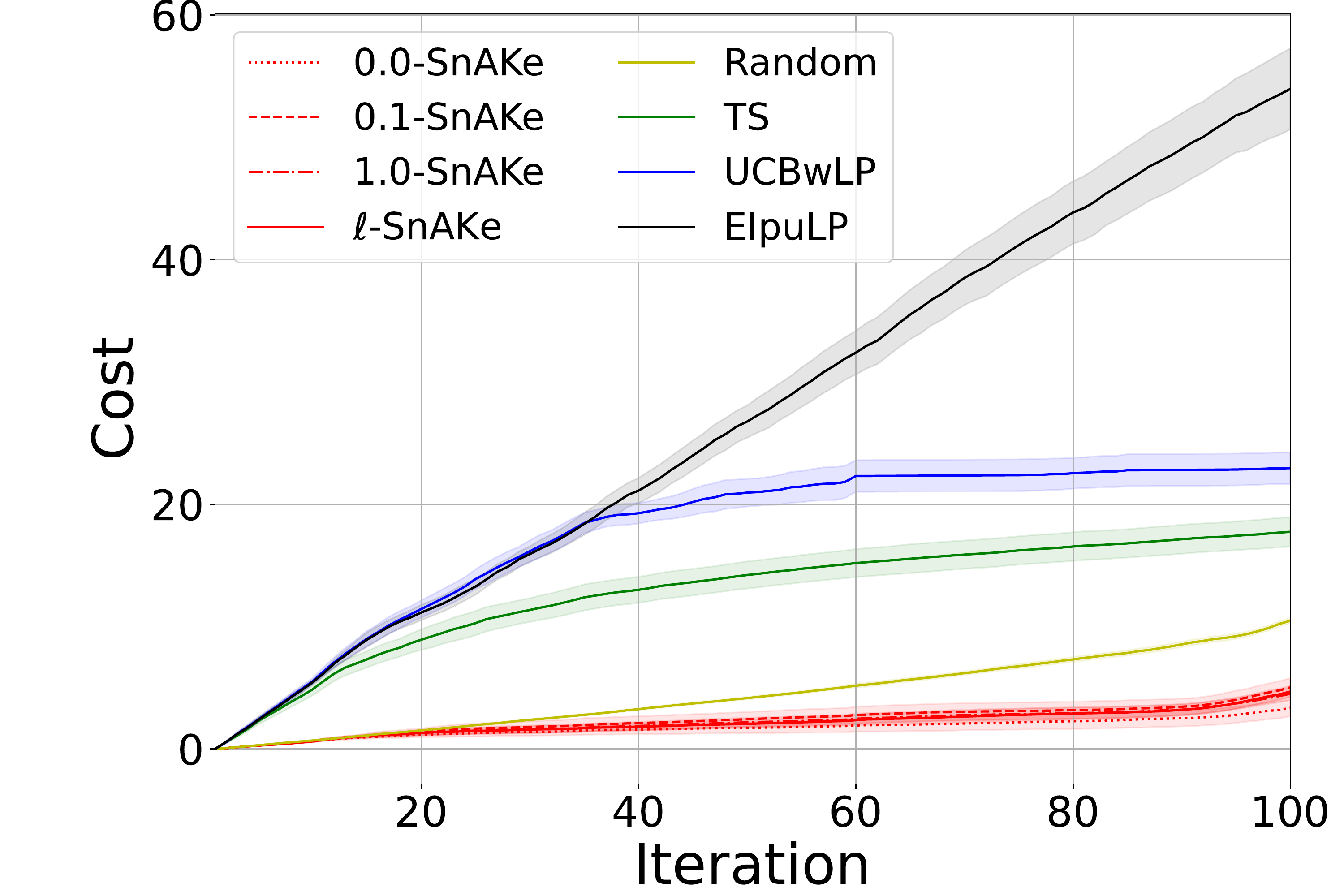}
	\caption{$T = 100$}
	\end{subfigure}
	\begin{subfigure}[t]{\textwidth}
	\includegraphics[width = 0.32\textwidth]{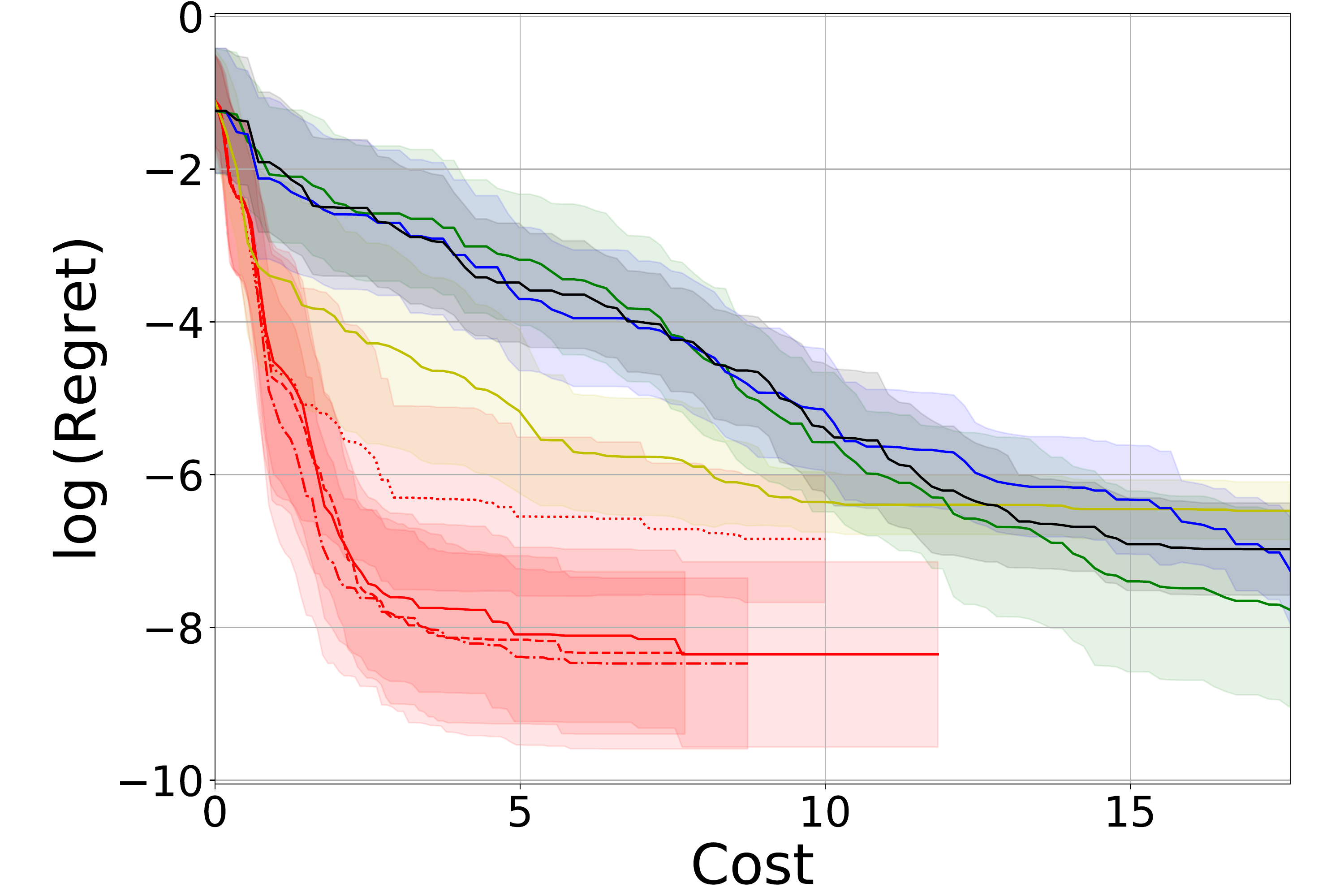}
	\includegraphics[width = 0.32\textwidth]{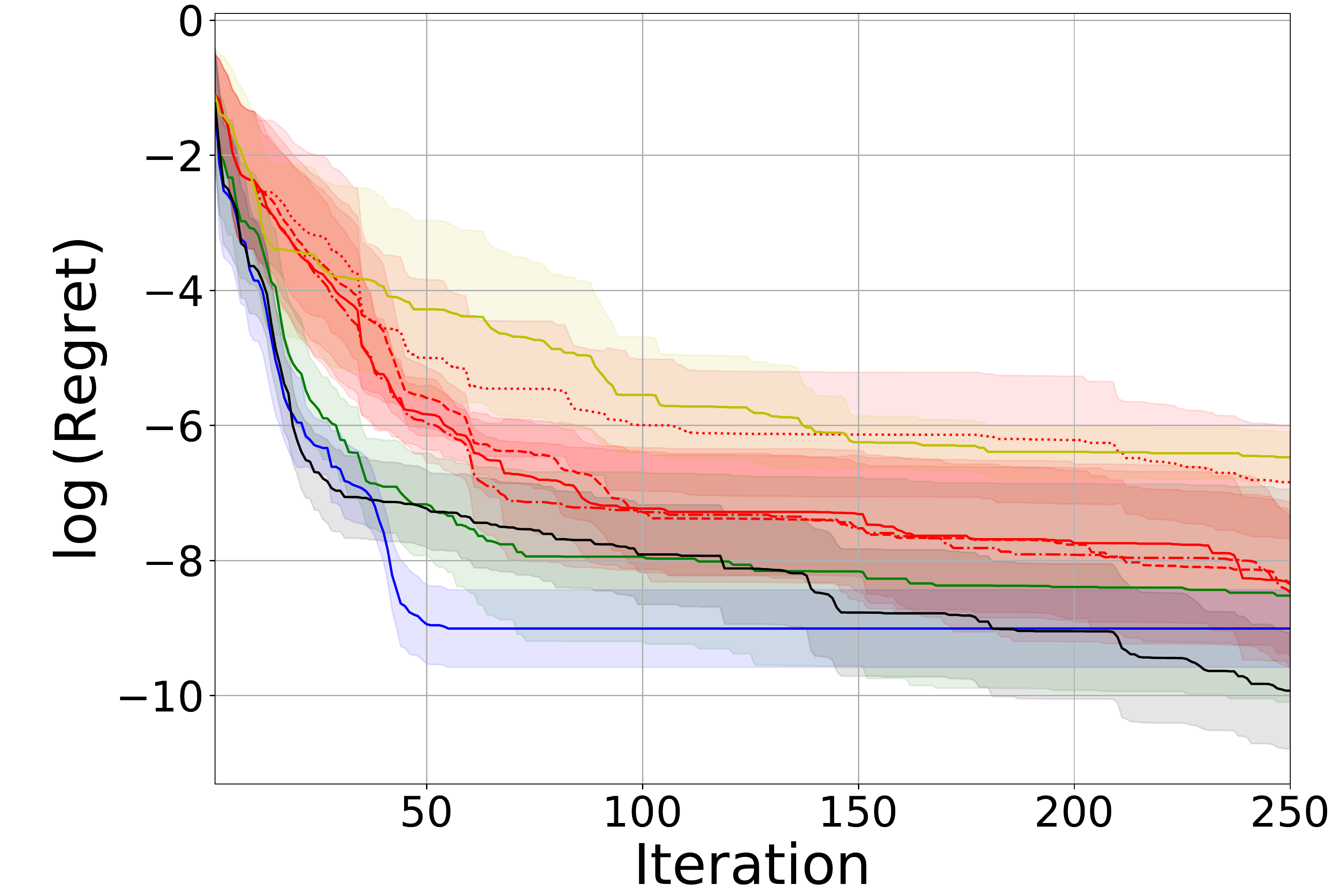}
	\includegraphics[width = 0.32\textwidth]{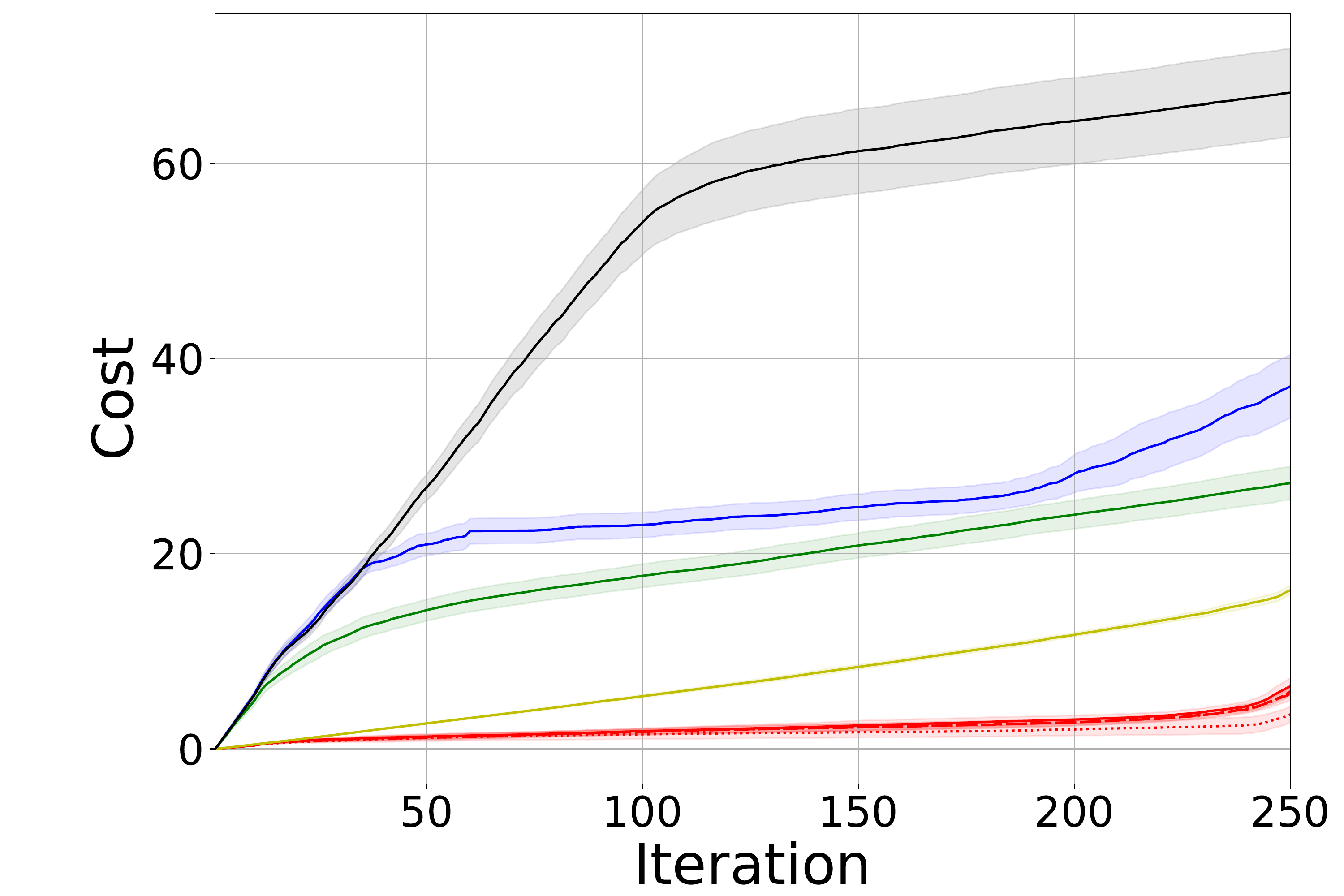}
	\caption{$T = 250$}
	\end{subfigure}
	\caption{Michaelwicz2D (Asynchronous),  $t_{delay}=10$. Each row represents a different budget. The left column shows the evolution of regret against the cost used. The middle column shows the evolution of regret with iterations, and the right columns show the evolution of the 2-norm cost. Surprisingly, EIpuLP achieves the best regret, but at considerable cost - suggesting local penalization is over-powering any cost-awareness. For low cost, SnAKe achieves much better regret.}
	\label{fig: michael_async_td_10}
\end{figure}

\begin{figure}[ht]
	\centering
	\begin{subfigure}[t]{\textwidth}
	\includegraphics[width = 0.32\textwidth]{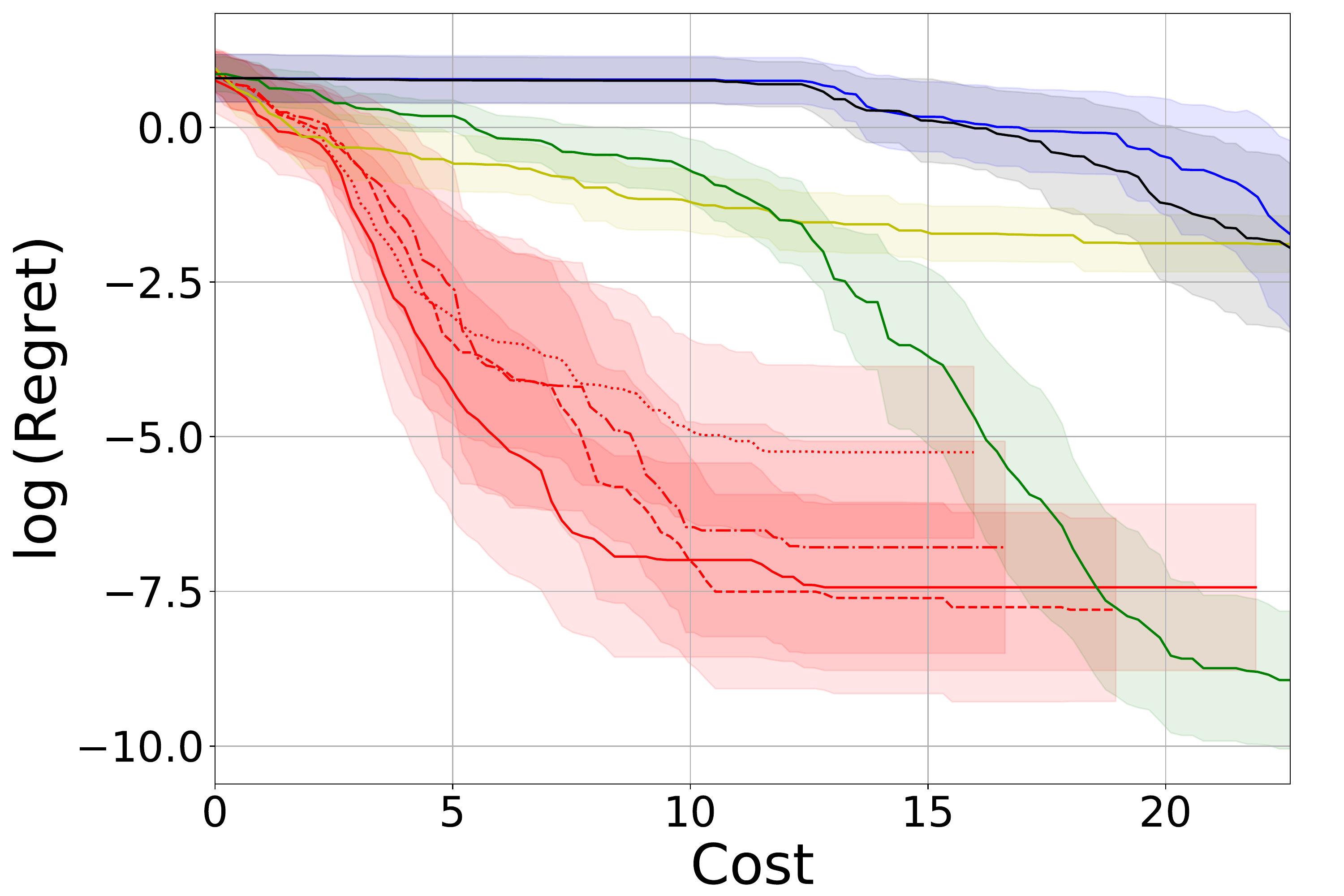}
	\includegraphics[width = 0.32\textwidth]{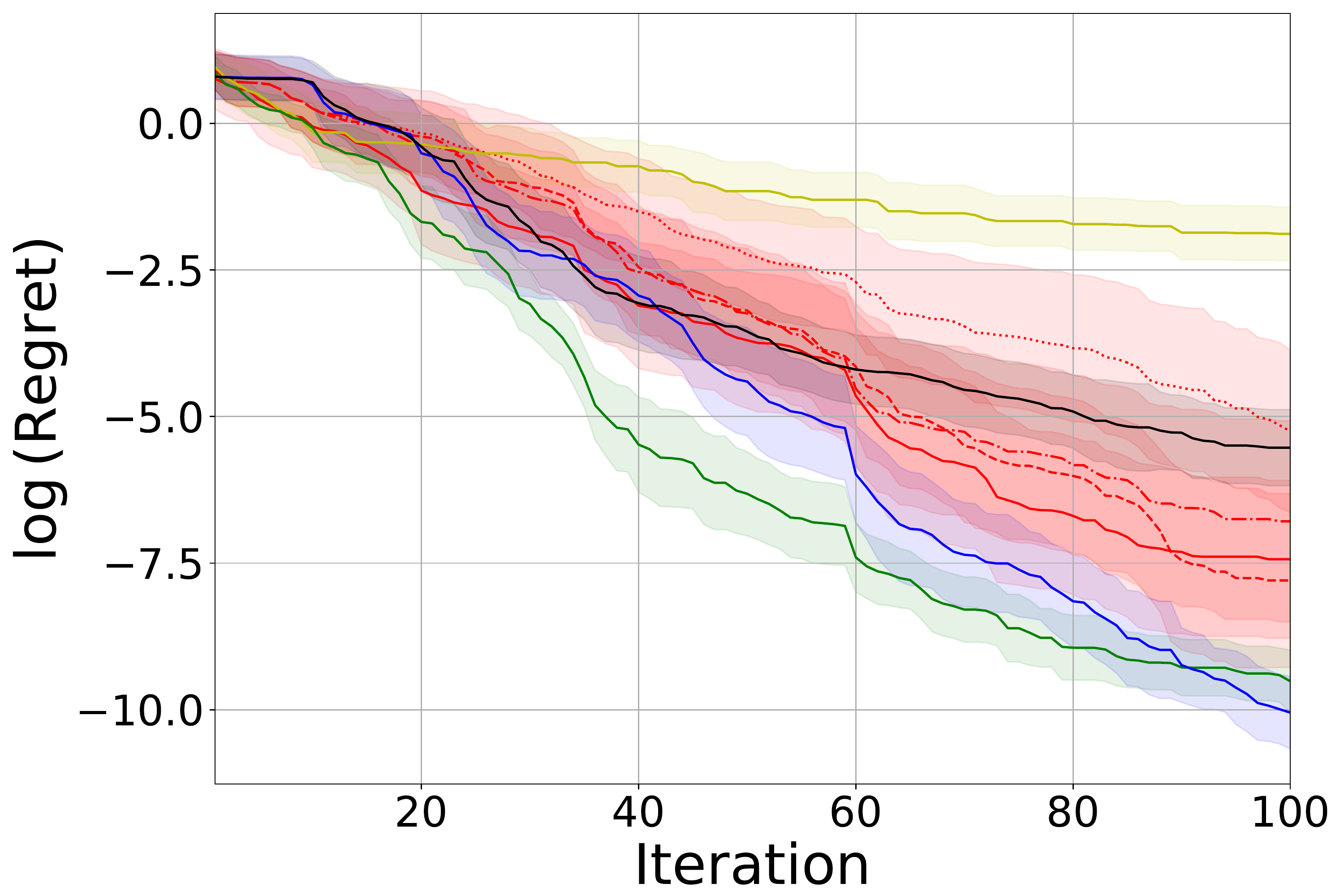}
	\includegraphics[width = 0.32\textwidth]{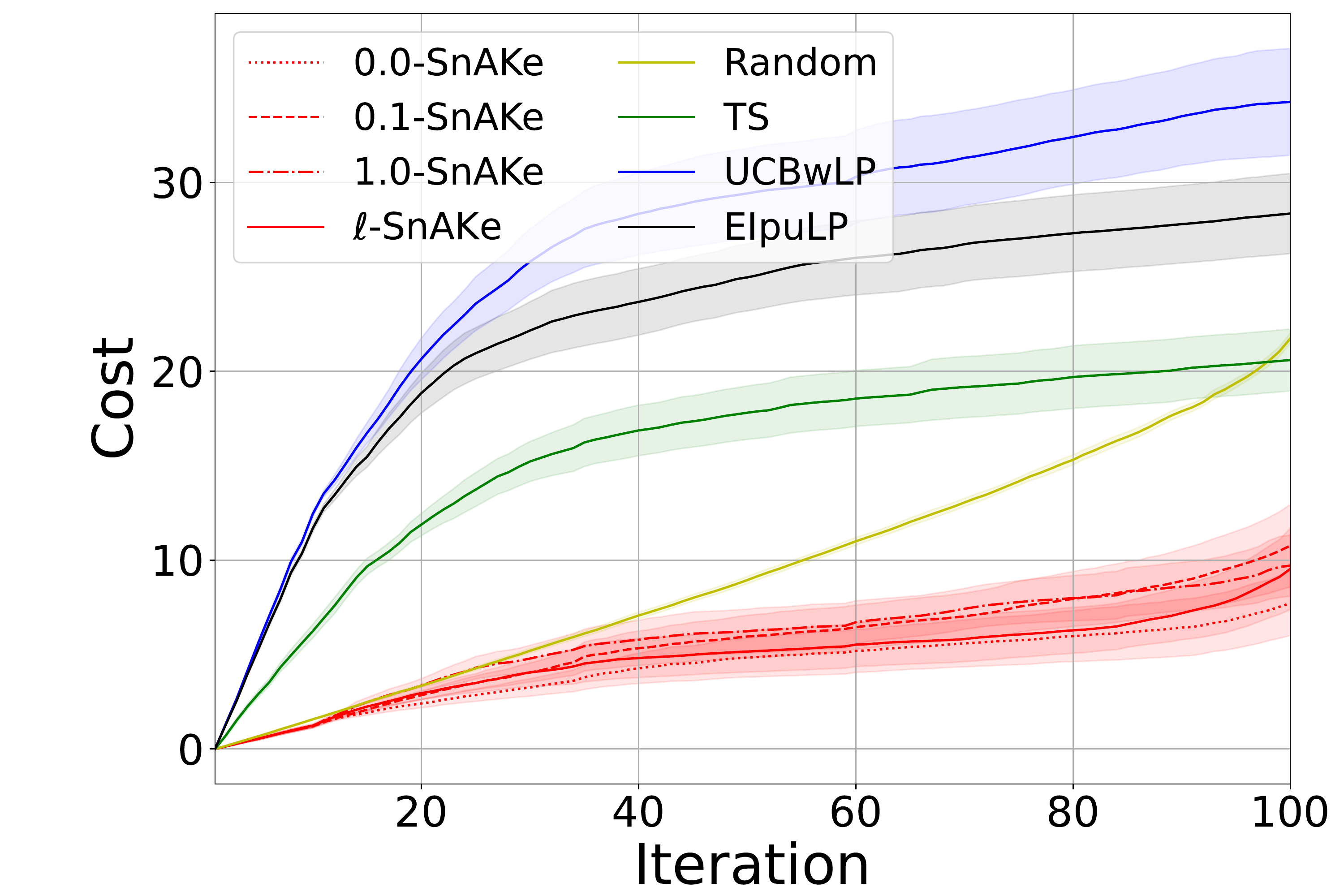}
	\caption{$T = 100$}
	\end{subfigure}
	\begin{subfigure}[t]{\textwidth}
	\includegraphics[width = 0.32\textwidth]{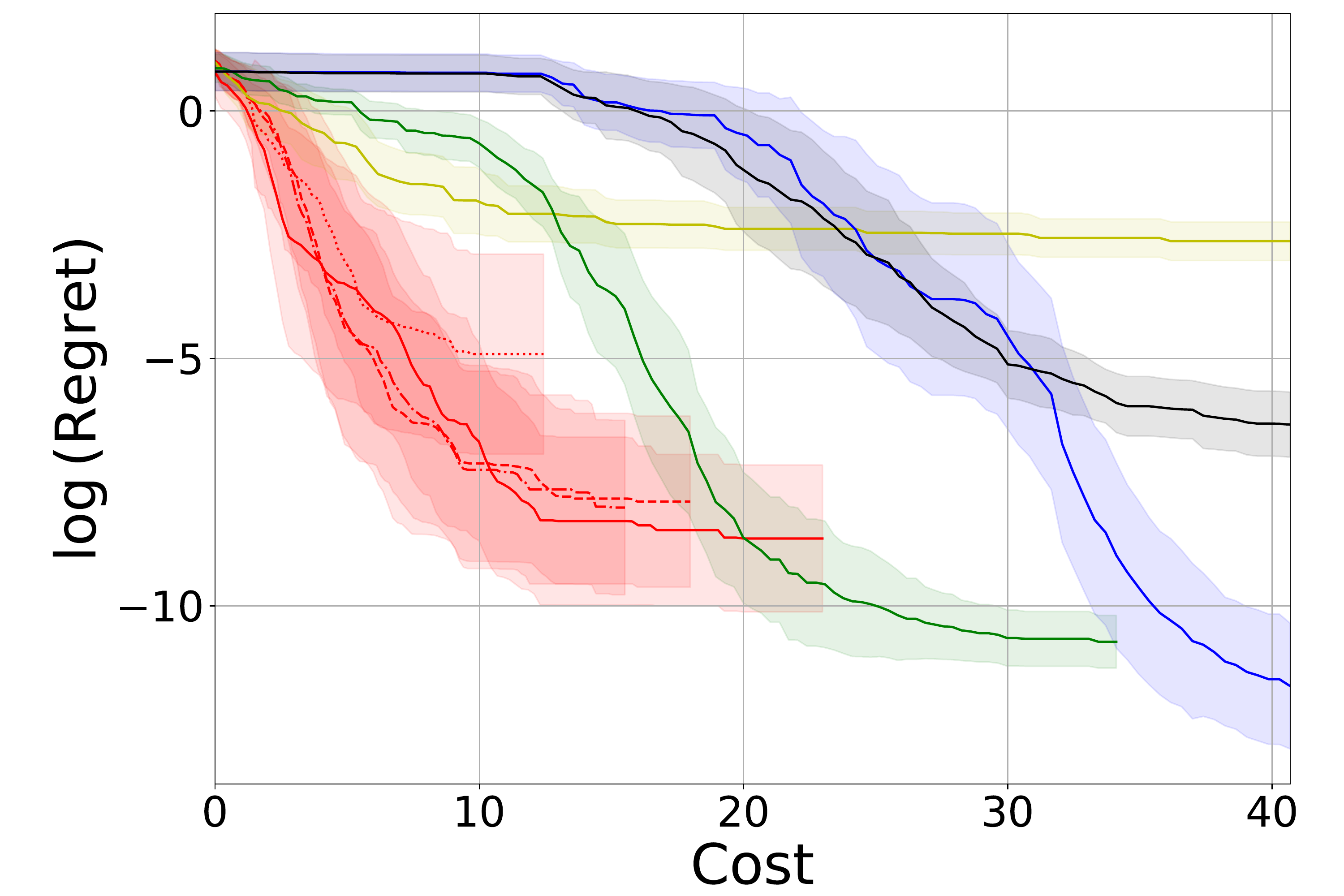}
	\includegraphics[width = 0.32\textwidth]{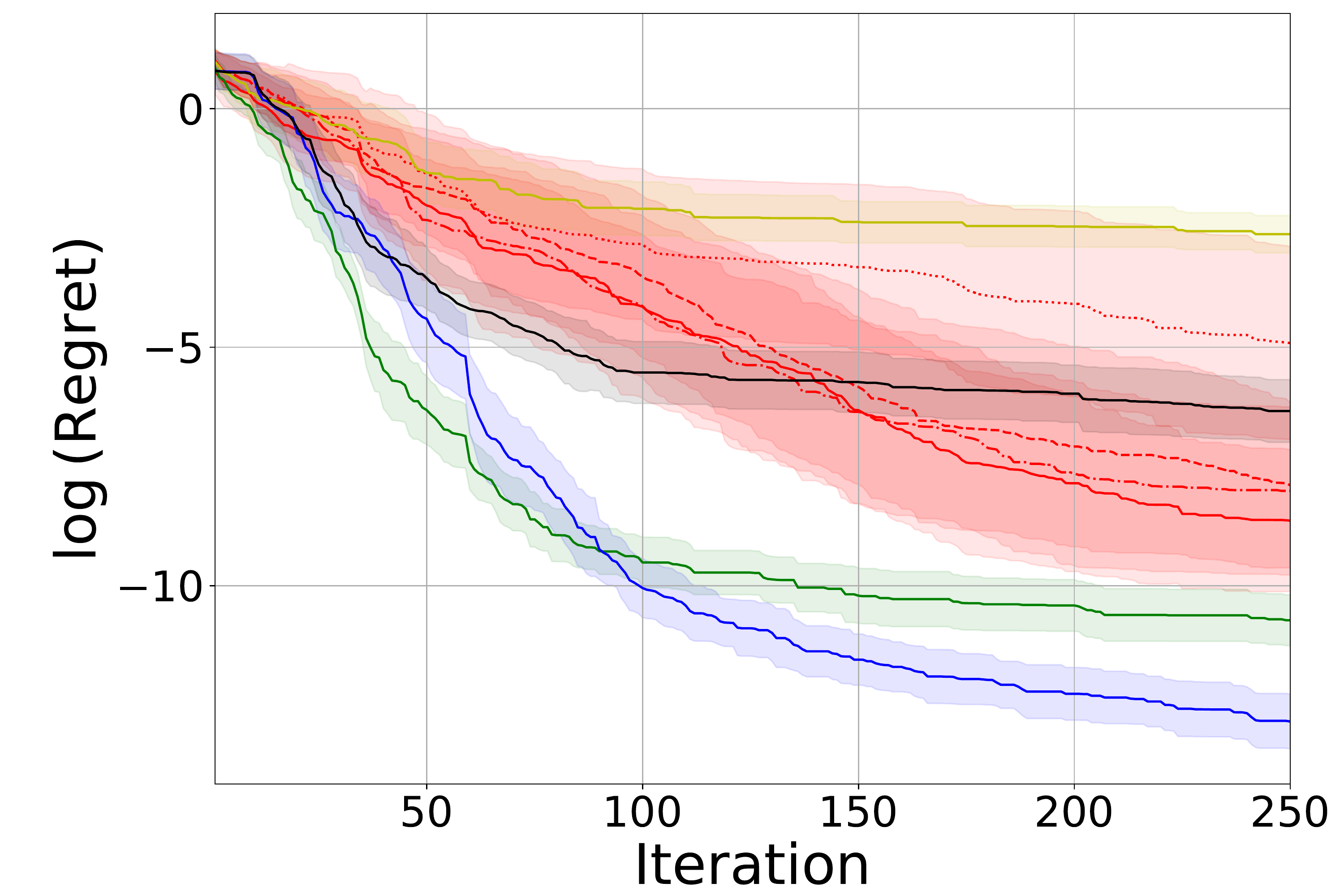}
	\includegraphics[width = 0.32\textwidth]{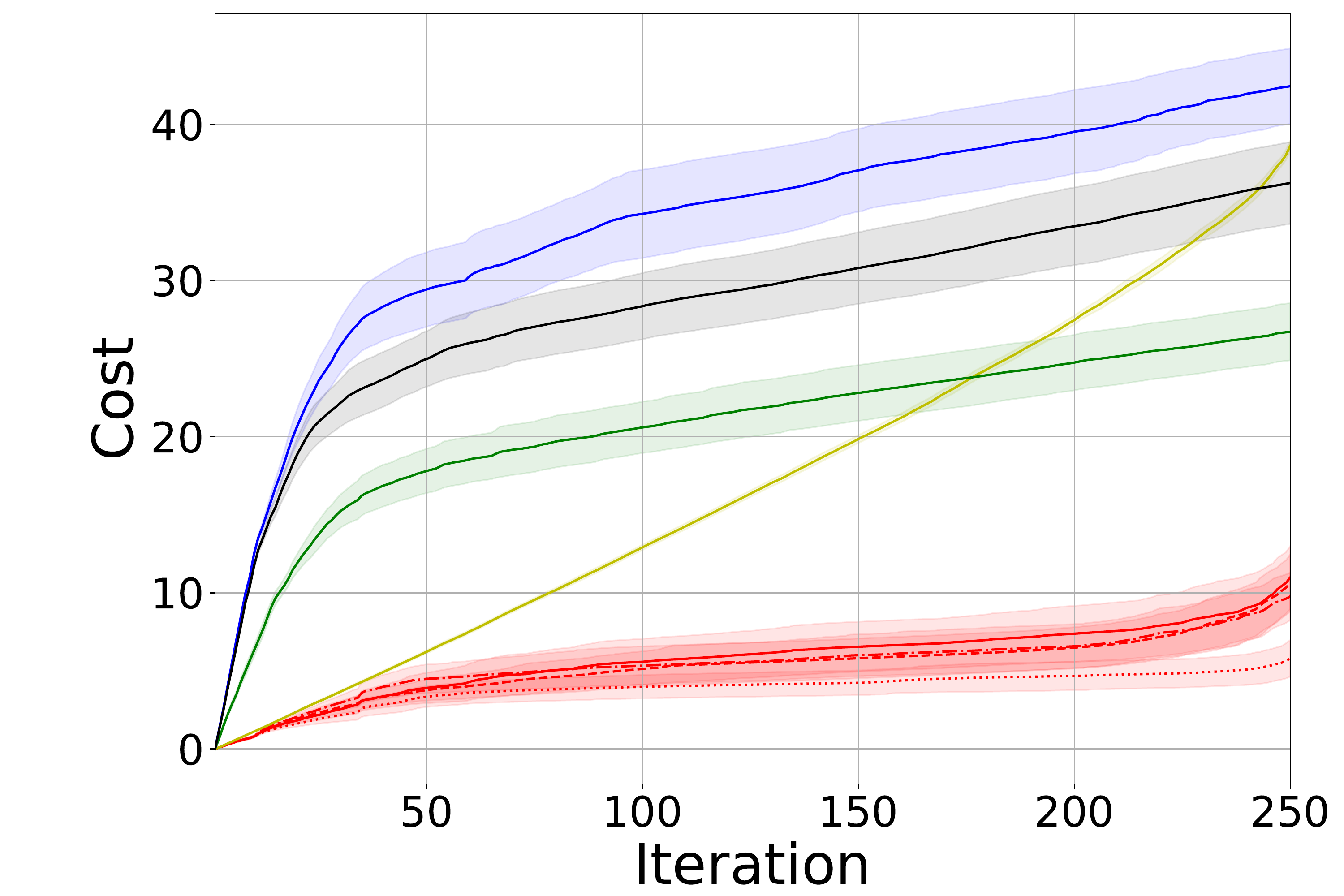}
	\caption{$T = 250$}
	\end{subfigure}
	\caption{Hartmann3D (Asynchronous),  $t_{delay}=10$. Each row represents a different budget. The left column shows the evolution of regret against the cost used. The middle column shows the evolution of regret with iterations, and the right columns show the evolution of the 2-norm cost. SnAKe achieves the best regret for low cost, with Thompson Sampling also giving a good performance. For the full optimization, UCBwLP achieves the best regret, at the expense of four times the cost of SnAKe. EIpuLP performs poorly, again, it seems Local Penalization is over-powering the cost term.}
	\label{fig: hartmann_3d_async_10}
\end{figure}

\begin{figure}[ht]
	\centering
	\begin{subfigure}[t]{\textwidth}
	\includegraphics[width = 0.32\textwidth]{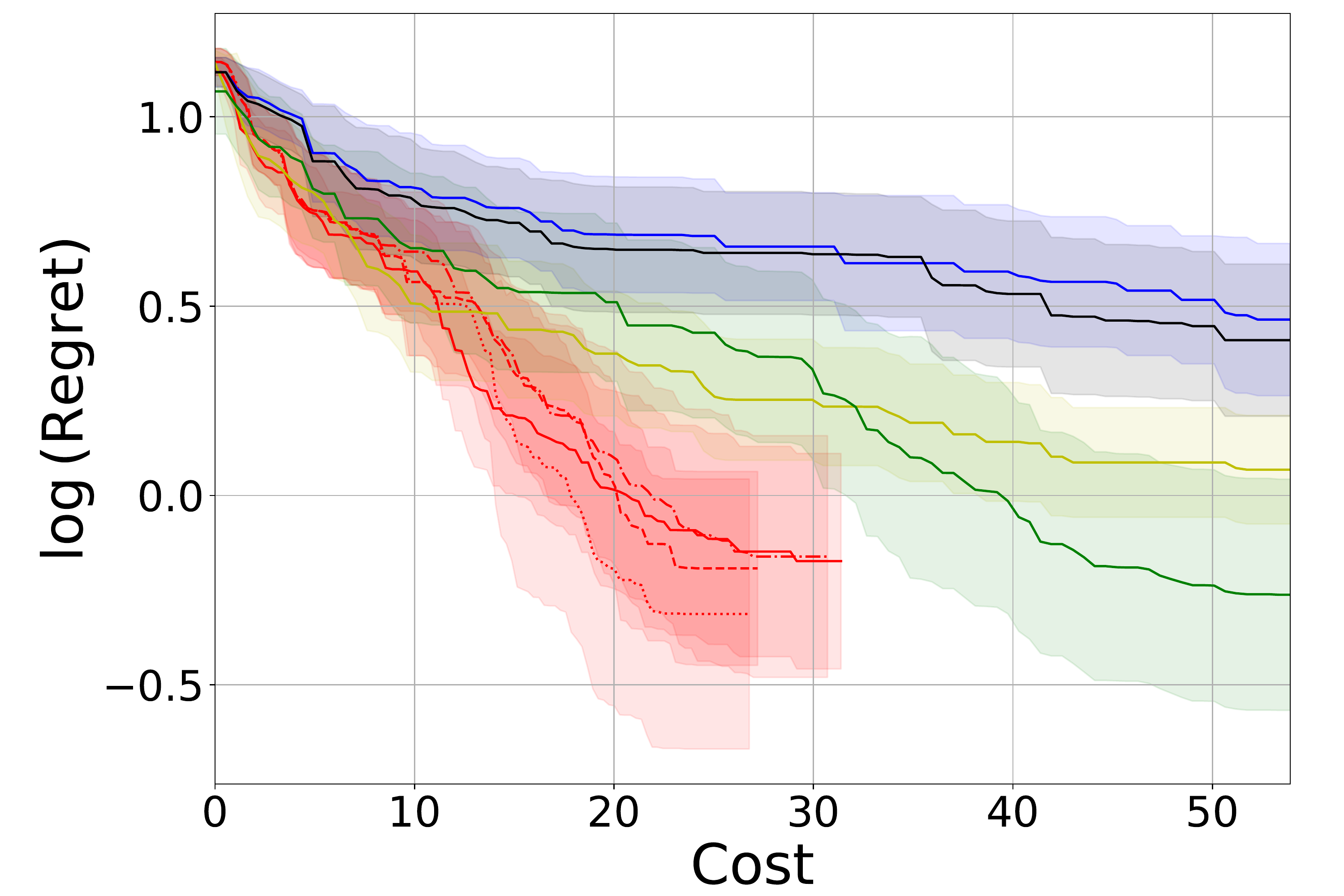}
	\includegraphics[width = 0.32\textwidth]{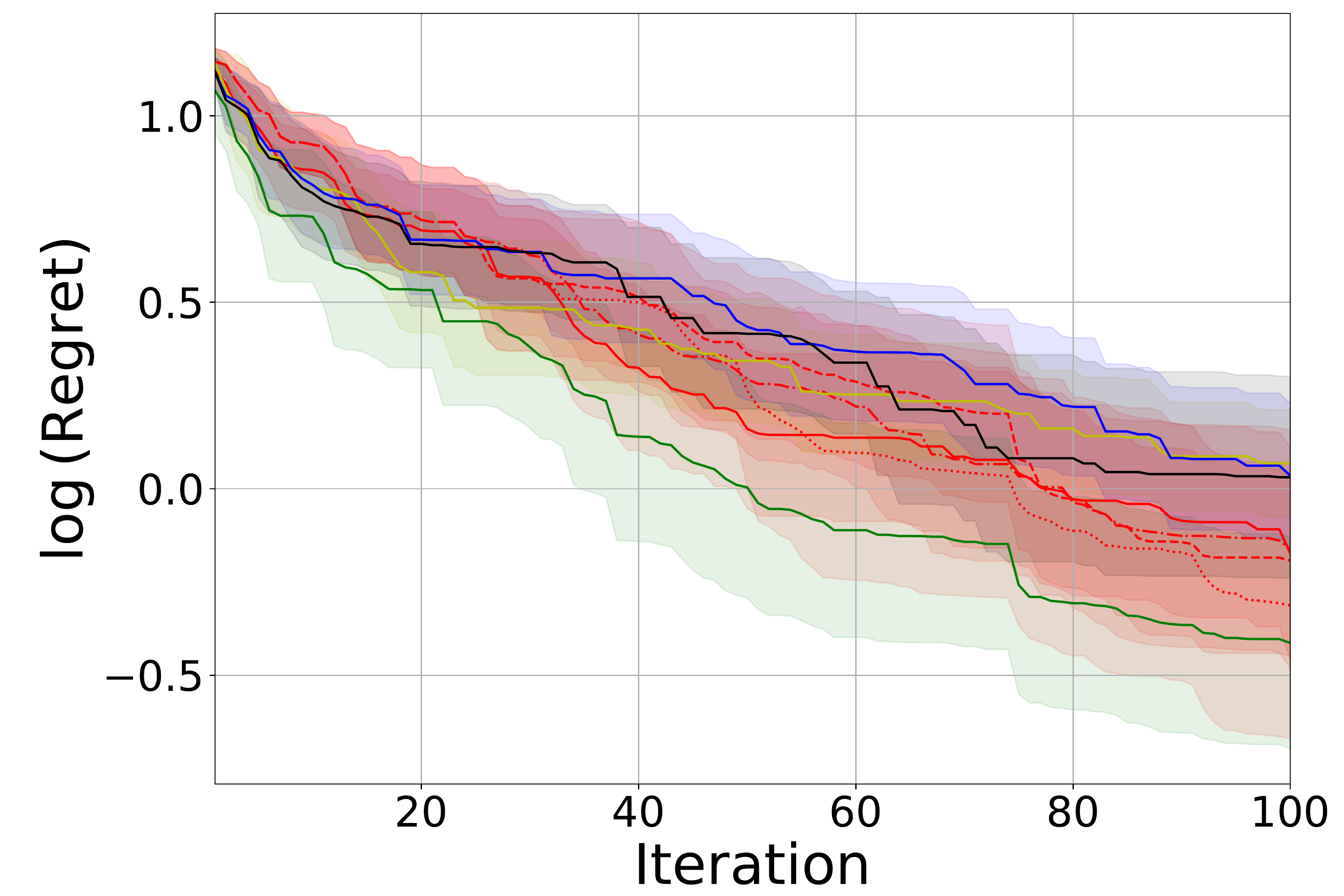}
	\includegraphics[width = 0.32\textwidth]{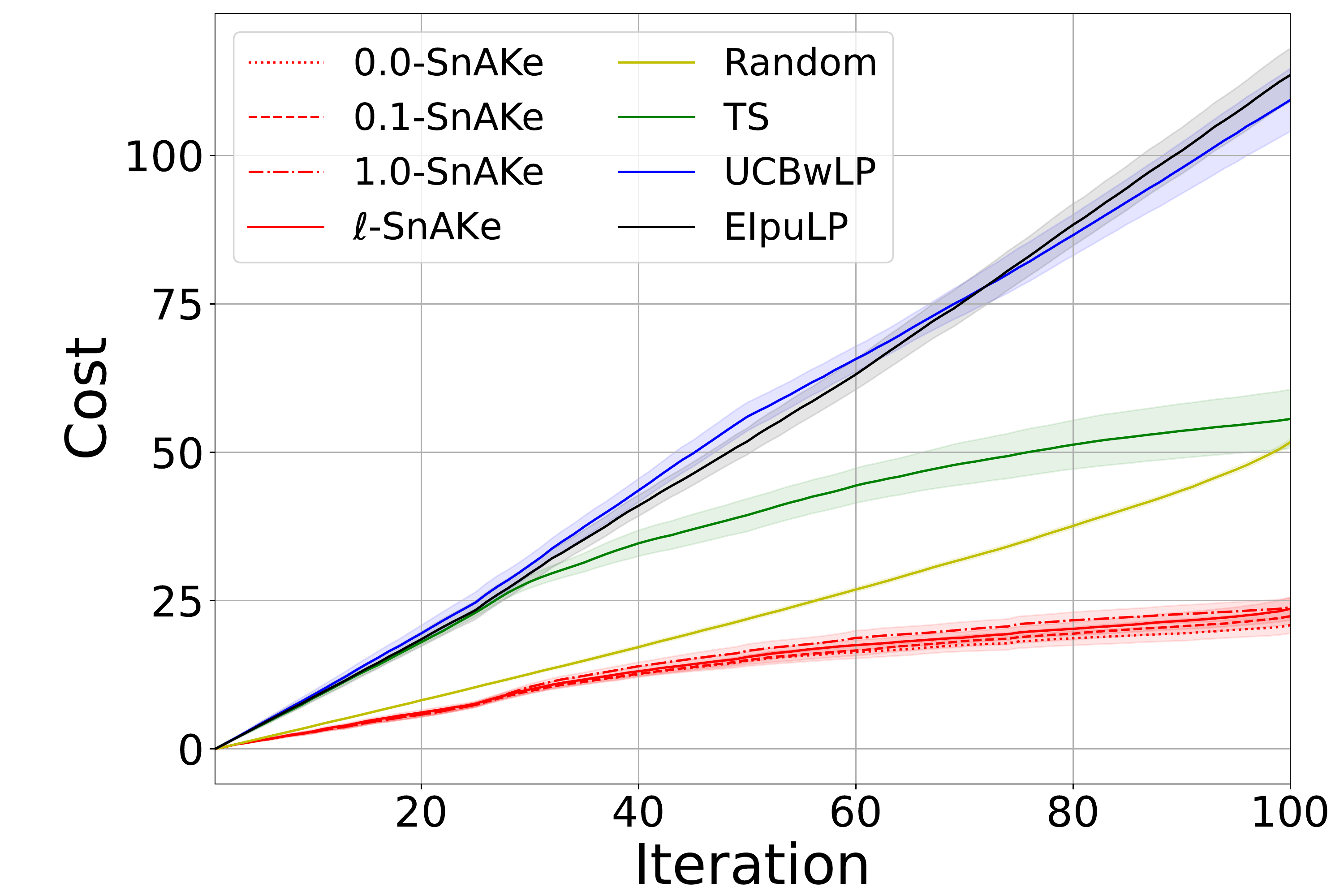}
	\caption{$T = 100$}
	\end{subfigure}
	\begin{subfigure}[t]{\textwidth}
	\includegraphics[width = 0.32\textwidth]{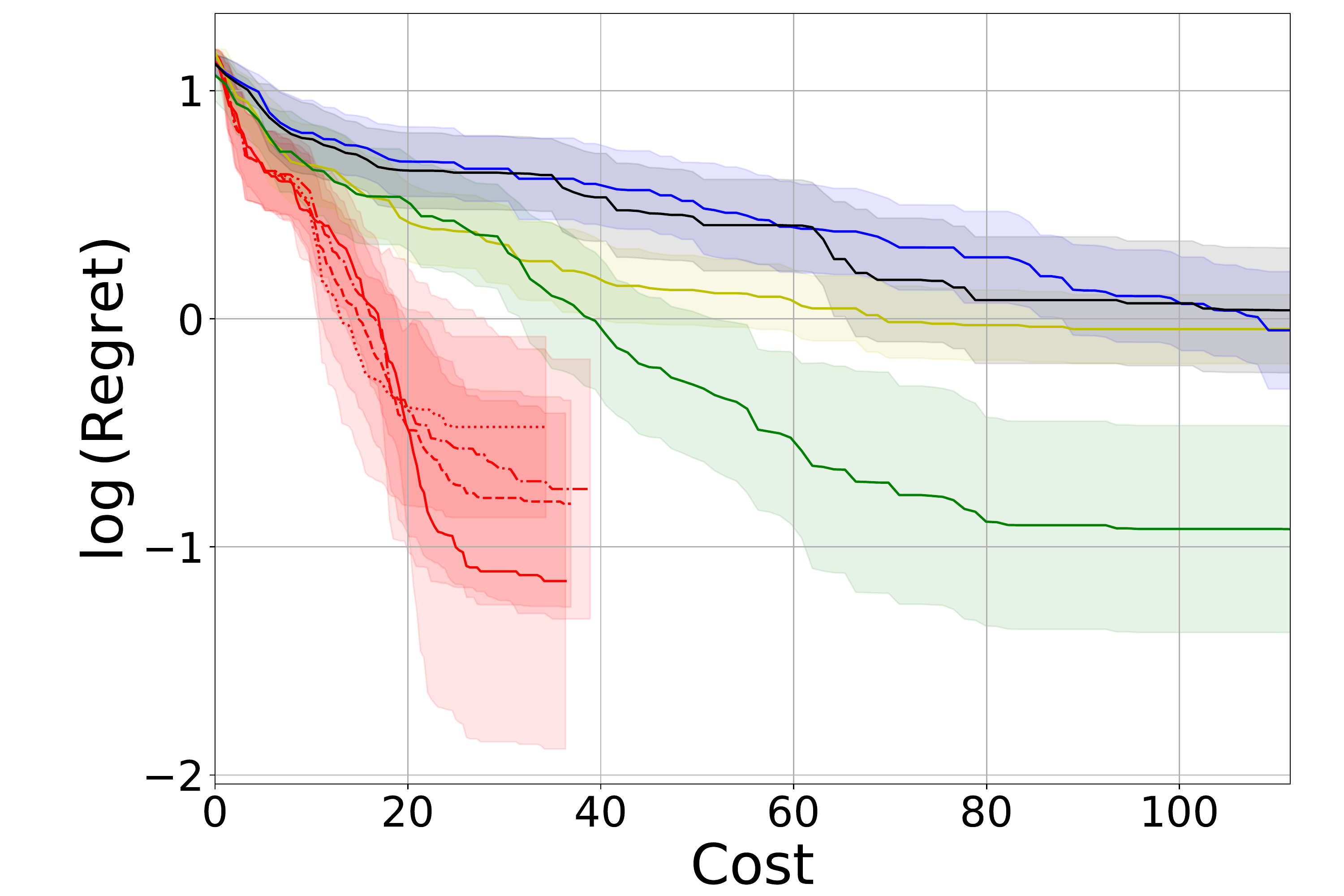}
	\includegraphics[width = 0.32\textwidth]{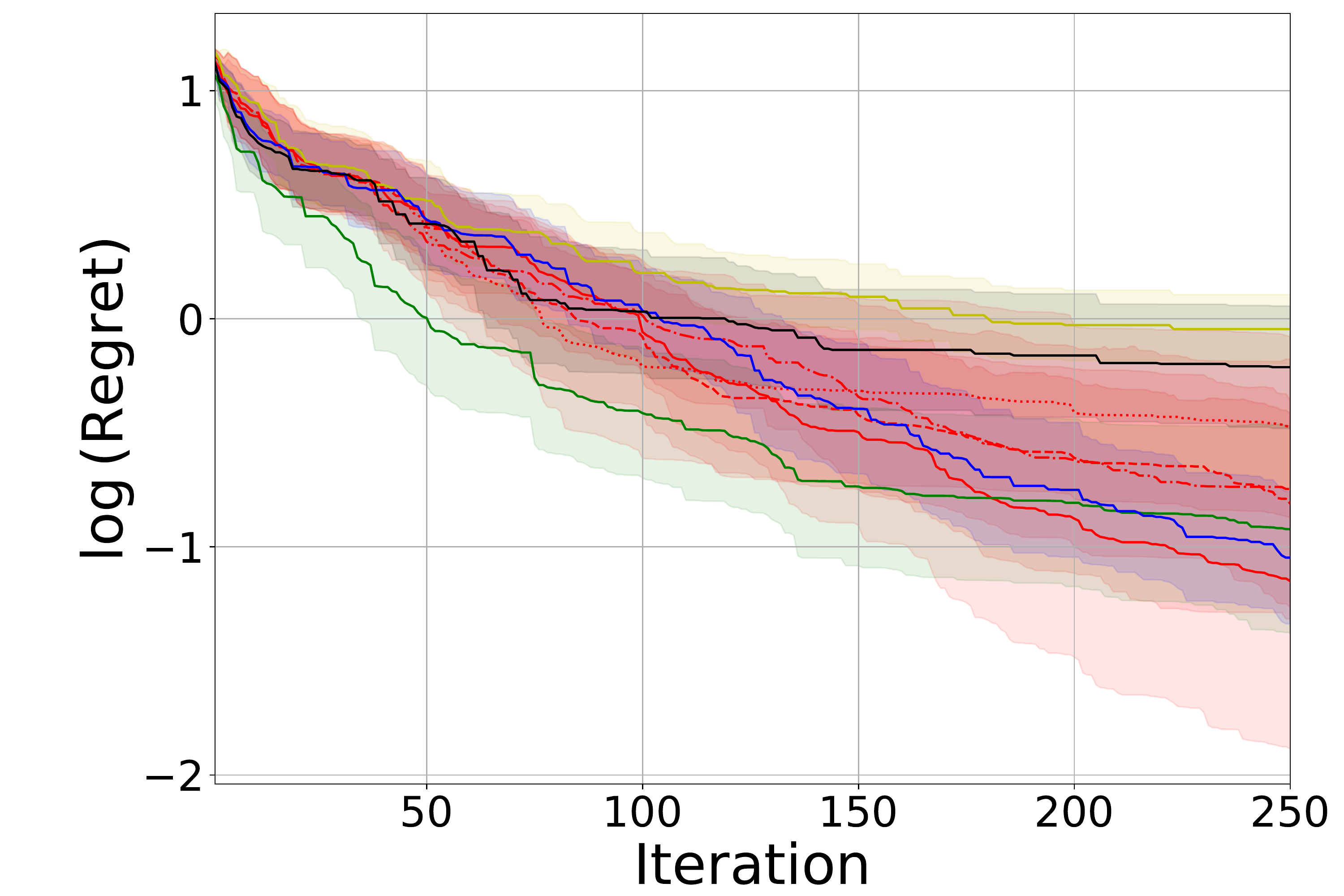}
	\includegraphics[width = 0.32\textwidth]{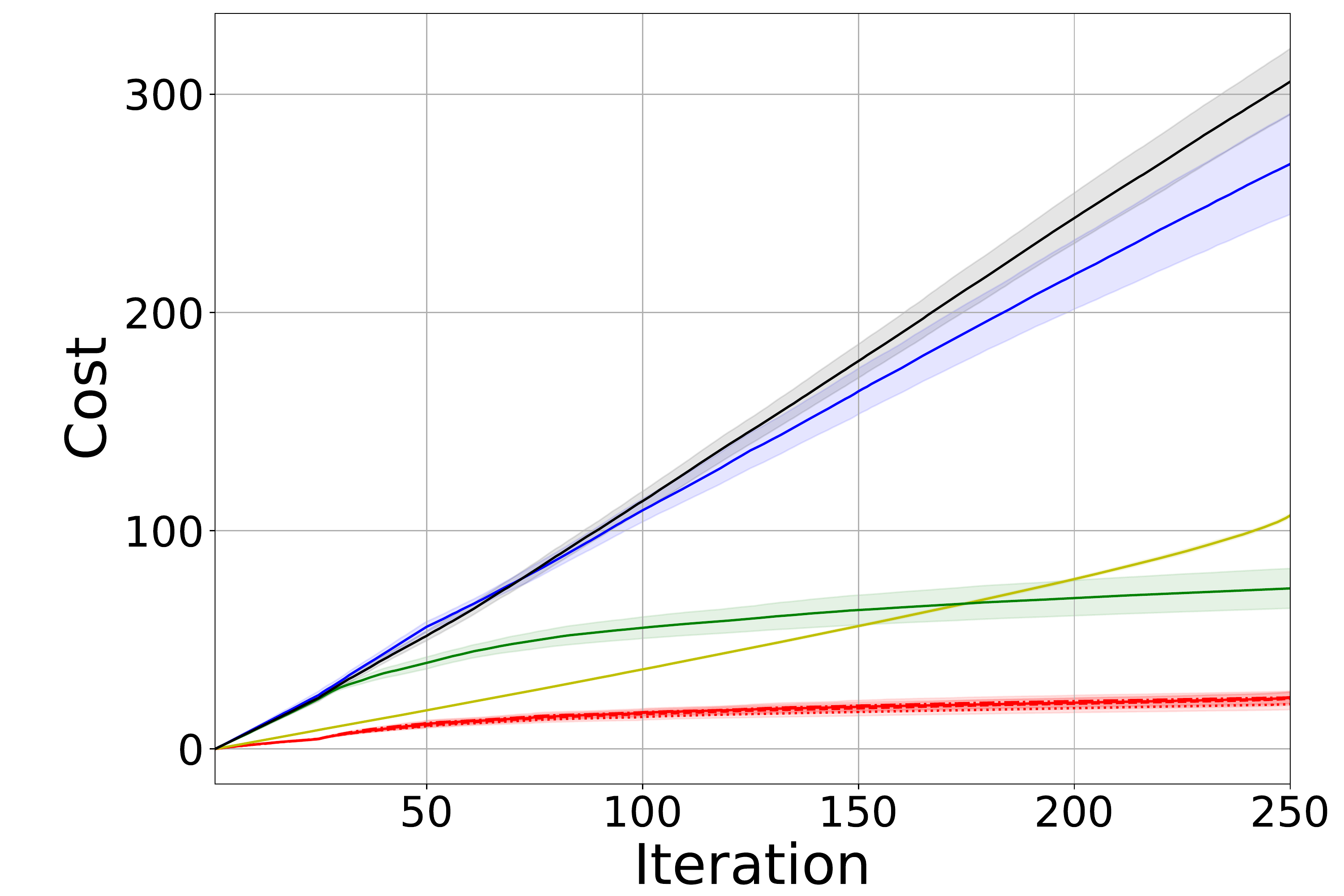}
	\caption{$T = 250$}
	\end{subfigure}
	\caption{Hartmann6D (Asynchronous),  $t_{delay}=25$. Each row represents a different budget. The left column shows the evolution of regret against the cost used. The middle column shows the evolution of regret with iterations, and the right columns show the evolution of the 2-norm cost. Similar results to other Hartmann benchmarks, see Figure \ref{fig: hartmann_3d_async_10}.}
	\label{fig: hartmann6d_25_async}
\end{figure}

We include the full  graphs of the asynchronous Bayesian Optimization experiments. Each row represents a different budget. The left column shows the evolution of regret against the cost used, the middle column shows the evolution of regret with iterations, and the right columns show the evolution of the 2-norm cost. The results encompass Figures \ref{fig: branin2d_10_async} to \ref{fig: hartmann6d_25_async}. The caption in each figure tells us the benchmark function being evaluated, and the time-delay for getting observations back. Each experiment is the mean $\pm$ half the standard deviation of 10 different runs.

\begin{figure}[ht]
	\centering
	\begin{subfigure}[t]{\textwidth}
	\includegraphics[width = 0.32\textwidth]{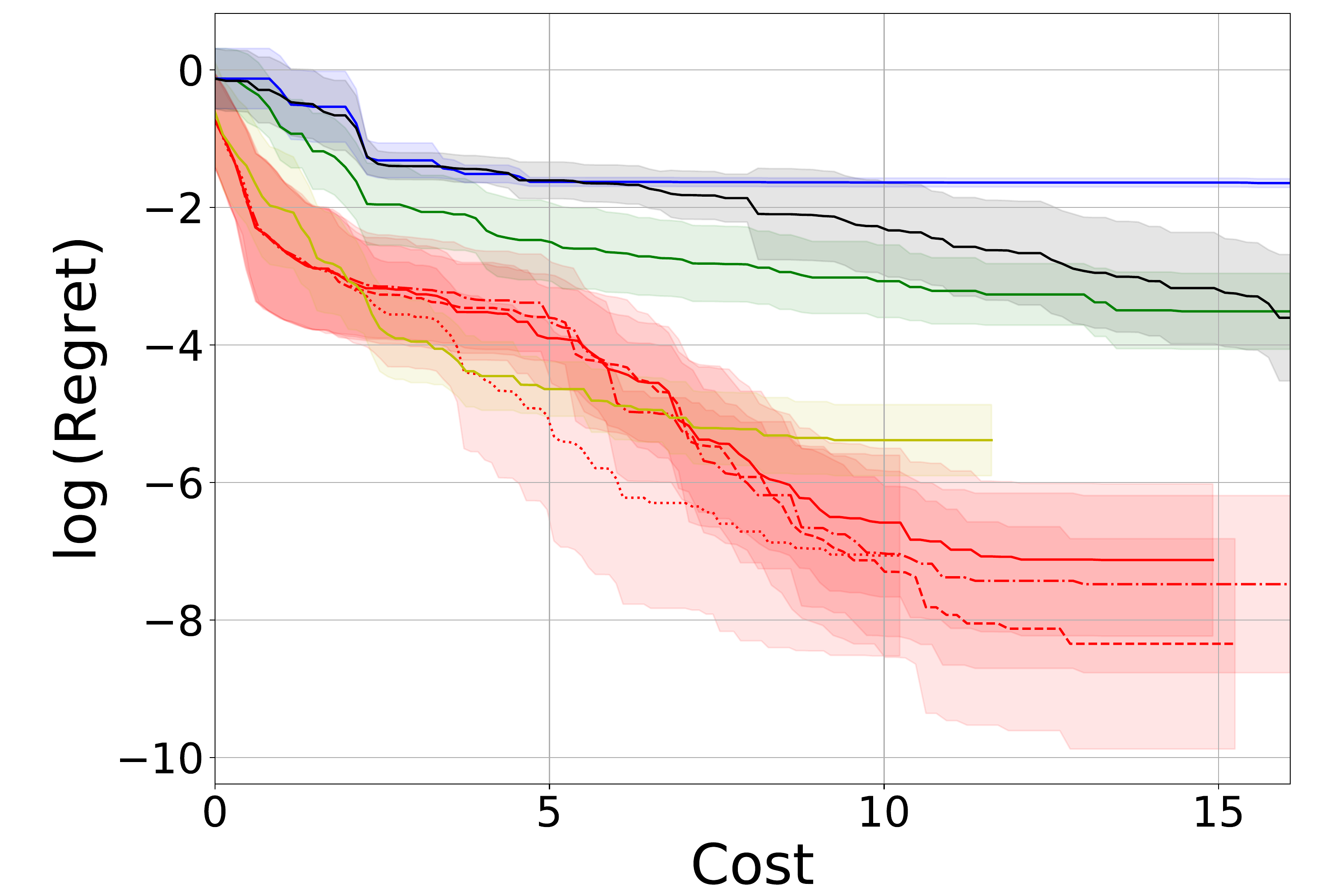}
	\includegraphics[width = 0.32\textwidth]{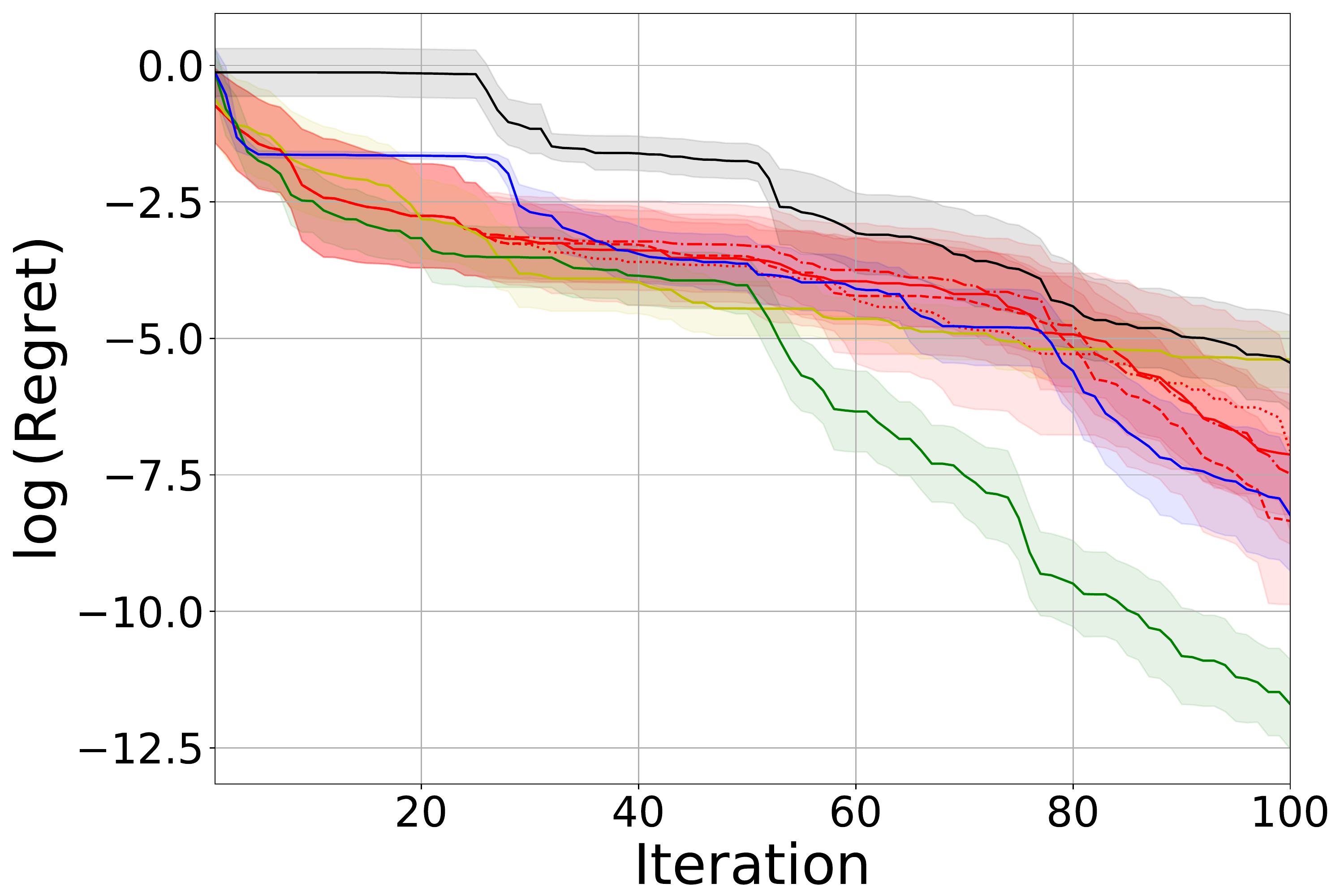}
	\includegraphics[width = 0.32\textwidth]{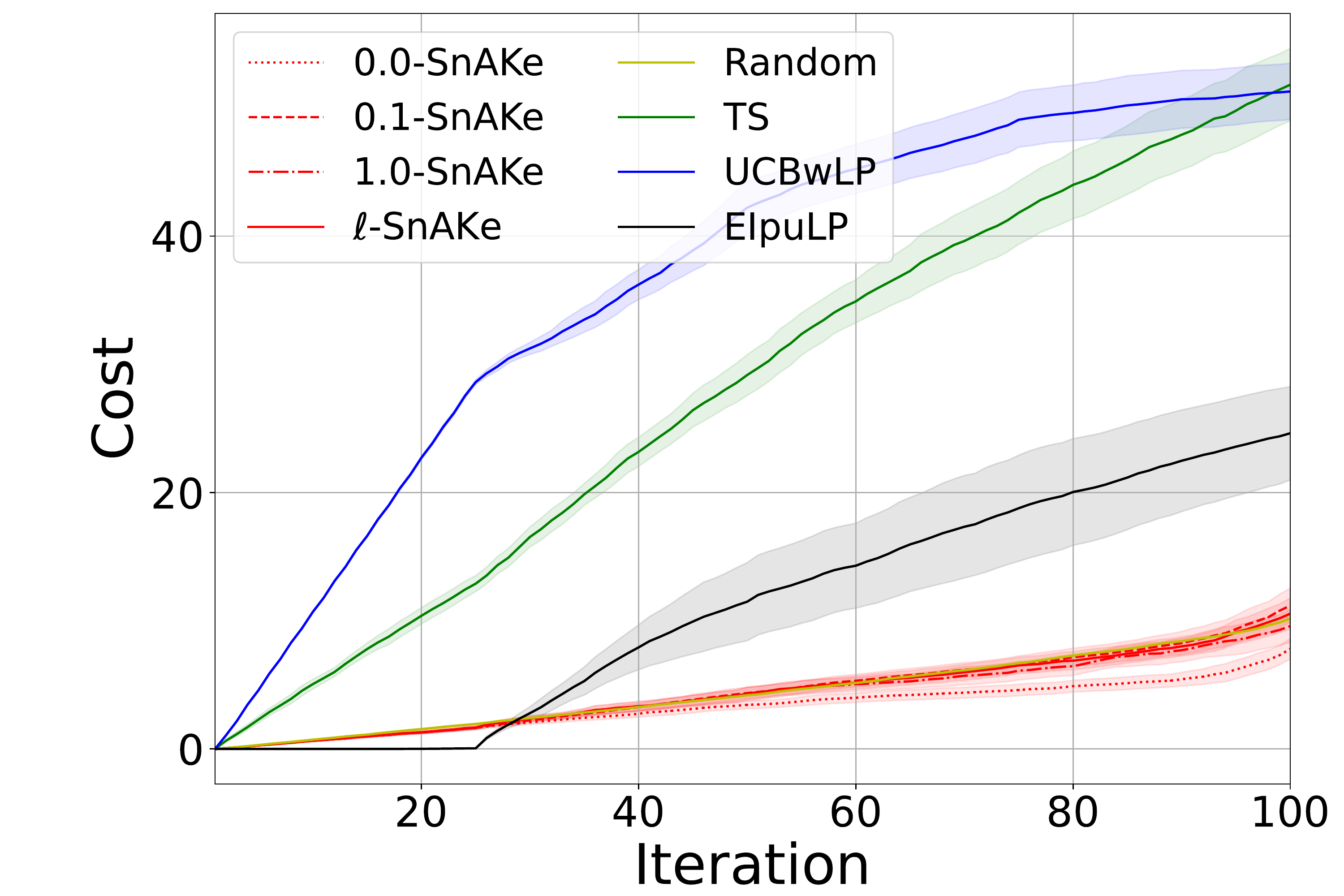}
	\caption{$T = 100$}
	\end{subfigure}
	\begin{subfigure}[t]{\textwidth}
	\includegraphics[width = 0.32\textwidth]{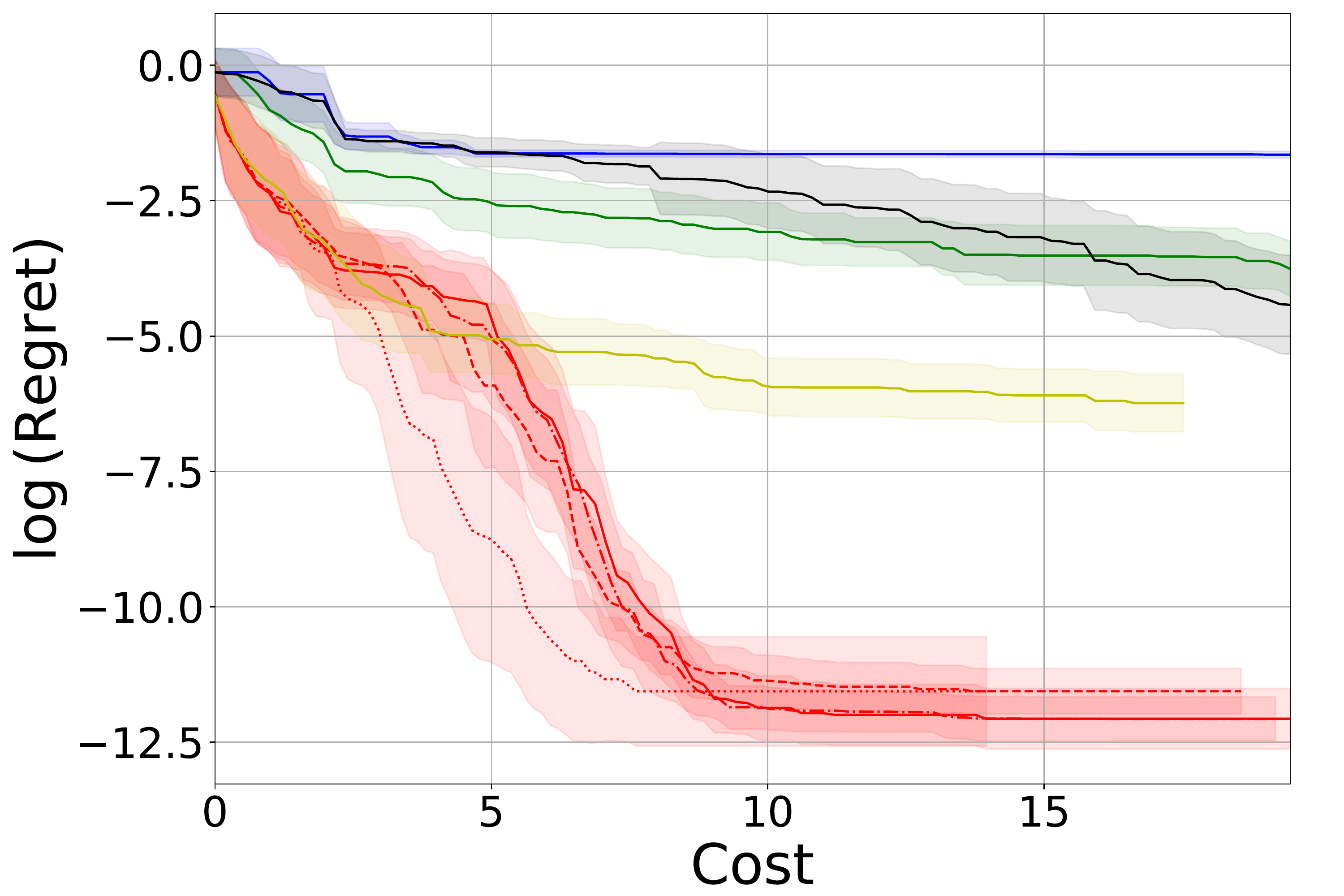}
	\includegraphics[width = 0.32\textwidth]{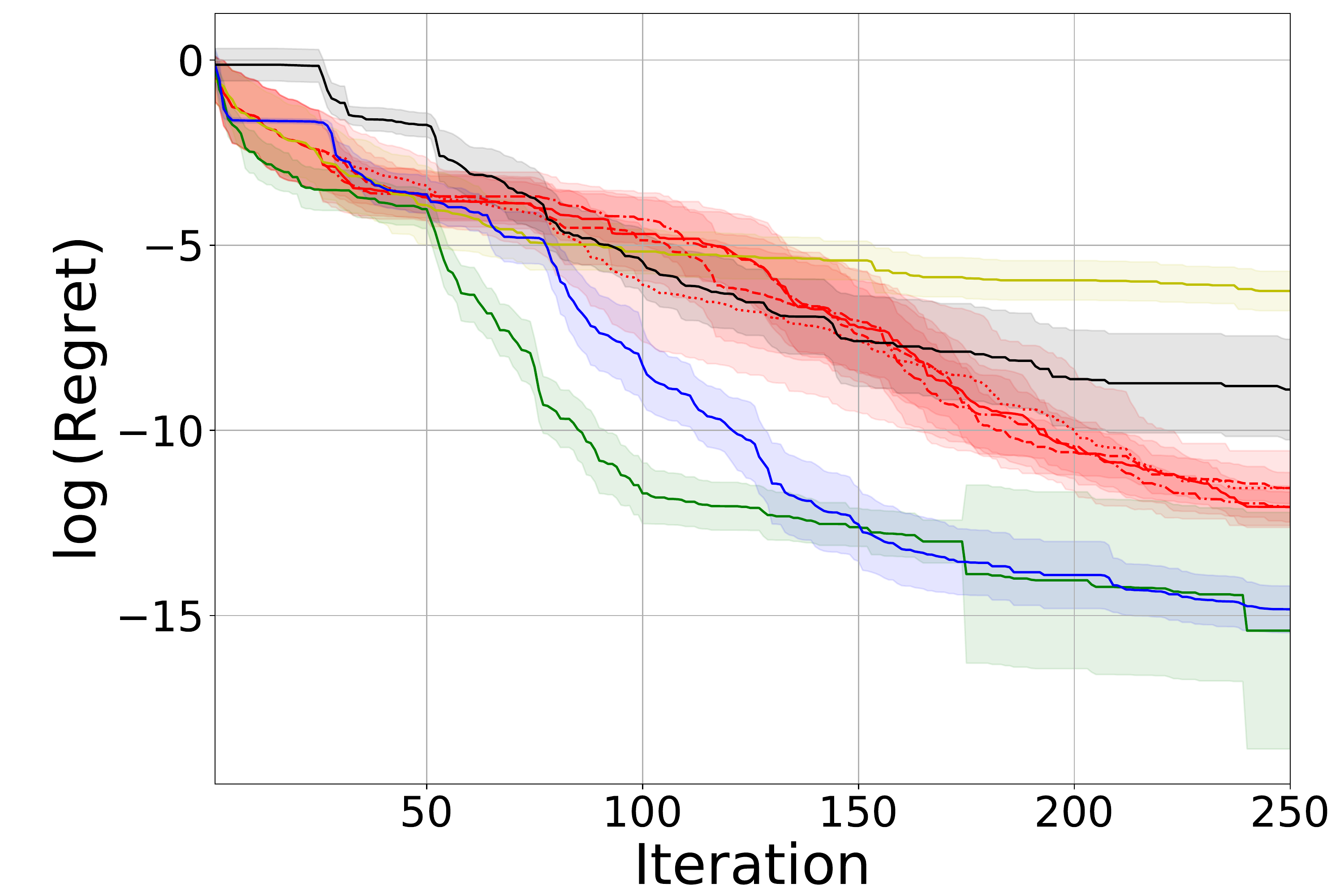}
	\includegraphics[width = 0.32\textwidth]{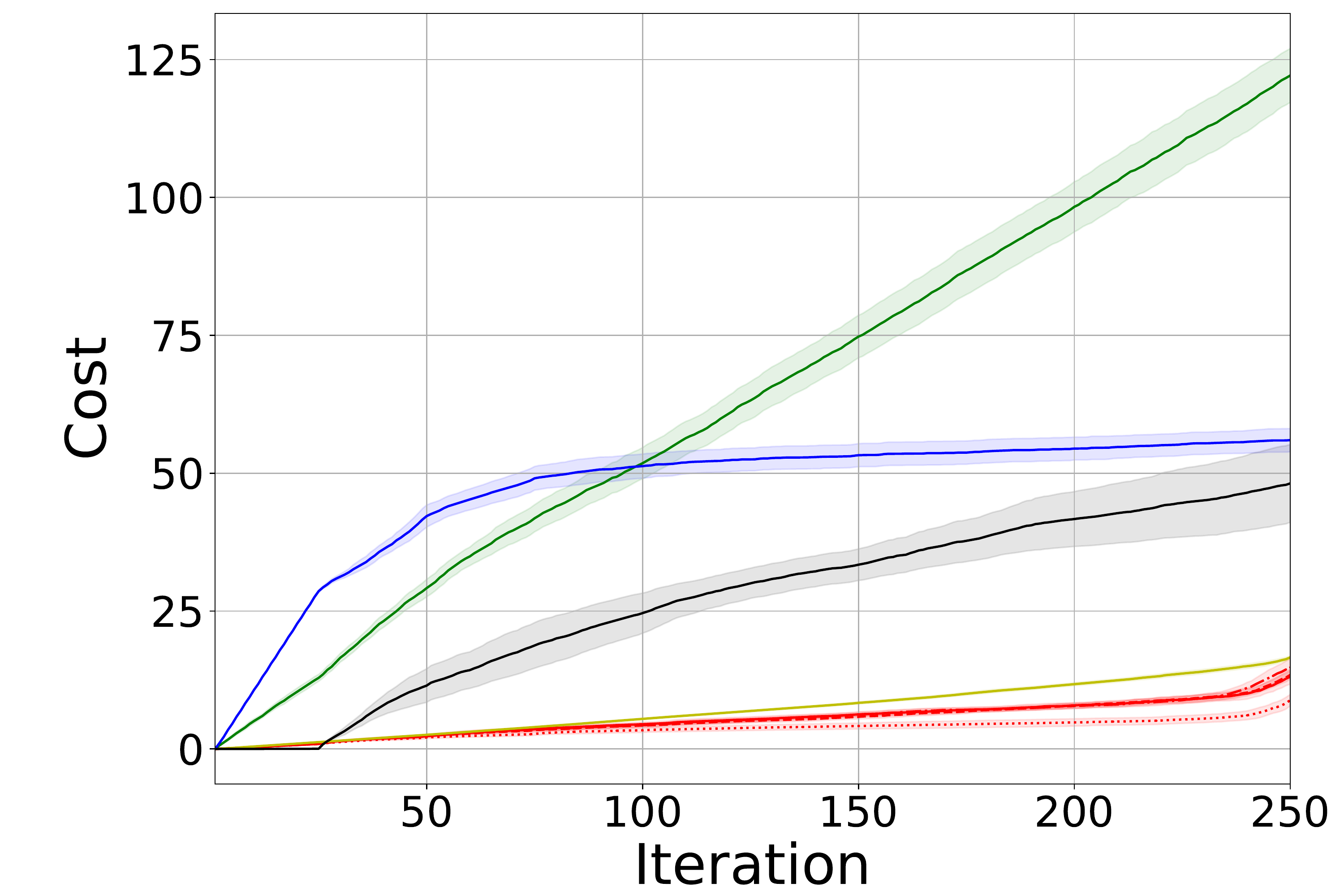}
	\caption{$T = 250$}
	\end{subfigure}
	\caption{Branin2D (Asynchronous),  $t_{delay}=25$. Each row represents a different budget. The left column shows the evolution of regret against the cost used. The middle column shows the evolution of regret with iterations, and the right columns show the evolution of the 2-norm cost. The results are similar to the shorter delay seen in Figure \ref{fig: branin2d_10_async}.}
\end{figure}

\begin{figure}[ht]
	\centering
	\begin{subfigure}[t]{\textwidth}
	\includegraphics[width = 0.32\textwidth]{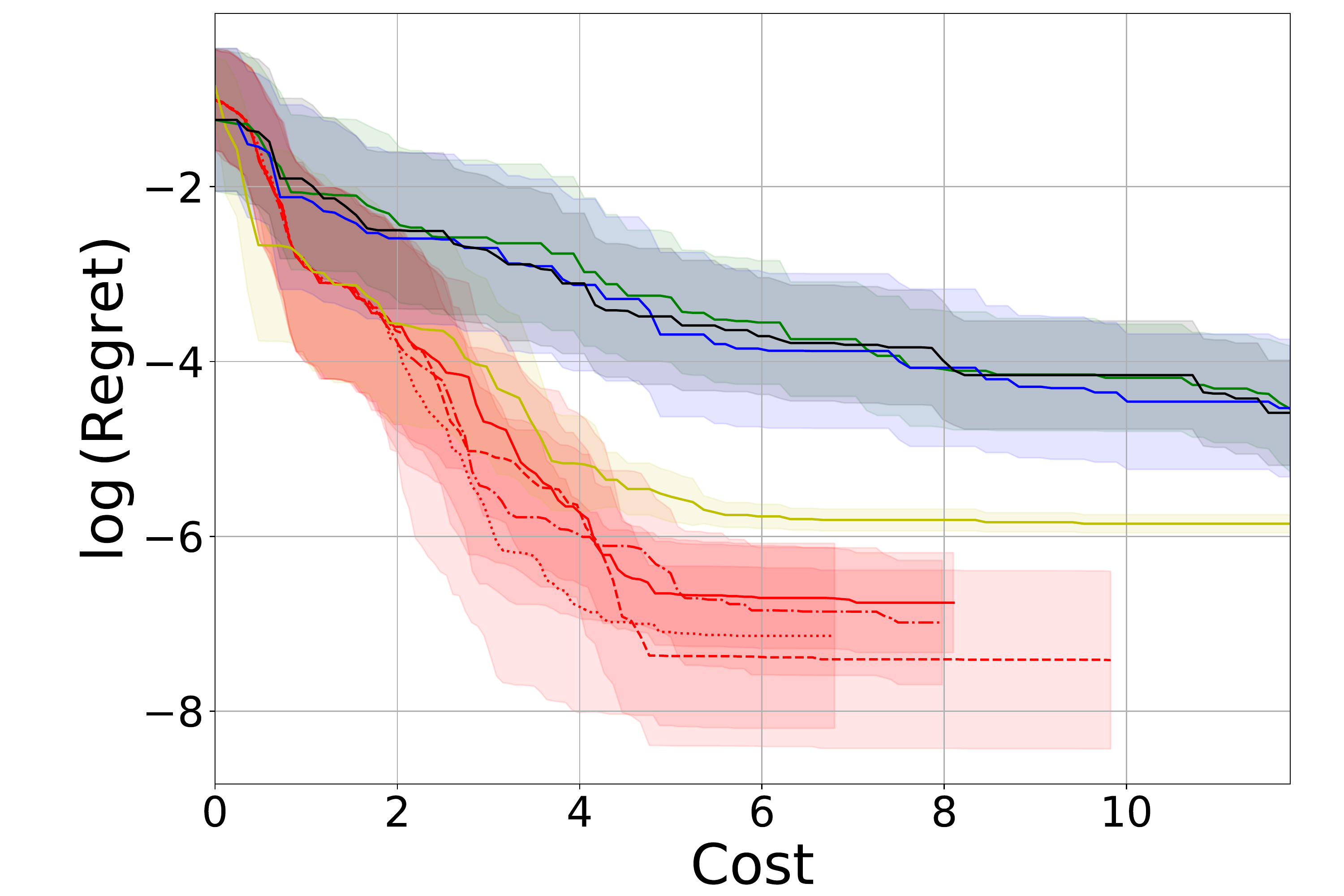}
	\includegraphics[width = 0.32\textwidth]{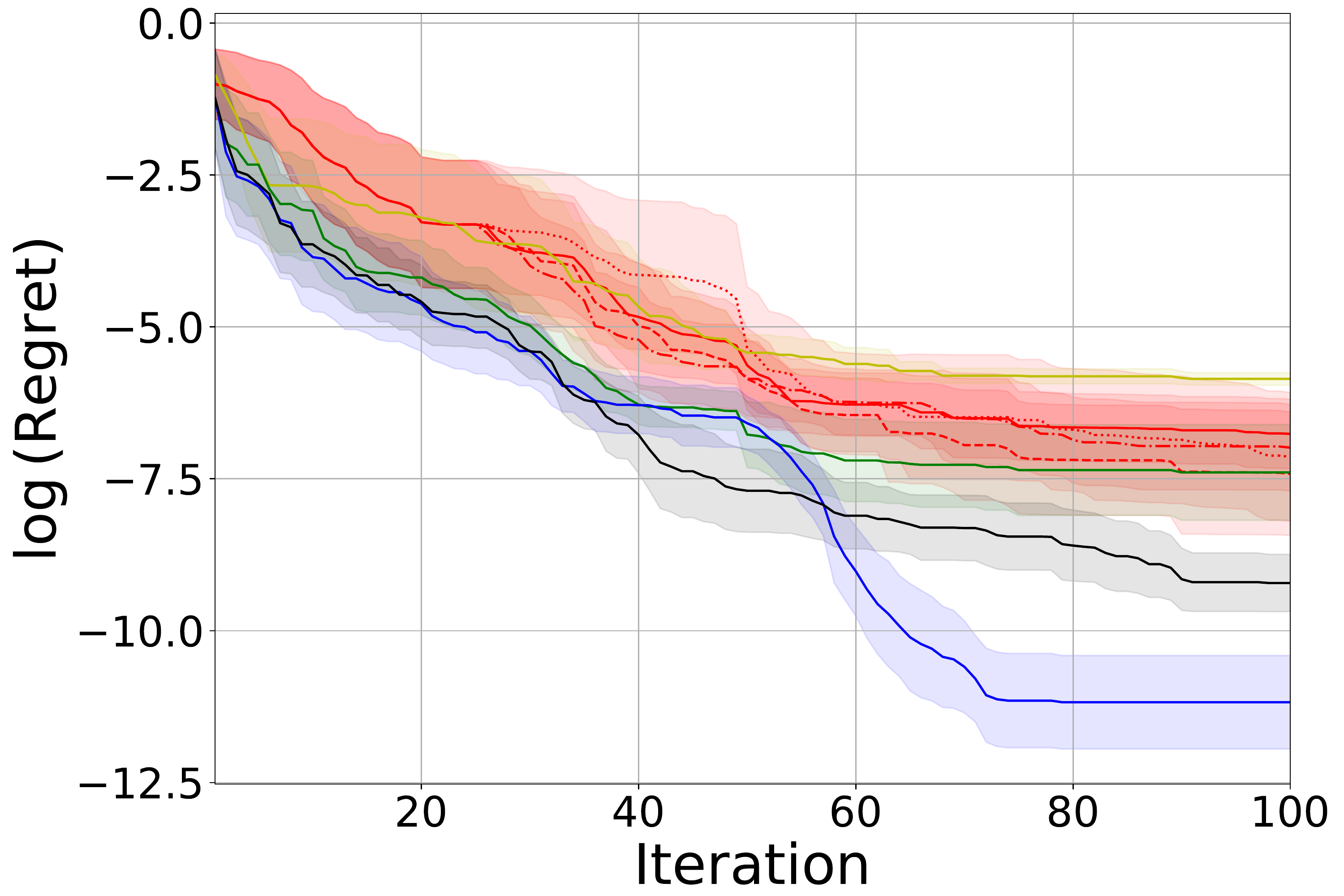}
	\includegraphics[width = 0.32\textwidth]{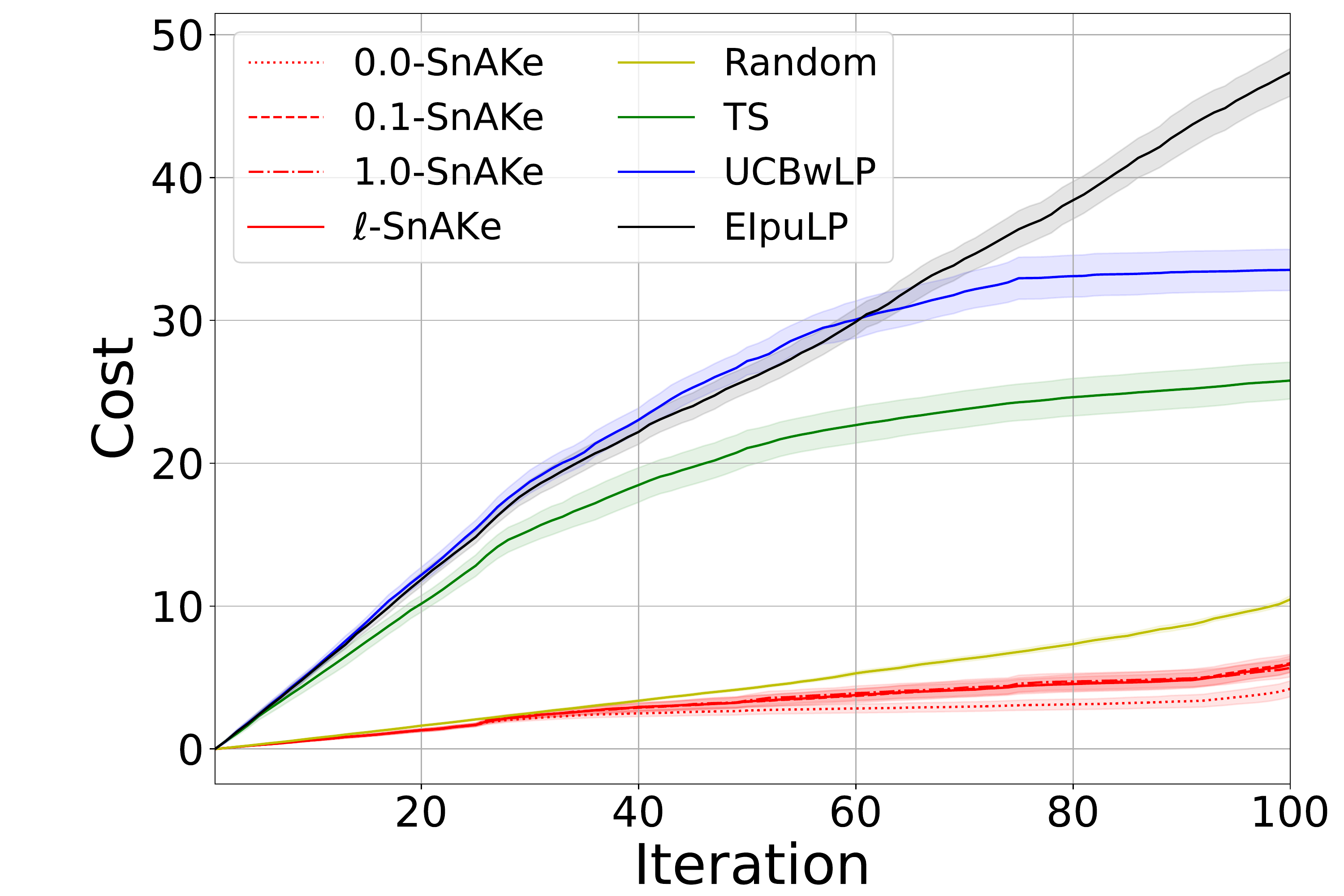}
	\caption{$T = 100$}
	\end{subfigure}
	\begin{subfigure}[t]{\textwidth}
	\includegraphics[width = 0.32\textwidth]{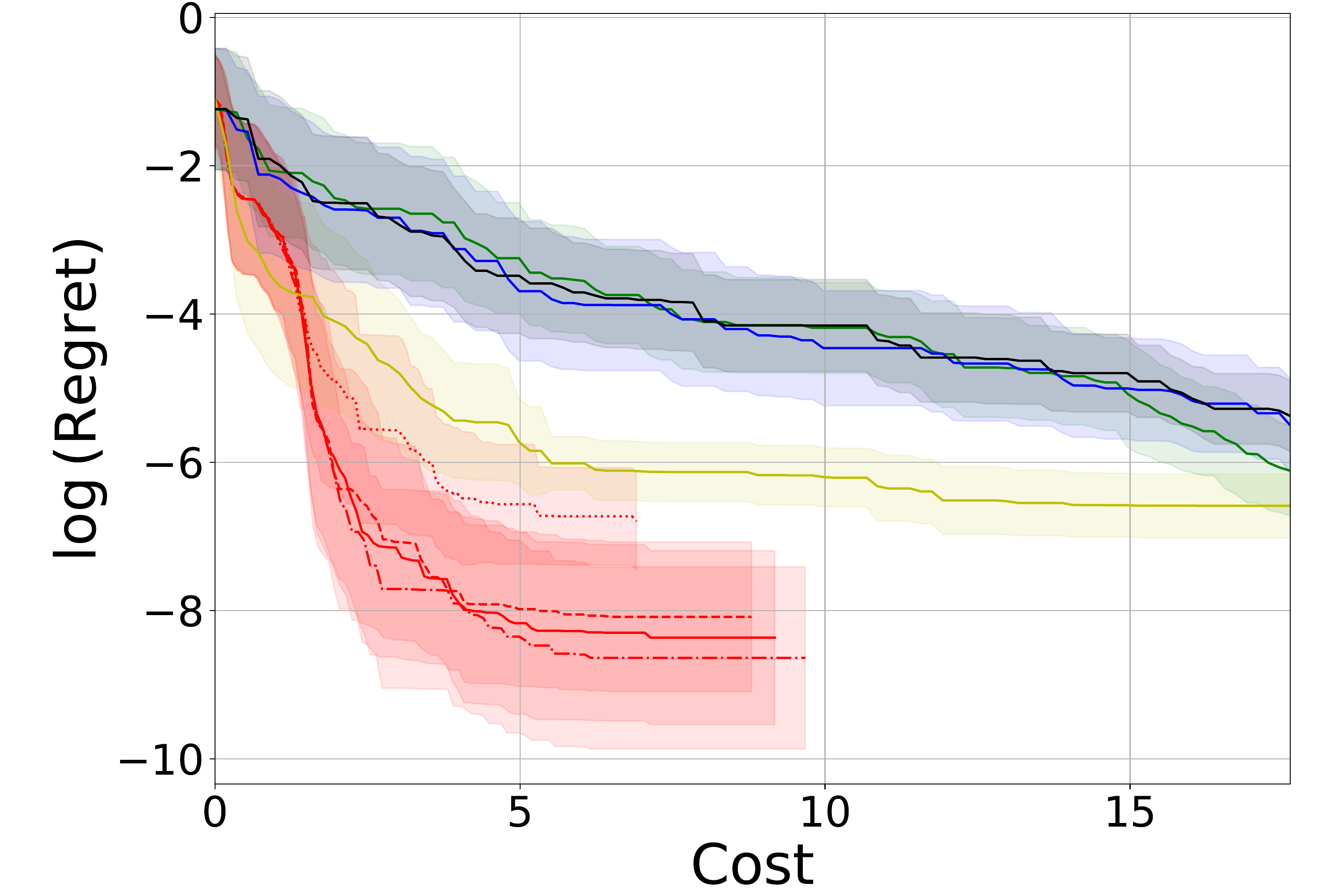}
	\includegraphics[width = 0.32\textwidth]{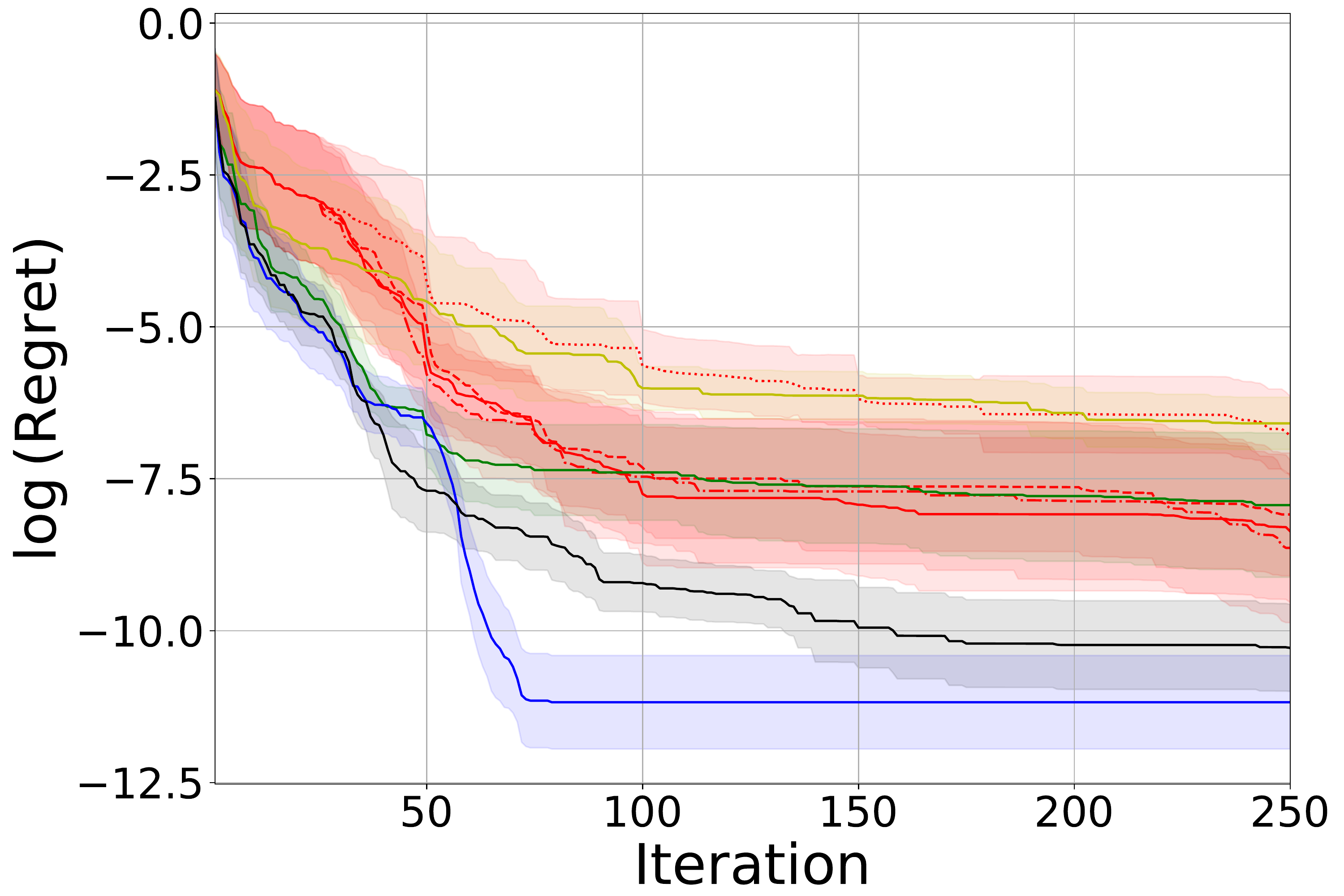}
	\includegraphics[width = 0.32\textwidth]{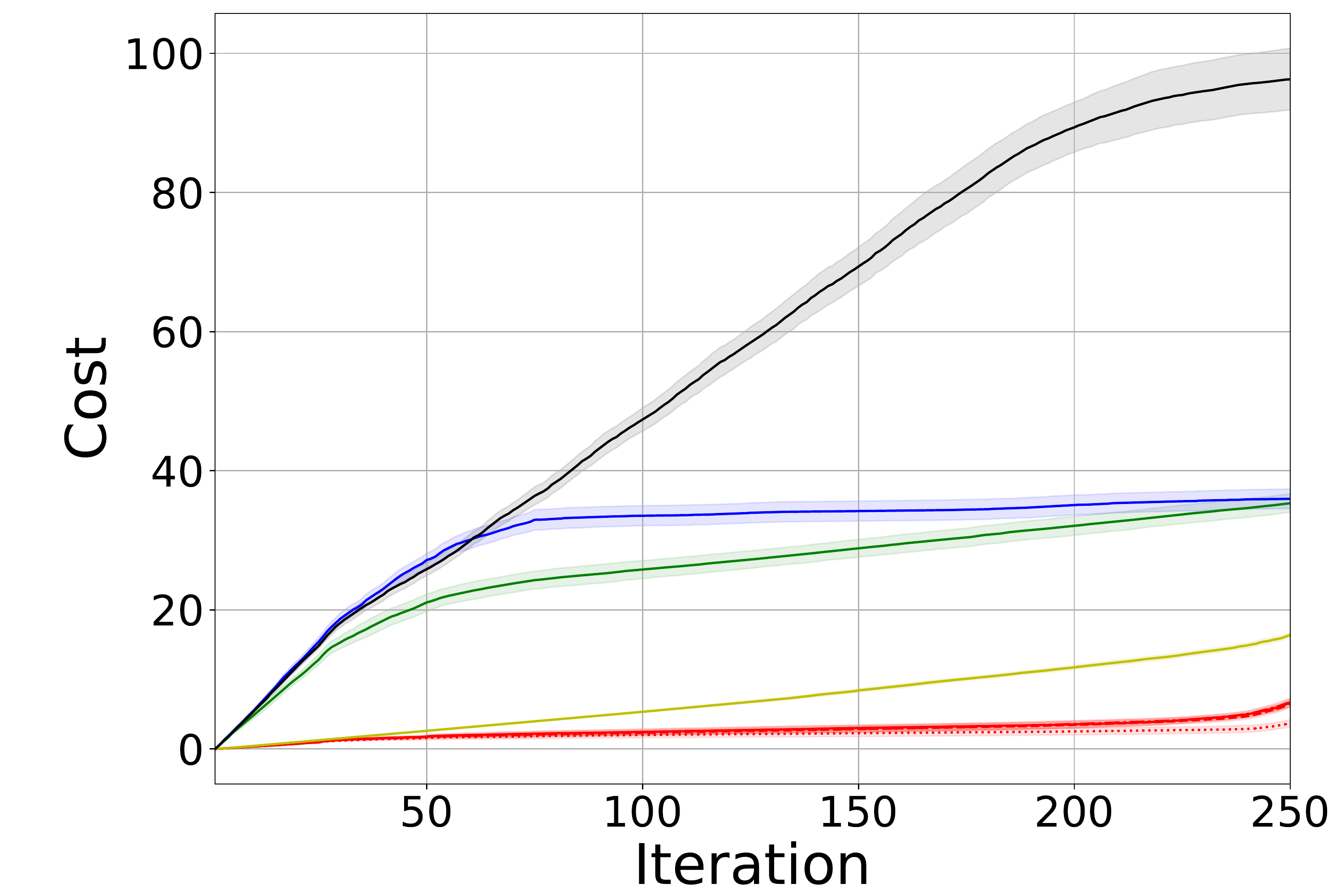}
	\caption{$T = 250$}
	\end{subfigure}
	\caption{Michaelwicz2D (Asynchronous),  $t_{delay}=25$. Each row represents a different budget. The left column shows the evolution of regret against the cost used. The middle column shows the evolution of regret with iterations, and the right columns show the evolution of the 2-norm cost. Similar results to shorter delay, see Figure \ref{fig: michael_async_td_10}.}
\end{figure}

\begin{figure}[ht]
	\centering
	\begin{subfigure}[t]{\textwidth}
	\includegraphics[width = 0.32\textwidth]{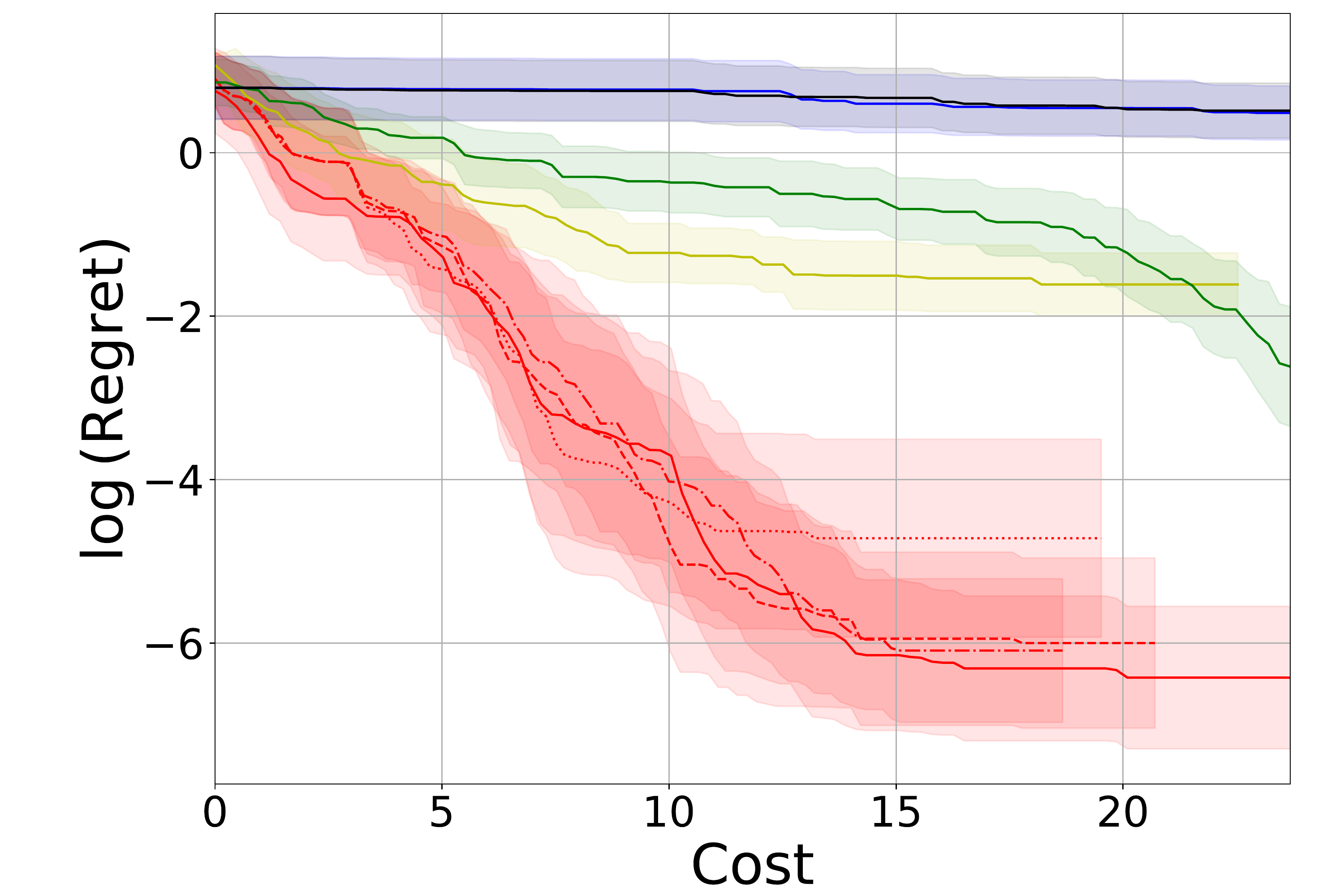}
	\includegraphics[width = 0.32\textwidth]{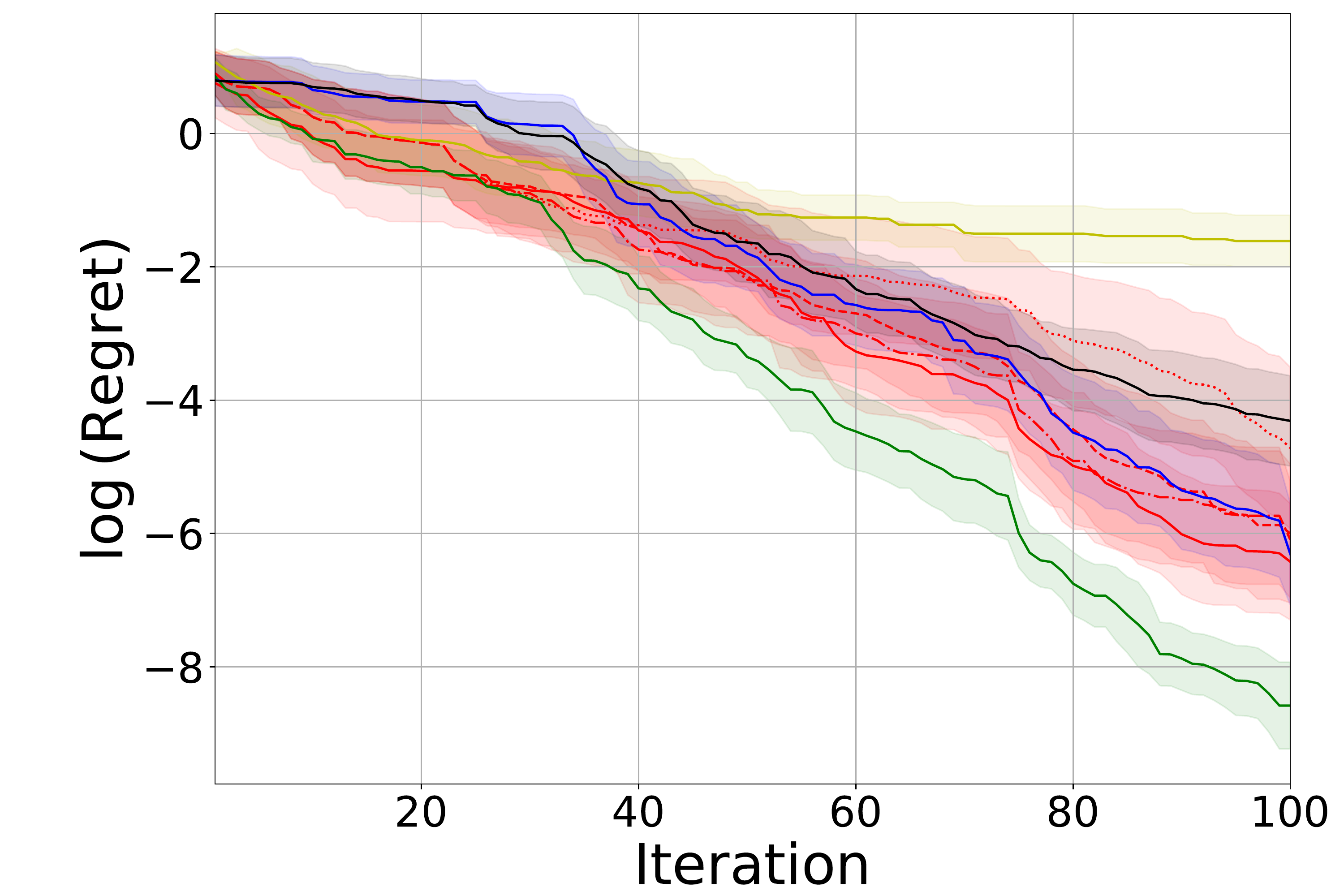}
	\includegraphics[width = 0.32\textwidth]{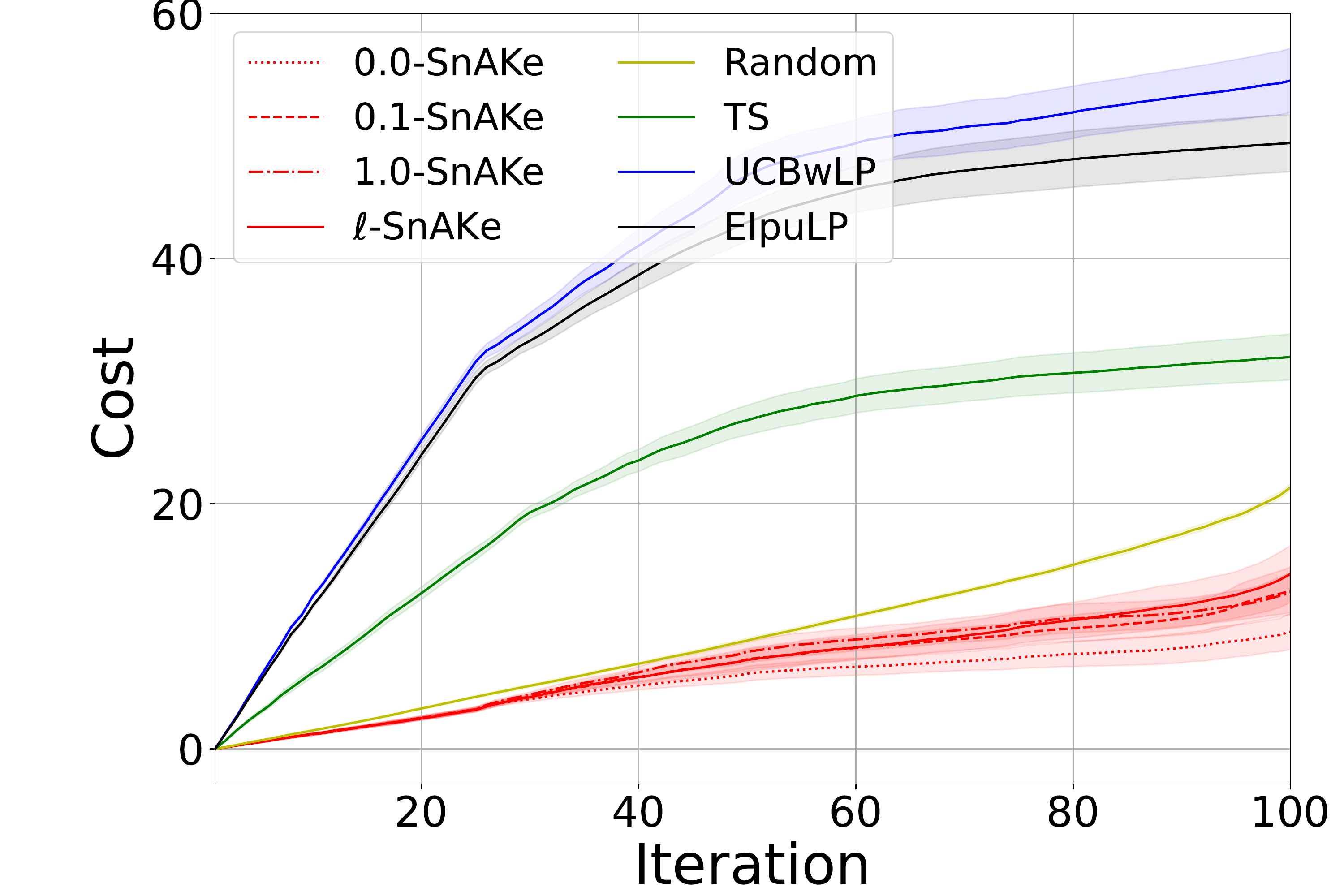}
	\caption{$T = 100$}
	\end{subfigure}
	\begin{subfigure}[t]{\textwidth}
	\includegraphics[width = 0.32\textwidth]{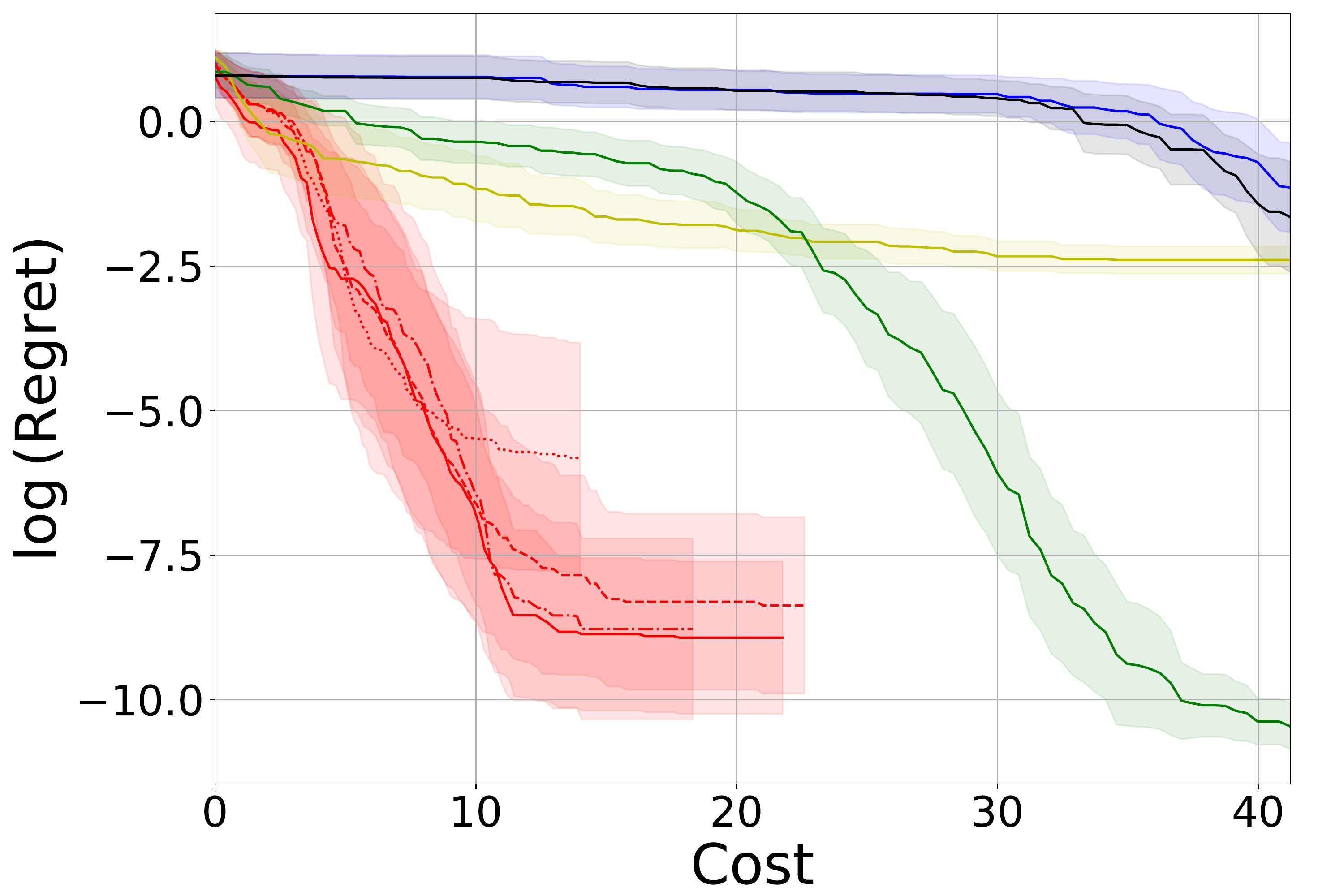}
	\includegraphics[width = 0.32\textwidth]{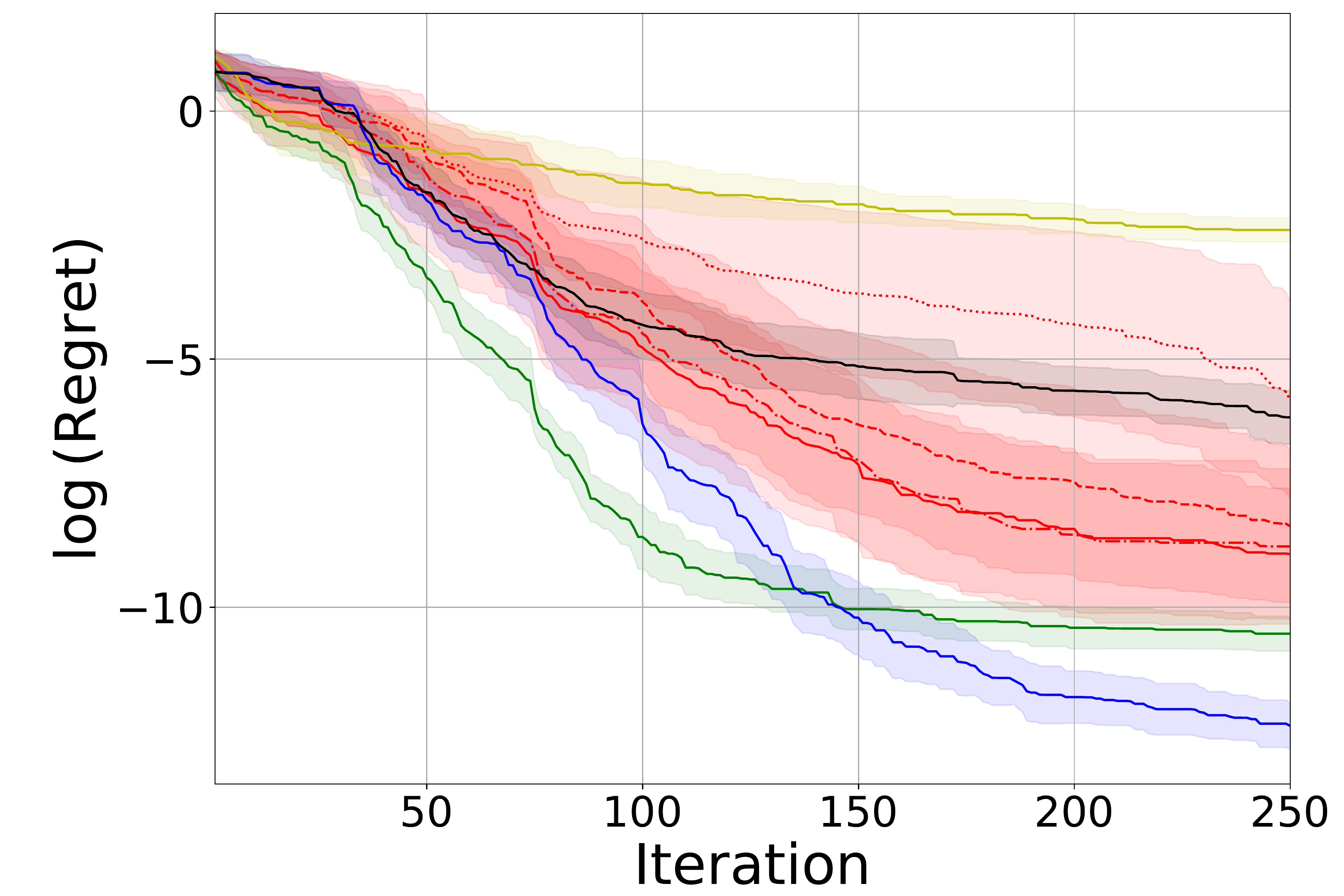}
	\includegraphics[width = 0.32\textwidth]{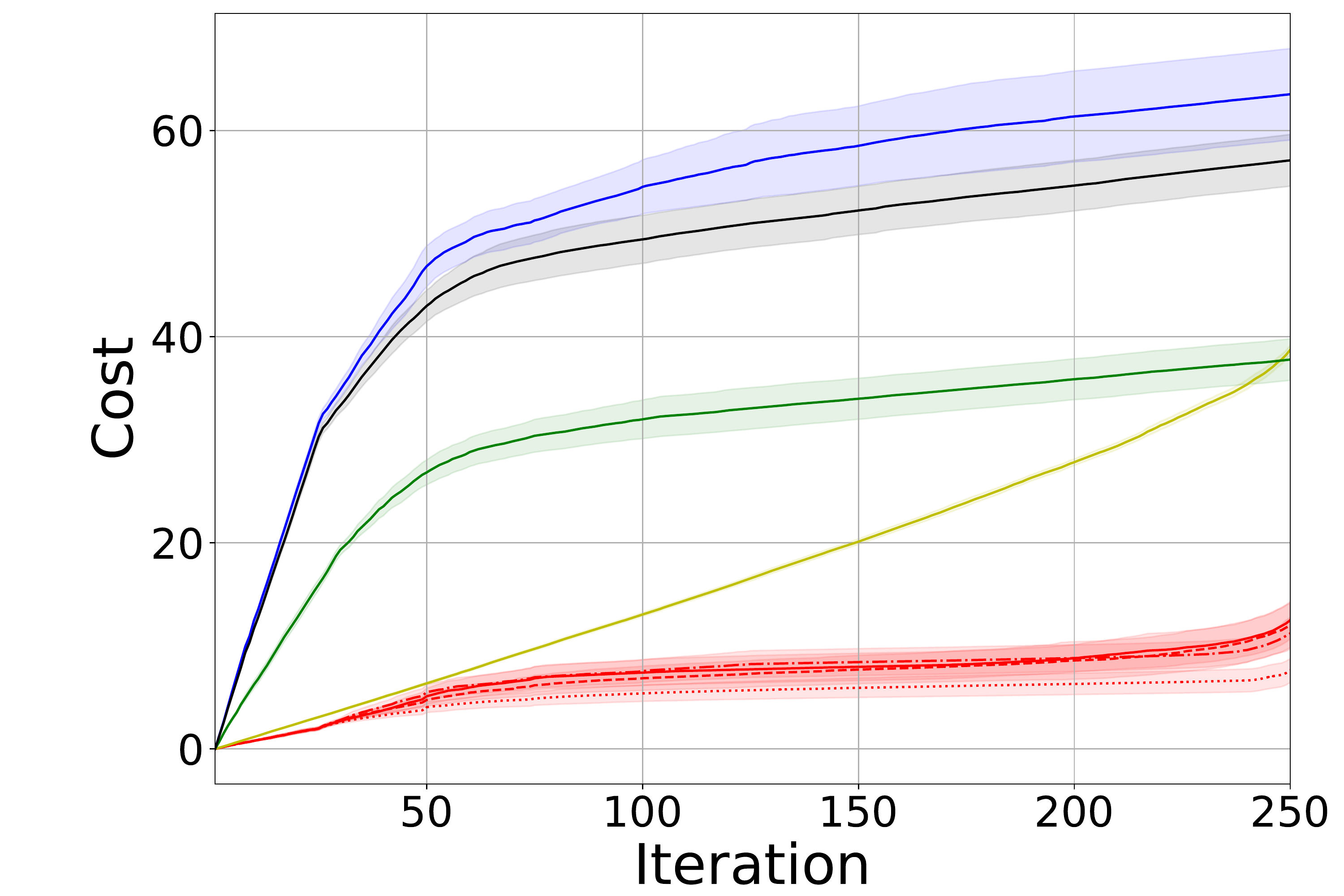}
	\caption{$T = 250$}
	\end{subfigure}
	\caption{Hartmann3D (Asynchronous),  $t_{delay}=25$. Each row represents a different budget. The left column shows the evolution of regret against the cost used. The middle column shows the evolution of regret with iterations, and the right columns show the evolution of the 2-norm cost. Similar results to the case with smaller delay, see Figure \ref{fig: hartmann_3d_async_10}.}
\end{figure}

\begin{figure}[ht]
	\centering
	\begin{subfigure}[t]{\textwidth}
	\includegraphics[width = 0.32\textwidth]{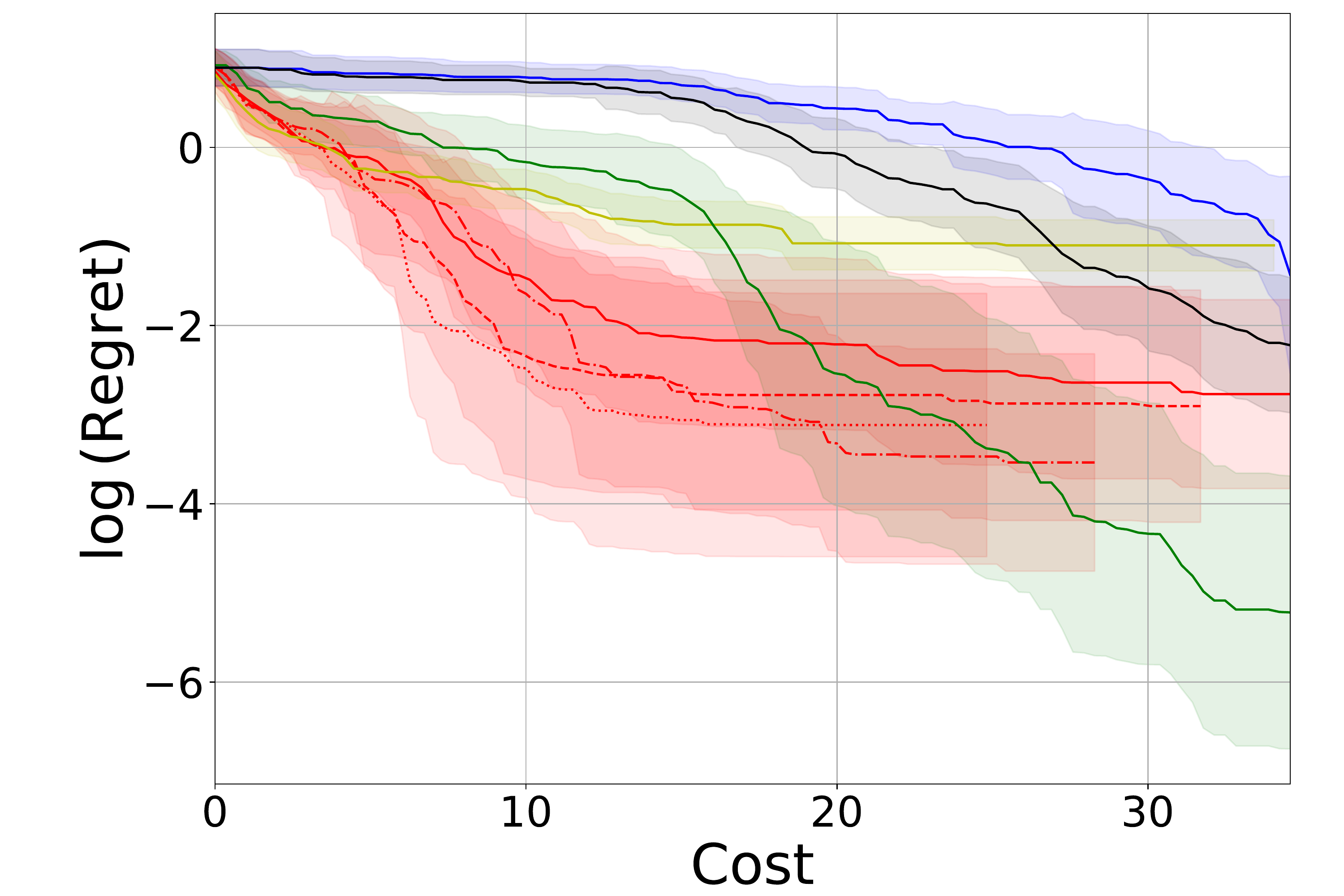}
	\includegraphics[width = 0.32\textwidth]{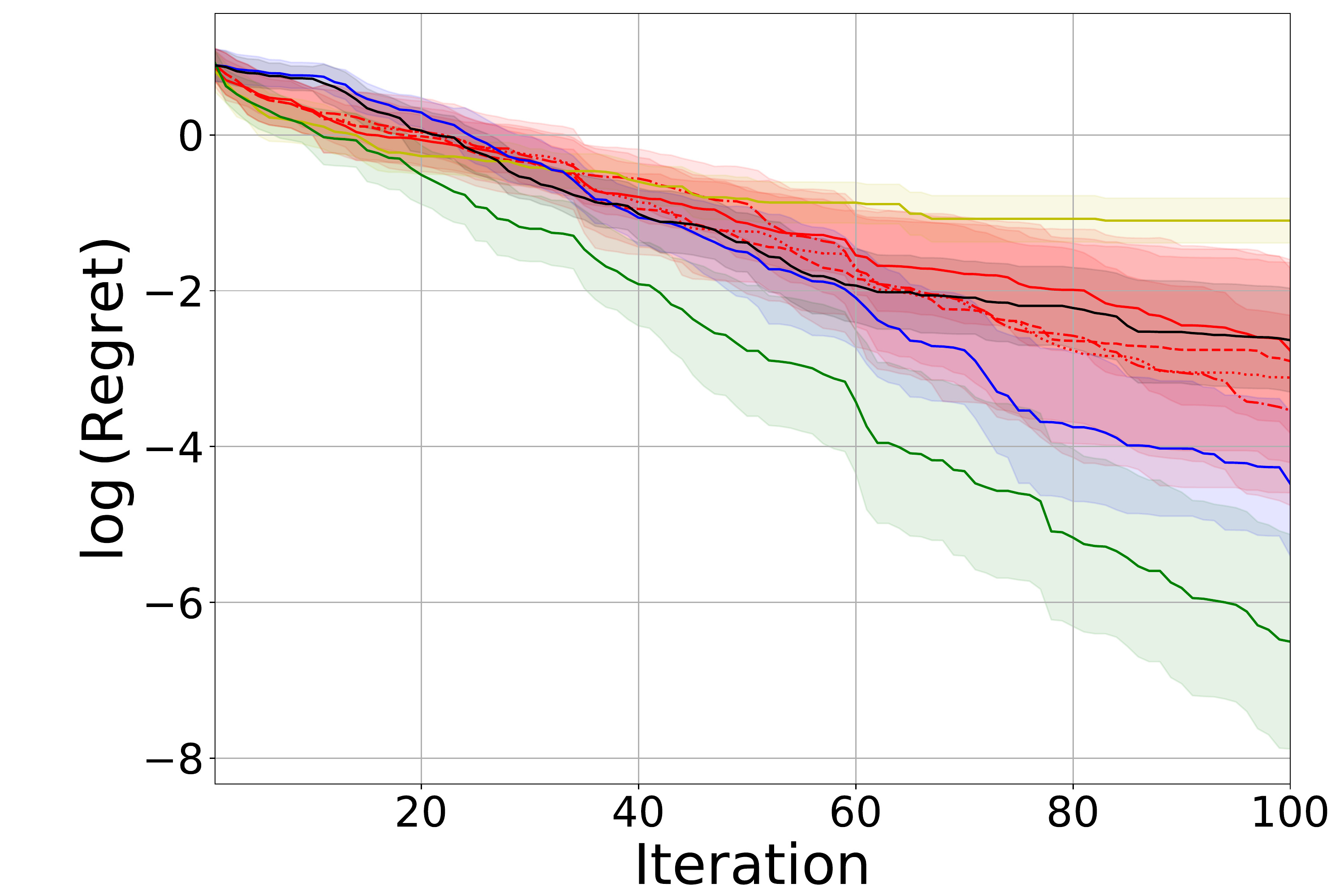}
	\includegraphics[width = 0.32\textwidth]{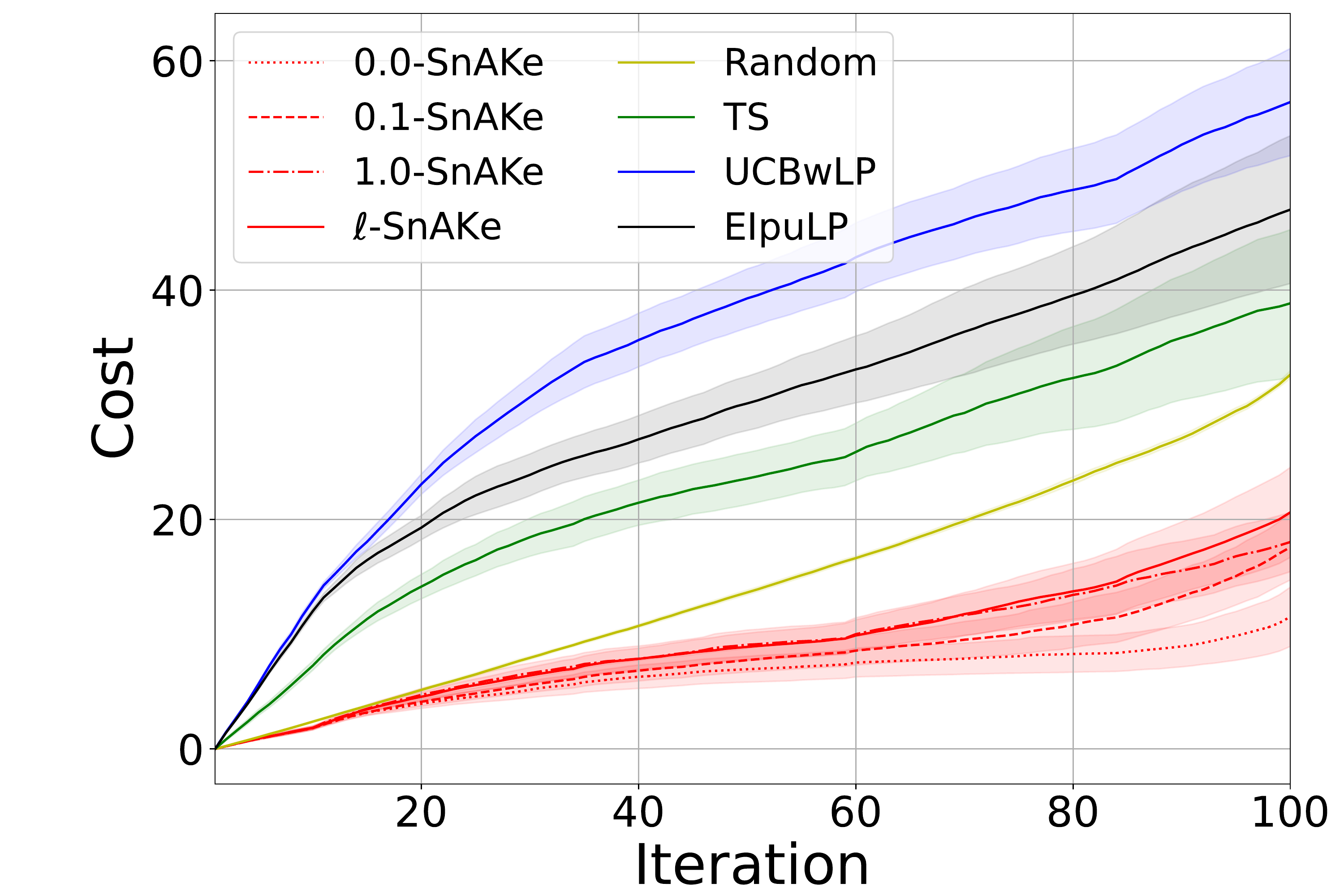}
	\caption{$T = 100$}
	\end{subfigure}
	\begin{subfigure}[t]{\textwidth}
	\includegraphics[width = 0.32\textwidth]{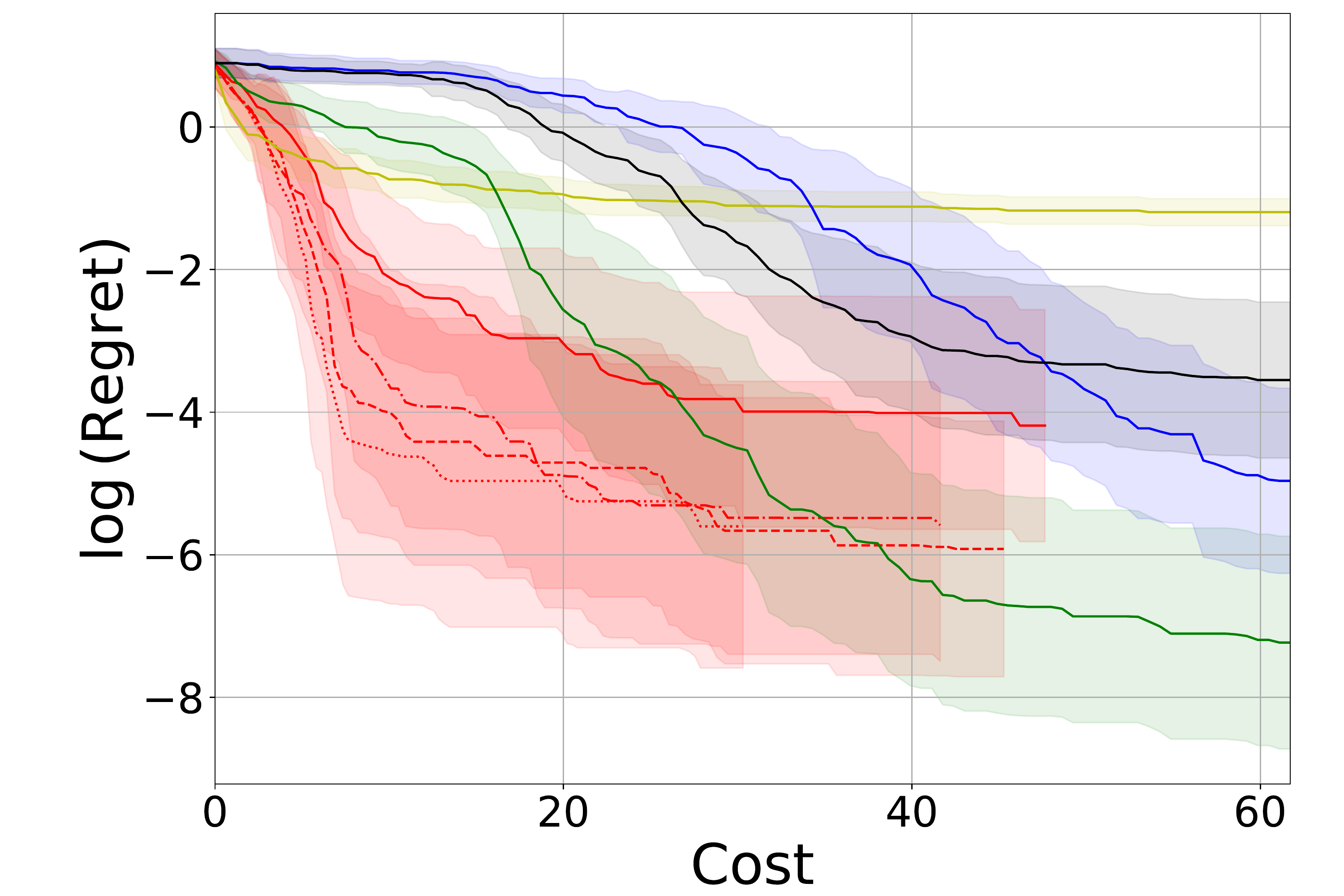}
	\includegraphics[width = 0.32\textwidth]{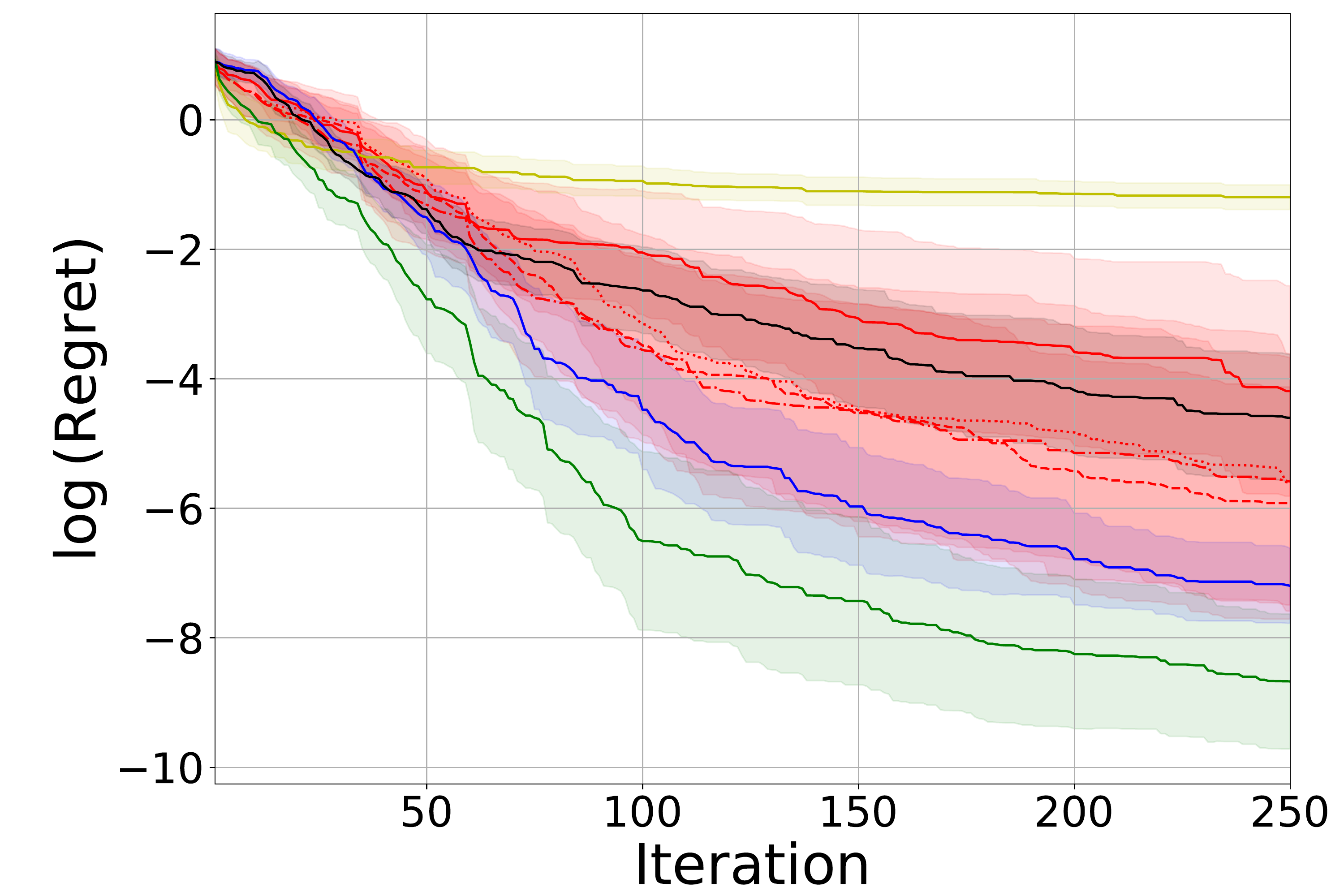}
	\includegraphics[width = 0.32\textwidth]{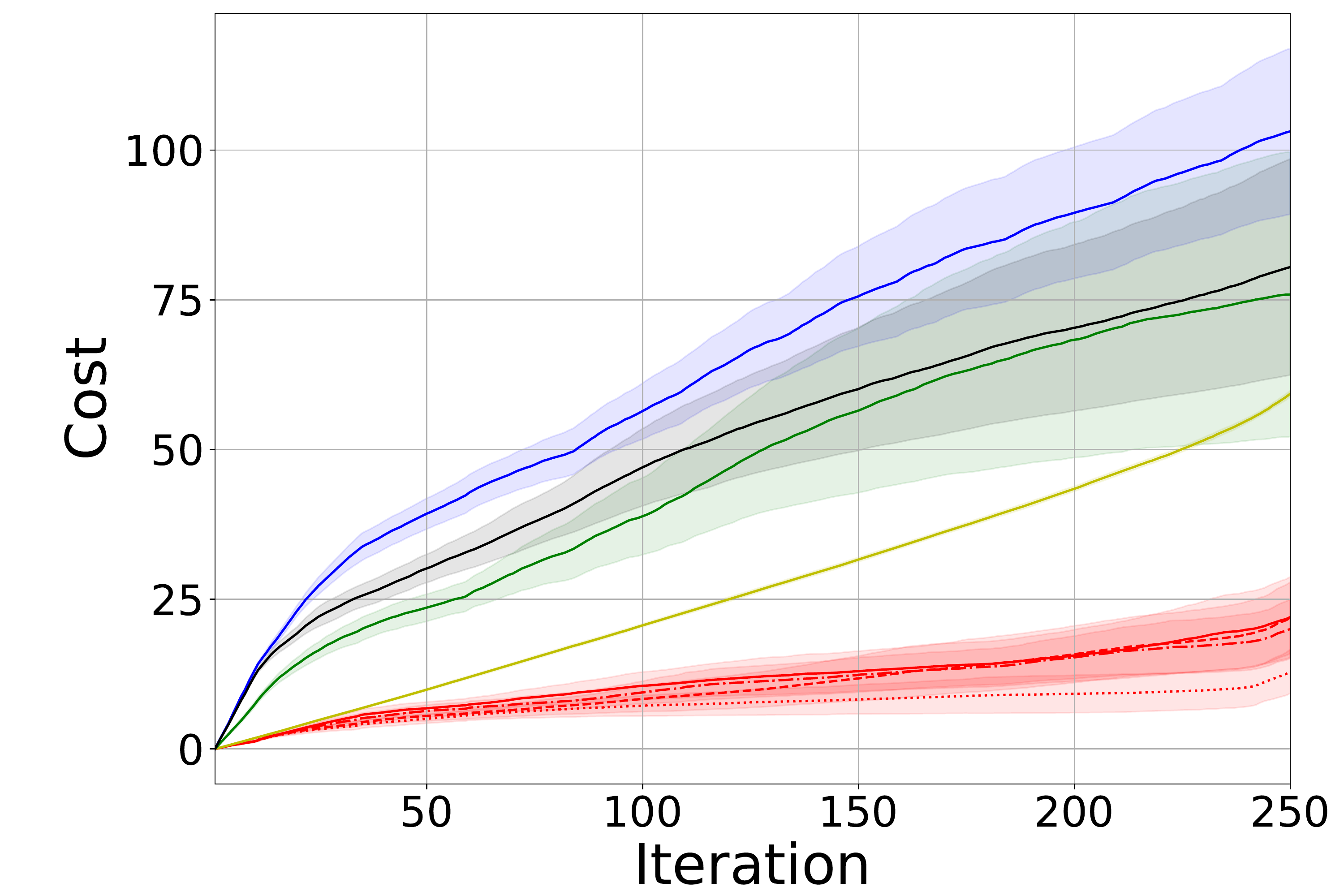}
	\caption{$T = 250$}
	\end{subfigure}
	\caption{Hartmann4D (Asynchronous),  $t_{delay}=10$. Each row represents a different budget. The left column shows the evolution of regret against the cost used. The middle column shows the evolution of regret with iterations, and the right columns show the evolution of the 2-norm cost. Similar results to other Hartmann benchmarks, see Figure \ref{fig: hartmann_3d_async_10}.}
\end{figure}

\begin{figure}[ht]
	\centering
	\begin{subfigure}[t]{\textwidth}
	\includegraphics[width = 0.32\textwidth]{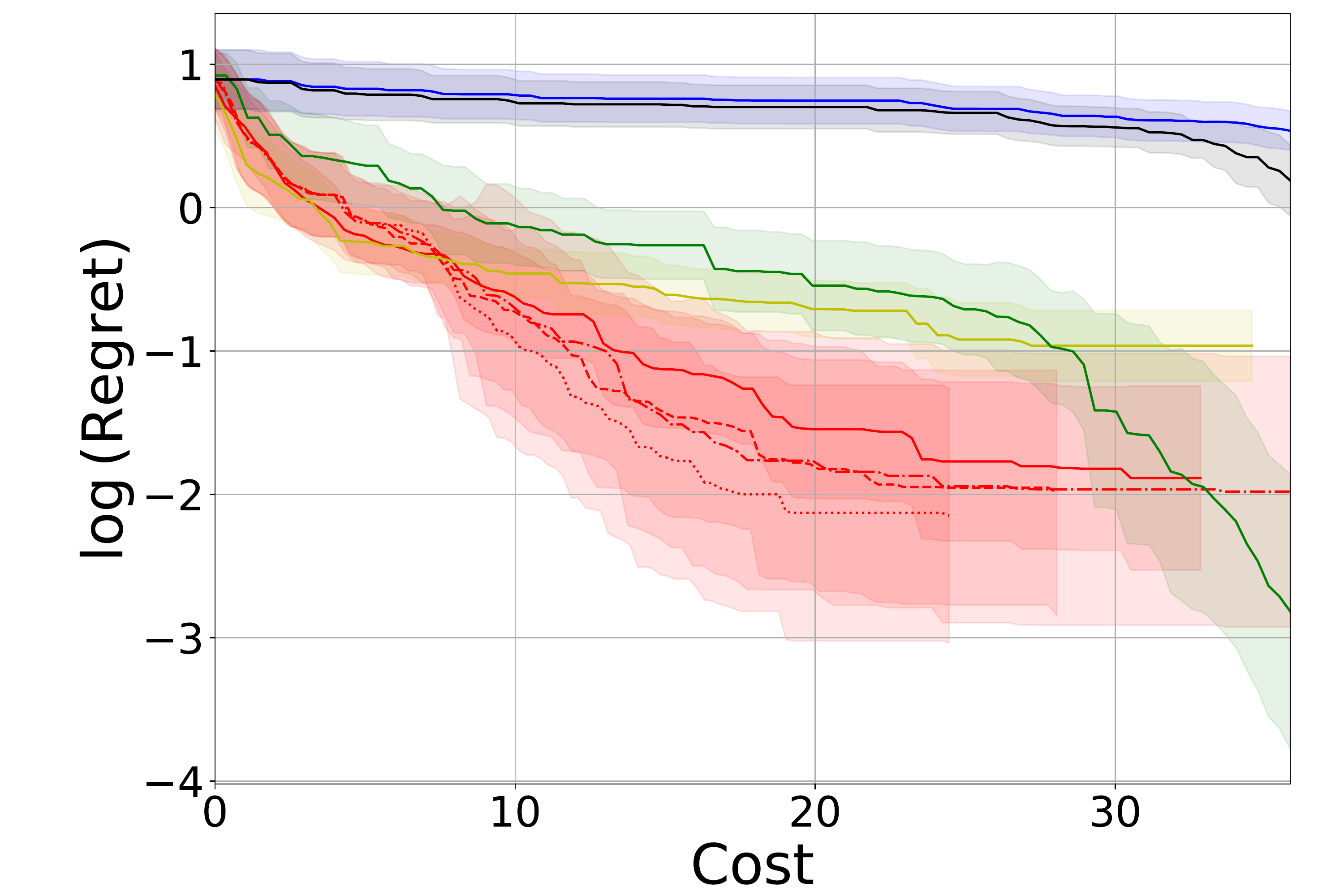}
	\includegraphics[width = 0.32\textwidth]{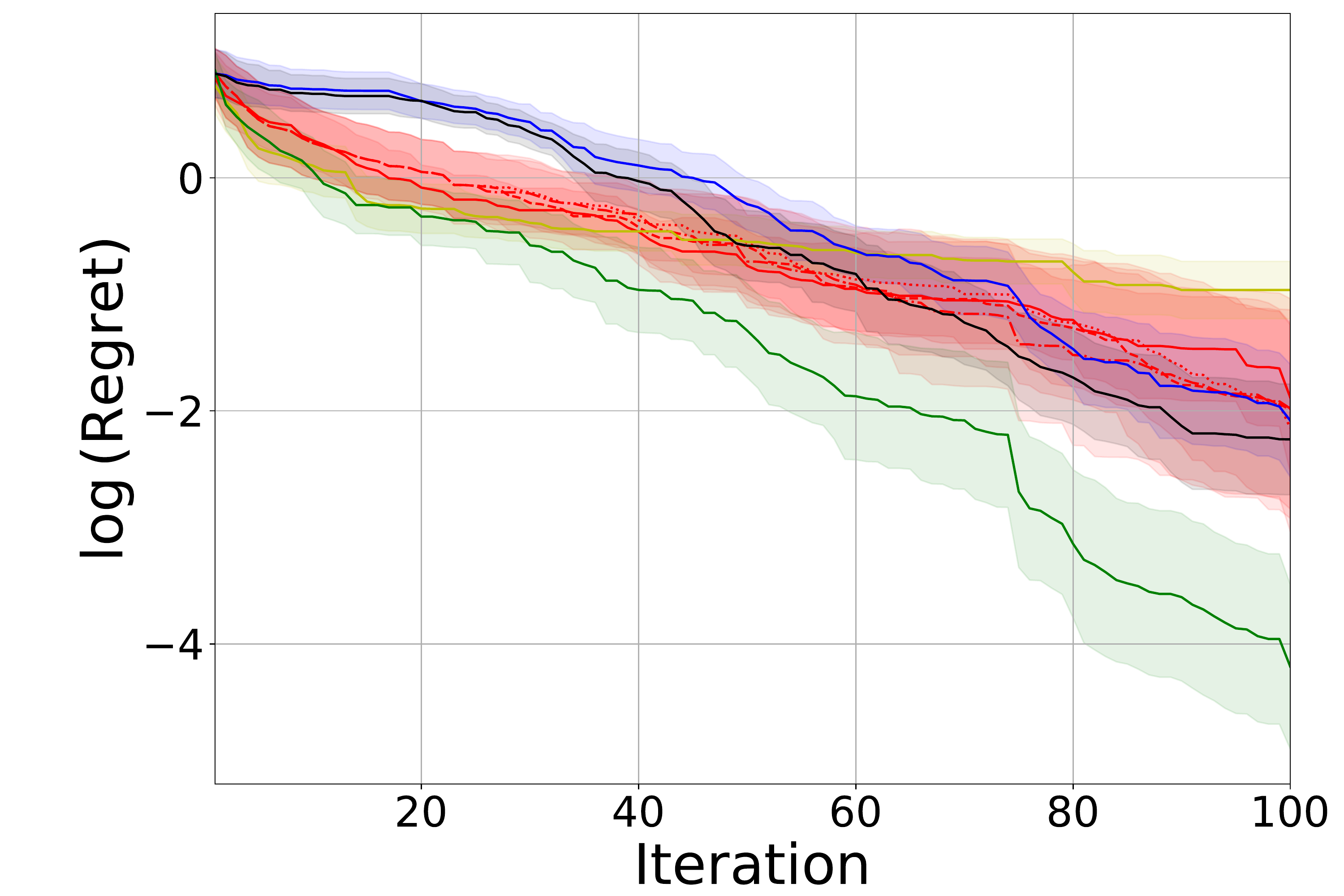}
	\includegraphics[width = 0.32\textwidth]{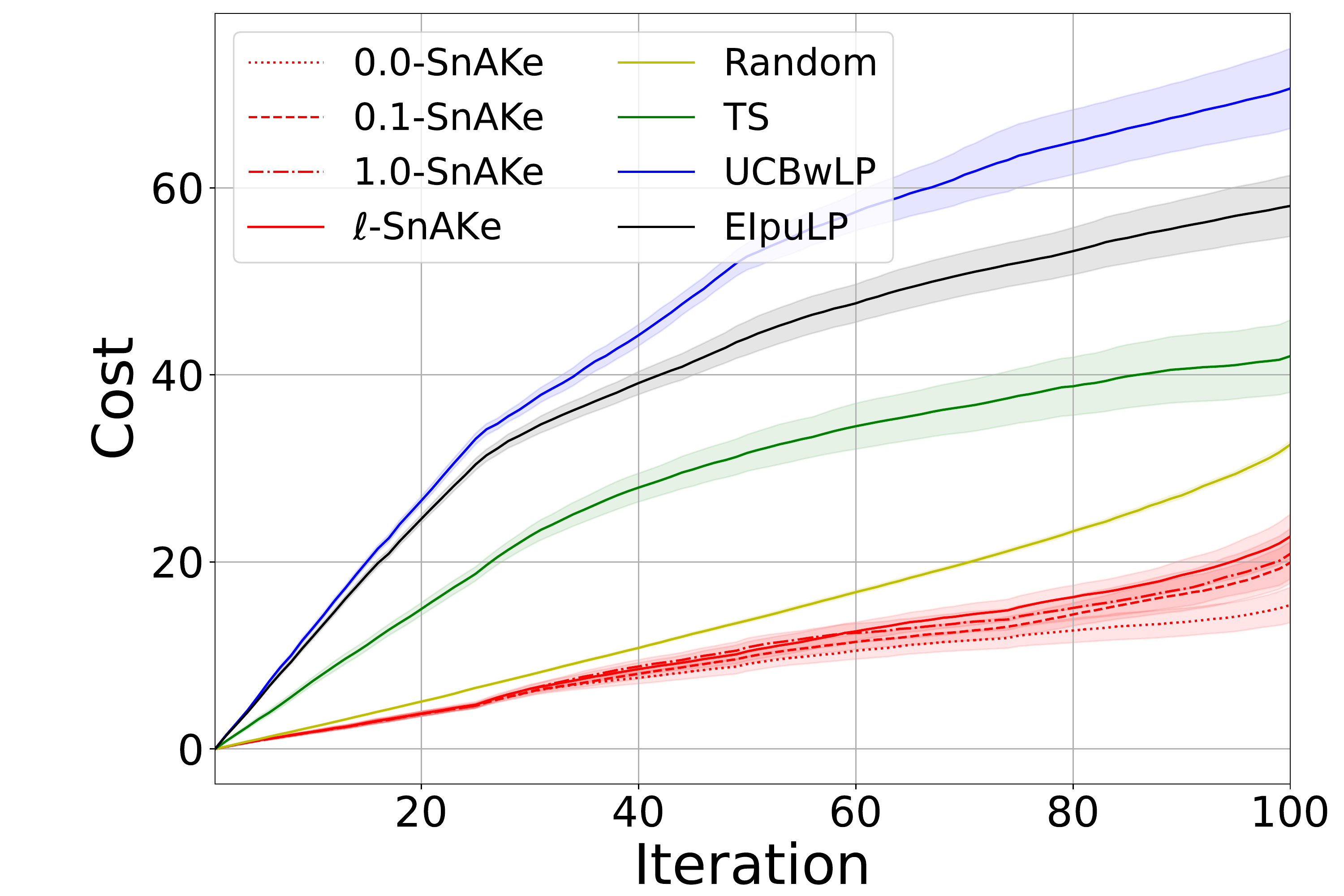}
	\caption{$T = 100$}
	\end{subfigure}
	\begin{subfigure}[t]{\textwidth}
	\includegraphics[width = 0.32\textwidth]{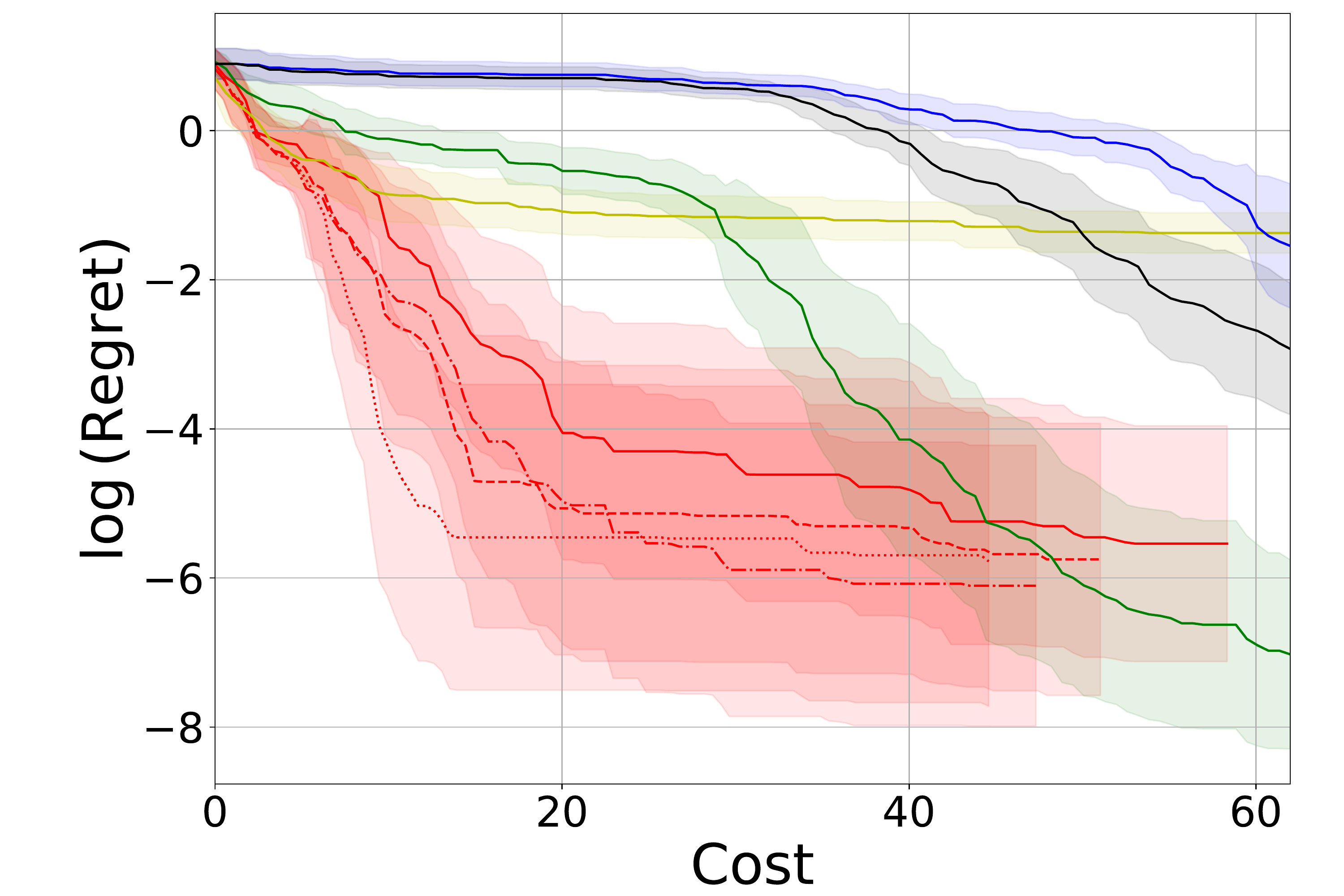}
	\includegraphics[width = 0.32\textwidth]{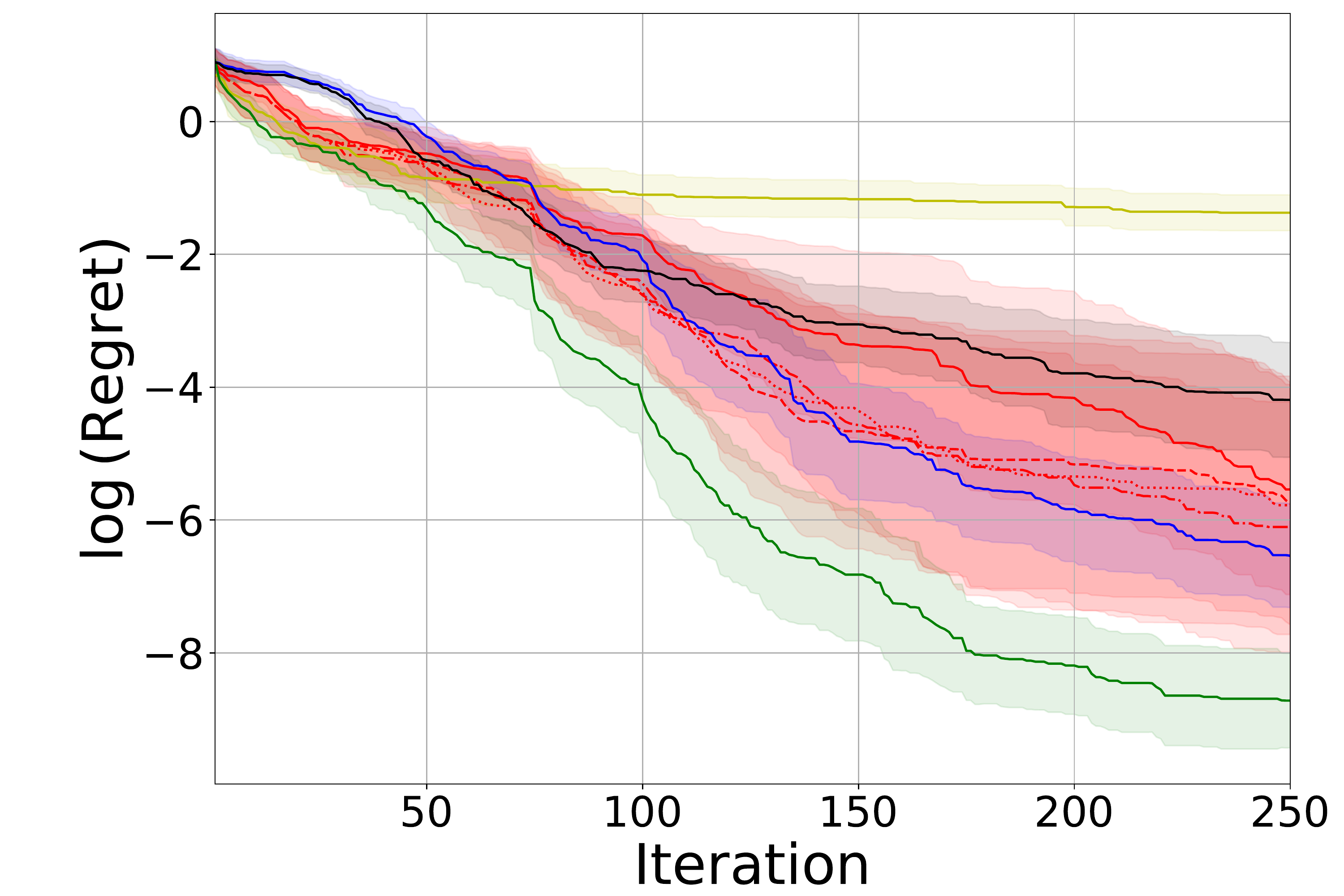}
	\includegraphics[width = 0.32\textwidth]{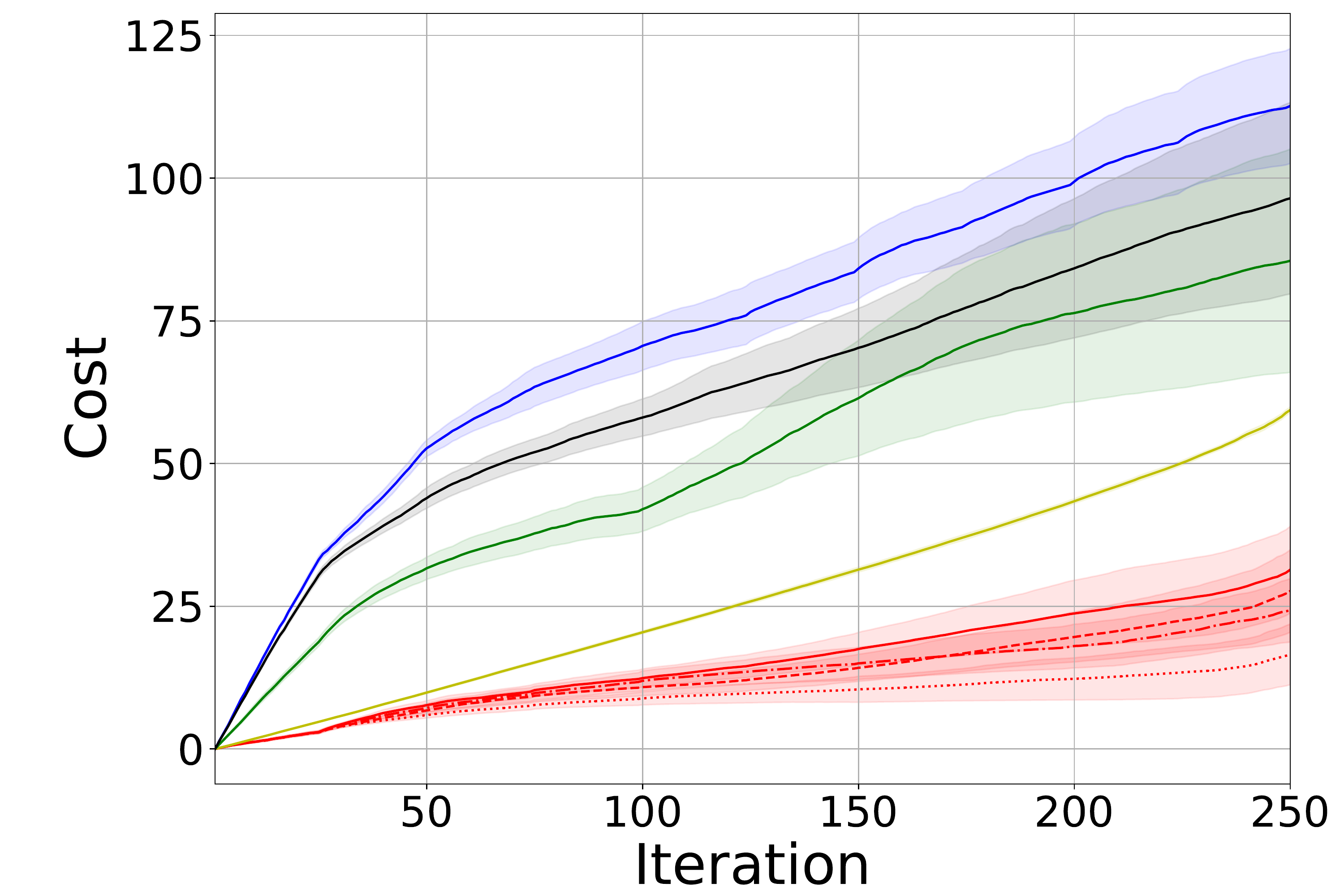}
	\caption{$T = 250$}
	\end{subfigure}
	\caption{Hartmann4D (Asynchronous),  $t_{delay}=25$. Each row represents a different budget. The left column shows the evolution of regret against the cost used. The middle column shows the evolution of regret with iterations, and the right columns show the evolution of the 2-norm cost. Similar results to other Hartmann benchmarks, see Figure \ref{fig: hartmann_3d_async_10}.}
\end{figure}

\begin{figure}[ht]
	\centering
	\begin{subfigure}[t]{\textwidth}
	\includegraphics[width = 0.32\textwidth]{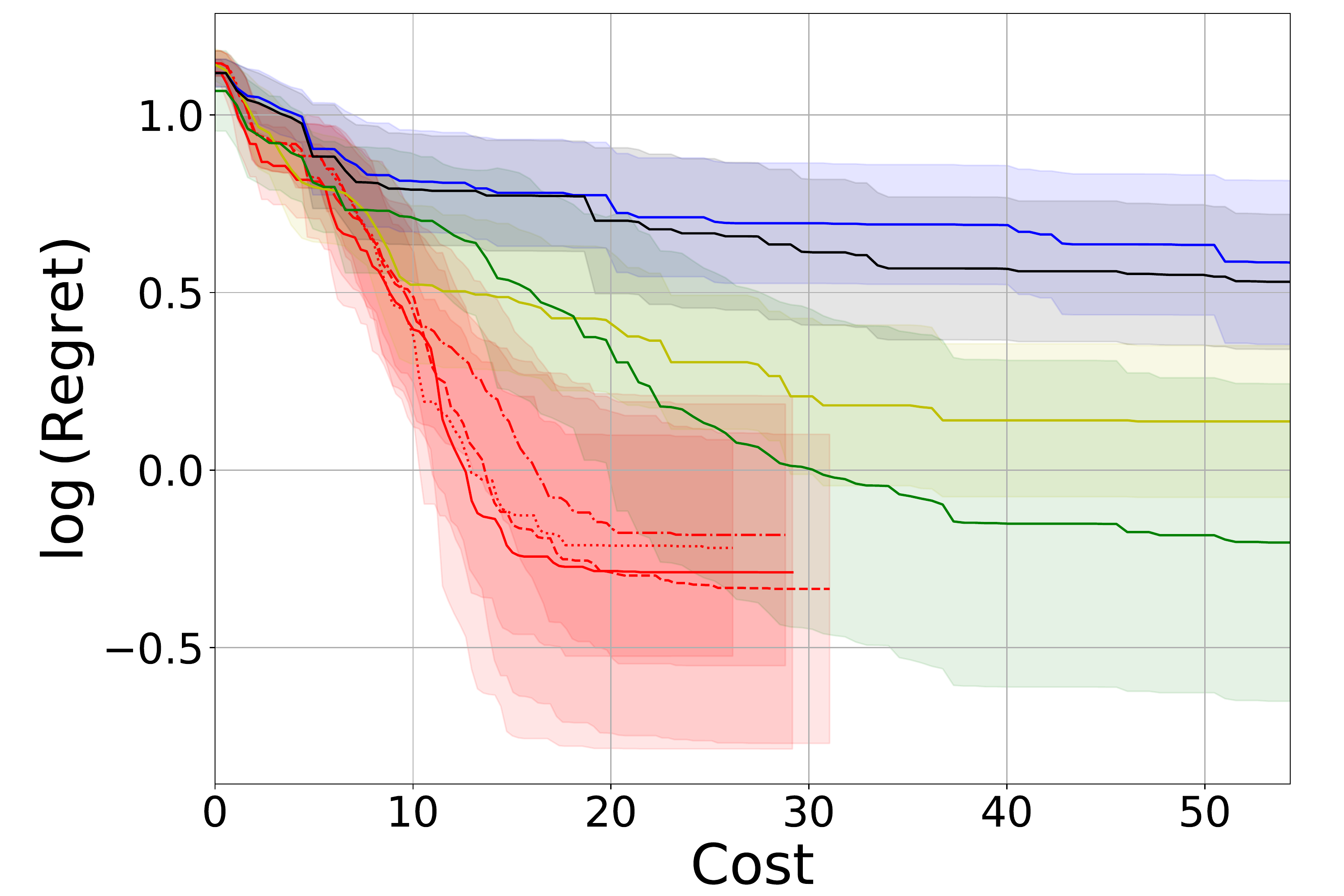}
	\includegraphics[width = 0.32\textwidth]{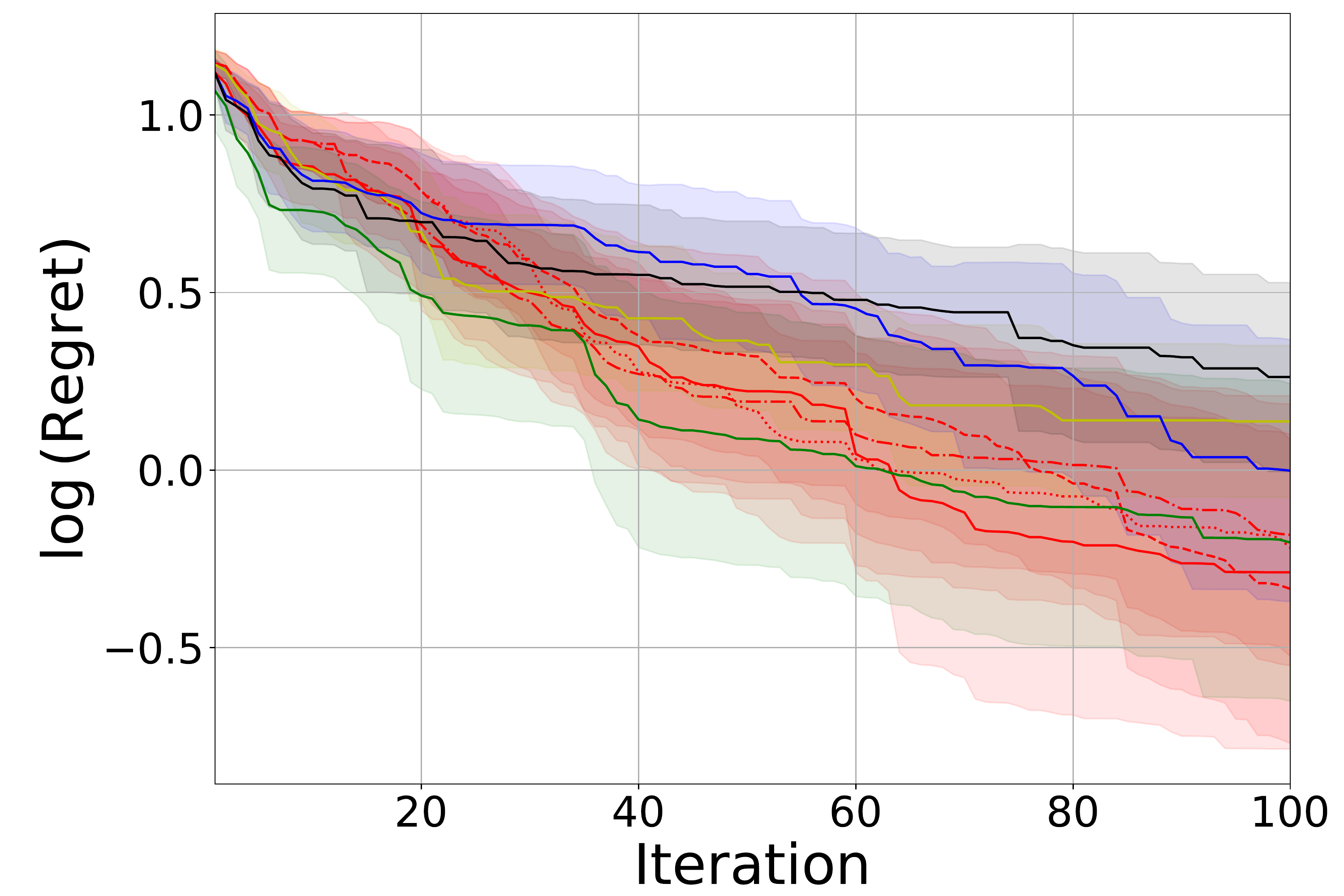}
	\includegraphics[width = 0.32\textwidth]{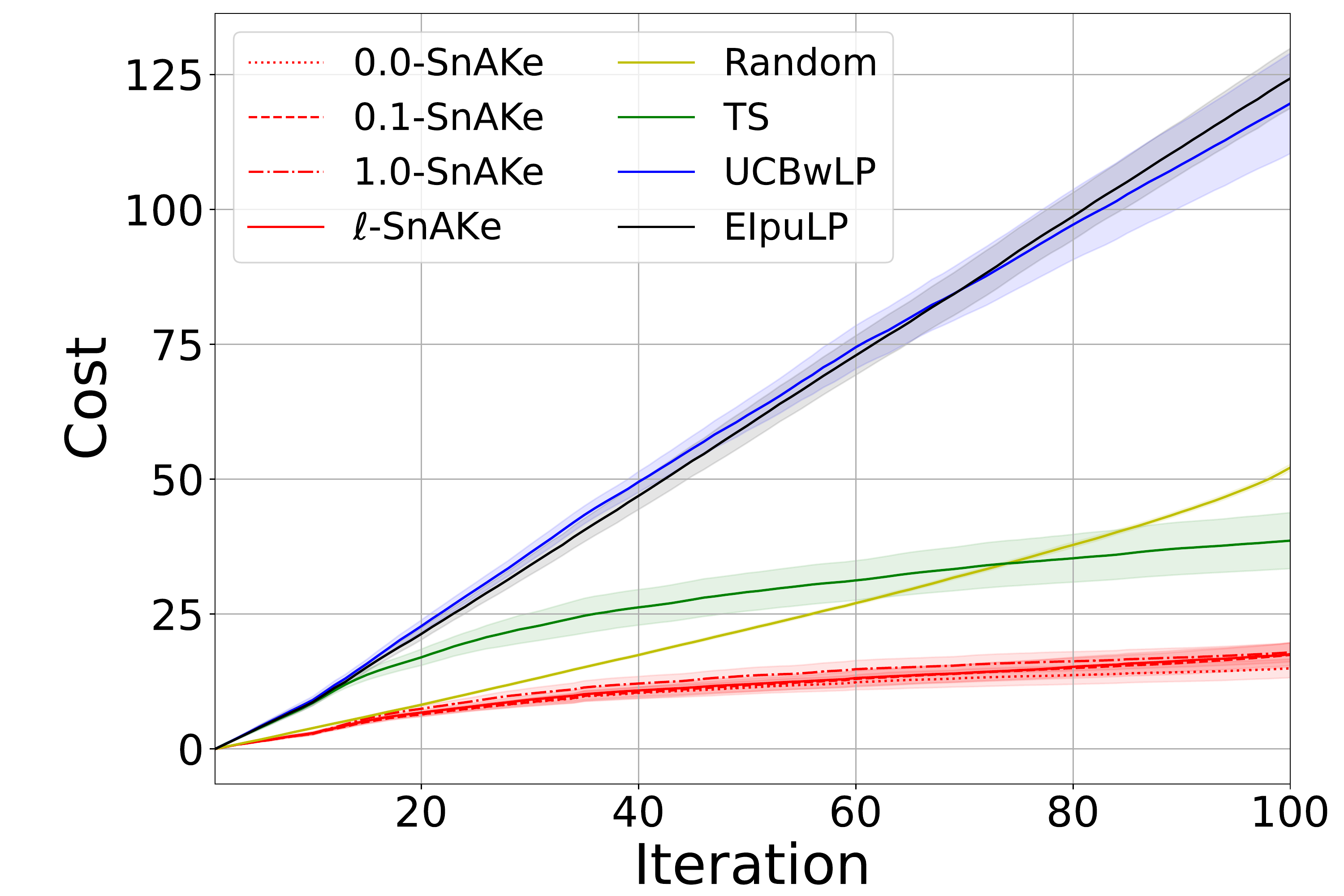}
	\caption{$T = 100$}
	\end{subfigure}
	\begin{subfigure}[t]{\textwidth}
	\includegraphics[width = 0.32\textwidth]{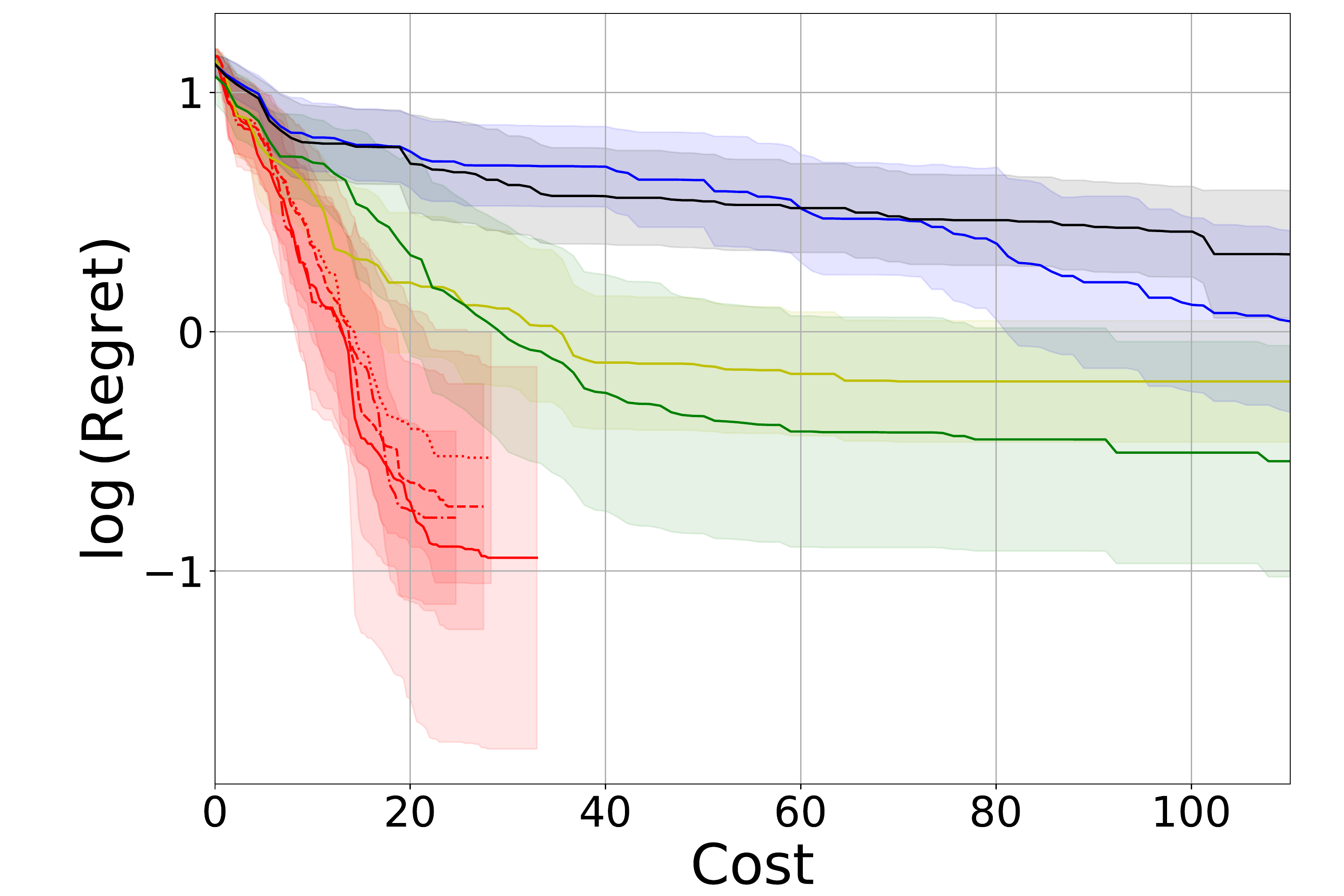}
	\includegraphics[width = 0.32\textwidth]{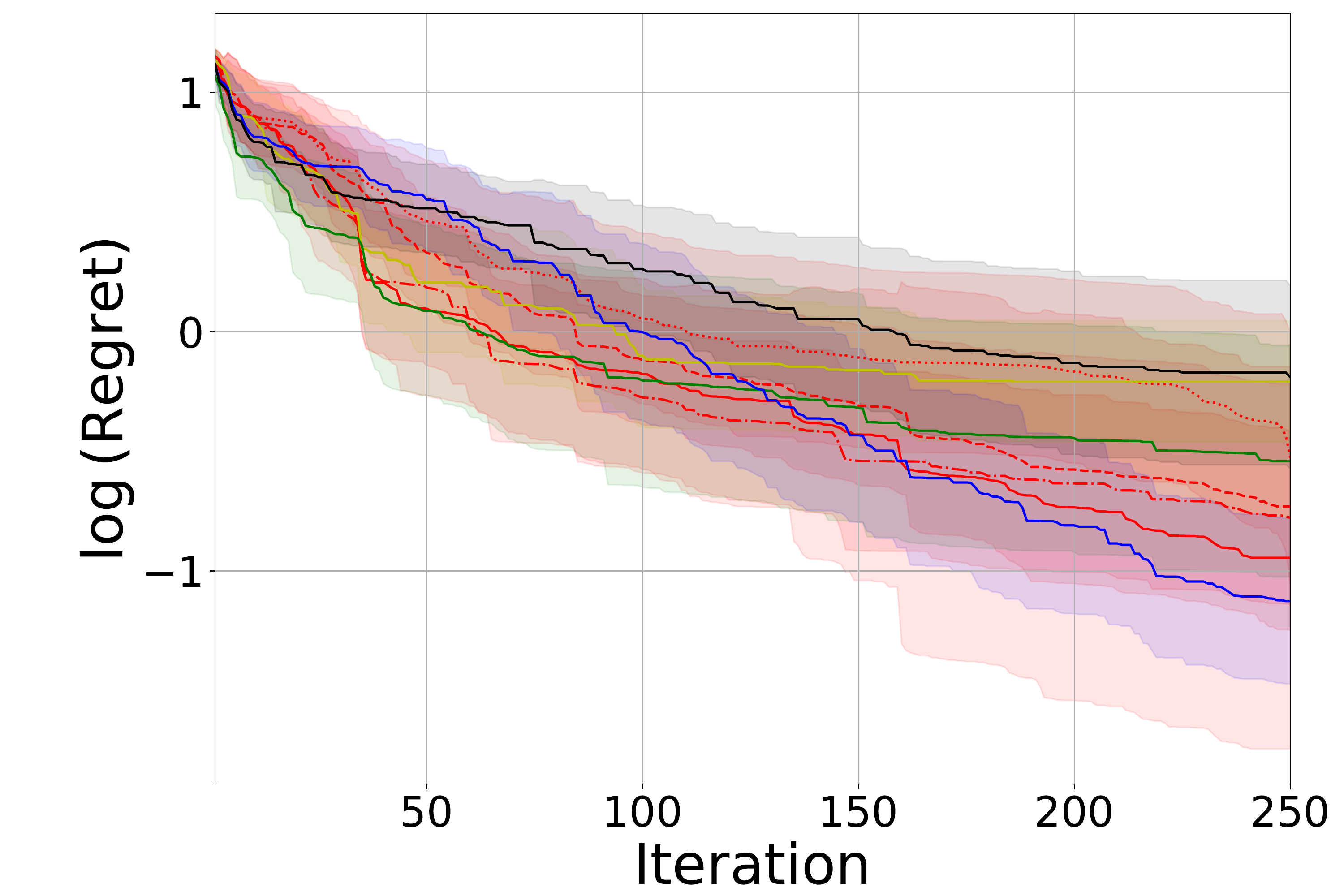}
	\includegraphics[width = 0.32\textwidth]{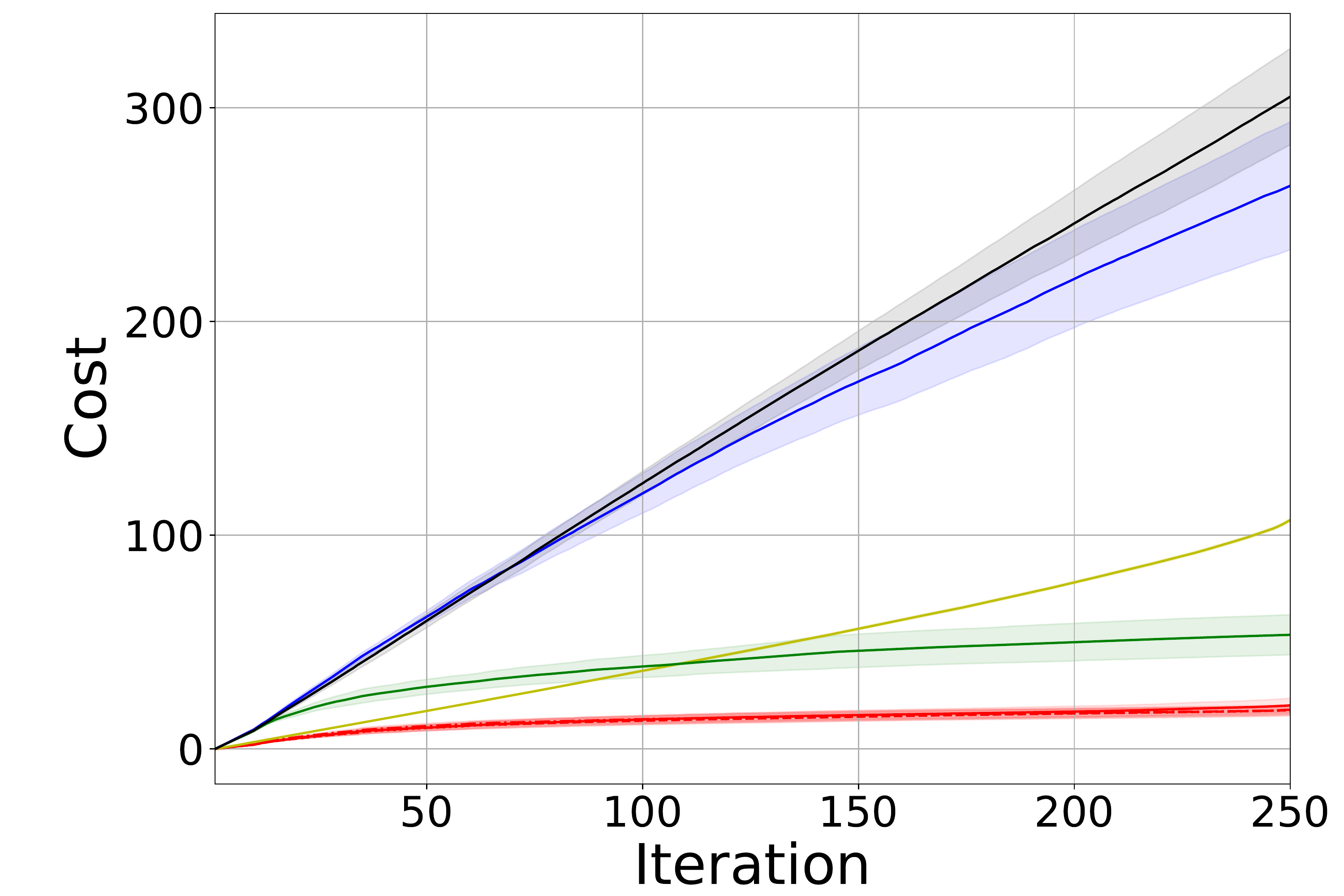}
	\caption{$T = 250$}
	\end{subfigure}
	\caption{Hartmann6D (Asynchronous),  $t_{delay}=10$. Each row represents a different budget. The left column shows the evolution of regret against the cost used. The middle column shows the evolution of regret with iterations, and the right columns show the evolution of the 2-norm cost. Similar results to other Hartmann benchmarks, see Figure \ref{fig: hartmann_3d_async_10}.}
\end{figure}

\subsection{Graphs for results of SnAr Benchmark (section \ref{subsec: snar})}

\begin{figure}[ht]
	\begin{subfigure}[t]{\textwidth}
	\centering
	\includegraphics[width=0.32\textwidth]{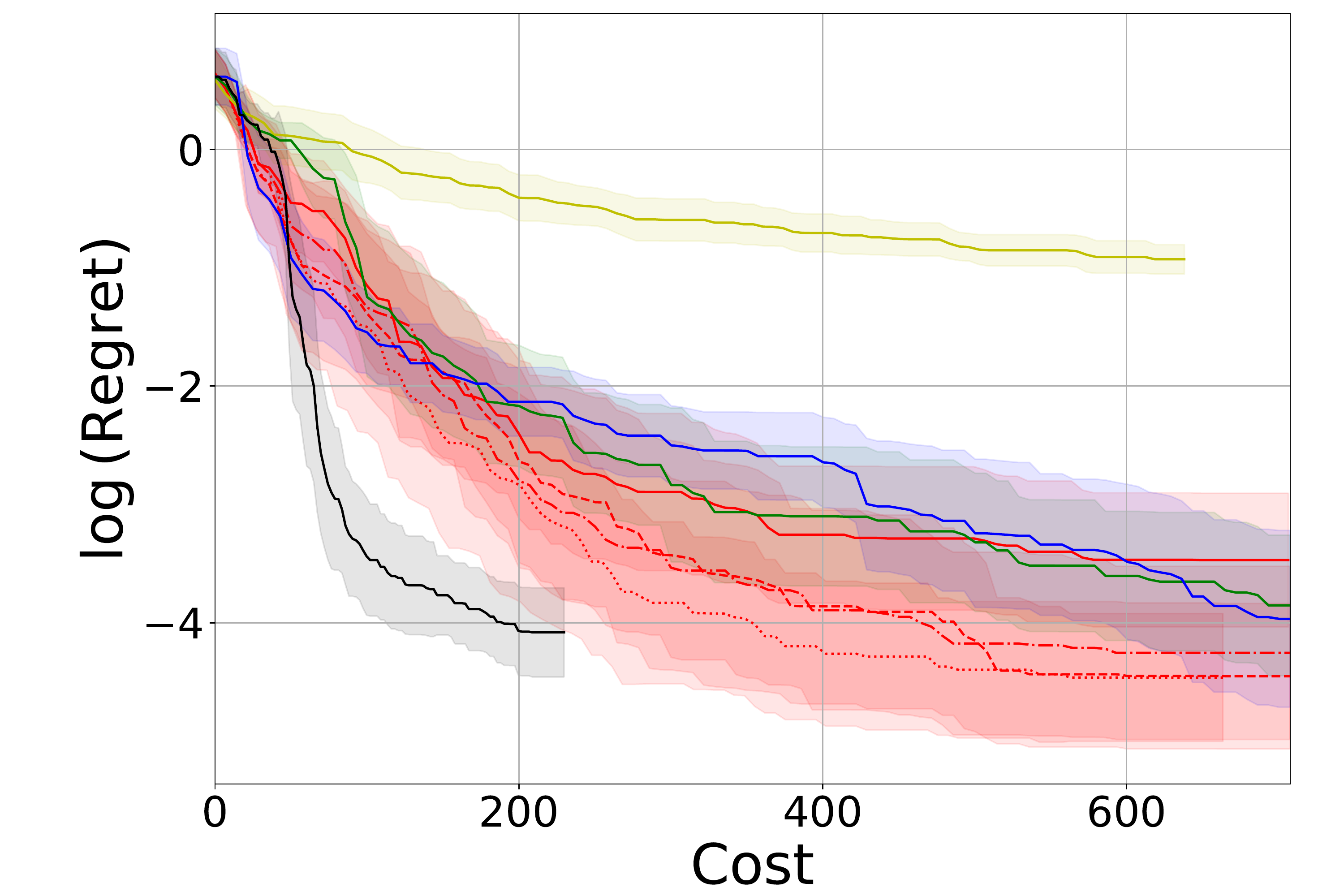}
	\includegraphics[width=0.32\textwidth]{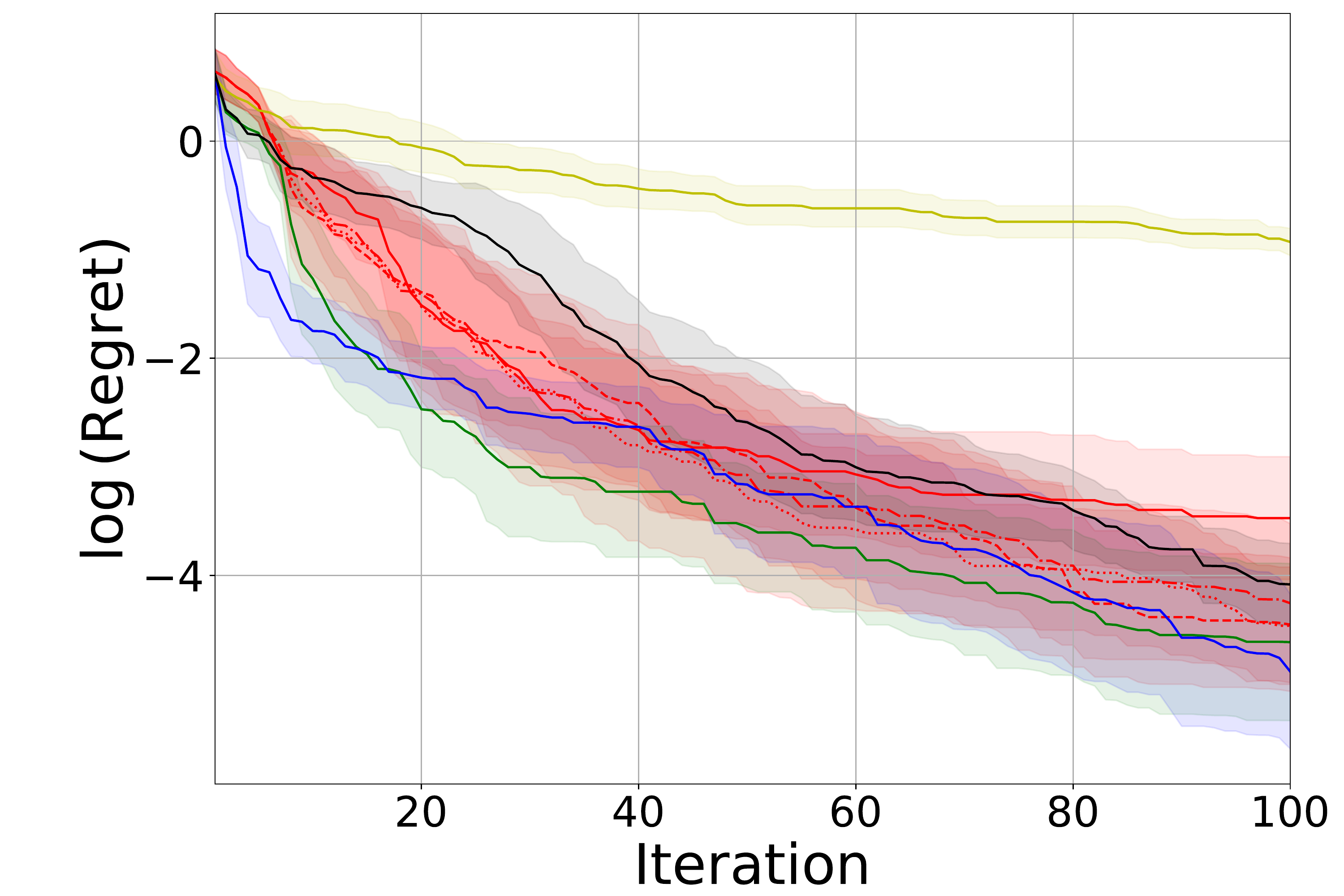}
	\includegraphics[width=0.32\textwidth]{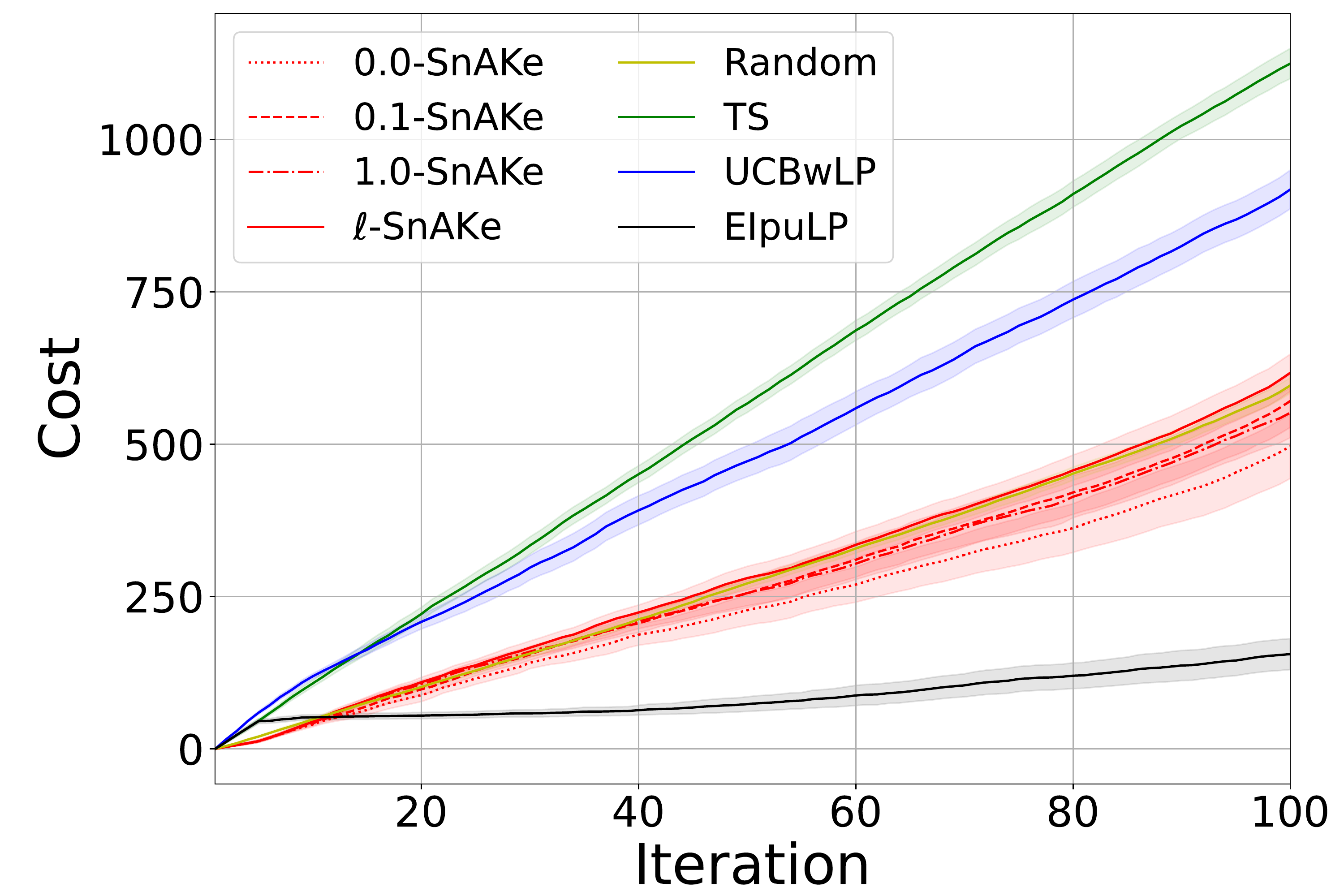}
	\caption{$t_{delay} = 5$}
	\end{subfigure}
	\hfill
	\begin{subfigure}[t]{\textwidth}
	\centering
	\includegraphics[width=0.32\textwidth]{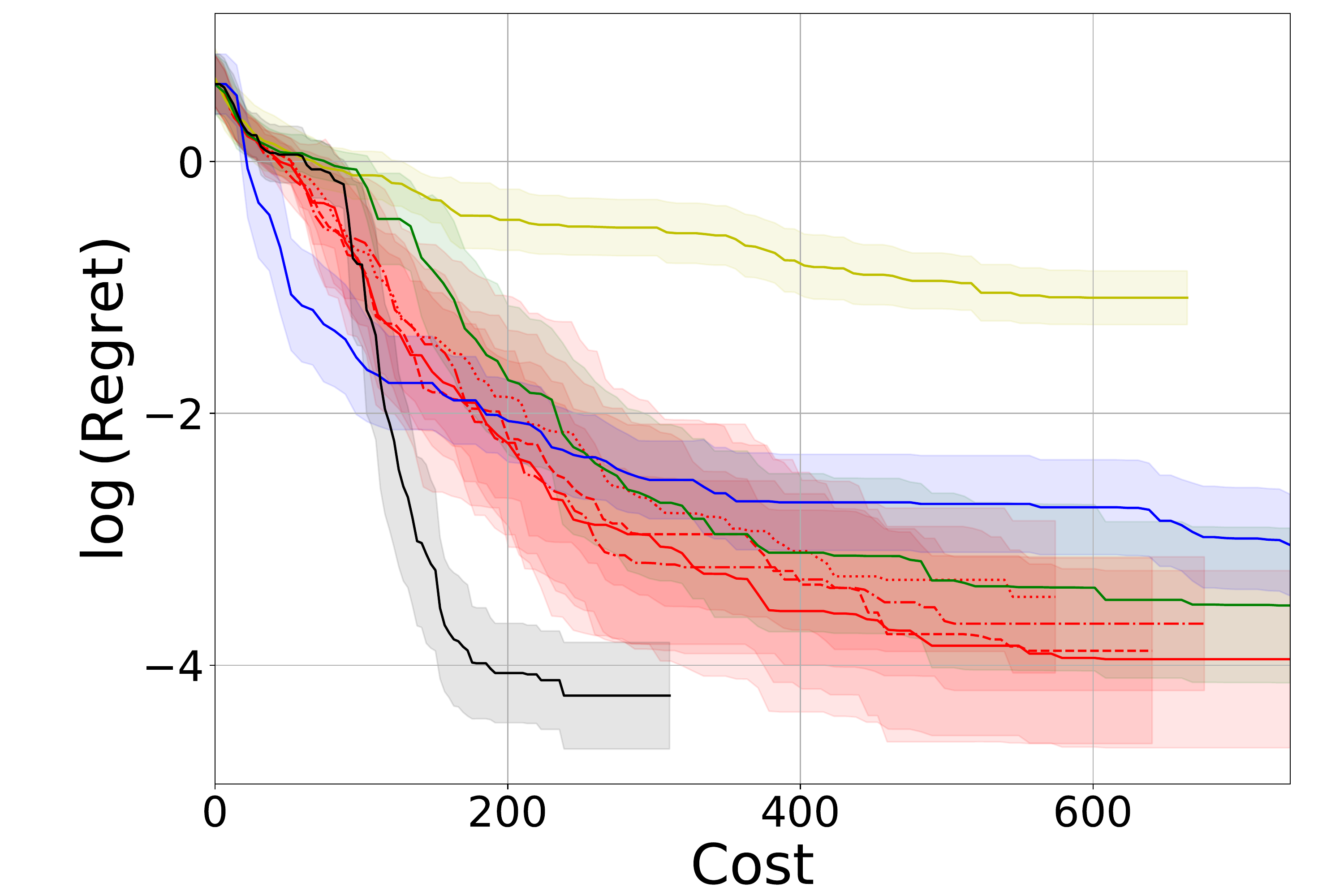}
	\includegraphics[width=0.32\textwidth]{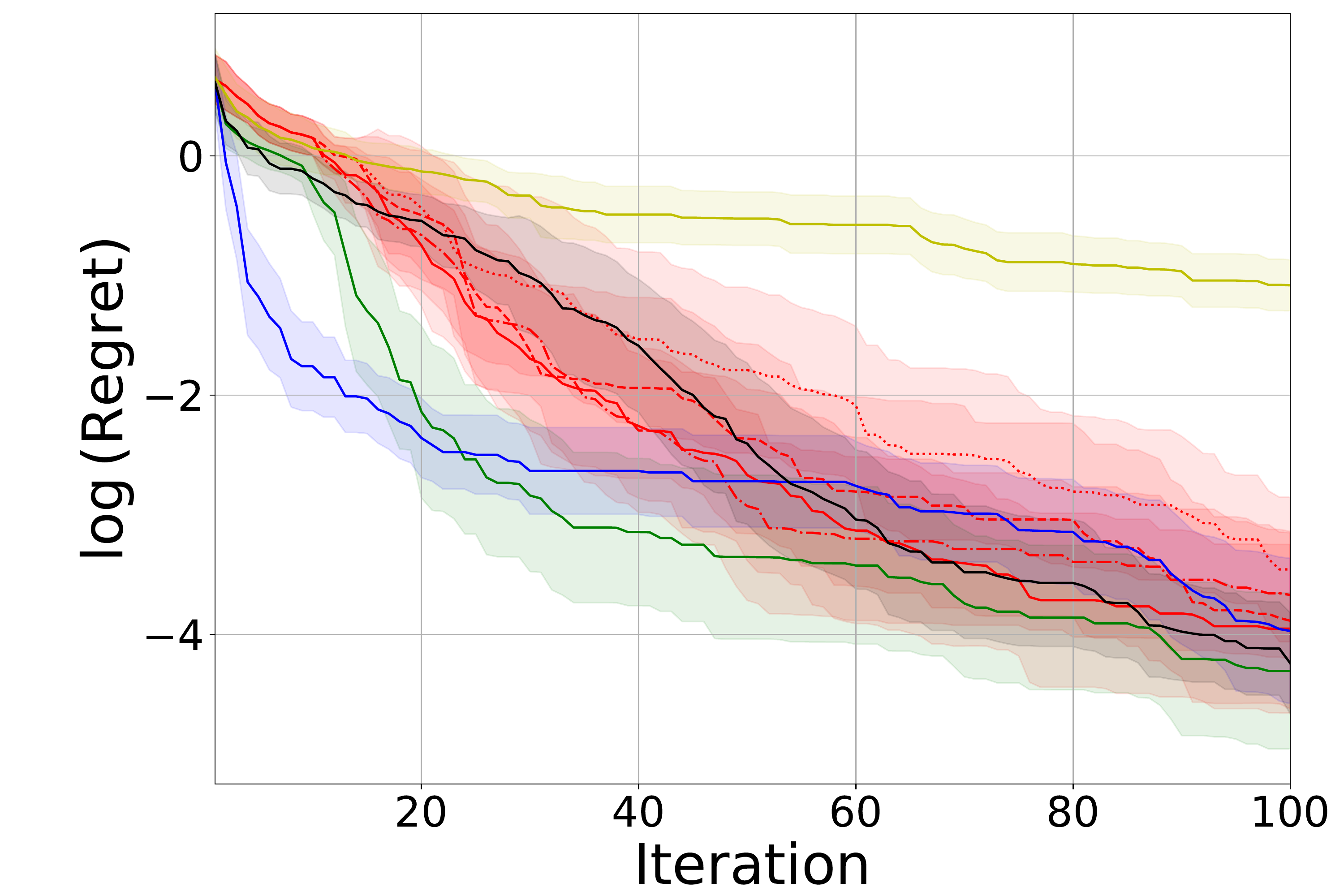}
	\includegraphics[width=0.32\textwidth]{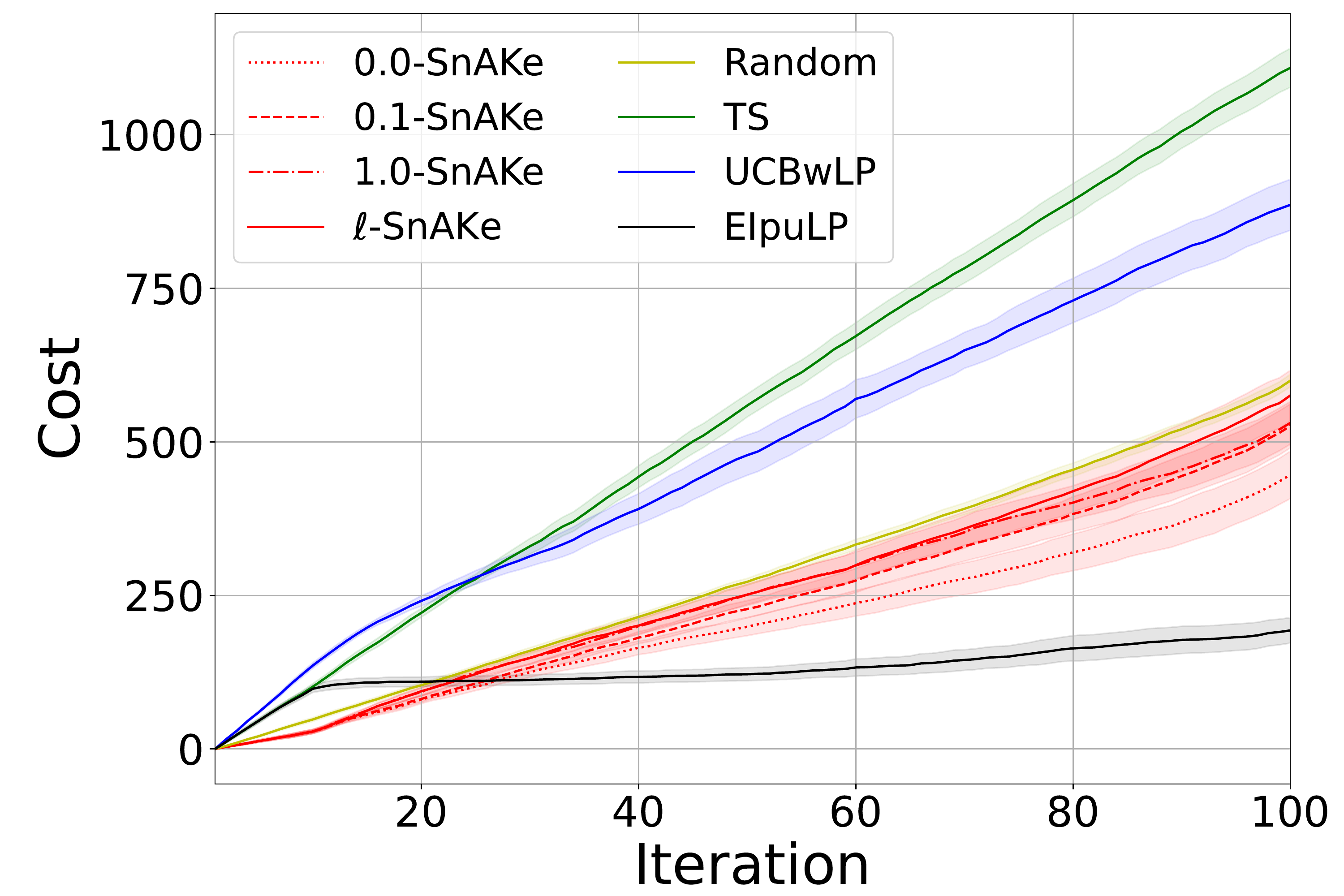}
	\caption{$t_{delay} = 10$}
	\end{subfigure}
	\hfill
	\begin{subfigure}[t]{\textwidth}
	\centering
	\includegraphics[width=0.32\textwidth]{Figures/Section4/25-Regret-Cost}
	\includegraphics[width=0.32\textwidth]{Figures/Section4/25-Regret}
	\includegraphics[width=0.32\textwidth]{Figures/Section4/25-Cost}
	\caption{$t_{delay} = 25$}
	\end{subfigure}
	\hfill
	\begin{subfigure}[t]{\textwidth}
	\centering
	\includegraphics[width=0.32\textwidth]{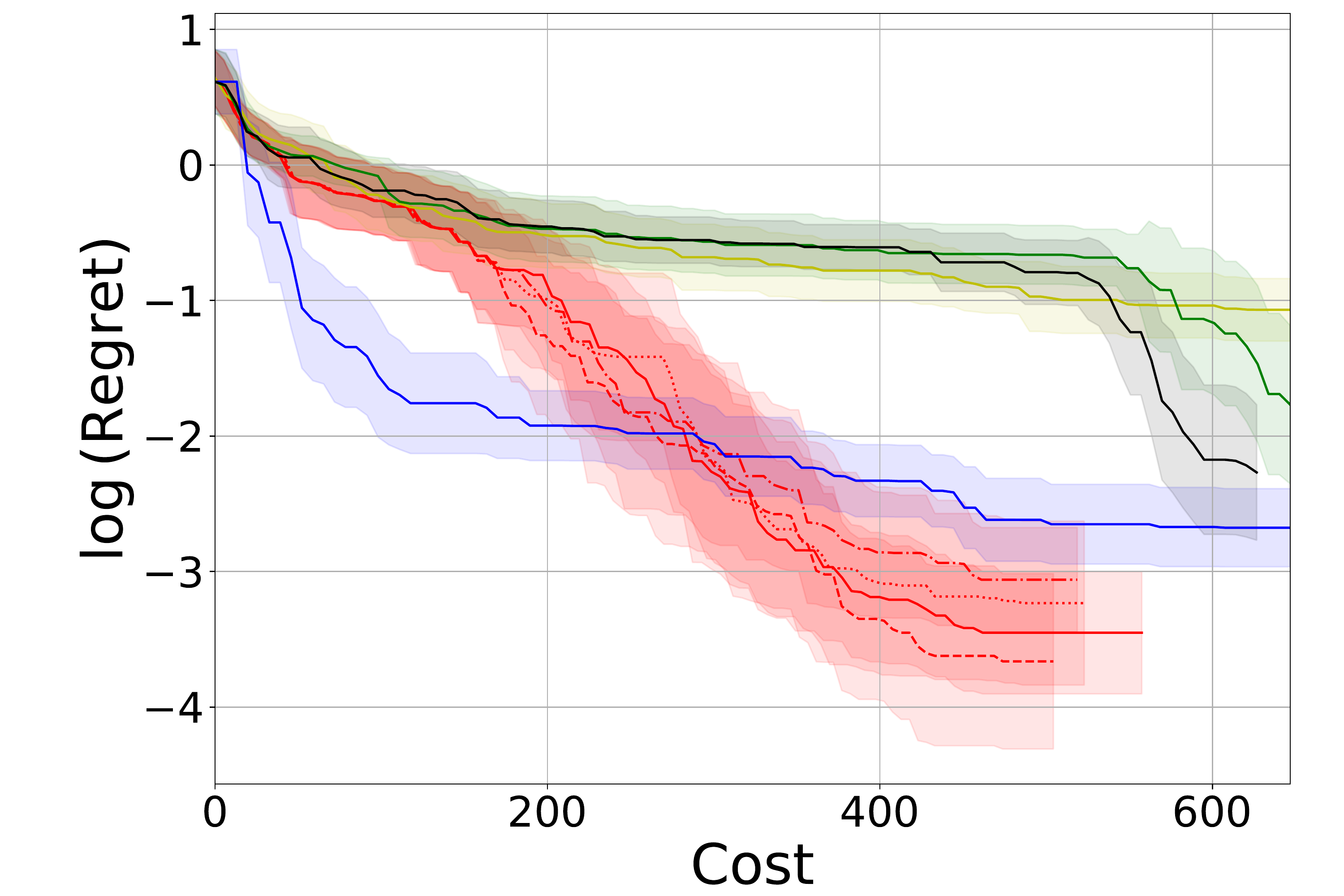}
	\includegraphics[width=0.32\textwidth]{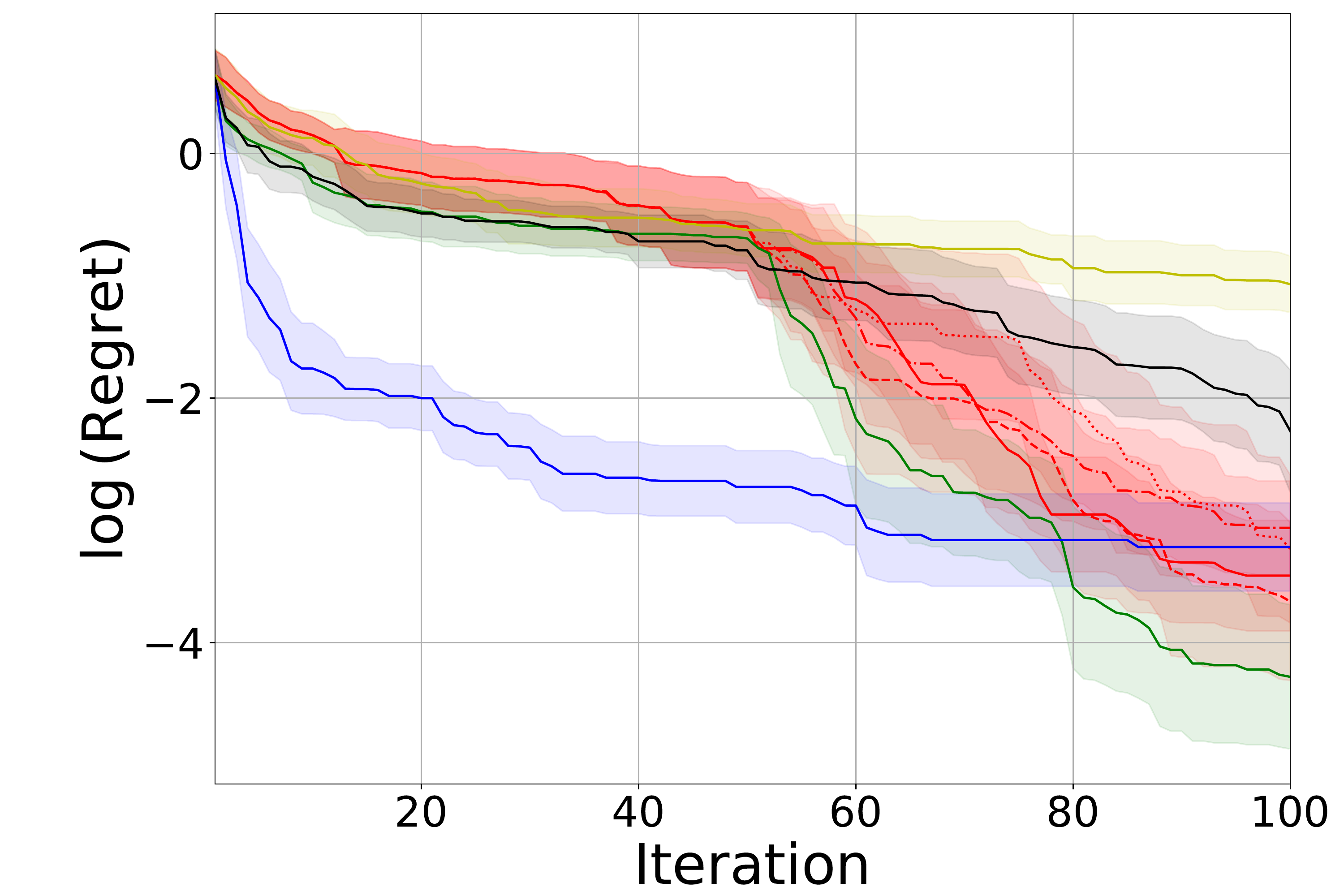}
	\includegraphics[width=0.32\textwidth]{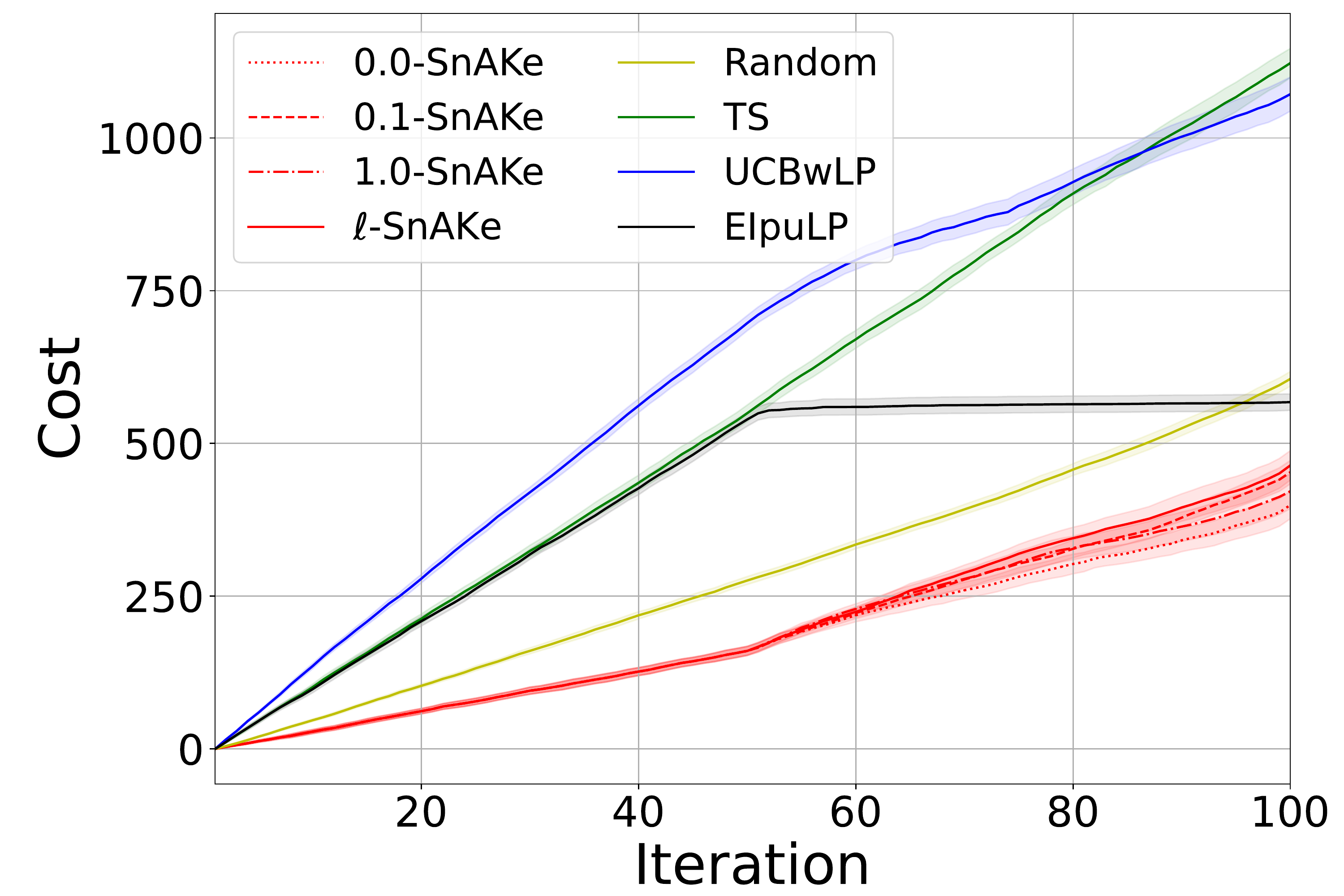}
	\caption{$t_{delay} = 50$}
	\end{subfigure}
	\caption{SnAr benchmark (Asynchronous) with $T = 100$. Each row represents a different $t_{delay}$. The left column shows the evolution of regret against the cost used. The middle column shows the evolution of regret with iterations, and the right columns show the evolution of the cost as defined in Section \ref{subsec: snar}. SnAKe achieves the better regret than classical BO algorithms at low cost for all budgets. EIpuLP performs well for small delays, but poorly for larger delays.}
	\label{fig: snar_async_graphs}
\end{figure}

\begin{figure}[ht]
	\begin{subfigure}[t]{\textwidth}
	\centering
	\includegraphics[width=0.32\textwidth]{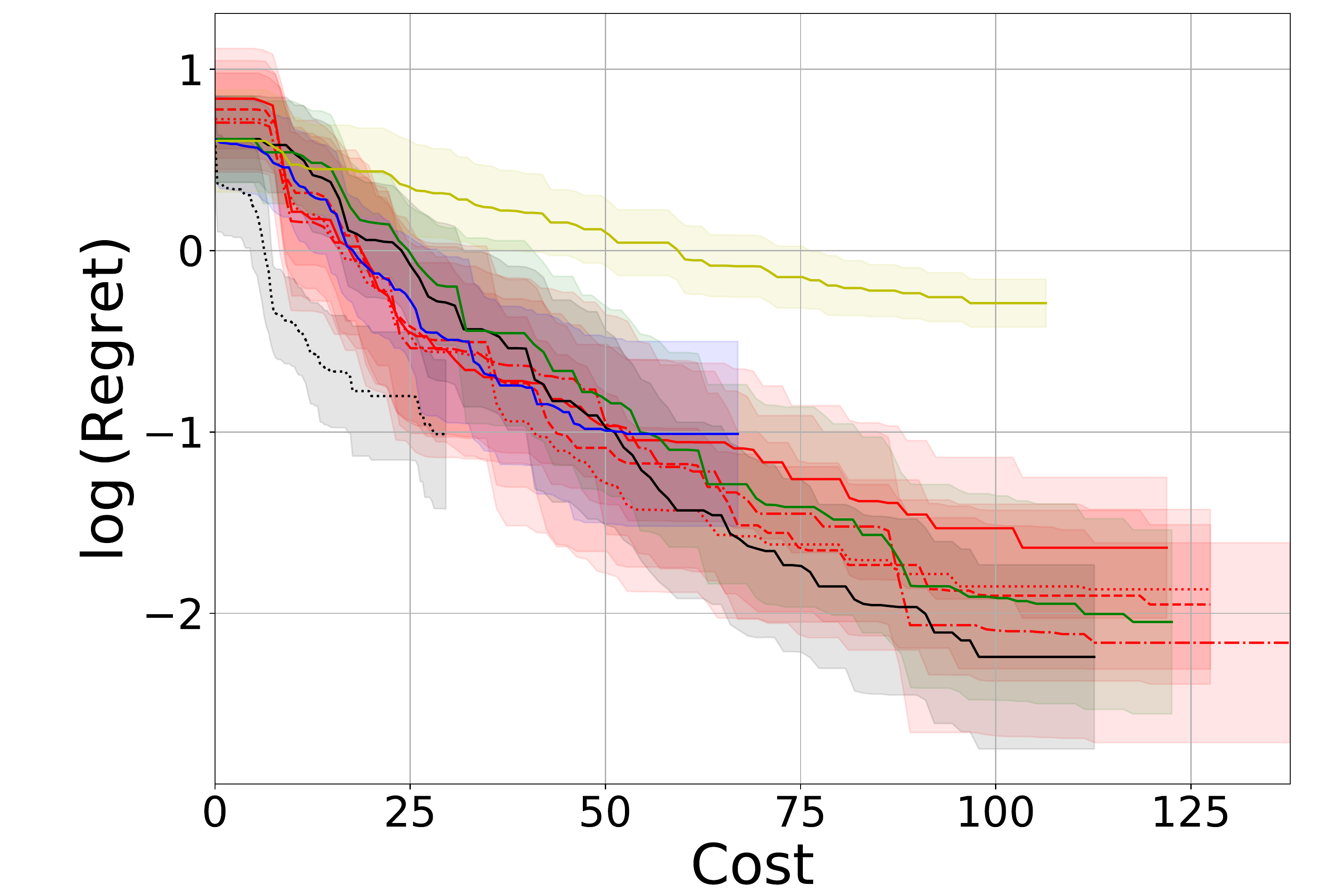}
	\includegraphics[width=0.32\textwidth]{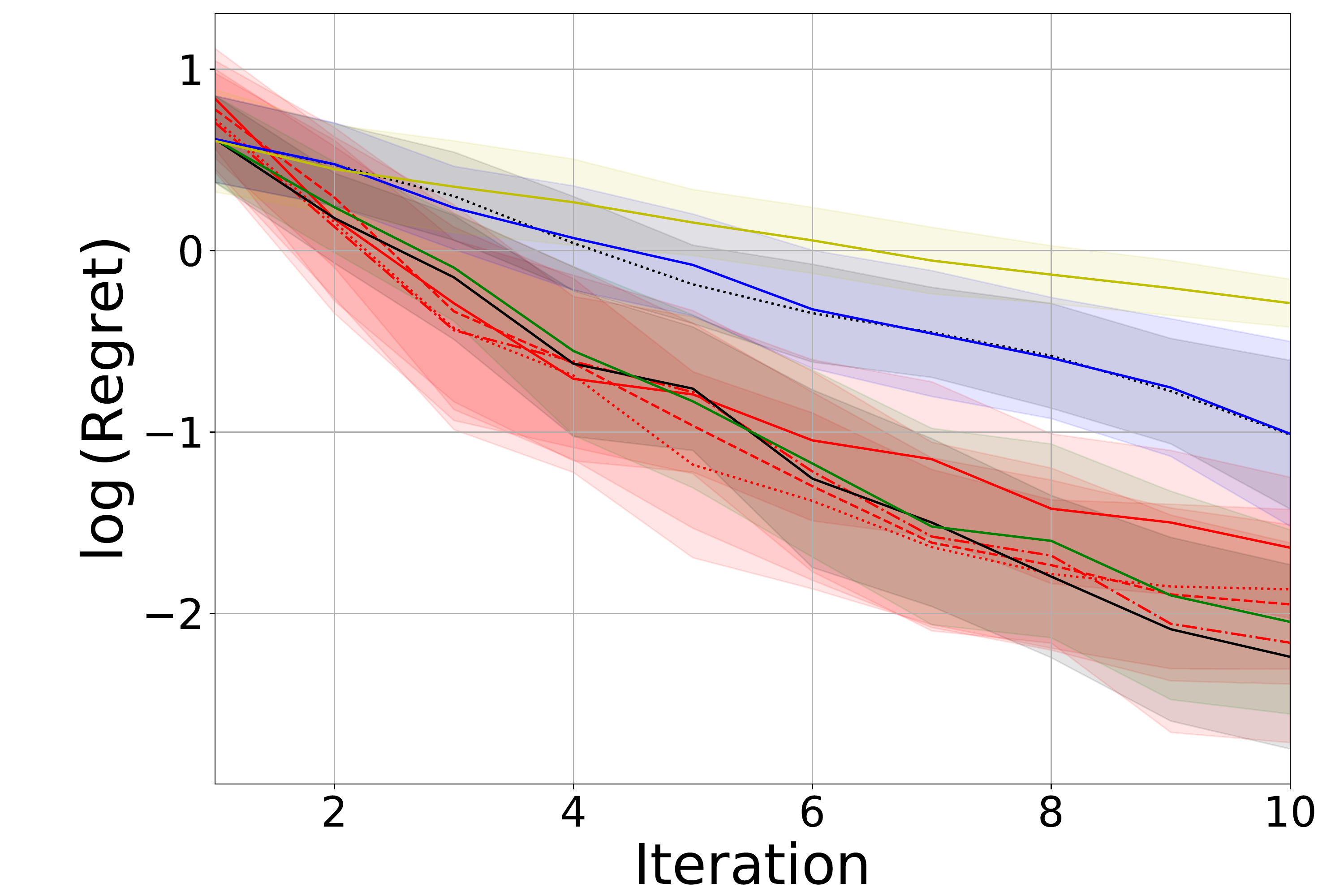}
	\includegraphics[width=0.32\textwidth]{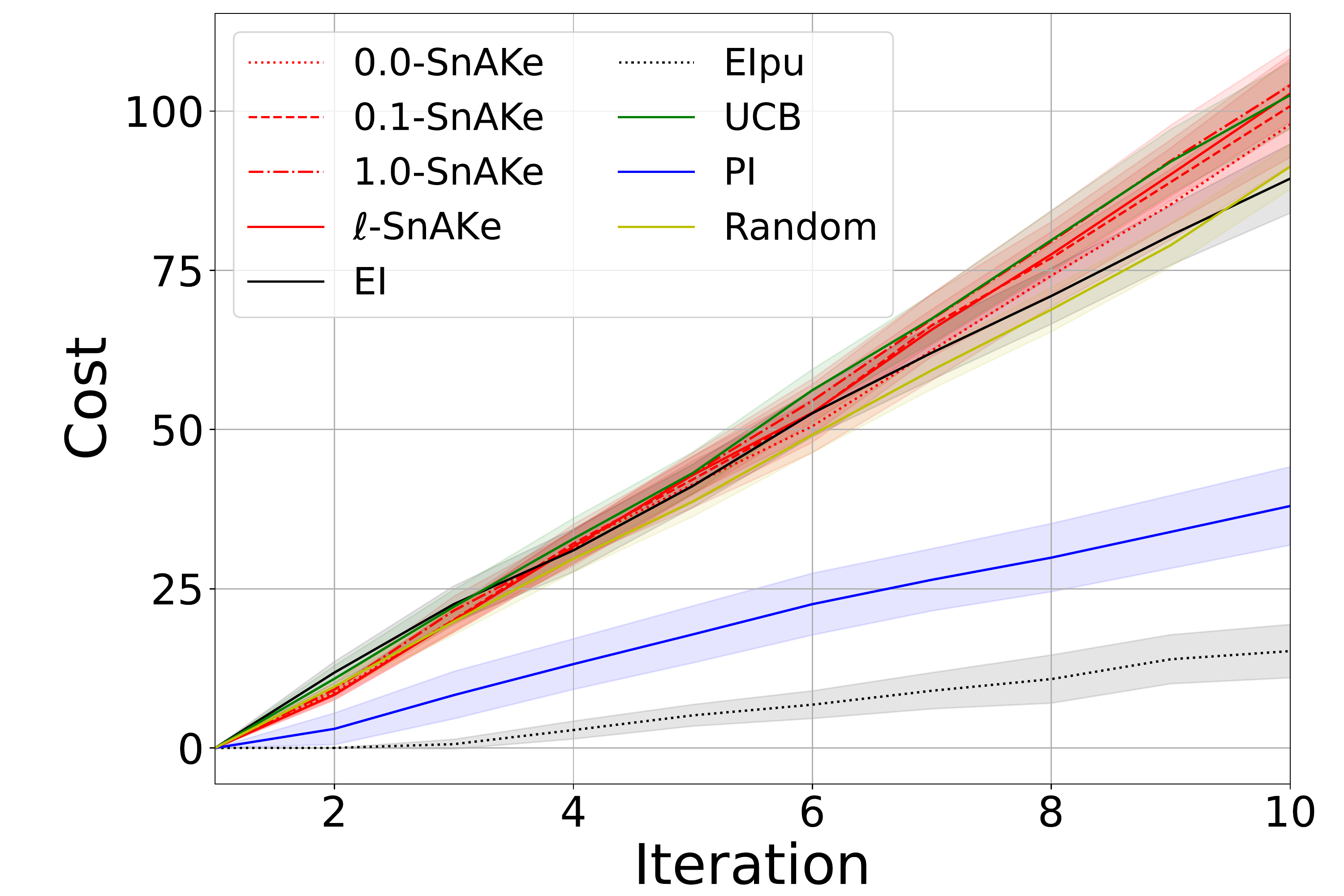}
	\caption{$T = 10$}
	\end{subfigure}
	\hfill
	\begin{subfigure}[t]{\textwidth}
	\centering
	\includegraphics[width=0.32\textwidth]{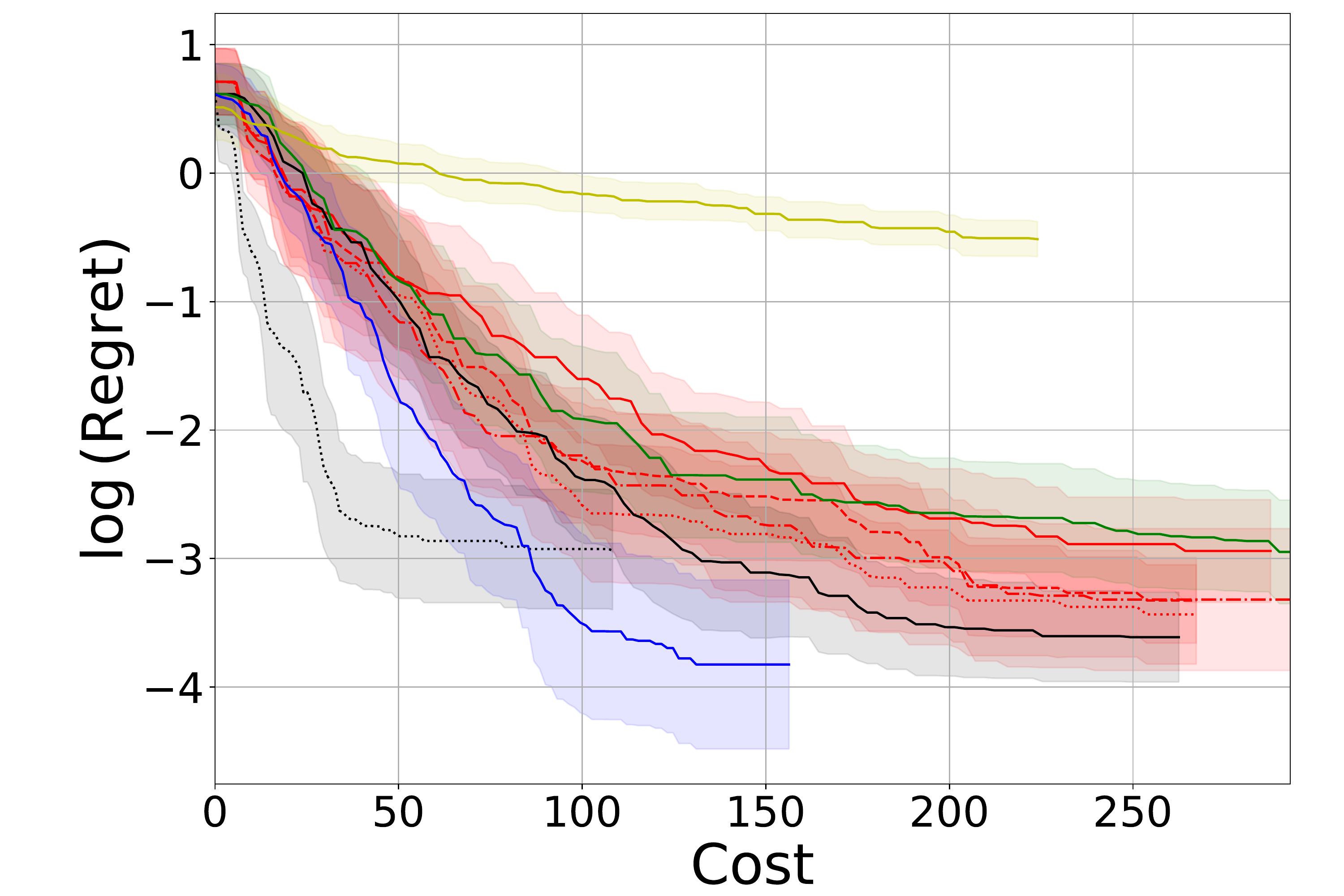}
	\includegraphics[width=0.32\textwidth]{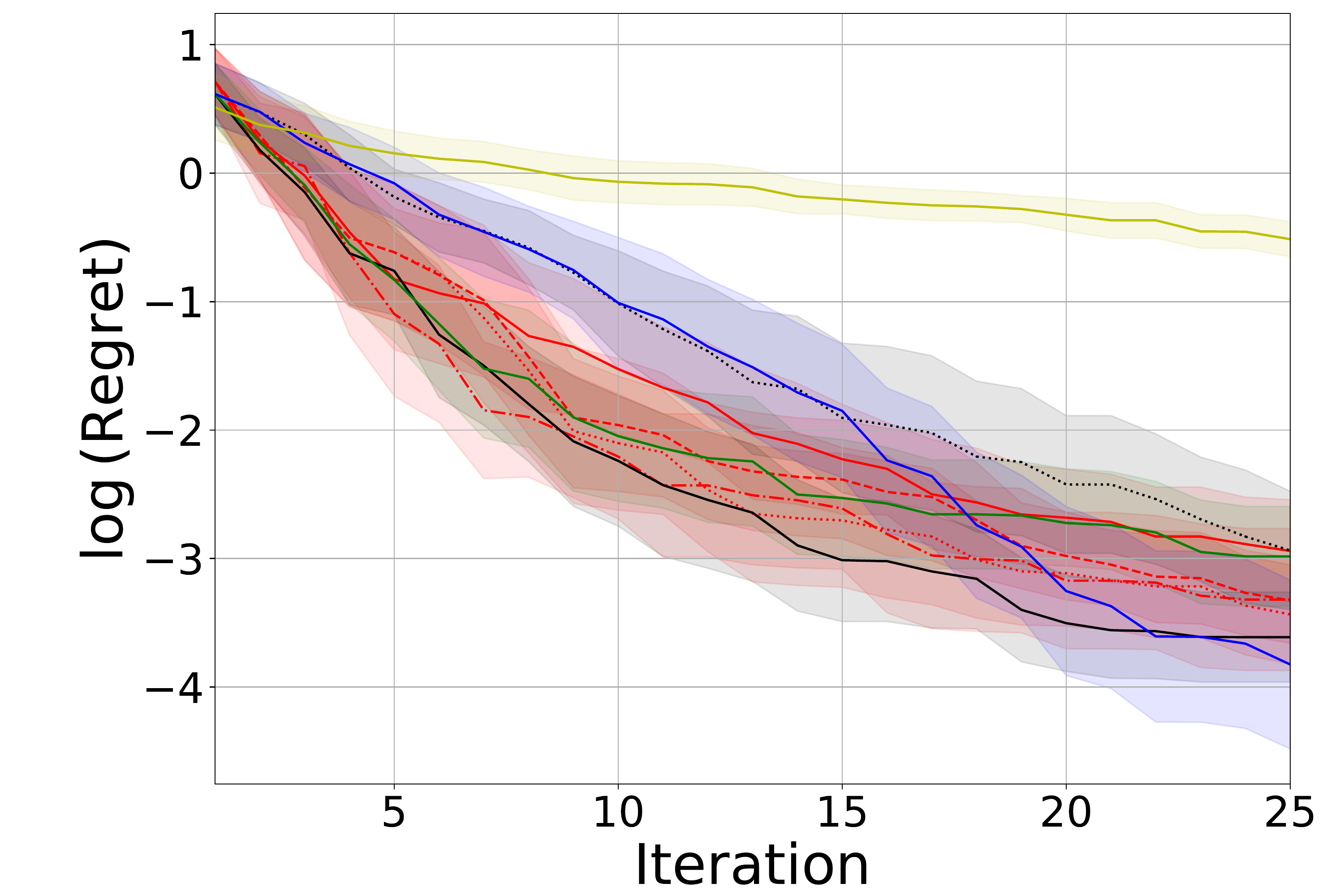}
	\includegraphics[width=0.32\textwidth]{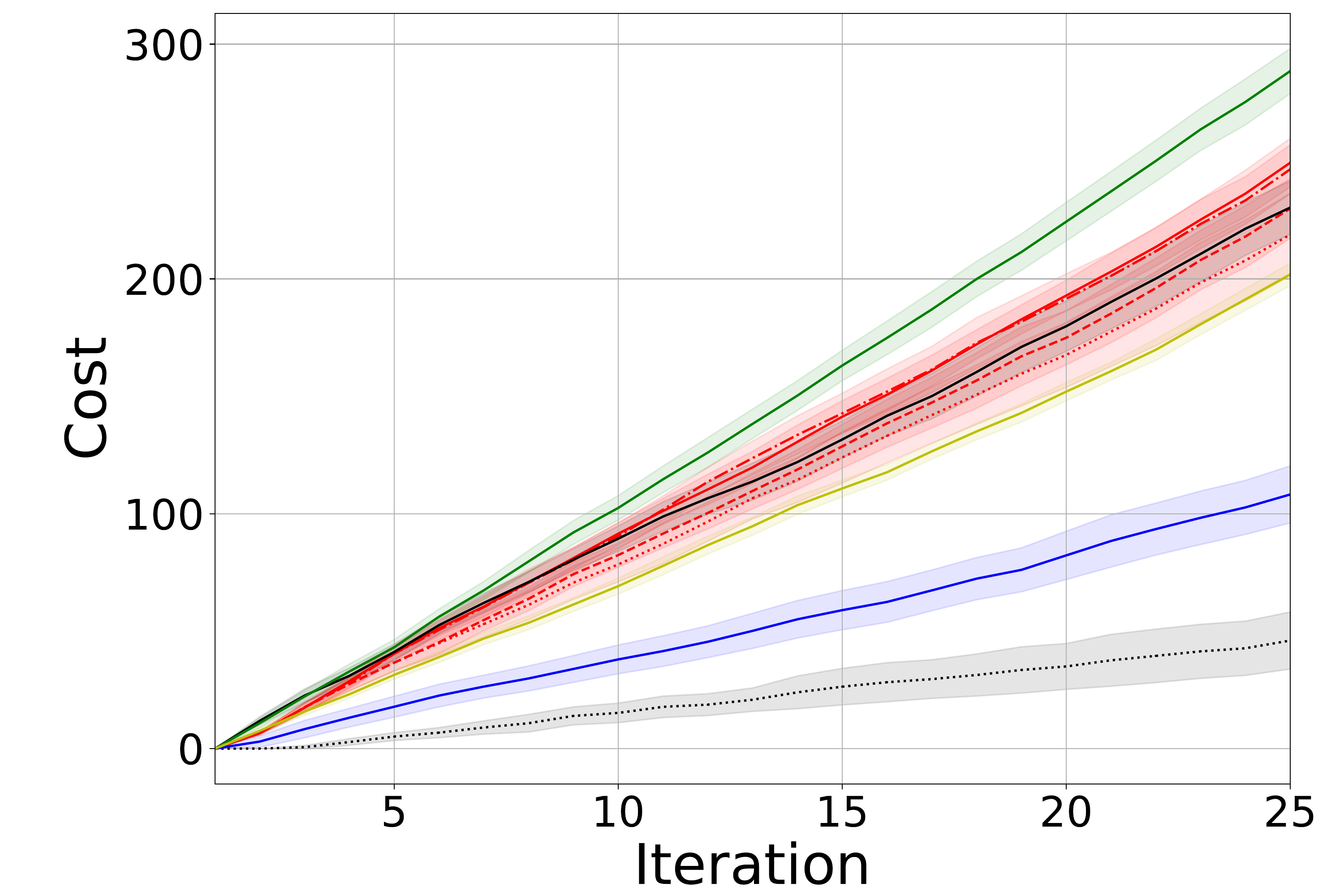}
	\caption{$T = 25$}
	\end{subfigure}
	\hfill
	\begin{subfigure}[t]{\textwidth}
	\centering
	\includegraphics[width=0.32\textwidth]{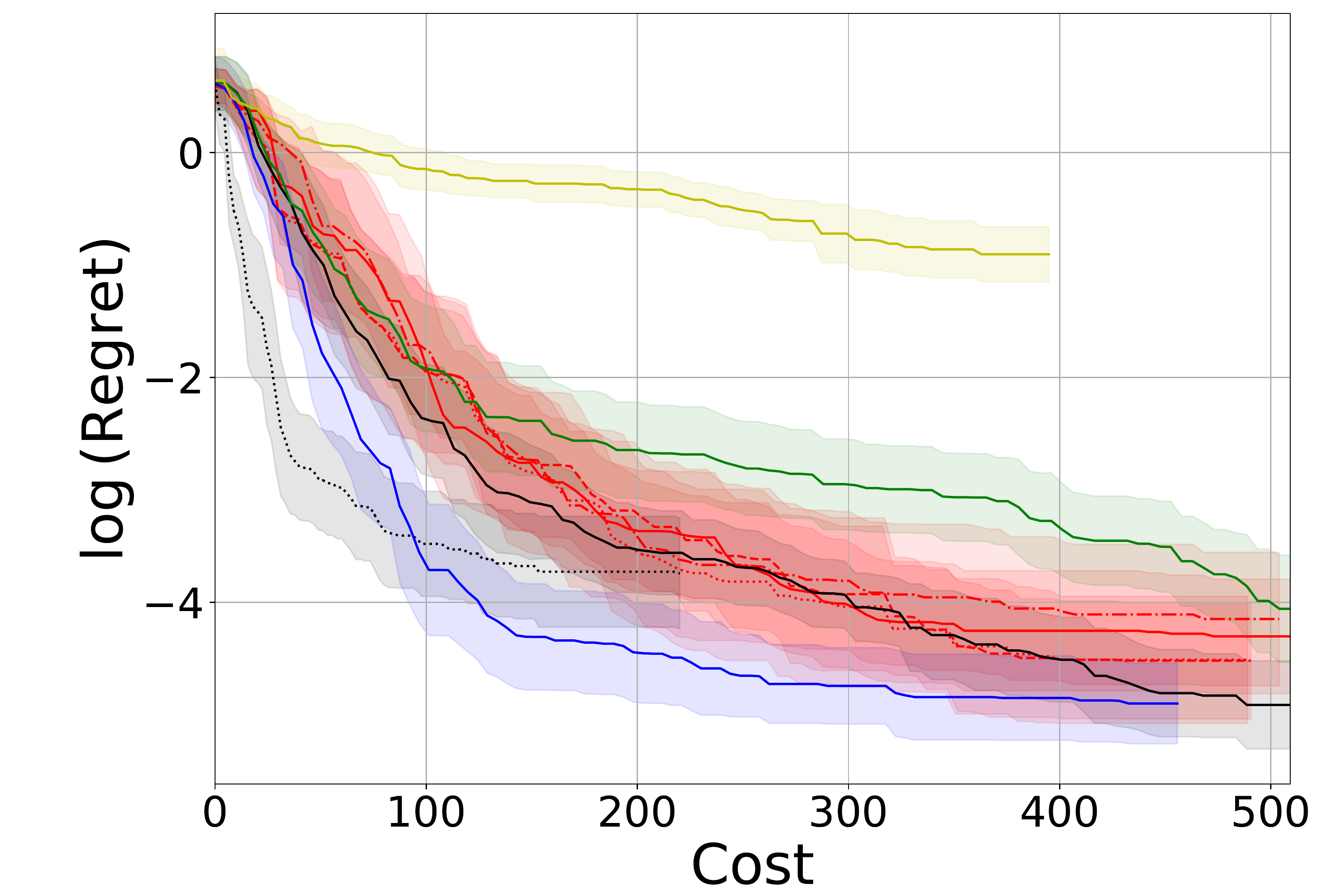}
	\includegraphics[width=0.32\textwidth]{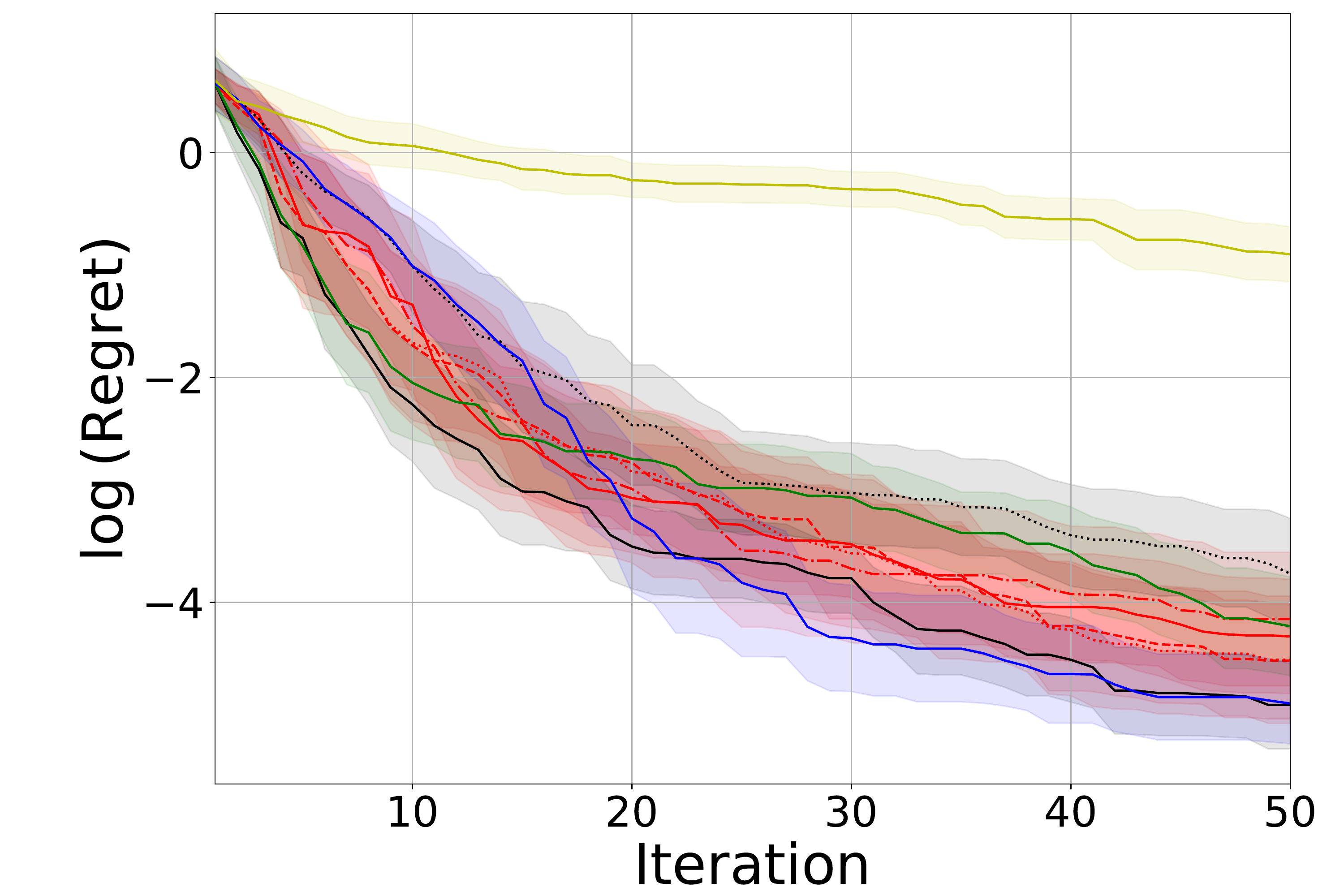}
	\includegraphics[width=0.32\textwidth]{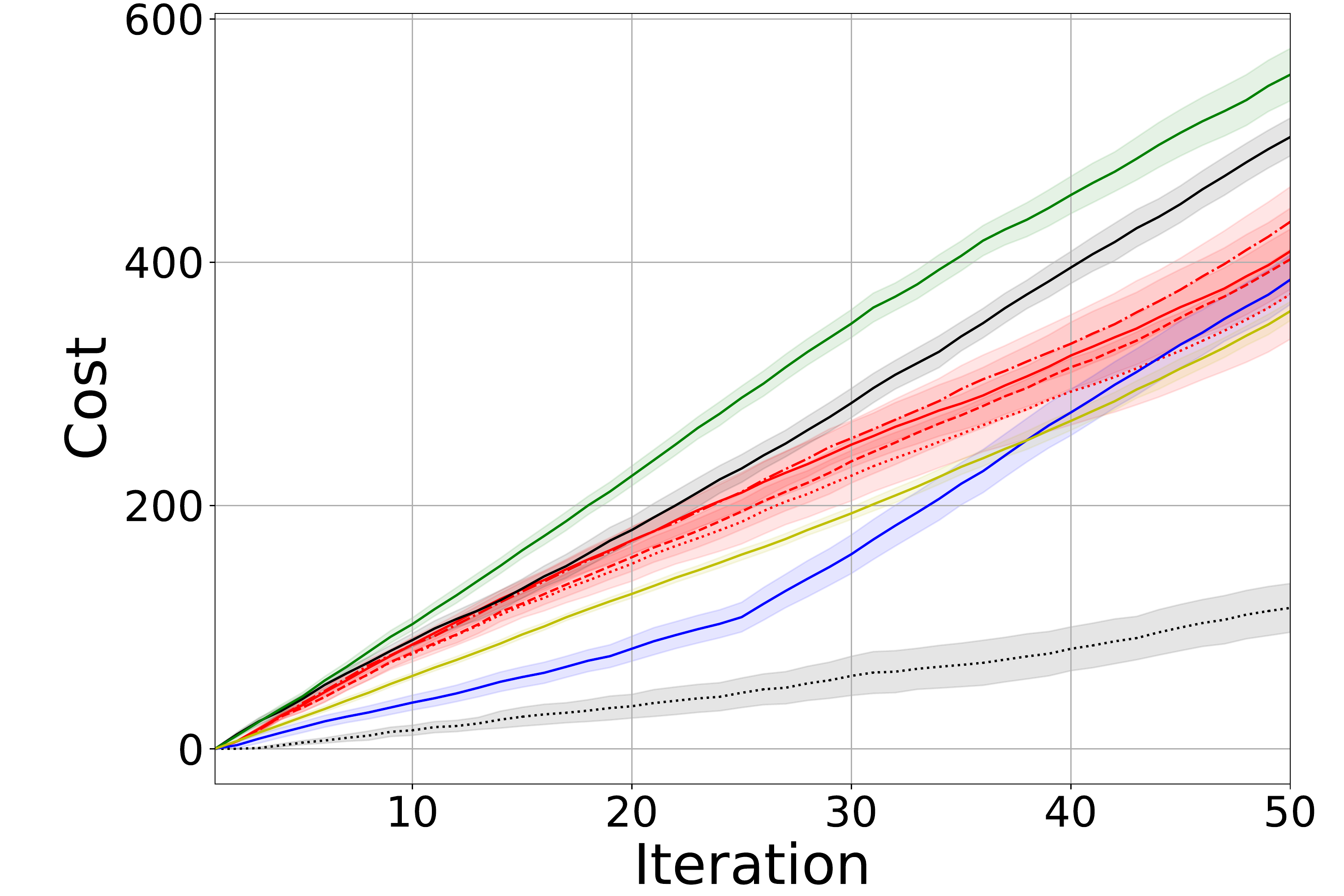}
	\caption{$T = 50$}
	\end{subfigure}
	\hfill
	\begin{subfigure}[t]{\textwidth}
	\centering
	\includegraphics[width=0.32\textwidth]{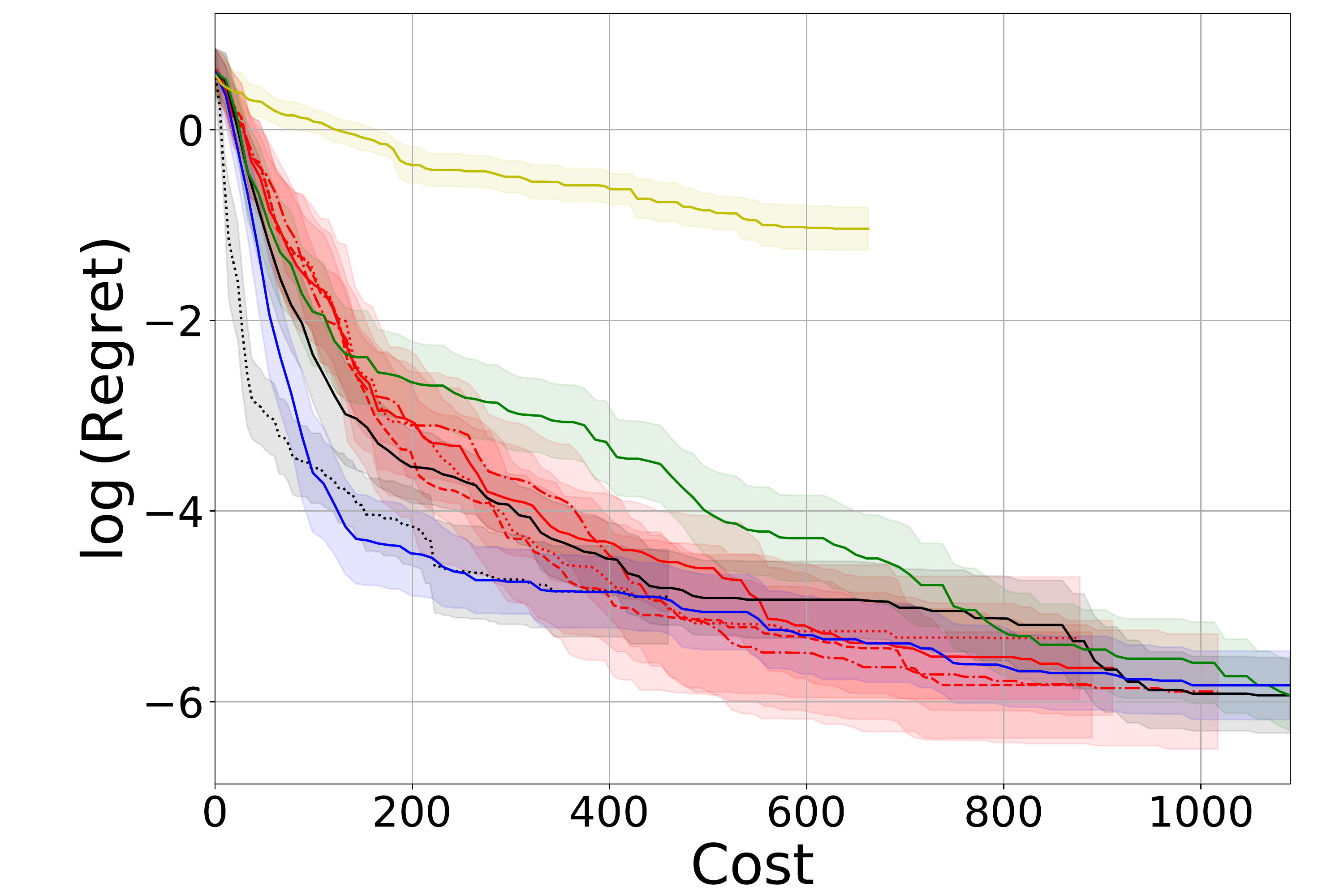}
	\includegraphics[width=0.32\textwidth]{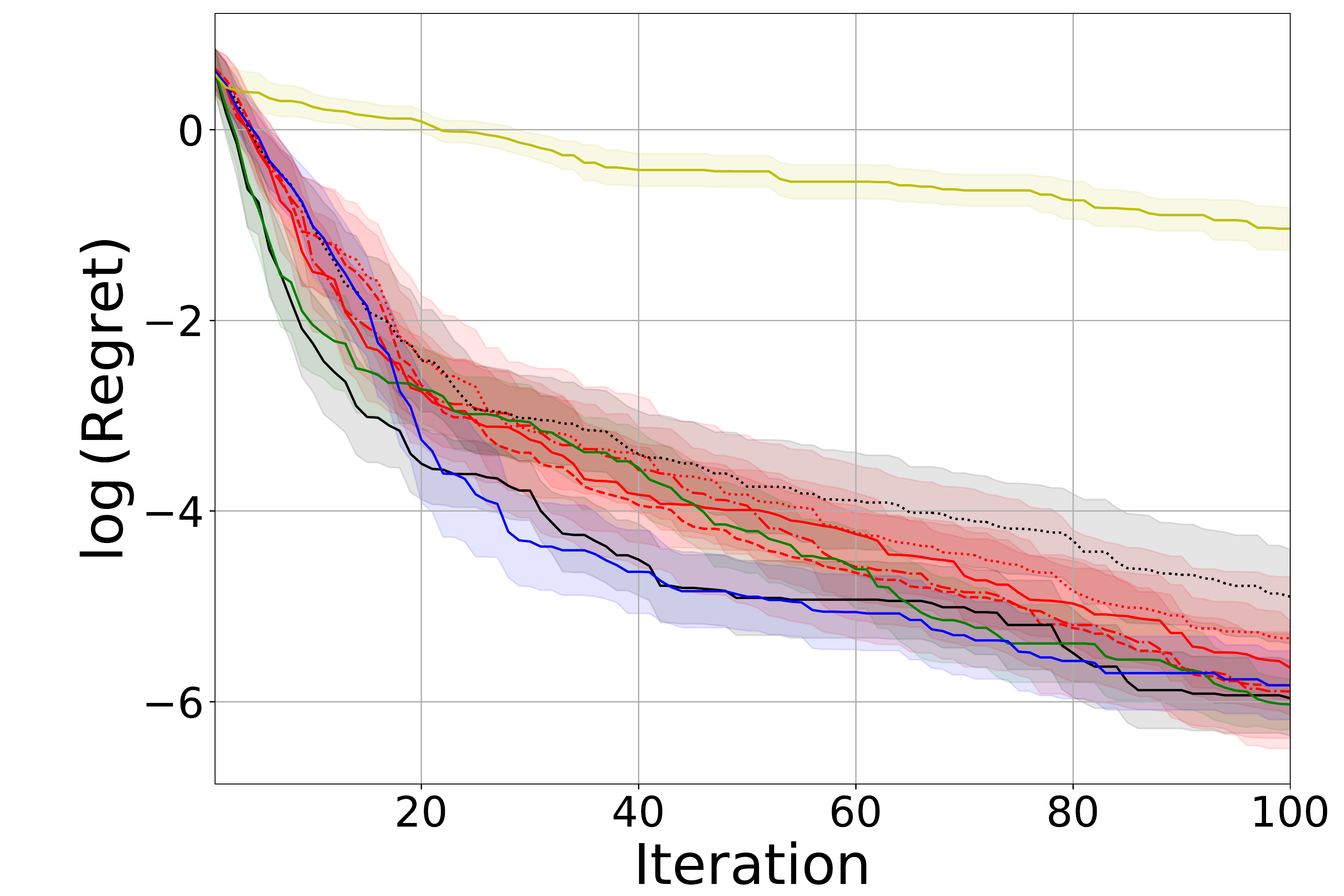}
	\includegraphics[width=0.32\textwidth]{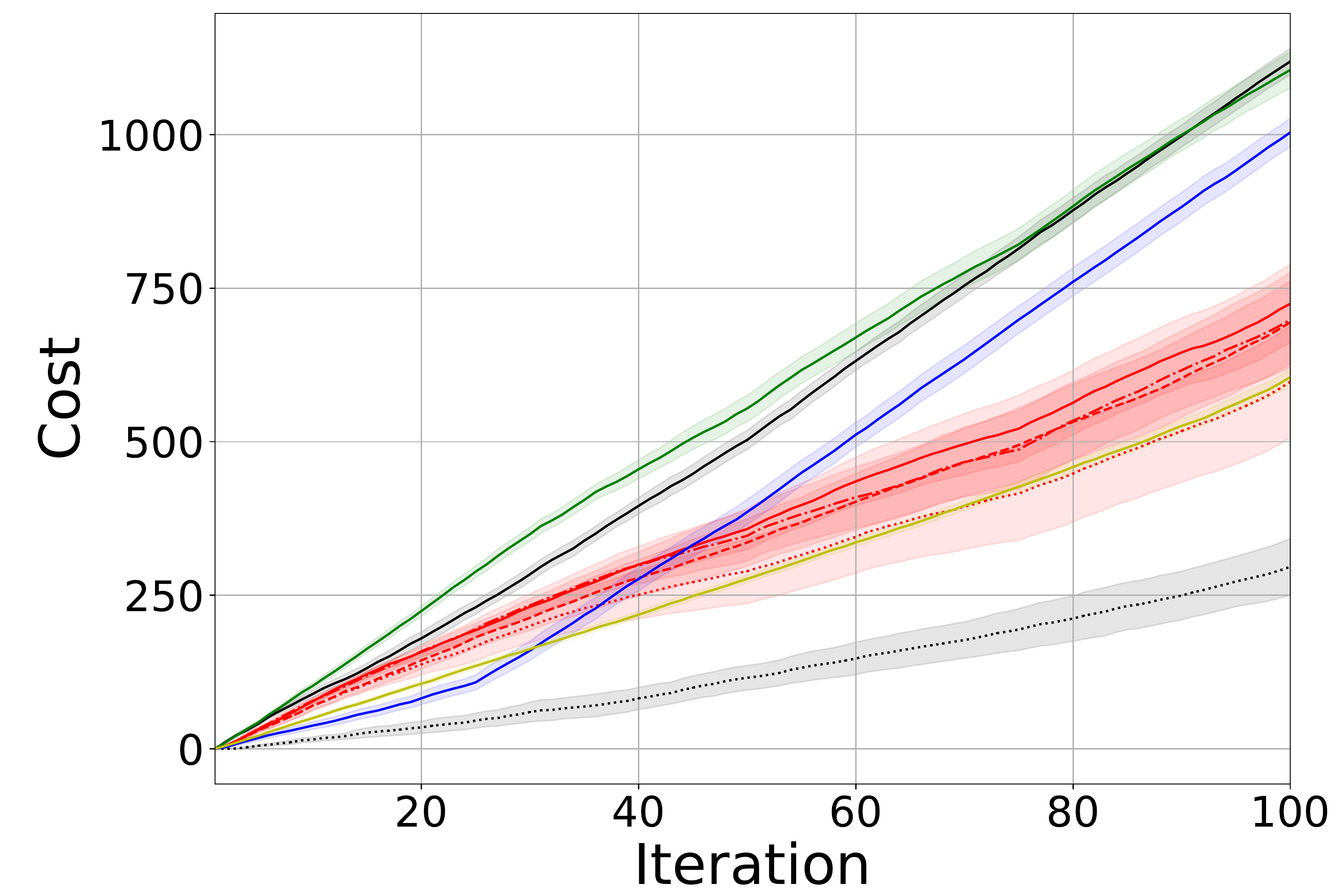}
	\caption{$T = 100$}
	\end{subfigure}
	\caption{SnAr benchmark (synchronous, $t_{delay} = 1$). Each row shows a different budget. The left column shows the evolution of regret against the cost used. The middle column shows the evolution of regret with iterations, and the right columns show the evolution of the cost as defined in Section \ref{subsec: snar}. SnAKe is the only method achieving low regret and low cost especially for larger budgets. EIpu generally achieves low cost but poor regret.}
	\label{fig: snar_graphs}
\end{figure}

Figure \ref{fig: snar_async_graphs} includes the whole set of results of the SnAr benchmark in the asynchronous setting. Figure \ref{fig: snar_graphs} includes the results for the synchronous setting. Each experiment is the mean $\pm$ half the standard deviation of 10 different runs.

\end{document}